\documentclass[times,twocolumn,final]{elsarticle}
\usepackage{medima}
\usepackage{framed,multirow}

\usepackage{amsmath,amssymb,amsfonts}
\usepackage{algorithmic}
\usepackage{graphicx}
\usepackage{textcomp}
\usepackage{xcolor}
\usepackage{acronym}
\usepackage{hyperref}
\usepackage{booktabs}
\usepackage{import}
\usepackage{gensymb}

\usepackage[utf8]{inputenc}
\usepackage{pgfplots}
\DeclareUnicodeCharacter{2212}{−}
\usepgfplotslibrary{groupplots,dateplot}
\usetikzlibrary{patterns,shapes.arrows}
\pgfplotsset{compat=newest}

\usepackage{enumitem} 

\usepackage{makecell}
\newcommand{\pro}{\textcolor{green}{~~\llap{\textbullet}~~}}
\newcommand{\con}{\textcolor{red}{~~\llap{\textbullet}~~}}

\definecolor{newcolor}{rgb}{.8,.349,.1}

\journal{Medical Image Analysis}

\newcommand{\discuss}[1]{#1}
\newcommand{\discussed}{$\bigstar$}

\newcommand{\metrics}{\textbf{Metrics:}}
\newcommand{\CCBY}{licensed under CC BY 4.0}
\newcommand{\screenshot}[3][noref]{\begin{figure}[tb]
		\centering
		\includegraphics[width=\linewidth]{#2}
		\caption{#3}
  \label{fig:#1}
\end{figure}}
\newcommand{\invivo}{{\em in vivo}}
\newcommand{\exvivo}{{\em ex vivo}}

\newcommand{\Exvivo}{{\em Ex vivo}}

\begin{document}
\verso{Adam Schmidt \textit{et~al.}}

\begin{frontmatter}
\title{Tracking and Mapping in Medical Computer Vision: A Review}
	\author[1]{Adam \snm{Schmidt}\corref{cor1}}
	\cortext[cor1]{Corresponding author: 
		}
	\ead{adamschmidt@ece.ubc.ca}
	\author[3]{Omid \snm{Mohareri}}
	\author[3]{Simon \snm{DiMaio}}
	\author[2]{Michael C. \snm{Yip}}
	\author[1]{Septimiu E. \snm{Salcudean}}

	\address[1]{The University of British Columbia, 2329 West Mall, Vancouver V6T 1Z4, BC}
	
	\address[2]{The University of California, 9500 Gilman Dr, La Jolla, California 92093, USA}
	\address[3]{Advanced Research, Intuitive Surgical, 1020 Kifer Rd, Sunnyvale, California 94086, United States}
	
	\received{16 October 2023}
	\finalform{8 Feb 2024}
	\accepted{29 Feb 2024}

\begin{abstract}

As computer vision algorithms increase in capability, their applications in clinical systems will become more pervasive.
	These applications include: diagnostics, such as colonoscopy and bronchoscopy; guiding biopsies, minimally invasive interventions, and surgery; automating instrument motion; and providing image guidance using pre-operative scans.
	Many of these applications depend on the specific visual nature of medical scenes and require designing algorithms to perform in this environment.

	In this review, we provide an update to the field of camera-based tracking and scene mapping in surgery and diagnostics in medical computer vision.
    We begin with describing our review process, which results in a final list of 515 papers that we cover.
	We then give a high-level summary of the state of the art and provide relevant background for those who need tracking and mapping for their clinical applications.
	After which, we review datasets provided in the field and the clinical needs that motivate their design.
	Then, we delve into the algorithmic side, and summarize recent developments.
	This summary should be especially useful for algorithm designers and to those looking to understand the capability of off-the-shelf methods.
        We maintain focus on algorithms for deformable environments while also reviewing the essential building blocks in rigid tracking and mapping since there is a large amount of crossover in methods. With the field summarized, we discuss the current state of the tracking and mapping methods along with needs for future algorithms, needs for quantification, and the viability of clinical applications.
	We then provide some research directions and questions.
    We conclude that new methods need to be designed or combined to support clinical applications in deformable environments, and more focus needs to be put into collecting datasets for training and evaluation.
\end{abstract}

\end{frontmatter}

\section{Introduction}
To begin, we define camera-based tracking and mapping in medical computer vision (MCV).
In tracking, methods observe the environment using a camera and estimate the motion and position of objects in it.
This motion includes that of the camera, instruments, or tissue in the environment.
In mapping, methods take in data and create a persistent underlying representation that can be used for other applications.
This underlying representation essentially provides a memory state to tracking and mapping methods.
In mosaicking for example, the map is an image, while in Simultaneous Localization and Mapping (SLAM) it is often a point cloud.
Tracking and mapping often go hand in hand.
By tracking and mapping, we mean methods that both create a map of the scene and perform tracking on the same scene.
We focus on methods that utilize camera-based imaging devices such as endoscopes, bronchoscopes and cytoscopes.
However, it is important to note that while some methods employ these devices, others, such as microscopy, utilize techniques that are outside the scope of our discussion.

To motivate the importance of these methods in medical applications, we will provide some brief background.
For medical intervention involving cameras, be it controlling a robot, scanning autonomously, or using scans to guide the surgeon, it is extremely important to know where tissue is, and how it is moving.
The case is the same for diagnostics: in colonoscopy and bronchoscopy it is important to localize (find the position of) the camera to enable accurate surveys of the tissue and guide biopsies.
This is important in easing the process for clinicians, in addition to improving outcomes for patients (e.g., for colonoscopy: earlier detection of cancers, and for minimally invasive surgery (MIS): better margins in tumor resection).

Tracking and mapping in MCV poses specific challenges, some of which are outlined next.
Many organs, such as the colon, can have low texture, and this makes matching points between images difficult~\citep{widyaWholeStomach3D2019}.
Even if textured, fluid on tissue surfaces reflects light.
For endoscopes with a collocated light, this creates reflections when the tissue is normal to the camera~\citep{makkiEllipticalSpecularityDetection2023}.
This causes saturated brightness patches in images that then need to be masked~\citep{zhao3DEndoscopicDepth2022}.
Organs deform, and their deformation can occur when they are out of view, which makes map creation difficult~\citep{azagraEndoMapperDatasetComplete2022}.
A changing environment requires priors to estimate what is happening out of frame, but modelling these priors proves difficult~\citep{schuleModelbasedSimultaneousLocalization2022}.
Blood and fluids in endoscopy can blur or smudge the camera and affect the video data~\citep{richterAutonomousRoboticSuction2021}.
Smoke created during electrocautery changes the depth estimation problem from one where there is a clear path for light rays to one in which the volume of smoke has to be removed~\citep{liuSurfaceDeformationTracking2023}.
Addressing these difficulties is important to create successful tracking and mapping algorithms, and could help to improve patient outcomes, ease clinical tasks, and reduce cost of care.

Research in medical computer vision is fast-moving given the concepts, and progress, it shares with multiple intersecting fields.
Some of these fields include: human tracking, SLAM for robotics and self driving cars, mosaicking, panorama creation, neural rendering, and point tracking.
The medical field has requirements for precise tracking of points, models that deal well with deformation, and means to generate useful results with small amounts of training data.
As demonstrated with the difficulties mentioned in the last paragraph, the main difference between algorithms in MCV and other algorithms is that the objects under observation in MCV are different from those outside the body.
Since medical data often has a distinct appearance, it is important to address this with a specific framing of losses and models suited to the medical environment that includes the priors within.
This could be done via training on medical data, designing specific losses, or building models with the priors embedded into the model itself.
Thus, in this review, we limit the search to applications in the medical field that use cameras for measurement.
With the specifics of MCV now noted, it is still important to consider relevant works in the non-medical computer vision field, 
especially since this is where many of the new developments and algorithms come from, with adaptation to deal with the specifics of medical data.
Work that lies outside of our search will be mentioned if relevant, or if it uses technical concepts built on medical applications.
Additional research that is useful but not currently used is listed in Section~\ref{sec:discussuseful}.

Our review begins with a detailed explanation of the review process (Section~\ref{sec:reviewprocess}) 
where we explain our literature search process
(Section~\ref{sec:litsearch}) followed by detailing prior relevant reviews and what makes our review necessary (Section~\ref{sec:priorreviews}).
Then, we summarize a broad list of medical specialties and the relevant algorithms that are useful for each of them in Section~\ref{sec:specialties}.
This should give algorithm designers, researchers, and clinicians a high-level overview of the clinical applications along with some example algorithmic needs.
Following that, we explain the datasets relevant to MCV in Section~\ref{sec:datasets}, which are of great importance for both training and evaluating algorithms.
In Section~\ref{sec:methods}, we delve deep into the algorithms and cover relevant works to help the reader understand the benefits, approaches, and design decisions for the applications that were mentioned in Section~\ref{sec:specialties}.
The flowchart in Fig.~\ref{fig:flowchart} provides a high-level overview of the relevant methods.
Finally, in Section~\ref{sec:discussion}, we provide a discussion on the features and drawbacks of algorithms, along with future needs and discussion points as we draw connections between the different algorithms.
We follow up this discussion with some questions and needs that still need to be addressed in tracking and mapping in MCV.
Finally, we conclude and summarize the state of the field.

\textbf{For the researcher looking for inspiration:} We recommend reading about datasets (Section~\ref{sec:datasets}) and methods (Section~\ref{sec:methods}), followed by the discussion (Section~\ref{sec:discussion}).
This provides an overview of where there are research gaps along with ideas for future research.

\textbf{For the engineer looking to implement or use an algorithm:} We recommend reading Section~\ref{sec:specialties} along with Section~\ref{sec:methods}, and selecting methods according to their details and the algorithmic needs.

\textbf{For the clinician or researcher looking to understand the field:} We recommend reading the whole paper, and referring to Fig.~\ref{fig:flowchart} for guidance on how methods interrelate.

\section{Review Process}
\label{sec:reviewprocess}

\subsection{Literature Search}
\label{sec:litsearch}

We review all papers which perform any sort of camera-based mapping or tracking in medical computer vision (MCV).
These can include mosaicking (Section~\ref{sec:mosaicking}), depth estimation (Section~\ref{sec:depthmapping}), tissue tracking (Section~\ref{sec:tissuetracking}), structure from motion (SfM) (Section~\ref{sec:sfm}), shape from template (SfT) (Section~\ref{sec:sft}), simultaneous localization and mapping (SLAM) (Section~\ref{sec:slam}), and nonrigid variants (which are in explained in their respective sections).
Refer to the referenced sections for more details on each method.
We survey any of these methods that use a clinical camera (e.g. endo/colono/bronchoscope/etc).
With these specifics, we perform a SCOPUS~\citep{Scopus2024} search to get a preliminary initial paper list.
Our search term reflects our criterion:
\texttt{
	(( TITLE-ABS-KEY ( ( mosaicing, OR mosaicking, OR "simultaneous localization and mapping" OR slam, OR (surface* w/6 reconstruction) OR "structure from motion" OR sfm OR (stereo w/6 reconstruction) OR (tissue w/6 track*) OR ( deform AND tracking OR mapping ) OR ( deformable AND tracking OR mapping ) OR ( deformation AND tracking OR mapping ) OR ( deforming AND tracking OR mapping ) ) AND ( endoscop* OR bronchoscop* OR colonoscop* OR "surgical" OR surgery OR (capsule w/6 robot*) OR (capsule w/6 camera) ) ) )) AND ( LIMIT-TO ( SUBJAREA,"COMP" ) OR LIMIT-TO ( SUBJAREA,"ENGI" ) )}
This term is a combination of tracking and mapping terms (reconstruction, mosaicking, SLAM) paired with (\texttt{AND}) terms related to surgery or diagnostics such as endoscopy, capsule cameras, etc.
\texttt{w/6} searches for terms within 6 words of one another.
On July 15th, 2023, this search returned 1497 results.
After culling irrelevant results based off title and abstract we were left with 563 papers.
Culling irrelevant papers was performed by removing items which included:
\begin{itemize}[noitemsep]
	\item Surgeon performance evaluation works
	\item Registration of multimodal images as the paper's primary topic. eg. MR to CT. Image guidance with multimodal imagery which uses camera data is still included.
	\item Endoscope or camera system designs (structured light, Lidar, etc.)
	\item Non-medical applications (sewer/pipe defect mapping, metal analysis, human hand pose)
	\item Video retrieval
\item Segmentation methods
	\item OCT and pCLE
	\item Needle steering and guidance
	\item Simulation platforms
	\item Surgical interventions (e.g. clinical grafting methods for eye surgery)
\end{itemize}

After this initial cull, we filtered out the papers that could not be decided on based solely on the abstract.
This was performed via reading the paper itself, which reduced the list to a final count of 516 papers.
After this, we separated the papers into groupings by application and algorithms, which helped to define the structure of this review.
Additional frequently encountered citations were added, along with recent papers that cite prior review papers.
See Fig.~\ref{fig:histogram} for a histogram plotting the number of included publications over time, and Fig.~\ref{fig:sankey} for a figure summarizing the filtering process.

\begin{figure}[tb]
	\centering
\pgfplotsset{compat=1.11,
    /pgfplots/ybar legend/.style={
    /pgfplots/legend image code/.code={\draw[##1,/tikz/.cd,yshift=-0.25em]
        (0cm,0cm) rectangle (3pt,0.8em);},
   },
}
\begin{tikzpicture}

\definecolor{chocolate2267451}{RGB}{226,74,51}
\definecolor{dimgray85}{RGB}{85,85,85}
\definecolor{gainsboro229}{RGB}{229,229,229}
\definecolor{lightgray204}{RGB}{204,204,204}
\definecolor{steelblue52138189}{RGB}{52,138,189}
\definecolor{bla}{RGB}{70,70,70}

\begin{axis}[
	/pgf/number format/.cd,
	use comma,
	1000 sep={},
axis background/.style={fill=gainsboro229},
axis line style={black},
legend cell align={left},
legend style={
  fill opacity=0.8,
  draw opacity=1,
  text opacity=1,
  at={(0.03,0.97)},
  anchor=north west,
  draw=lightgray204,
  fill=gainsboro229
},
tick align=outside,
tick pos=left,
title={Publications by year},
xlabel=\textcolor{dimgray85}{Year},
xmajorgrids,xminorgrids,
xmin=1998, xmax=2024,
xtick style={color=dimgray85},
xtick distance=4,
ylabel=\textcolor{dimgray85}{Number of Publications},
ymajorgrids,
ymin=0, ymax=48.3,
ytick style={color=dimgray85},
area legend,
]
\draw[draw=bla,fill=chocolate2267451,fill opacity=0.5,very thin] (axis cs:1998,0) rectangle (axis cs:1999,1);
\addlegendimage{ybar,ybar legend,draw=none,fill=chocolate2267451,fill opacity=0.5,very thin}
\addlegendentry{All}

\draw[draw=bla,fill=chocolate2267451,fill opacity=0.5,very thin] (axis cs:1999,0) rectangle (axis cs:2000,1);
\draw[draw=bla,fill=chocolate2267451,fill opacity=0.5,very thin] (axis cs:2000,0) rectangle (axis cs:2001,2);
\draw[draw=bla,fill=chocolate2267451,fill opacity=0.5,very thin] (axis cs:2001,0) rectangle (axis cs:2002,2);
\draw[draw=bla,fill=chocolate2267451,fill opacity=0.5,very thin] (axis cs:2002,0) rectangle (axis cs:2003,2);
\draw[draw=bla,fill=chocolate2267451,fill opacity=0.5,very thin] (axis cs:2003,0) rectangle (axis cs:2004,11);
\draw[draw=bla,fill=chocolate2267451,fill opacity=0.5,very thin] (axis cs:2004,0) rectangle (axis cs:2005,7);
\draw[draw=bla,fill=chocolate2267451,fill opacity=0.5,very thin] (axis cs:2005,0) rectangle (axis cs:2006,4);
\draw[draw=bla,fill=chocolate2267451,fill opacity=0.5,very thin] (axis cs:2006,0) rectangle (axis cs:2007,10);
\draw[draw=bla,fill=chocolate2267451,fill opacity=0.5,very thin] (axis cs:2007,0) rectangle (axis cs:2008,11);
\draw[draw=bla,fill=chocolate2267451,fill opacity=0.5,very thin] (axis cs:2008,0) rectangle (axis cs:2009,18);
\draw[draw=bla,fill=chocolate2267451,fill opacity=0.5,very thin] (axis cs:2009,0) rectangle (axis cs:2010,12);
\draw[draw=bla,fill=chocolate2267451,fill opacity=0.5,very thin] (axis cs:2010,0) rectangle (axis cs:2011,25);
\draw[draw=bla,fill=chocolate2267451,fill opacity=0.5,very thin] (axis cs:2011,0) rectangle (axis cs:2012,20);
\draw[draw=bla,fill=chocolate2267451,fill opacity=0.5,very thin] (axis cs:2012,0) rectangle (axis cs:2013,31);
\draw[draw=bla,fill=chocolate2267451,fill opacity=0.5,very thin] (axis cs:2013,0) rectangle (axis cs:2014,31);
\draw[draw=bla,fill=chocolate2267451,fill opacity=0.5,very thin] (axis cs:2014,0) rectangle (axis cs:2015,25);
\draw[draw=bla,fill=chocolate2267451,fill opacity=0.5,very thin] (axis cs:2015,0) rectangle (axis cs:2016,30);
\draw[draw=bla,fill=chocolate2267451,fill opacity=0.5,very thin] (axis cs:2016,0) rectangle (axis cs:2017,23);
\draw[draw=bla,fill=chocolate2267451,fill opacity=0.5,very thin] (axis cs:2017,0) rectangle (axis cs:2018,27);
\draw[draw=bla,fill=chocolate2267451,fill opacity=0.5,very thin] (axis cs:2018,0) rectangle (axis cs:2019,32);
\draw[draw=bla,fill=chocolate2267451,fill opacity=0.5,very thin] (axis cs:2019,0) rectangle (axis cs:2020,23);
\draw[draw=bla,fill=chocolate2267451,fill opacity=0.5,very thin] (axis cs:2020,0) rectangle (axis cs:2021,46);
\draw[draw=bla,fill=chocolate2267451,fill opacity=0.5,very thin] (axis cs:2021,0) rectangle (axis cs:2022,46);
\draw[draw=bla,fill=chocolate2267451,fill opacity=0.5,very thin] (axis cs:2022,0) rectangle (axis cs:2023,42);
\draw[draw=bla,fill=chocolate2267451,fill opacity=0.5,very thin] (axis cs:2023,0) rectangle (axis cs:2024,28);
\draw[draw=bla,fill=steelblue52138189,fill opacity=0.5,very thin] (axis cs:1998,0) rectangle (axis cs:1999,0);

\addlegendimage{ybar,ybar legend,draw=none,fill=steelblue52138189,fill opacity=0.5,very thin}
\addlegendentry{Neural}

\draw[draw=bla,fill=steelblue52138189,fill opacity=0.5,very thin] (axis cs:1999,0) rectangle (axis cs:2000,0);
\draw[draw=bla,fill=steelblue52138189,fill opacity=0.5,very thin] (axis cs:2000,0) rectangle (axis cs:2001,0);
\draw[draw=bla,fill=steelblue52138189,fill opacity=0.5,very thin] (axis cs:2001,0) rectangle (axis cs:2002,0);
\draw[draw=bla,fill=steelblue52138189,fill opacity=0.5,very thin] (axis cs:2002,0) rectangle (axis cs:2003,0);
\draw[draw=bla,fill=steelblue52138189,fill opacity=0.5,very thin] (axis cs:2003,0) rectangle (axis cs:2004,0);
\draw[draw=bla,fill=steelblue52138189,fill opacity=0.5,very thin] (axis cs:2004,0) rectangle (axis cs:2005,0);
\draw[draw=bla,fill=steelblue52138189,fill opacity=0.5,very thin] (axis cs:2005,0) rectangle (axis cs:2006,0);
\draw[draw=bla,fill=steelblue52138189,fill opacity=0.5,very thin] (axis cs:2006,0) rectangle (axis cs:2007,0);
\draw[draw=bla,fill=steelblue52138189,fill opacity=0.5,very thin] (axis cs:2007,0) rectangle (axis cs:2008,0);
\draw[draw=bla,fill=steelblue52138189,fill opacity=0.5,very thin] (axis cs:2008,0) rectangle (axis cs:2009,0);
\draw[draw=bla,fill=steelblue52138189,fill opacity=0.5,very thin] (axis cs:2009,0) rectangle (axis cs:2010,0);
\draw[draw=bla,fill=steelblue52138189,fill opacity=0.5,very thin] (axis cs:2010,0) rectangle (axis cs:2011,0);
\draw[draw=bla,fill=steelblue52138189,fill opacity=0.5,very thin] (axis cs:2011,0) rectangle (axis cs:2012,0);
\draw[draw=bla,fill=steelblue52138189,fill opacity=0.5,very thin] (axis cs:2012,0) rectangle (axis cs:2013,1);
\draw[draw=bla,fill=steelblue52138189,fill opacity=0.5,very thin] (axis cs:2013,0) rectangle (axis cs:2014,0);
\draw[draw=bla,fill=steelblue52138189,fill opacity=0.5,very thin] (axis cs:2014,0) rectangle (axis cs:2015,0);
\draw[draw=bla,fill=steelblue52138189,fill opacity=0.5,very thin] (axis cs:2015,0) rectangle (axis cs:2016,1);
\draw[draw=bla,fill=steelblue52138189,fill opacity=0.5,very thin] (axis cs:2016,0) rectangle (axis cs:2017,0);
\draw[draw=bla,fill=steelblue52138189,fill opacity=0.5,very thin] (axis cs:2017,0) rectangle (axis cs:2018,2);
\draw[draw=bla,fill=steelblue52138189,fill opacity=0.5,very thin] (axis cs:2018,0) rectangle (axis cs:2019,4);
\draw[draw=bla,fill=steelblue52138189,fill opacity=0.5,very thin] (axis cs:2019,0) rectangle (axis cs:2020,1);
\draw[draw=bla,fill=steelblue52138189,fill opacity=0.5,very thin] (axis cs:2020,0) rectangle (axis cs:2021,5);
\draw[draw=bla,fill=steelblue52138189,fill opacity=0.5,very thin] (axis cs:2021,0) rectangle (axis cs:2022,4);
\draw[draw=bla,fill=steelblue52138189,fill opacity=0.5,very thin] (axis cs:2022,0) rectangle (axis cs:2023,9);
\draw[draw=bla,fill=steelblue52138189,fill opacity=0.5,very thin] (axis cs:2023,0) rectangle (axis cs:2024,7);
\end{axis}

\end{tikzpicture}
 \caption{Histogram of publications after final filtering. This search was performed on July 15th, 2023, and thus not all publications from 2023 are necessarily included in this study. Neural denotes publications which have a title or abstract containing: CNN, GNN, or neural network.}
\label{fig:histogram}
\end{figure}
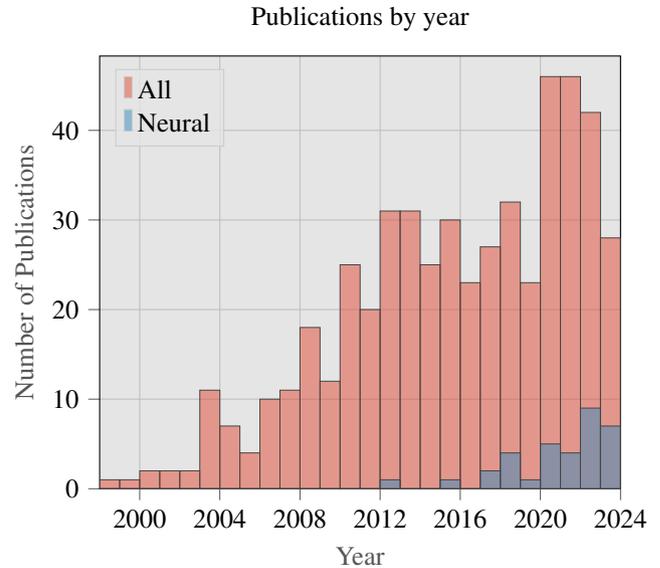\begin{figure}[tb]\includegraphics[width=\linewidth]{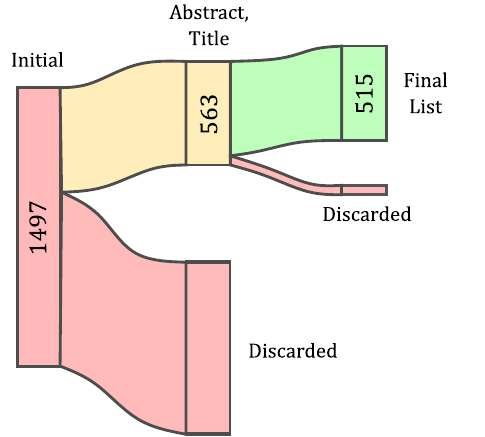}
	\caption{A Sankey diagram of the paper filtering process.}
    \label{fig:sankey}
\end{figure}

\subsection{Prior Reviews}
\label{sec:priorreviews}
To justify the necessity of this review and assert proper coverage in our list of included papers, we also performed a comprehensive search through all reviews from the last decade in the field.
By noting prior reviews, we help to motivate the need for a recent review in medical camera tracking and mapping.

In 2013,~\citet{maier-heinOpticalTechniques3D2013} provide an in-depth review of optical techniques for surface reconstruction covering: stereo, structured light, SfM, SLAM, Time-of-Flight, models, toolkits, and intra-operative registration.
Much has happened since then with the adoption of machine learning.
More recently, providing more detail on devices, ~\citet{fuFutureEndoscopicNavigation2021} reviews devices in optical and fluorescence imaging, along with providing a brief coverage of surgical tool tracking and SLAM methods.

Focusing on image stitching, surface reconstruction and view enhancement, \citet{bergenStitchingSurfaceReconstruction2016} provide a review that covers technology readiness and provide a useful classification of different methods and their clinical feasibility.
At a similar time,~\citet{linVideobased3DReconstruction2016} review the complementary problem of deformation recovery and surface reconstruction.  They concluded that deformation recovery and localization remain an open challenge.

In surgical data management and processing, \cite{munzerContentbasedProcessingAnalysis2018} focus on content-based processing (specularity removal, compression, retrieval) methods for endoscopic images.
Later,~\citet{maier-heinSurgicalDataScience2022} review the field of surgical data science, detailing infrastructure, data annotation, and analytics.

In augmented reality (AR),~\citet{bernhardtStatusAugmentedReality2017} provide an in depth review of the uses of augmented reality in laparoscopic surgery, which serves to motivate many image guidance applications. \citet{qianReviewAugmentedReality2020} provide a review of AR for robotic assisted surgery, summarizing methods and AR content used for each application (e.g. heart model, kidney, pre-op imaging).
\cite{malhotraAugmentedRealitySurgical2023} further review AR for surgical navigation but do not delve into models or deformation.
More broadly,~\cite{chadebecqArtificialIntelligenceAutomation2023} provide a review of artificial intelligence and automation in surgery, with a summary including robotic control, and other applications.

With clinical focus,~\citet{schneiderPerformanceImageGuided2021} perform a systematic review on image guided liver surgery, focusing on interventions.
They provide motivation for improving image guidance, and thus tracking algorithms as well.  \citet{acidiAugmentedRealityLiver2023} survey clinical applications of AR in liver surgery, concluding that the application of AR is limited due to insufficient precision, but stating that it is likely to become more effective with increased usage.
These reviews are of specific relevance to algorithmic applications in image guidance.

In summary, these reviews either cover specific subfields or do not have more recent technical details on deformation models or neural networks that are used for tracking and mapping.
In contrast to the mentioned reviews, we will be more algorithmically focused without constraining our discussions to devices or sensors such as Time-of-Flight.
Thus, our review fills the position as: a guide for recent algorithmic advances through the entire tracking and mapping process, a coverage of quantification and data, and finally a thorough discussion of needs for this field in the future.

\section{Medical Specialties and Relevant Applications}
\label{sec:specialties}
\begin{figure*}[tb]\includegraphics[width=\textwidth]{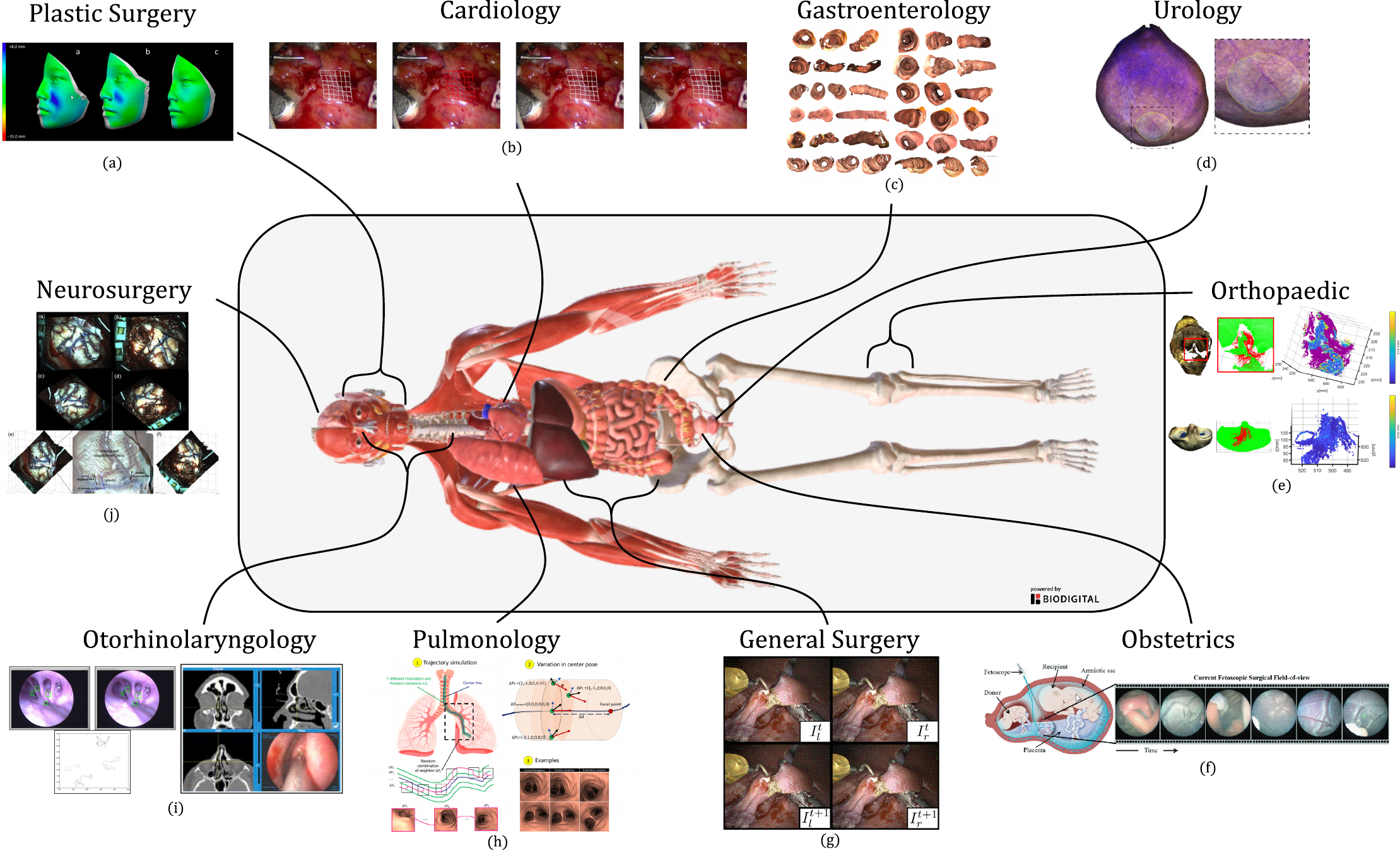}
	\caption{A body model with medical specialties that use tracking and mapping overlaid. Images are adapted from: (a). ~\cite{basergaEfficacyAutologousFat2020},
	(b). ~\cite{richaRobust3DVisual2011},
	(c). ~\cite{maRNNSLAMReconstructing3D2021},
	(d). ~\cite{soperSurfaceMosaicsBladder2012},
	(e). ~\cite{marmolDenseArthroSLAMDenseIntraArticular2019},
	(f). ~\cite{banoPlacentalVesselguidedHybrid2023},
	(g). ~\cite{schmidtSENDDSparseEfficient2023},
	(h).~\cite{borrego-carazoBronchoPoseAnalysisData2023},
	(i). ~\cite{burschkaScaleinvariantRegistrationMonocular2005},
	(j). ~\cite{jiCorticalSurfaceShift2014}.
	Permissions:
	(a, f). \CCBY{}.
	(b, c, h, i, j). reprinted with permission from Elsevier.
	(d, e). reprinted with permission from IEEE.
	(g). reprinted with permission from authors.
    The 3D body model is generated and used with permission from BioDigital.}
    \label{fig:bodymodel}
\end{figure*}

In this section, we briefly cover different medical specialties that have clinical applications requiring tracking or mapping.
Alongside this, we summarize the algorithms that are relevant to said specialty.
This should serve as a quick reference of sample works for clinicians and those implementing algorithms.
To those researching algorithms and MCV, this should serve as a overview of how broadly applicable some of the methods can be.
We separate the sections by clinical application: cardiology, orthopaedic, obstetrics, otorhinolaryngology (ear, nose, throat (ENT)), plastic surgery, pulmonology, gastroenterology, neurosurgery, urology, and general surgery.
For locating which body regions are relevant to each specialty, along with what the data looks like, the diagram in Fig.~\ref{fig:bodymodel} should be of use.
For every mentioned application, see the flowchart in Fig.~\ref{fig:flowchart} for a description of the algorithm and its dependencies.
For a quick summary of specialties, Table~\ref{tab:interventions} provides an overview.
We note that ophthalmology and dermatology also could have relevant applications, but we do not provide sections as there limited works with algorithmic focus at this point in time.

\subsection{Orthopaedics}
In orthopaedics, guiding navigation via aligning models to the camera feed requires localizing the arthroscope's position relative to the body.
This helps achieve the clinical objectives of better registration for implants and bone reconstruction in orthopaedic surgery~\citep{marmolEvaluationKeypointDetectors2017,marmolDenseArthroSLAMDenseIntraArticular2019,zhangSLAMTKARealtimeIntraoperative2022}, or automating interventions such as milling of bone.
The algorithms useful for this field are rigid mapping (SLAM, SfM), and of course all dependent algorithms (see Fig.~\ref{fig:flowchart}).

\subsection{Obstetrics}
In obstetrics, twin-to-twin transfusion syndrome is treated via anastomosing placental vessels between twins.
Visualizing the surface of the placenta is difficult due to a small field of view, thus algorithms look to extend the field of view via mosaicking~\citep{liGloballyOptimalFetoscopic2021, banoDeepSequentialMosaicking2019a,banoDeepLearningbasedFetoscopic2020}.
Additionally in obstetrics,~\citet{desmetEvaluatingPotentialBenefit2019} show that pelvic repair could benefit from stereo reconstruction via enabling better visualization than a 2D screen.
Thus, the relevant algorithms are mosaicking and stereo reconstruction.

\subsection{Otorhinolaryngology (ENT)}
In otorhinolaryngology, enabling tracheal robot steering using cameras on the tip of a robotic device could help ease deployment and avoid damage to critical structures~\citep{girerdAutomaticTipSteeringConcentric2020}.
Additionally, maps of the nasal passage can help in sinus surgery by registering pre-operative data to aid in avoiding critical structures~\citep{liuDenseDepthEstimation2020a}.
In these environments, SLAM, SfM, feature description, and depth estimation are of particular importance.

\subsection{Plastic Surgery}
Plastic surgery includes, but is not limited to, reconstructive operations on the face.
Predicting facial outcome for planning in maxillofacial surgery requires surface reconstruction~\citep{buchartHybridVisualizationMaxillofacial2009}.
Deformation modelling is also important to help create accurate craniofacial models for surgery~\citep{suputra3DLaplacianSurface2020}.
Stereo reconstruction can also be used for efficiently evaluating grafting outcomes after surgery~\citep{basergaEfficacyAutologousFat2020}.
Thus, both stereo reconstruction and nonrigid reconstruction are useful for plastic surgery.

\subsection{Neurosurgery}
In neurosurgery, brain shift between the time when the MRI scan is acquired and surgery affects the usability of the MRI-determined landmarks.
Deformable tracking is important here since the brain can undergo complex nonrigid deformation~\citep{hartkensMeasurementAnalysisBrain2003a}.
Therefore, methods that can visually track the surface of the brain could prove useful for deforming the preoperative scan accordingly~\citep{demomiMethodAssessmentTimevarying2016}.
This is especially true if the tracking can be performed using camera video without markers~\citep{jiangMarkerlessTrackingBrain2016}.
Recently, in neurosurgery, convolutional neural networks (CNNs) have been used to quantify vascular structures and track regions~\citep{martinUsingArtificialIntelligence2023}.

\subsection{Gastroenterology}

In gastroenterology, medical computer vision is useful for extending the camera's field of view with the purpose of ensuring coverage in colonoscopy screening.
This enables better detection of polyps or cancer by helping all regions to be seen and surveyed~\citep{maRealTime3DReconstruction2019, zhang3DReconstructionDeformable2021,turanNonrigidMapFusionbased2017}.
It is similarly helpful for stomach reconstruction where it can again help to detect ulcers or cancer.
To enable the reconstruction of a 3D surface such as the colon, successful localization of the camera is key~\citep{widyaStomach3DReconstruction2021}.
Thus, methods that are important in this field are SfM, NRSfM, SLAM, NR SLAM, and mosaicking.
These environments are nonrigid, so the accuracy of rigid methods when they are applied depends on the rigidity of capture and length of video.

\subsection{Cardiology}
In cardiology, being able to compensate for motion during heart surgery is a promising application of medical computer vision.
This is called motion compensation, where the goal is to give the surgeon the impression that the heart is stationary by moving the camera observing the heart in a synchronized manner with the heart motion and moving the robotic instruments relative to the heart's surface.
This requires accurately measuring the motion of the heart surface, which has been addressed algorithmically~\citep{richaRobust3DVisual2011,schoobStereoVisionbasedTracking2017}.
Tissue tracking, stereo reconstruction and deformable SLAM are the particular methods that are useful for this.

\subsection{Pulmonology}
In pulmonology, the primary image modality using medical computer vision is bronchoscopy.
In bronchoscopy, a camera is inserted into the lungs.
Thus, depth estimation and mapping are important for visually guiding the scope to a nodule biopsy~\citep{visentini-scarzanellaDeepMonocular3D2017,wangVisualSLAMbasedBronchoscope2020} rather than using fluoroscopy (live X-ray) or CT which expose the patient to ionizing radiation.
Thus, SLAM and deformable SLAM methods are of specific importance, as we would like to recover the pose of the bronchoscope to then be able to correctly localize the biopsy site.

\subsection{Urology}
Bladder cancer screening can require surveying the entire bladder to ensure all lesions can be found.
Therefore, creating panoramas could help aid in diagnostics~\citep{soperSurfaceMosaicsBladder2012}.
Designing algorithms to aid navigation can also make procedures easier by providing a map when inspecting the kidneys or ureters.
Kidney stone removal is an application of flexible ureteroscopy where it can be hard to orient the instrument.
SLAM methods have been introduced here as potential solutions~\citep{fuVisualelectromagneticSystemNovel2021, olivamazaORBSLAM3basedApproachSurgical2023}.
Thus, mosaicking and deformable SLAM are of relevance in urology.

\subsection{General Surgery}
In minimally invasive surgery, tracking and mapping would help improve surgical perception, thus providing better image guidance.
This could help to: improve margins in surgery by deforming pre-operative scans to track the movement of tissue, enable autonomous scanning and suturing, and ease proctoring~\citep{maier-heinComparativeValidationSingleshot2014, chadebecqArtificialIntelligenceAutomation2023}.
In this field, the main algorithmic applications are mosaicking, NR SLAM, and Nonrigid SfM, with NR SLAM being the one suitable for use during surgery, since it is real-time.

\begin{table*}
	\caption{Medical specialties, their clinical uses for camera-based medical computer vision, and their algorithmic needs. Some select references are in the final column. Terms: Automation (Auto.), Tissue Tracking (tis. track.), Depth Mapping (DM), Image Guidance (Guid.), Nonrigid SLAM (NR SLAM), Surface Reconstruction (Recon.), Mosaicking (Mos.), Motion Compensation (Mo. Comp.), Measurement (Meas.), Twin to Twin Transfusion Syndrome (TTTS). Fields: Orthopaedic (Ortho.), Obstetrics (Obste.), Plastic surgery (Plastics.), Neurosurgery (Neuro.), Gastroenterology (Gastro.), Cardiology (Cardio.), Pulmonology (Pulmo.), Urology (Uro.), Otorhinolaryngology (ENT).}
\centering
\begin{tabular}{lll}\toprule
	Location & Clinical Use & Algorithms\\
	\cmidrule{1-3}
	Ortho. & Auto., Guid. & SfM, SLAM ~\citep{marmolEvaluationKeypointDetectors2017,maKneeArthroscopicNavigation2020,zhangSLAMTKARealtimeIntraoperative2022}\\
	Obste. & TTTS., Pelv. surg. & DM, Mos. ~\citep{desmetEvaluatingPotentialBenefit2019,banoDeepLearningbasedFetoscopic2020}\\
	ENT & Guid., Auto., & SfM, SLAM ~\citep{girerdAutomaticTipSteeringConcentric2020, liuDenseDepthEstimation2020a}\\
	Plastics. & Recon. & DM, NRSfM ~\citep{suputra3DLaplacianSurface2020,basergaEfficacyAutologousFat2020}\\
	Neuro. & Img. Guid. & NRSLAM, tis. track.~\citep{jiangMarkerlessTrackingBrain2016,martinUsingArtificialIntelligence2023}\\
	Gastro. & Recon., Diag. & Mos., (NR)SfM, (NR)SLAM ~\citep{maRealTime3DReconstruction2019,widyaStomach3DReconstruction2021}\\
	Cardio. & Mo. Comp., Meas. & DM, tis. track., NRSLAM ~\citep{richaRobust3DVisual2011,schoobStereoVisionbasedTracking2017}\\
	Pulmo. & Diag., Biopsy & NRSLAM ~\citep{visentini-scarzanellaDeepMonocular3D2017,wangVisualSLAMbasedBronchoscope2020}\\
	Uro. & Cancer, Uretoscopy & Mos., NRSfM, NRSLAM ~\citep{soperSurfaceMosaicsBladder2012,olivamazaORBSLAM3basedApproachSurgical2023}\\
	Gen. surg. & Auto., Guid., Meas., Recon. & DM, tis. track., NRSLAM ~\citep{maier-heinComparativeValidationSingleshot2014,chadebecqArtificialIntelligenceAutomation2023}\\
	\bottomrule
\end{tabular}
\label{tab:interventions}
\end{table*}

\section{Datasets}

\label{sec:datasets}
\subsection{Introduction}

In this section, we detail datasets that have been released and are available for quantifying tracking and mapping methods in MCV.
As a sample, some of these datasets include labelled data for evaluating: image stitching, stereo estimation, reconstruction, or tracking.
Datasets which are for segmentation or classification are excluded.
Datasets which have been used for tracking but do not have labels will be mentioned in brief.

We begin in Section~\ref{sec:unlabelled} with a summary of datasets that do not have any ground truth which are primarily useful for training unsupervised methods.
In Section~\ref{sec:pseudotruth}, we delve into datasets with algorithmically generated ground truth. 
The algorithmically generated datasets are in their own section because they depend on the reconstruction accuracy of stereo algorithms or SfM and are limited to be at best as good as the classical reconstruction methods used to create them.
We then follow this up with summarizing simulated ground truth datasets, generated via rendered 3D models, in Section~\ref{sec:simulated}, and physical phantoms, e.g., silicone tissue models, in Section~\ref{sec:phantoms}.
We finally close with ground truth that uses real tissue in Section~\ref{sec:realtissue}.
By real tissue, we mean tissue from animal or human sources.
We note that the truly ideal data would be both \invivo{}, and human.

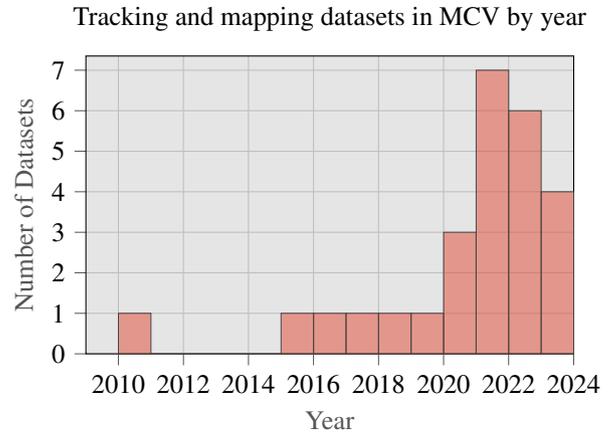
\begin{figure}[tb]
	\centering
\begin{tikzpicture}

\definecolor{chocolate2267451}{RGB}{226,74,51}
\definecolor{dimgray85}{RGB}{85,85,85}
\definecolor{gainsboro229}{RGB}{229,229,229}
\definecolor{lightgray204}{RGB}{204,204,204}
\definecolor{bla}{RGB}{70,70,70}

\begin{axis}[
/pgf/number format/.cd,
	use comma,
	1000 sep={},
axis background/.style={fill=gainsboro229},
axis line style={black},
legend style={
  fill opacity=0.8,
  draw opacity=1,
  text opacity=1,
  at={(0.03,0.97)},
  anchor=north west,
  draw=lightgray204,
  fill=gainsboro229
},
height=5.5cm,
width=8cm,
tick align=outside,
tick pos=left,
title={Tracking and mapping datasets in MCV by year},
xlabel=\textcolor{dimgray85}{Year},
xmajorgrids,xminorgrids,
xmin=2009, xmax=2024,
xtick style={color=dimgray85},
xtick distance=2,
ylabel=\textcolor{dimgray85}{Number of Datasets},
ymajorgrids,
ymin=0, ymax=7.35,
ytick distance=1,
ytick style={color=dimgray85}
]

\draw[draw=bla,fill=chocolate2267451,fill opacity=0.5,very thin] (axis cs:2010,0) rectangle (axis cs:2011,1);
\draw[draw=bla,fill=chocolate2267451,fill opacity=0.5,very thin] (axis cs:2011,0) rectangle (axis cs:2012,0);
\draw[draw=bla,fill=chocolate2267451,fill opacity=0.5,very thin] (axis cs:2012,0) rectangle (axis cs:2013,0);
\draw[draw=bla,fill=chocolate2267451,fill opacity=0.5,very thin] (axis cs:2013,0) rectangle (axis cs:2014,0);
\draw[draw=bla,fill=chocolate2267451,fill opacity=0.5,very thin] (axis cs:2014,0) rectangle (axis cs:2015,0);
\draw[draw=bla,fill=chocolate2267451,fill opacity=0.5,very thin] (axis cs:2015,0) rectangle (axis cs:2016,1);
\draw[draw=bla,fill=chocolate2267451,fill opacity=0.5,very thin] (axis cs:2016,0) rectangle (axis cs:2017,1);
\draw[draw=bla,fill=chocolate2267451,fill opacity=0.5,very thin] (axis cs:2017,0) rectangle (axis cs:2018,1);
\draw[draw=bla,fill=chocolate2267451,fill opacity=0.5,very thin] (axis cs:2018,0) rectangle (axis cs:2019,1);
\draw[draw=bla,fill=chocolate2267451,fill opacity=0.5,very thin] (axis cs:2019,0) rectangle (axis cs:2020,1);
\draw[draw=bla,fill=chocolate2267451,fill opacity=0.5,very thin] (axis cs:2020,0) rectangle (axis cs:2021,3);
\draw[draw=bla,fill=chocolate2267451,fill opacity=0.5,very thin] (axis cs:2021,0) rectangle (axis cs:2022,7);
\draw[draw=bla,fill=chocolate2267451,fill opacity=0.5,very thin] (axis cs:2022,0) rectangle (axis cs:2023,6);
\draw[draw=bla,fill=chocolate2267451,fill opacity=0.5,very thin] (axis cs:2023,0) rectangle (axis cs:2024,4);
\end{axis}

\end{tikzpicture}
 \caption{Histogram of publicly available datasets usable for camera-based tracking and mapping in MCV from datasets mentioned in Table~\ref{tab:datasets}.}
\label{fig:datasethistogram}
\end{figure}

We separate the datasets into these classes as different data types can be vulnerable to different biases.
For example, simulation or phantom data might not carry over to real tissue data.
Alongside the sections, we have a table of datasets to reference in Table~\ref{tab:datasets}, and a figure showing their availability over time in Fig.~\ref{fig:datasethistogram}.
This table should provide a means to get a high-level summary of different algorithmic approaches and dataset generation.
For our discussion on the datasets, please go to Section~\ref{sec:discussdata} near the end of this review.

\begin{table*}[tb]
	\caption{Datasets released with applications in tracking and mapping. Reconstruction (Recon.), Abdomen (Abd.), Fetoscopy (Fet.). Truth abbreviations: Bounding boxes (BB), Visible (Vis), Infrared (IR), Depth Mapping: DM., Transfer learning (Transf.). R/S/P: (Real Tissue/Simulated (rendered)/Phantom).}
\centering
\begin{tabular}{llllll}\toprule
	Dataset	& R/S/P & Location	& Rigid	& Truth	&	Use\\
	\cmidrule{1-6}
	\citet{stoyanovRealtimeStereoReconstruction2010,prattDynamicGuidanceRobotic2010} & P & Heart & Nonrigid & Stereo (CT) & Recon.\\
 	\citet{maier-heinCrowdtruthValidationNew2015} & R & Abd. & Nonrigid & Annotated & Stereo Recon.\\ 

	\citet{yeOnlineTrackingRetargeting2016} & R & Abd. & Nonrigid & Annotated (BB) & Tracking, Retargeting\\
 
 	\citet{visentini-scarzanellaDeepMonocular3D2017} & S & Lung & Rigid & Sim/Phantom CT & DM, Transf., Navigation\\
  
	\citet{penzaEndoAbSDatasetEndoscopic2018} & P & Abd. & Rigid & Laser Scan & Stereo Recon.\\ 

	\citet{rauImplicitDomainAdaptation2019} & S & Colon & Rigid & Simulation & Stereo Recon.\\
 
	\citet{liSuperSurgicalPerception2020} & R & Abd. & Nonrigid & Annotated & Tracking\\
	\citet{fultonComparingVisualOdometry2020} & P & Colon & Nonrigid & Phantom, Pose & Localization, Navigation\\ 
 	\citet{banoDeepPlacentalVessel2020} & R & Fet. & Nonrigid & Annotated Sem. Labels & Mosaicking\\
	\citet{banoFetRegPlacentalVessel2021} & R & Fet. & Nonrigid & Annotated Sem. Labels & Mosaicking\\
  
 	\citet{zhang3DReconstructionDeformable2021} & S & Colon & Nonrigid & Simulation & Deformable 3D Recon.\\
	\citet{recasensEndoDepthandMotionReconstructionTracking2021} & R & Abd. & Nonrigid & LibELAS & Training/Tracking\\
	\citet{ozyorukEndoSLAMDatasetUnsupervised2021} & R & GI & Rigid & Scanner & SLAM, DM\\
	\citet{ozyorukEndoSLAMDatasetUnsupervised2021} & S & GI & Rigid & Rendering & SLAM, DM\\
	\citet{xiRecoveringDense3D2021} & R & Abd. & Rigid & Neural & Monocular Recon.\\
	\citet{zhangTemplateBased3DReconstruction2021} & S & Colon & Rigid & Sim & SLAM, DM\\
	\citet{allanStereoCorrespondenceReconstruction2021} & R & Abd. & Rigid & Structured Light & Stereo Recon.\\
 
	\citet{edwardsSERVCTDisparityDataset2022} & R & Abd. & Rigid & CT & Stereo Recon.\\
	\citet{guyQualitativeComparisonImage2022} & S & Abd. & Rigid & Simulation & Stitching\\ \citet{rauBimodalCameraPose2022} & S & Colon & Rigid & Simulation & Depth and SLAM\\ \citet{azagraEndoMapperDatasetComplete2022} & S & Colon & Nonrigid & Simulation & SLAM\\
	\citet{azagraEndoMapperDatasetComplete2022} & R & Colon & Nonrigid & Colmap & SLAM\\
	\citet{bobrowColonoscopy3DVideo2022} & P & Colon & Rigid & Phantom & Recon., Localization\\
	\citet{cartuchoSurgTChallengeBenchmark2024} & R & Abd. & Nonrigid & Annotated & Tracking\\
	\citet{hayozLearningHowRobustly2023} & R & Abd. & Nonrigid & Kinematics & Rel. pose est.\\
	\citet{linSemanticSuPerSemanticawareSurgical2023} & R & Abd. & Nonrigid & Vis Markers & Tracking, Recon.\\
	\citet{schmidtSTIRSurgicalTattoos2023} & R & Abd. & Nonrigid & IR Markers & Tracking, Recon.\\
\bottomrule
\end{tabular}
\label{tab:datasets}
\end{table*}

\subsection{Unlabelled Datasets}
\label{sec:unlabelled}
As described in our review process (Section~\ref{sec:reviewprocess}), we focus on literature related to tracking and mapping.
We exclude unlabelled datasets that are useful in other domains, or are designed for tasks such as segmentation, since they are seldom used in tracking work.
Starting with datasets that are often used, the Hamlyn Centre datasets include many unlabelled sequences from procedures using both monocular and stereo cameras~\citep{mountneyThreedimensionalTissueDeformation2010}, in addition to some stereo sequences with a deforming heart~\citep{stoyanovSoftTissueMotionTracking2005}.
Additionally, they provide some datasets designed for qualitatively evaluating tissue tracking in varied environments and with different artifacts such as smoke, blood, and lens smudges~\hbox{\citep{giannarouProbabilisticTrackingAffineinvariant2013}}.
They also provide a dataset of unlabelled stereo image pairs for the purpose of evaluating unsupervised methods using photometric reconstruction error~\citep{yeSelfSupervisedSiameseLearning2017}. 
Photometric reconstruction evaluates how accurately a depth estimation works for reproducing an image by using photometric error, which compares colors at image pixels, but does not provide actual measurements of reconstruction accuracy.
The Hamlyn datasets that provide labels will be referenced in later sections.

\subsection{Algorithmic Ground Truth Datasets}
\label{sec:pseudotruth}
By algorithmic ground truth, we mean data that is generated via a reconstruction algorithm and can act as a pseudo ground truth.
Reconstruction algorithms include stereo algorithms, SLAM, or SfM.
By using reconstruction algorithms to generate ground truth data, we have to assume that they are accurate.
This limits the performance evaluation of new algorithms. 
For example, a classical SfM method will only obtain sparse points in a rigid manner and does not deal with lighting effects such as specularities, and thus cannot be used to robustly evaluate a new method that addresses these issues. 

Many works have used stereo depth networks to evaluate accuracy, with EndoDepthAndMotion~\citep{recasensEndoDepthandMotionReconstructionTracking2021} being one.
They release a dataset with ground truth generated by LibELAS~\citep{geigerEfficientLargeScaleStereo2011} in abdominal sequences.
This dataset is intended for training depth models and evaluating tracking methods.
In another dataset,~\citet{xiRecoveringDense3D2021} generate pseudo-ground truth using autoencoders.
They design a network for monocular depth learning along with a method for point cloud completion.
They evaluate their algorithm on the EndoAbS~\citep{penzaEndoAbSDatasetEndoscopic2018} dataset, and then release the point clouds created with their network. 

\subsection{Simulated data}
\label{sec:simulated}
MCV scenes can be generated by rendering from simulation. 
Recent methods have been improving the photorealism of these simulations, bringing simulation closer to the true environment.
Some of these methods use CT scans and phantoms, but they are still grouped into  being simulated if they use rendered data for ground truth.
In~\citet{visentini-scarzanellaDeepMonocular3D2017}, the authors generated 32 video sequences with ground truth depth and rendering in a simulated bronchoscopy.
These sequences are generated using a rigid realistic lung phantom with rendering performed using a model from paired CT scans.
The rendering contains frames that act as depth ground truth.
To align the physical phantom with the simulation model, they use SLAM and follow it with  Iterative Closest Point (ICP) alignment.
The dataset is designed for transfer learning in depth networks for modelling from rendered to real tissue and vice-versa, and for depth estimation and mapping in bronchoscopy.
\citet{rauImplicitDomainAdaptation2019} also release rendered ground truth depth frames in monocular colonoscopy that are generated via simulation based on CT scans.

Since it is very difficult to obtain ground truth in colonoscopy due to the nonrigid environment of the colon,~\citet{zhang3DReconstructionDeformable2021} opt to use simulated colonoscopies.
To construct realistic models, they texture four different CT scans by applying colors and lighting parameters to a mesh.
Then, to simulate nonrigid motion, they deform their simulated tubular colon model about the centerline.
They release depth maps and monocular frames from their dataset for evaluation of reconstruction algorithms.
They also release a similar dataset with fifteen colon models, generated from a rigid model~\citep{zhangTemplateBased3DReconstruction2021}.
Instead of using a monocular camera, this dataset provides stereo pairs, and includes ground truth camera poses.

Moving on to systems for simulation in minimally invasive surgery (MIS), VisionBlender~\citep{cartuchoVisionBlenderToolEfficiently2021} propose and publish code for creating simulated endoscopic data along with a utility for creating depth maps, optical flow, poses, and normals.
Later, in a similar vein of simulation, but for image stitching instead of depth estimation and flow,~\citet{guyQualitativeComparisonImage2022} generate a dataset for evaluating image stitching.
Specifically, they look to merge images taken at the same time in multi-camera setups.
They look to address difficulties that occur in stitching such as the duplication of or disappearing of objects in the surgical field.
Their simulation framework can generate tools and organs with varying camera models and is shown in Fig.~\ref{fig:guy2022}.
In C3VD (Colonoscopy 3D Video Dataset),~\citet{bobrowColonoscopy3DVideo2022} release many video sequences with video from 3D printed phantoms alongside sequences from the rendered simulated models.
In SimCol,~\citet{rauBimodalCameraPose2022} release another colonoscopy dataset but with the additions of monocular pose and depth images.
This can help evaluate SLAM and depth mapping frameworks, although this dataset does not include deformation.
Alongside this submission, the authors propose a novel pose estimation network.
Reconstructions using their framework are demonstrated in Fig.~\ref{fig:rau2022}.
They provide depth, pose, flow, and the 3D models as a part of their dataset.

\screenshot[guy2022]{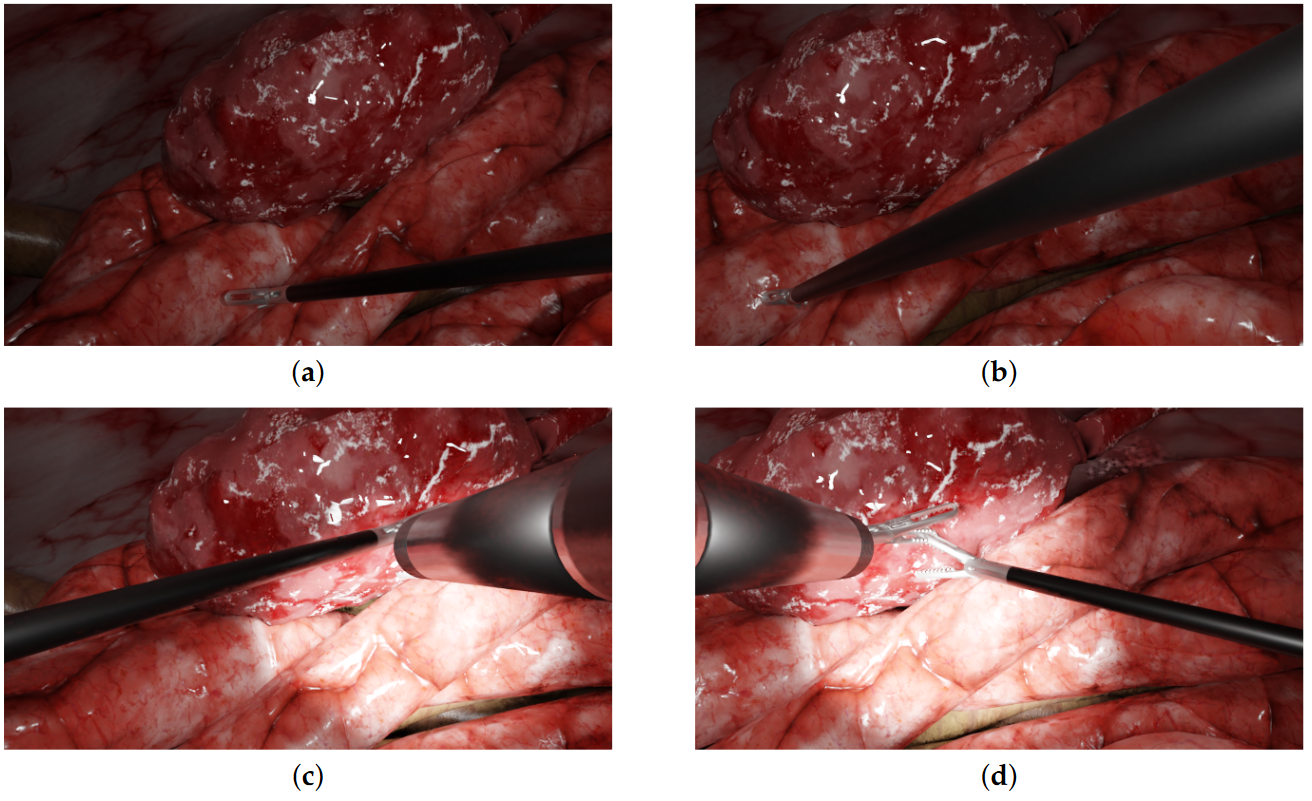}{Dataset simulation framework for multi-camera systems to evaluate image stitching algorithms. From~\cite{guyQualitativeComparisonImage2022} \CCBY{}} 

\screenshot[rau2022]{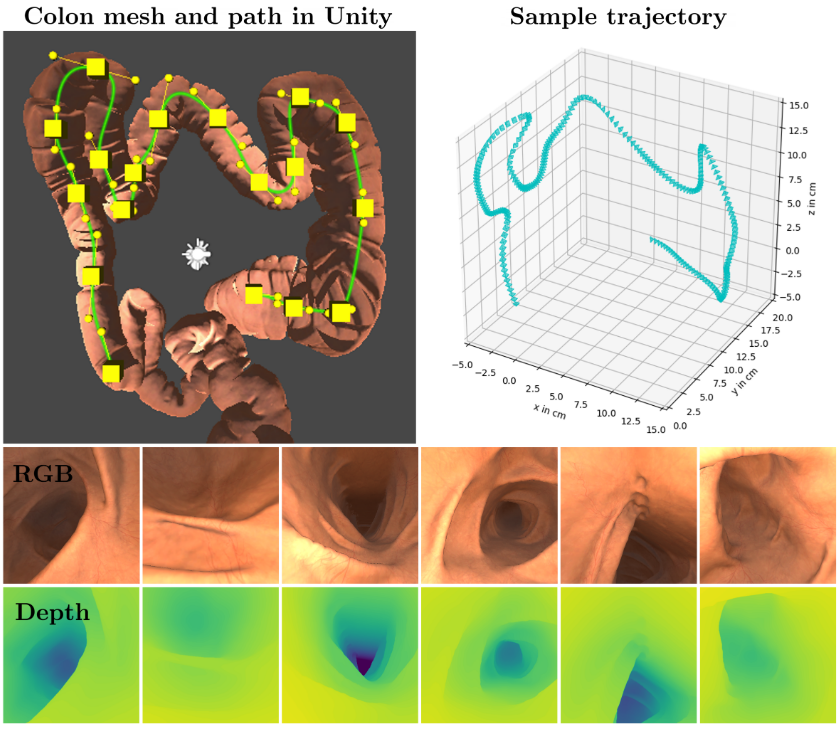}{A colon reconstruction framework for generating sythetic data. The framework renders depth and RGB frames of a colon model in Unity. From~\citet{rauBimodalCameraPose2022} \CCBY{}}

In the EndoMapper dataset~\citep{azagraEndoMapperDatasetComplete2022}, data is presented from both real tissue and simulated scenarios.
The real tissue dataset comprises videos and camera calibrations without ground truth labels.
For the real tissue dataset, they provide some algorithmically generated ground truth via 3D reconstructions generated with COLMAP~\citep{schonbergerPixelwiseViewSelection2016a, schonbergerStructureFromMotionRevisited2016}--a publicly available library for generating point clouds using SfM.
This data is released in partial colon segments, since SfM can fail in colonoscopy on larger environments.
For the simulated section of their dataset release, they artificially deform their model to better represent motions of a real colon.
In the simulated dataset, they release depth, video frames and the camera trajectory (pose over time).
The dataset is available with a release request for nonprofit institutions.

\subsection{Phantoms}
\label{sec:phantoms}
These datasets are designed to quantify performance using phantoms, which are physically printed or sculpted models of organs or different environments.
One of the Hamlyn datasets~\citep{stoyanovRealtimeStereoReconstruction2010,prattDynamicGuidanceRobotic2010} includes a beating heart phantom.
Using CT, 3D ground truth is created and then registered to the stereo camera.
This dataset can be used for evaluating stereo algorithms and tracking performance.
In EndoAbS~\citep{penzaEndoAbSDatasetEndoscopic2018}, release a dataset for evaluating stereo reconstruction which comprises 120 stereo pairs with camera calibration.
Their ground truth is generated on abdominal organ phantoms using a laser scanner.
They collect stereo frames over multiple different distances, lighting, and smoke conditions. \citet{fultonComparingVisualOdometry2020} release a dataset with a deformable phantom colon.
The ground truth they provide is camera pose generated via a magnetic tracker.
They collect sequences with multiple different levels of deformation.
They additionally survey the performance of different visual odometry (VO) systems in correctly estimating pose using their dataset.
\cite{edwardsSERVCTDisparityDataset2022} introduce a methodology for evaluating stereo algorithms via paired CT scans.
Their dataset comprises 16 stereo image pairs of varying organ phantoms along with CT-generated 3D ground truth.

\subsection{Real Tissue}
\label{sec:realtissue}
In summarizing real tissue datasets, we include work that focuses on surgical tissue and organs, both \invivo{} and \exvivo{}.
Beginning with tissue tracking and deformable mapping datasets,~\citet{maier-heinCrowdtruthValidationNew2015} introduce crowdsourcing to address labelling of endoscopic data in which users track salient points and label them using software.
The authors release a methodology for generating validation sets.
Alongside the methodology, they release a set of one hundred annotated stereo pairs.
\citet{yeOnlineTrackingRetargeting2016} also release a dataset for evaluating tracking in endoscopy with data generated via user labelling, where users label the bounding boxes of tracked regions throughout a video clip.
SuPer~\citep{liSuperSurgicalPerception2020} and SurgT~\citep{cartuchoSurgTChallengeBenchmark2024} perform a similar user labelling procedure for tissue in stereo endoscopy.
Later, Semantic SuPer~\citep{linSemanticSuPerSemanticawareSurgical2023} uses green pins to mark the tissue surface rather than requiring user labelling.
With Surgical Tattoos in Infrared (STIR),~\citet{schmidtSTIRSurgicalTattoos2023} introduce a dataset for evaluating tissue tracking, SLAM, and reconstruction methods.
It comprises labeled points in infrared and stereo video clips, with the benefit being that it neither requires software labelling, nor visible markers that can affect algorithm evaluation.

For pose estimation and depth mapping in the gastrointestinal tract, EndoSLAM~\citep{ozyorukEndoSLAMDatasetUnsupervised2021} provides a rigid dataset with video from many different capsule cameras and endoscopes.
The ground truth is obtained as point clouds generated from a 3D scanner that are aligned to the camera frame with ICP.
\Exvivo{} sequences are acquired by attaching tissue to a foam scaffold.
Fig.~\ref{fig:ozyoruk2021} demonstrates their collection methodology.
They provide a separate synthetic dataset as well.

Addressing accurate depth generation in MIS, SCARED~\citep{allanStereoCorrespondenceReconstruction2021} provides a dataset of depth maps calculated using structured light.
This is provided alongside stereo endoscopic videos.
Focusing on pose,~\citet{hartwigMITISLAMBenchmark2022} release the MITI dataset which includes stereo video and camera pose collected during a surgical intervention.
The pose is calculated using an IMU (inertial measurement unit) and infrared (IR) markers.
StereoMIS~\citep{hayozLearningHowRobustly2023} also address the problem of quantifying pose estimation, focusing specifically on estimating relative pose between images.
Their dataset releases relative pose calculated using kinematics alongside stereo videos from porcine models.

In a different vein,~\citet{banoDeepPlacentalVessel2020} provide segmentation of hundreds of frames of vessels in fetoscopic procedures for mosaicking.
This dataset is extended to a multi-center dataset with thousands of labeled frames by~\citet{banoFetRegPlacentalVessel2021}, and is used for a challenge~\citep{banoPlacentalVesselguidedHybrid2023}.
This work is a helpful reference for other segmentation datasets and methods relevant to mosaicking.

As seen, the datasets using real tissue vary in the actual ground truth they provide.
These include using pose, depth, or motion as truth, along with using different methods for collecting each type.

\screenshot[ozyoruk2021]{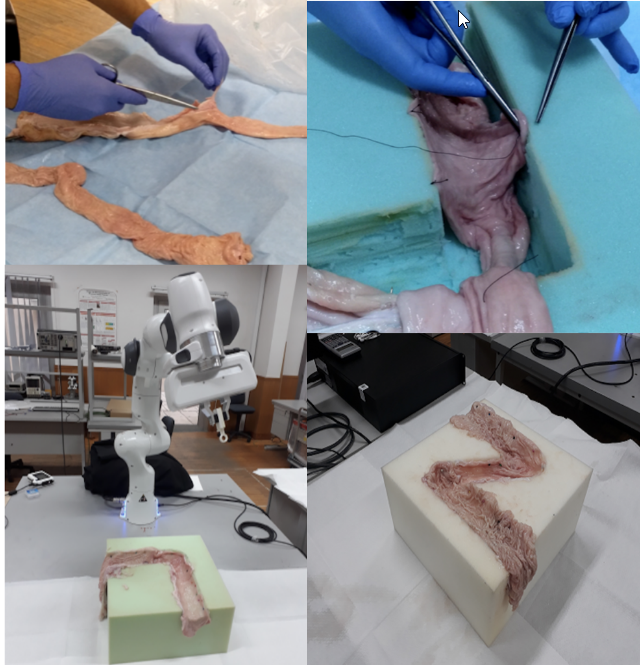}{The dataset collection methodology figure for the EndoSLAM dataset. Porcine tissue is sewn to foam scaffolds, and then scanned with a 3D scanner. From~\citep{ozyorukEndoSLAMDatasetUnsupervised2021} reprinted with permission from Elsevier.}

\section{Algorithms}

\label{sec:methods}
\begin{figure*}[tb]\includegraphics[width=\textwidth]{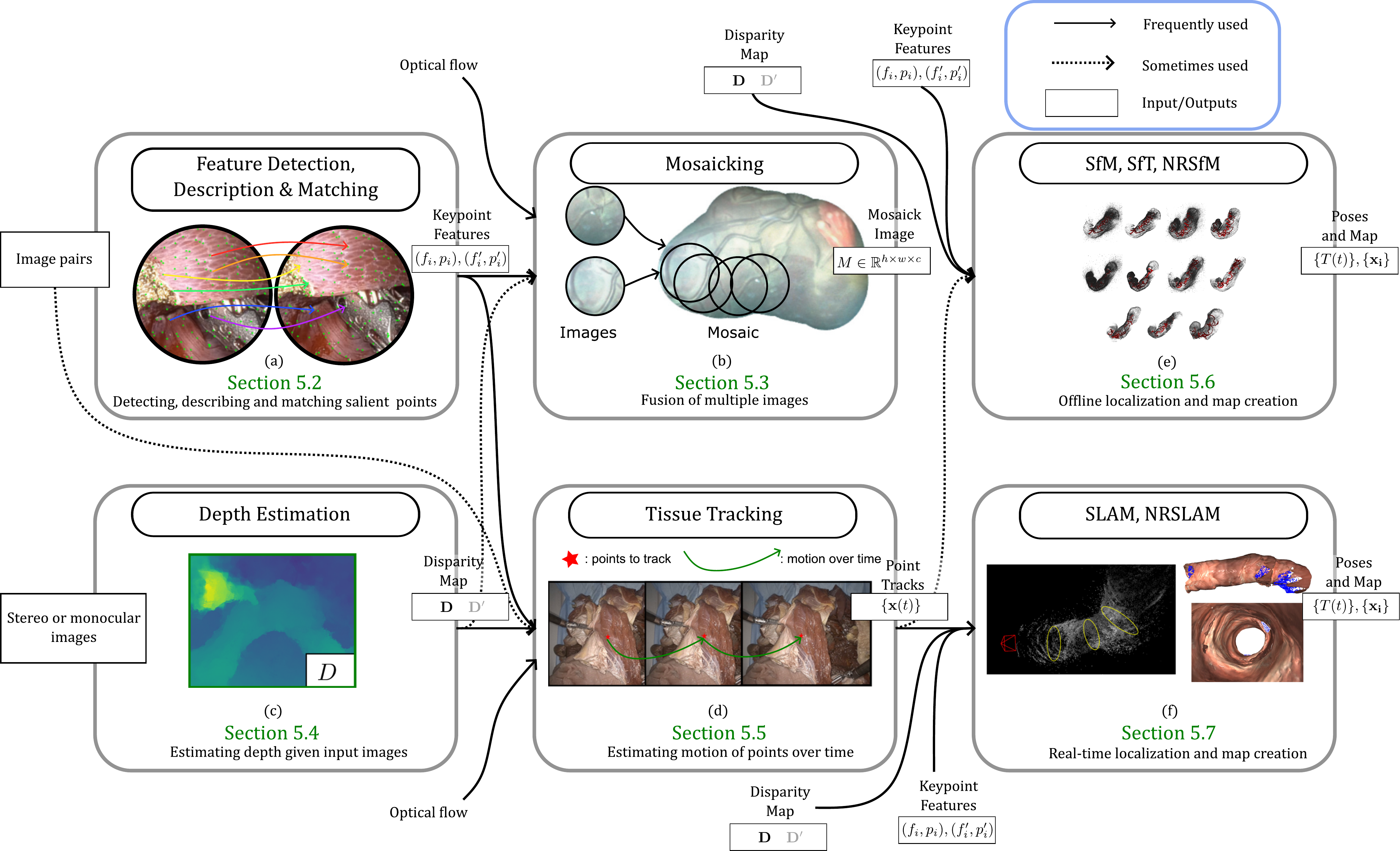}
	\caption{A flowchart of the different methods, their inputs, and their outputs. (b). Adapted and edited from~\cite{banoPlacentalVesselguidedHybrid2023} \CCBY. (d). adapted from~\cite{schmidtSTIRSurgicalTattoos2023} \CCBY. (e). Rigid SfM from~\cite{widyaWholeStomach3D2019} \CCBY. (f). Rigid SLAM in the colon adapted from~\citep{maRNNSLAMReconstructing3D2021} with permission from Elsevier.}
    \label{fig:flowchart}
\end{figure*}

In this section we begin with the important technical building blocks in tracking and mapping, and then move into more complex methods that manage deformation.
First, we detail references for metrics commonly used in the field in Section~\ref{sec:metrics}, summarized in Table~\ref{tab:metrics}.
Each following subsection includes a bolded paragraph, (\metrics{}), denoting the specific metrics used for each method described.
Then, to review the methods, feature detection, description, and matching in MCV are covered in Section~\ref{sec:features}.
Following this, mosaicking, in which features are used to fuse images into panoramas is covered in Section~\ref{sec:mosaicking}.
In Section~\ref{sec:depthmapping}, we cover depth mapping which calculates the 3D position of 2D image pixels.
Then, we summarize surgical tissue tracking, which looks to track points in the surgical scene, in Section~\ref{sec:tissuetracking}.
For tracking while using a map as well, see the later section on SLAM.
After which, in Section~\ref{sec:sfm} we explain rigid and nonrigid (NR) Structure from Motion (SfM), and Shape-from-* methods that estimate shape using a model or a set of points.
Finally, in Section~\ref{sec:slam} we cover rigid and nonrigid (NR) SLAM which aim to create a real-time map from a video of the surgical environment.
In the SLAM section, we also include related methods that address the mapping problem without a localization focus.
We note that for select methods that can perform 3D tracking and deformation we include small summaries of takeaways in Table~\ref{tab:proscons}.
Citations that are in this table are denoted with a (\discussed{}) alongside their mention in the text.

See Fig.~\ref{fig:flowchart} for an illustration of how all these methods depend on one another.

\subsection{Metrics}
\label{sec:metrics}

\begin{table*}[tbp]
	\caption{A summary of some metrics used in medical computer vision. GT: ground truth}
	\centering
	\begin{tabular}{ll}
		\toprule
		Metric & Description\\
		\cmidrule{1-2}

		IOU & Intersection over union. A measure of intersection of two segments. \\
		PSNR & Peak signal-to-noise ratio. A ratio that measures estimator quality/noise, measured in decibels. \\
		SSIM & Structural similarity index measure. A perception-based measure for comparing image pairs. \\
		LPIPS & Learned perceptual image patch similarity metric. Trained on a real-world dataset~\citep{zhangUnreasonableEffectivenessDeep2018}. \\
		MSE & Mean squared error. Average squared error between two sets of paired measurements.\\
		RMSE & Root mean squared error. Square root of mean-squared error between two sets of paired points. \\
		MAE & Mean absolute error. Mean of absolute error between two sets of paired points. \\
		MedAE & Median absolute error. Median of absolute error between two sets of paired points. \\
		Chamfer Distance & Averaged distance between two point sets. Distance is calculated between each point and its nearest. \\
		mAA & Mean average accuracy. Accuracy averaged over multiple thresholds. \\
		mAP & Mean average precision. Precision averaged over multiple thresholds. \\
		ATE & Absolute trajectory error. Translational trajectory difference between two camera trajectories. \\
		PCK & Percentage of correct keypoints. \\
		Forward-backward & A measure of track drift/stability~\citep{kalalForwardBackwardErrorAutomatic2010}. \\
		MMA & Mean Matching Accuracy. The average percentage of correct keypoint matches~\citep{dusmanuD2NetTrainableCNN2019}. \\
		PCK & Percentage of correct keypoints. \\
		RPE & Relative Pose Error~\citep{sturmBenchmarkEvaluationRGBD2012}. \\
		epe. & Endpoint error. Euclidean distance between the estimated and GT end points in tracking. \\
		Abs Rel. & Absolute relative difference. The absolute distance between GT and estimated depth divided by GT.\\
		Sq Rel. & Squared relative difference. The squared version of Abs Rel.\\
		$\delta$ & Accuracy at a certain threshold.\\
		\bottomrule
	\end{tabular}
	\label{tab:metrics}
\end{table*}

Many of the methods we cover choose to evaluate their performance with varying metrics.
Here is a brief guide for where to refer for more detail on these metrics.
For more detail on image analysis metrics, refer to~\citet{maier-heinMetricsReloadedRecommendations2023}.
For mosaicking-specific, ones, see~\citet{banoPlacentalVesselguidedHybrid2023}.
For tissue tracking, refer to STIR~\citet{schmidtSTIRSurgicalTattoos2023}, and point tracking benchmarks~\citet{doerschTAPVidBenchmarkTracking2022}.
For pose estimation, the image matching challenge is a good reference~\citet{jinImageMatchingWide2021}.
For SLAM systems, refer to~\citet{sturmBenchmarkEvaluationRGBD2012}.
For depth estimation, refer to the SCARED challenge~\citet{allanStereoCorrespondenceReconstruction2021}.
A text description of metrics that we mention is in Table~\ref{tab:metrics}.

In terms of evaluation, comparison between methods can be difficult, and for specific results, we ask readers to refer to the SurgT challenge for tissue tracking~\citep{cartuchoSurgTChallengeBenchmark2024}, FetReg~\citep{banoPlacentalVesselguidedHybrid2023} for mosaicking, and SCARED~\citep{allanStereoCorrespondenceReconstruction2021} for depth estimation.

\subsection{Feature Description and Detection}
\label{sec:features}
\subsubsection{Introduction}

The purpose of image features is to provide a numerical means to create correspondences between images.
Therefore, having well-defined image features has been a well-established goal for the purpose of enabling methods in tracking deformation.
Image features assign numerical vectors to positions and can be either sparse or dense.
By comparing these vectors, features can be matched to create data correspondences.
The feature error comparison vectors (similarity scores) are used to create data association terms, which are terms in the cost function for optimization models such as SLAM or relative pose estimation.
In this section we summarize sparse features (Section~\ref{sec:sparsefeature})) and feature matching (Section~\ref{sec:matching}), followed by dense features (Section~\ref{sec:densefeature})) used in MCV.
Sparse features are often used for image alignment/mapping, or other problems that require computational efficiency.
The follow-up task of feature matching is often performed only for sparse features.
For dense features, instead of using feature matching methods, we can perform a search over the entire image since the features are calculated on a regular image grid.
Dense features are calculated over a whole image grid and provide higher resolution at the cost of efficiency.

\metrics{}
Feature detection and description often evaluate their performance for downstream tasks, since features are seldom used on their own.
The downstream tasks can include pose estimation, in which mAA and relative pose error (RPE) are used.
If the downstream application evaluates tissue tracking, then metrics such as endpoint error are used.
This metric can be used when evaluating frame-to-frame matching, SfM, or SLAM works.
Forward-backward error can be used to estimate feature robustness for matching forward and backward in time.
Downstream accuracy estimates for SLAM systems such as tracking loss are also sometimes used.
MMA/PCK is also used, which evaluates matching accuracy over multiple thresholds, and requires ground truth feature matches.

\subsubsection{Sparse Features}
\label{sec:sparsefeature}
Sparse features are generated by two components: \textit{detection} and \textit{description}.
\textit{Detection} is the process of finding locations \(p_i\) for each keypoint \(i\) in an image \(I\).
\textit{Description} assigns each keypoint a d-dimensional numerical vector \(f_i \in \mathbb{R}^d\), which could also be binary.
SIFT~\citep{loweObjectRecognitionLocal1999a}, SURF~\citep{baySpeededUpRobustFeatures2008}, and ORB~\citep{rubleeORBEfficientAlternative2011a} are examples of classical descriptors. Classical in this sense means they are hand engineered and use intensity histograms, decision trees, etc., to create the numerical descriptor values.
Classical descriptors are still frequently used in SLAM works~\citep{lamarcaDefSLAMTrackingMapping2021,songMISSLAMRealTimeLargeScale2018}.
Early descriptors for surgical environments used feature histograms and decision trees along with LK (Lucas-Kanade) optical flow~\citep{mountneySoftTissueTracking2008}. \citet{giannarouAffineinvariantAnisotropicDetector2009} proposed an affine-invariant detector that detects points over scales, assigning ellipses to them to better deal with varying angle and scale.
Classical descriptors do remain in use with many applications, such as registration of pre-operative brain images to a camera~\citep{jiangRobustAutomatedMarkerless2015}.
On usage in the brain,~\citet{jiangMarkerlessTrackingBrain2016} use segmented Frangi features~\citep{frangiMultiscaleVesselEnhancement1998} -- which detect tube-like structures -- for non-rigid registration of brains using vessel/sulci surface features.
Classical features have also been evaluated in arthroscopy, which deals with a fairly rigid environment~\citep{marmolEvaluationKeypointDetectors2017}.

Moving onto neural applications, there are learned sparse features that are applicable to surgery. For example, ReTRo~\citep{schmidtRealTimeRotatedConvolutional2021a} proposes a lightweight real-time descriptor, trained using camera-pose self-supervision~\citep{wangLearningFeatureDescriptors2020} in surgical environments.
The authors use classical motivations to train a neural network that samples and rotates like ORB.
Although this does not include tissue deformation in training pairs, it contains the same point from different views.
To work in deformable spaces, although not trained on surgical data,~\citet{potjeEnhancingDeformableLocal2023} propose training deformable features using data augmentation with a thin plate spline. 
More specifically to surgery,~\citet{barbedTrackingAdaptationImprove2023} present a SuperPoint~\citep{detoneSuperPointSelfSupervisedInterest2018} style descriptor and detector which uses a COLMAP~\citep{schonbergerStructureFromMotionRevisited2016} reconstruction for training.
Rather than depending on homographies (a re-projection that treats an image as planar), they propose tracking adaptation, which trains on the re-projections of the 3D points.
This should help the descriptors perform in surgical environments.
With the same goal of improving performance in surgical environments,~\cite{karaogluRIDESelfSupervisedLearning2023} note that the surgical environments differ from real-world images which are often oriented vertically, and they propose RIDE which builds rotation equivariance into the network design.

\subsubsection{Feature Matching}
\label{sec:matching}
After obtaining a sparse set of features \(\{\mathit{f}_i\}\) with their positions \(\{p_i \in \mathbb{R}^2\}\) in an image, we need to match them to corresponding features and positions in the other image \(I'\): \(\{f'_j\}\), \(\{p'_j\}\).
This is often done via performing a dot product between features, \(f_i \cdot f'_j = c\), to obtain a similarity score.
Modern methods can better match features using more than just their descriptor values.
By using descriptors along with the relative motion estimated motion, for example, they can design efficient and accurate feature matching schemes.
GMSMatch (Grid-Based Motion Statistics~\citep{bianGMSGridBasedMotion2020}) aids matching by using heuristics based on the motion of surrounding matches.
Recent neural network-based matchers such as SuperGlue~\citep{sarlinSuperGlueLearningFeature2020} and LightGlue~\citep{lindenbergerLightGlueLocalFeature2023} learn matching based on a graph neural network of points.
The principle of these modern methods is to take in a point and, rather than brute-force match, use the motion of the surrounding points, their features, or both.
For example, if a match is in a different motion direction than all of its surrounding matches then it can be discarded.
GMSMatch uses a heuristic for this, while SuperGlue uses a learned graph neural network trained using homographies and large-scale outdoor depth reconstruction scenes.
These methods could be trained for surgical environments as well if they are provided ground truth, or robust reconstructions such as SfM.

For match filtering specifically in surgical environments,~\citet{chuEndoscopicImageFeature2020} use A-SIFT descriptors for laparoscopy and gastroscopy.
They assume that features move smoothly and slowly in these environments and perform match filtering via expectation maximization (EM).
Again, using EM,~\citet{zhangRobustFeatureMatching2023} refine matches under the assumption that the environment can be represented with Dual-Quaternion Blending (DQB).
This does not allow for discontinuities or transformations that do not fit the smooth DQB deformation field.
We note that detected points need not necessarily be discarded entirely, because they can still provide useful information, e.g., texture. 
Since feature matching is very sensitive to position, 
in~\citet{schmidtFastGraphRefinement2022a}, the authors chose to keep all keypoint matches, but train to refine (instead of discard) the detections to best improve downstream photometric reconstruction using graph neural networks (GNNs).

\subsubsection{Research in Dense Image Descriptors}
\label{sec:densefeature}

Research in dense image descriptors is less common, likely due to computational costs.
Indirectly, some models could be said to create dense descriptors (e.g. the stereo or optical flow networks mentioned later), but these models directly use the features as part of the model, so it comes down to a question of semantics.
Models that are designed primarily as a means for feature description, e.g., ~\citet{liuExtremelyDensePoint2020}, train a CNN-based descriptor model in a novel way.
They use SfM to generate ground truth for their dense descriptor.
To find matches, the detected point searches for matches over the entire image.
On a system with 4 NVIDIA Tesla M60 GPUs, this takes \(\sim37\)ms to match a set of descriptors on a \(256 \times 320\) image.
Since this is a convolutional search method, we can expect the costs to scale by the amount of additional keypoints and the increase in image size.
For a full-resolution image (\(1024 \times 1280\)), we can expect it to take anywhere from \(\sim 150\)ms if the CNN is the bottleneck (16x the data) to \(\sim 2400\)ms if the bottleneck is in the matching step (16x the matches and 16x the data).

\subsection{Mosaicking}
\label{sec:mosaicking}

\subsubsection{Introduction}
Mosaicking is the process of creating a compound image using a collection of images over time.
This is performed by first matching and aligning similar features in the images by warping the images.
This is often followed by color-correcting via blending, introduced by \citet{burtMultiresolutionSplineApplication1983}, and still in frequent use today (including in MCV).
Having a mosaicked image, \(M\), can help in interventions or diagnostics where the camera only provides a small field of view.
Mosaicking can be done in a 2D manner, or in 3D on a surface such as a sphere or mesh.
Although they still use mosaicking, we omit works in pCLE and microscopy (as per our literature search methodology), to maintain our focus on work that uses video images for tracking and mapping.
For more information,~\citet{banoChapter15Image2024} provide a very recent summary chapter on mosaicking.

\metrics{}
In terms of metrics, mosaicking is often evaluated in two ways: using segmentation accuracy, via IOU and mean IOU; and using texture accuracy, via n-frame SSIM~\citep{banoFetRegPlacentalVessel2021, banoPlacentalVesselguidedHybrid2023}.
Texture accuracy is often used for evaluating registration quality, since a measure of how well images align is desired.
For evaluating segmentation, methods use IOU since it provides a metric for how well segmented regions overlap.

\subsubsection{Mosaicking in MCV}
In an early work on retinal and catadioptric endometrial videos, ~\citet{seshamaniRealtimeEndoscopicMosaicking2006a} propose using mosaicking to create a broader field of view.
They do this by aligning images photometrically with an affine transformation for each image, and provide an algorithm that can run at native camera frame rate (30fps).
Mosaicking using images from fibroscopes is challenging because of the many artifacts and specularities present.
To address this,~\citet{atasoyGlobalApproachAutomatic2008a} propose a method that uses SIFT features to match between images.
They additionally solve for a global alignment, where the relative transformation is calculated by optimizing over all frames.
This better allows consensus and reduces the drift that can be caused when just aligning on a frame-to-frame basis, since errors can compound.
This is similar to bundle adjustment in SfM and SLAM (as we shall see in Section~\ref{sec:sfm}).
They evaluate their method on \exvivo{} kidney tissue.
In order to account for image differences, they merge images with multi-band blending~\citep{burtMultiresolutionSplineApplication1983} which partitions the images to remove low frequency variations while preserving high frequency details.
A similar method is proposed and evaluated on \invivo{} experiments with applications to bladder mosaicking in urology~\citep{miranda-lunaMosaicingBladderEndoscopic2008a}.
For endoscopy,~\citet{bergenFeaturebasedRealtimeEndoscopic2009} generate a mosaick using a Kanade-Lucas Tomasi tracker (KLT) with RANSAC (Random Sample Consensus) for outlier removal.
A homography transformation is estimated between frames, and specularity removal is performed via masking.
In cytoscopy, mosaicking using dense cross-correlation is also used for aligning images, and the results are evaluated clinically in research by~\citet{hernandez-mierFastConstructionPanoramic2010}.
Here, the tissue surface can be ill-featured, and this can make robust matching difficult.
Because of poor features, it can be helpful to perform mosaicking in the fluorescent imaging spectrum, where features of interest, such as tumors, are better illustrated.
By mosaicking on fluorescent images, diagnostics can be improved by providing a wider field of view~\citep{behrensRealtimeImageComposition2011}.
Even though most mosaicking work has been performed in 2D, images can be projected and blended on 3D surfaces as well.
For example, 3D spherical models have been created to provide 360\degree{} views of the bladder~\citep{soperSurfaceMosaicsBladder2012}.
None of these methods account for re-aligning features when a camera loops back to the same location, and~\citet{weibelGraphBasedConstruction2012a} solve this by providing a way to close and align loops via detecting when features are seen again.
For more details on stitching and mosaicking,~\cite{bergenStitchingSurfaceReconstruction2016} provide an in-depth review of works before 2016.

More recently, mosaicking methods have begun utilizing machine learning.
\citet{banoDeepSequentialMosaicking2019a} use a CNN-based homography estimation network that takes in image pairs and estimates a homography between them.
They adapt it to fetoscopy via adding data augmentation along with outlier rejection for artifacts such as specularity.
\cite{banoDeepPlacentalVessel2020} have also designed CNN-segmentation models for vessel-based mosaicking. 
Recently, they found that a combination of deep learning for homography along with matching vessel segmentation maps creates a hybrid method that outperforms either method on their own~\citep{banoPlacentalVesselguidedHybrid2023}.
See Fig.~\ref{fig:bano2023} for their mosaicking architecture.

The concepts of loop closure and pose graphs from SLAM are also used in fetoscopic mosaicking.
A pose graph is a connected set of camera locations with measurements or co-observance of features acting as connections.
When used with loop closure, it allows for better global alignment and bundle adjustment~\citep{liGloballyOptimalFetoscopic2021}.
By combining a neural method along with the idea of a pose graph in endoscopy,~\citet{liRobustEndoscopicImage2023} mosaick using both neural optical flow and SIFT keypoint matches by optimizing the image transformations in an underlying pose graph.

\screenshot[bano2023]{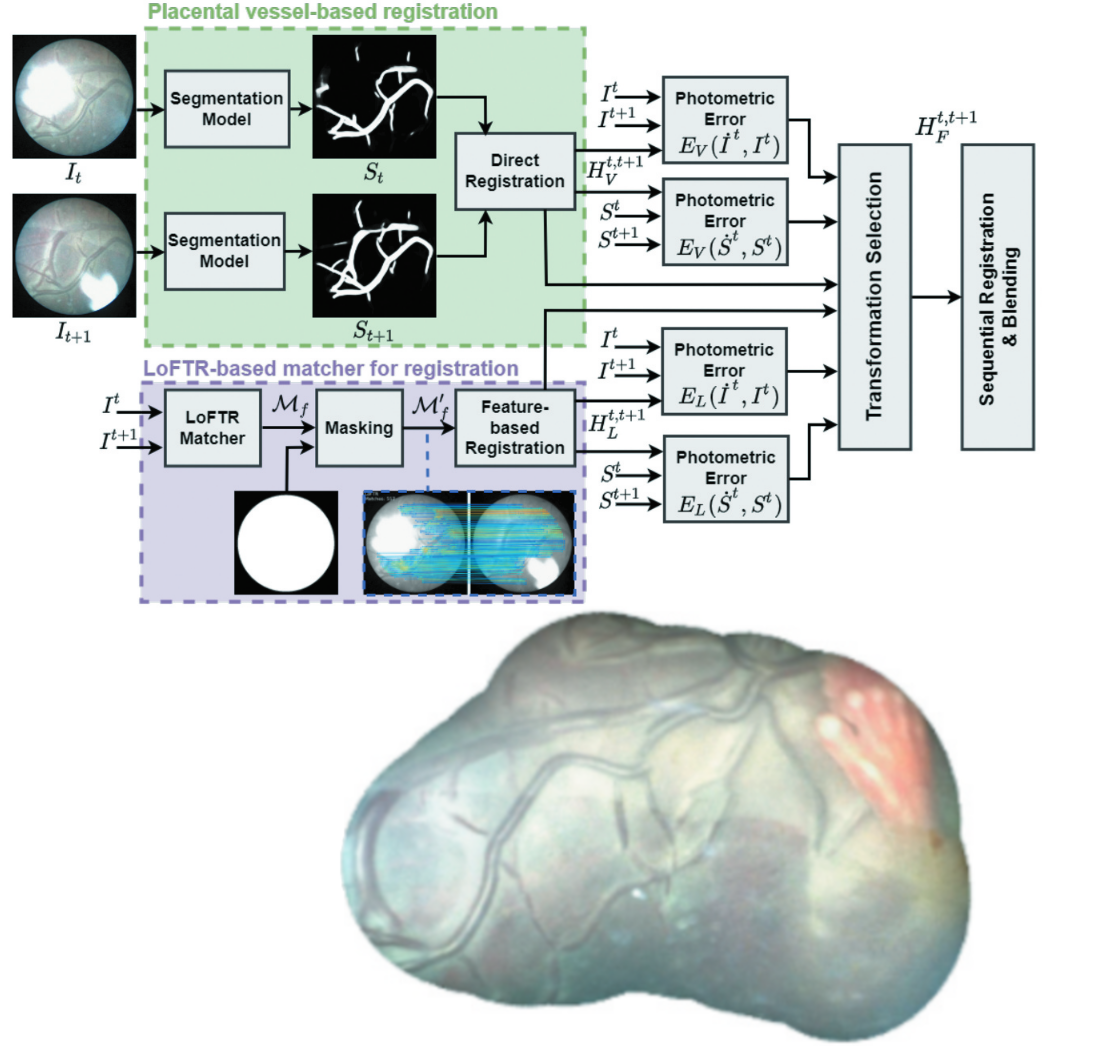}{Placental fetoscopic mosaicking using a combination of vessel segmentation and dense feature matching. From~\cite{banoPlacentalVesselguidedHybrid2023} \CCBY}

\subsection{Stereo and Monocular Depth Estimation}
\label{sec:depthmapping}

\subsubsection{Introduction}
In order to track a point in 3D, its depth must be known.
This requires a depth estimation network.
In stereo formulations, this network estimates the disparity value of each pixel in the image, denoted as a disparity map, \(\mathbf{D} \in \mathbb{R}^{h \times w \times 1}\).
In monocular methods the formulation is similar, except the scale is unknown.
The disparity is the relative difference in pixels from a point to the camera center between each image in the stereo pair.
This disparity of a point can then be used along with the camera matrix to calculate the 3D position of that point. Many depth estimation methods exist that have been applied, and are still used in endoscopy such as the CNN-based GA-Net~\citep{zhangGANetGuidedAggregation2019}, RAFT-Stereo~\citep{lipsonRAFTStereoMultilevelRecurrent2021}, or the classical LibELAS~\citep{geigerEfficientLargeScaleStereo2011}.
These methods come from computer vision and are relevant to both the surgical and non-surgical applications.
Additionally, we note that although the methods we review in this section calculate depth densely, some SLAM or SfM methods instead efficiently back-project the points from feature matching to estimate their 3D position sparsely, although this does not provide the smoothness regularization that dense methods do.

\metrics{}
For evaluating depth estimation networks, RMSE and log RMSE are frequently used when depth ground truth is available.
Mean absolute distance (MAD) is also used.
To accomodate for possible scale changes and estimate relative error in the case of monocular stereo, Abs Rel and Sq Rel provide percentage metrics.
$\delta$ reports accuracy at varying thresholds.
When ground truth is unavailable, the quality of photometric reconstruction using reprojection from one frame to another is evaluated with SSIM or PSNR.

\subsubsection{Stereo Depth Mapping}
Stereo disparity estimation algorithms work as follows.  
For each pixel in one image of the stereo pair (\(I_{left}\)), a search is performed along the epipolar (usually horizontal) line to find the most similar pixel in the other image, (\(I_{right}\)).
A patch-based similarity metric is often used.  This can be built into a neural network or performed classically using an optimization framework.
Neural networks are often trained using an image reconstruction loss, 
which measures how well the network warps the left image into the right image using image-level photometric errors such as L1 distance and structural similarity index measure (SSIM).
This is an indirect, or unsupervised, approach since when training on datasets in MCV we seldom have ground truth depth.
Since it is indirect method, visual effects in MCV such as specularity will cause artifacts in the algorithmic reconstruction.
Simulated ground truth, or ground truth generated using scans is also feasible for training models without requiring photometric supervision.

To begin, we will summarize early work in depth mapping and help frame how the field has changed.
Motion compensation and instrument stabilization are frequently mentioned goals in endoscopy, and depth mapping is necessary for this.
~\citet{stoyanovDense3DDepth2004} propose a depth estimation method that could be used to estimate motion by calculating depth over time.
Depth is solved for by using multi-resolution Normalized Cross Correlation (NCC) between rectified images, and BFGS (Broyden-Fletcher-Goldfarb-Shanno) as the iterative optimization algorithm. Later,~\citet{loBeliefPropagationDepth2008} propose a hybrid that combines stereo depth and Shape from Shading (SfS).
SfS uses lighting cues to estimate the normal of a point in space.
For example, a viewing ray that intersects with the tissue surface normal to the light source will appear brighter.
The authors use a Markov Random Field (MRF) to fuse these methods, where the SfS measurement and the stereo measurements influence the true depth in a Bayesian form.
Nearby points on the depth map grid are also connected in this model to provide a smoothness constraint.

With a focus on increased efficiency, sparse means for depth prediction have also been proposed.
~\citet{stoyanovRealtimeStereoReconstruction2010} use a sparse set of feature matches to propagate measurements around said matches.
The method can take any set of feature descriptors and matches (\((f_i, p_i), (f'_i, p'_i)\)) as input, and then propagates depth around each match according to color and distance difference.
The very popular LibELAS (Efficient Large-scale Stereo~\citep{geigerEfficientLargeScaleStereo2011}) work uses a similar idea with more details on refinement in neighborhoods around each sparse match.
They propose a model for the probability distribution of a depth point given (conditioned on) sparse matches of point features (support points), and image features.
With this Bayesian model, they can propose a procedure for estimating depth.
The model takes in matches which use Sobel features as their descriptors.
These act as support points.
To densify the matches onto an image grid, they refine the estimated points in regions surrounding the support points by fitting them to the maximum probability in their model, which combines the distance from support points with a regularizing distance to keep pixel estimates close to the neighboring support points.
LibELAS~\citep{geigerEfficientLargeScaleStereo2011} is often used for pseudo ground truth in surgical tracking and mapping~\citep{recasensEndoDepthandMotionReconstructionTracking2021,gomezrodriguezTrackingMonocularCamera2022,gomez-rodriguezSDDefSLAMSemiDirectMonocular2021}.

Some classical computer vision methods have been adjusted for surgical video.
\citet{changRealtimeDenseStereo2013} use ZNCC (Zero-mean Normalized Cross Correlation) to help accommodate for brightness differences when comparing patches along an edge. They evaluate their method using 3D data from CT scans.

With machine learning beginning to make an impact,~\citet{luoDetailsPreservedUnsupervised2019} train an encoder/decoder model for each image in a stereo pair, fusing the results from each view afterwards.
They train using proxy labels from classical stereo algorithms along with an image reconstruction loss for enforcing left-right consistency.
By warping the left image according to the left disparity map, it should look visually similar to the right image.
They measure performance using CT ground truth on heart phantoms from the Hamlyn dataset~\citep{stoyanovRealtimeStereoReconstruction2010,prattDynamicGuidanceRobotic2010} and compare performance against pseudo-ground truth from classical algorithms.
As mentioned in Section~\ref{sec:datasets}, a drawback of using the pseudo-ground truth means it is not possible to see if the method outperforms classical methods. 

More recently, StaSiSNet~\citep{bardozzoStaSiSNetStackedSiamese2022} use a Siamese network for real-time depth estimation.
On another note, since accurate calibration is essential for quality depth estimation,
\citet{luoUnsupervisedLearningDepth2022} propose a machine learning method that can deal with imperfect rectification that can occur due to an incorrect stereo camera model.
They first use a network to perform vertical correction estimation to better align the image pair so the epipolar lines match.
They follow the vertical correction with a disparity estimation CNN which uses a Generative Adversarial Network (GAN) to differentiate between warped stereo frames from left to right (and right to left) and the true frame on the right (left).
They add a mask based on the residual between the reconstructed image and the true image to reduce the influence from outlier points such as specularities.
Even more specific to surgical tissues, and, specifically, their contiguity,~\citet{zhao3DEndoscopicDepth2022} estimate depth by incorporating a constraint that takes into account the surface smoothness in camera space (3D) instead of just using image-space based photometric loss.
Like many other methods, they run a specularity removal step. For quantifying their method, they use the EndoDepthAndMotion~\citep{recasensEndoDepthandMotionReconstructionTracking2021} dataset for ground truth, which in turn uses LibELAS.
Also dealing with uncalibrated stereo pairs,~\cite{yangDenseDepthEstimation2021} use an optical flow network with photometric losses and sparse feature matching loss to learn an optical flow network that is used for depth estimation on uncalibrated images. Finally,~\citet{weiStereoDenseScene2023} use a pre-trained HSM-Net and then fine tune it on the SERV-CT dataset. Their goal is to perform localization and 3D reconstruction of dense scenes.

Coming back to earlier work which used hybrids of methods,~\citet{caoAlgorithmStereoVision2022} combine SfS with a classical stereo algorithm for stereo MIS, again demonstrating the benefits of joint methods. Pushing classical methods further forward,~\citet{songBDISBayesianDense2023} use conditional random fields and a coarse-to fine methodology, similar to LibELAS but with faster performance.
Their method does not require a GPU.
On an i5-9400 CPU, inference takes 72ms for \((1280,720)\) sized images.
For comparison, with the same setup along with an NVIDIA 1080 Ti, LibELAS takes 291ms~\cite{geigerEfficientLargeScaleStereo2011}, and PSMNet takes 566ms~\cite{changPyramidStereoMatching2018}).

Bringing in more modern machine learning, contrastive learning also improves endoscopic stereo when used in combination with photometric loss, outperforming other self-supervised models~\citep{tukraStereoDepthEstimation2022}.
Machine learning in MCV has also had a recent growth in the use of transformers, multitask models and foundation models as well.
\cite{longEDSSREfficientDynamic2021} use a stereo transformer to estimate depth, and then reconstruct a 3D scene with a surfel based model.
\cite{psychogyiosMSDESISMultitaskStereo2022} design a model that uses shared features for estimating both depth and instrument segmentation to result in improved performance.
In foundation models, specifically DINOv2~\citep{oquabDINOv2LearningRobust2023}, \cite{cuiSurgicalDINOAdapterLearning2024} adapt and fine tune DINOv2 using Low Rank Adaptation (LoRA,~\cite{huLoRALowRankAdaptation2021}) for surgery.
This is trained using a ground truth split from the SCARED dataset.

In brief, many different methods exist for stereo depth in MCV, with most of the baseline networks coming from the broader computer vision field.
The gap that MCV algorithms fill compared to broader CV is primarily how to train and design losses for the visual appearance in MCV scenes along with designing models to incorporate priors from this environment.

\subsubsection{Monocular Depth Mapping}
Monocular depth estimation uses images from a single camera alongside visual cues to estimate depth.
When there is no known reference distance (e.g., a camera baseline or instrument with known diameter), scale estimation is not performed.
Monocular depth estimation is necessary in bronchoscopy, for example, where the cameras are often monocular due to size constraints.
Addressing monocular reconstruction,~\citet{visentini-scarzanellaDeepMonocular3D2017} train a CNN to estimate relative depth (up to scale) by using ground truth renderings.
In essence, the network learns the visual cues from lighting to estimate depth, similar to SfS.
Of course, areas with no texture or lighting will have to be inpainted or estimated by infilling with whatever those regions looked like in training.
To train in a realistic environment,~\citet{liuSelfsupervisedLearningDense2018} train a monocular depth estimation network using SfM point clouds for ground truth.
This is a sparse, albeit accurate, form of supervision that works in rigid environments such as sinus surgery.
They later extend their loss formulation and demonstrate generalization to other environments~\citep{liuDenseDepthEstimation2020a}.
In MCV, there is often additional information given the camera and lighting that are present.
~\citet{batllePhotometricSingleviewDense2022} propose a monocular photometric reconstruction method, which uses known positions of the camera and light to model shape under a Lambertian assumption.
The Lambertian lighting model treats a surface as perfectly matte.
The surface's appearance is not view-dependent, unlike a mirror, for example.
Due to this assumption, pixels that do not follow the assumption have to be masked or adjusted, so they opt to remove specularities with in-painting.
They use an iterative optimization method to solve for depth.   Although their method is offline, it opens the door to model-based methods for MCV.
\citet{hanDepthAnythingMedical2024} investigate modern monocular models such as Depth Anything~\citep{yangDepthAnythingUnleashing2024}, noting its favorable inference, but motivate more fine-tuning and research in the medical field due to the model's similar performance to existing methods.

\subsection{Tissue Tracking}
\label{sec:tissuetracking}

\subsubsection{Introduction}
Tissue tracking entails methods that estimate motion of tissue surfaces or organs in MCV.
These are useful for any applications that require tracking of specific points or regions.
These applications include autonomous scanning, image guidance, automation, and measurement of marked points.
Tissue tracking often uses optical flow (dense) or temporal feature matching (sparse).
This can be paired with feature management to maintain features over time and to find features after they disappear.
We will briefly cover evaluation metrics, and then delve into the field.

\metrics{}
Tissue tracking methods often evaluate their work using the performance of tracking algorithms compared to ground truth.
The metrics used here include endpoint error, and accuracy at a threshold, $\delta$.
IOU or chamfer distance can also be used if tracking is evaluated on segments.
Forward-backward error, or cycle consistency, is also sometimes used for evaluating drift of trackers.
Metrics from the TAP-Vid metric for occlusion accuracy are important to reference for future quantification under drift and for long-term tracking~\citep{doerschTAPVidBenchmarkTracking2022}.

\subsubsection{Tissue Tracking in MCV}

We will begin with a summary of classical methods that are still used in MCV to this day.
Summarizing a classical computer vision-based tracking method,~\citet{lucasIterativeImageRegistration1981} introduce a tracker that, for each tracked patch, uses image similarity to find the best aligned patch and optimize its position until convergence using image intensity metrics (e.g. L1, Sum of Squared Differences (SSD)).
\citet{tomasi1991detection} extend this with salient detections, creating the frequently used Kanade--Lucas--Tomasi (KLT) tracker.

Turning our attention to surgical algorithms for tissue tracking,~\citet{richaEfficient3DTracking2008} perform tracking for motion compensation on beating heart surgery.
They use an underlying thin plate spline (TPS) model to fit motion.
In other work that does not require an underlying model,~\citet{yipTissueTrackingRegistration2012} (\discussed{}) maintain features over time using a STAR detector~\citep{agrawalCenSurECenterSurround2008} and BRIEF (Binary Robust Independent Elementary Features~\citep{hutchisonBRIEFBinaryRobust2010}) descriptor.
To perform their tracking in 3D, they match features between stereo pairs to triangulate points.
Using their method, they also propose a region tracking framework that allows tracking of user-selected regions.
Regions are then tracked with a rigid transformation according to the motion of feature points lying within them.
This is limited in cases with deformation, specularity, or occlusion.

For tracking with novel features designed for surgery,~\citet{giannarouProbabilisticTrackingAffineinvariant2013} track detected elliptical regions in real-time with an extended Kalman filter (EKF) to improve noise tolerance.
We note that tissue tracking methods are also useful for image guidance in other environments, such as in brain surgery for registration of MR images under brain shift.
\citet{jiCorticalSurfaceShift2014} track a cortical surface using LK optical flow and use stereo reconstruction to estimate 3D positions (Fig.~\ref{fig:ji2014}).

\screenshot[ji2014]{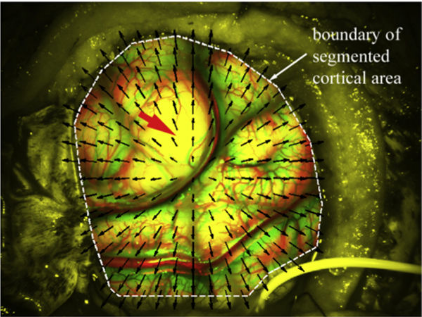}{Cortical surface tracking on an image deformed image with a radial motion field (red: deformed, green: undeformed). From~\citep{jiCorticalSurfaceShift2014}, reproduced with permission from Elsevier}

In less featured regions, sparse correspondences enable alignment between salient features while denser optical flow can prove useful in less textured regions.
Exploiting this idea,~\citet{duRobustSurfaceTracking2015} combine the benefits of sparse correspondences with LK optical flow.
For estimating displacement, they represent the scene using a triangular mesh.
They choose to use Sum of Conditional Variance (SCV) instead of SSD as their similarity metric for optical flow.
This enables better performance under non-linear variations in the images.
\citet{schoobStereoVisionbasedTracking2017} (\discussed{}) use a similar tracking algorithm, but with application in laser ablation for microsurgery.

Using the classic KLT,~\citet{penzaLongTermSafety2018} match regions with additional specularity filtering for tracking Safety Areas (structures to avoid damaging) in surgery.
Their method importantly estimates when tracking fails.
When tracking fails, they find SURF~\citep{baySURFSpeededRobust2006} matches in the image and compare to those in the lost region to re-localize. 
Failure estimation of tracked regions is performed via a hand-engineered probability dependent on features in the area, percentage of features lost, validity of the transformation, and standard deviation of the optical flow distribution.

In a different approach that uses neither LK optical flow nor sparse features,~\citet{collinsRobustRealtimeDense2016} estimate flow by solving for the deformation that a model's texture map must undergo to match the current image via rendering the model.
This requires a pre-acquisition of a model with texture and is close in principle to Shape-from-Template (SfT, Section~\ref{sec:sft}).

For applications which only require tracking a few points, tracking-by-detection can prove useful.
In tracking-by-detection, a location is set as the center of initial patch to track, and then this patch region is detected in following frames to perform tracking.
\citet{yeOnlineTrackingRetargeting2016} (\discussed{}) perform tracking-by-detection using a descriptor like the Haar descriptor~\citep{violaRapidObjectDetection2001}, searching in local windows around the patch for candidate matches.
They train in an unsupervised manner by sampling patches near each tracked patch as positive, and those far away as negatives.

With the advent of neural networks, CNN-based optical flow methods have begun to be used in MCV.
~\citet{ihlerSelfsupervisedDomainAdaptation2020} train a CNN using FlowNetL.
They fine tune their network in an unsupervised manner using synthetic image warps and a zero-flow regularization (the same image tested against itself should result in zero movement).
Since FlowNet is relatively efficient, their fine tuning enables a fast convolutional tissue tracking model.

Other fast methods include~\citet{schmidtFastGraphRefinement2022a}, where neural networks are used in a sparse manner.
A tracking algorithm is proposed that works by conditioning motion on a graph neural network of salient sparse correspondences.
The authors later extend their work with a recurrence model~\citet{schmidtRecurrentImplicitNeural2022}, and then to 3D (\discussed{}~\cite{schmidtSENDDSparseEfficient2023}).

Multiple tracking methods participated in the SurgT Challenge~\citep{cartuchoSurgTChallengeBenchmark2024}.
Here we will summarize the top three to show where these methods are.
The CSRT tracker~\citep{lukezicDiscriminativeCorrelationFilter2017}, which is classical and correlation based got third place.
For second place, Jmees~\citep{Jmees2024} build a correction framework on top of CSRT which adjusts scale, detects instrument occlusion, and uses template matching to verify validity.
For first place, ICVS-2Ai~\citep{2AI2024} built a tracker using a dense optical flow network built off ARFlow~\citep{liuLearningAnalogyReliable2020} and PWC-Net~\citep{sunPWCNetCNNsOptical2018} with smoothness regularization.

Returning to tracking-by-detection, but in a neural manner,~\citet{kamAutonomousSystemVaginal2023} present a neural network for detecting six points around a vaginal cuff for cuff closure using autonomous suturing.

Finally,~\citet{liuSurfaceDeformationTracking2023} (\discussed{}) use an MRF to mask surgical instruments, and they then perform tissue tracking with an underlying piecewise affine deformation model (triangular mesh) for representing motions.

\subsection{Structure-from-Motion (SfM), Nonrigid Structure-from-Motion (NRSfM), Shape-from-Template (SfT)}
\label{sec:sfm}

In this section, we will cover three sets of methods for offline reconstruction.
These are Structure from Motion, Nonrigid Structure from Motion, and Shape from Template.
Structure from motion (SfM) estimates a map in a rigid scene given a set of images (in our case).
Nonrigid structure from motion (NRSfM) does the same, except with an underlying map that can be non-rigid.
Shape from Template (SfT) uses a learned template.
The template's position can then be estimated and fitted given observations.
These methods all differ from depth estimation since they look to create and maintain a usable map over time.
For a detailed and wide survey in computer vision, see~\citet{tretschkStateArtDense2022} for a review on dense monocular non-rigid 3D reconstruction.
Here we will focus on the specific applications in MCV.

\metrics{}
These methods evaluate their performance using metrics for pose accuracy such as RPE, or for reconstruction accuracy such as RMSE on point clouds.
Qualitative visualizations are also frequently used for dynamic methods in this section due to the lack of ground truth data in these environments.

\subsubsection{Structure from Motion (SfM)}

SfM aims to reconstruct a rigid environment, often a set of points in 3D space, \(\{\mathbf{x}_i \in \mathbb{R}^3\}\) and estimate camera poses, \(\{T(t)\}\), given a set of images, \(\{I(t)\}\).
This process is performed offline with the images collected beforehand.
The map in this context could be a mesh or another representation, but in MCV, the map most commonly consists of 3D points alongside point features.
Refer to Section~\ref{sec:slam} (SLAM) for the real-time counterpart which, for the sake of efficiency, differs in optimization and mapping methods.
SfM is designed for rigid environments, and often entails optimizing a map and a set of poses in tandem until convergence.
SfM can be used for dataset generation, or for creating maps that surgeons can use for decision-making and planning.
Most methods in SfM for MCV use the same base algorithm but adjust algorithms and terms to suit the medical environment.
These modifications include methods such as outlier and specularity filtering.

We begin with an early example:~\citet{hu3DReconstructionInternal2007} propose to use SfM for creating a larger field-of-view for surgeons. To make these methods more robust by accounting for specularity and other artifacts,~\citet{huReconstruction3DSurface2012a} extend their work by adding outlier removal and bundle adjustment.
Bundle adjustment is an optimization that performs alignment of the 3D point positions \(\mathbf{x}_i\) and camera pose \(T(t)\) to reduce re-projective error under the camera projection using the pose, \(\Pi_{T(t)}\), of measured points, \(p_{i}(t)\), in each image.
This optimization is performed over time, indexed by \(t\), and map points, indexed by \(i\).
\begin{equation}
	\min_{T(t),\mathbf{x}_i}\sum_i \sum_t ||\Pi_{T(t)}\left(\mathbf{x}_i\right) - p_{i}(t)||^2
\end{equation}

Since endoscopic environments can often be ill featured, in addition to having artifacts,~\citet{widyaWholeStomach3D2019} increase visible features via dying tissue with indigo carmine (IC) dye, and they demonstrate the comparative performance increase by using dye for helping SfM reconstruction.
They remove outliers from the point cloud map that they generate with SfM by using local plane fitting.
Then they create a mesh of the SfM point cloud.
With this mesh, they can perform another outlier removal step for points that do not align well with the mesh.
Both these steps account for the physical surface consistency priors we often have in MCV.
To localize where the camera is in a current map, they use NetVLAD~\citep{arandjelovicNetVLADCNNArchitecture2016}, which is a CNN-based model that provides a distance metric between image pairs.
Then, given similar pairs, they can reconstruct higher detail images of these regions.
This process is shown in Fig.~\ref{fig:widya2019}.
This approach is computationally intensive and runs offline, so it can not be used for interactive clinical applications.
Interestingly, the authors then take the concept of IC-dye improving texture, and carry it on to design a model to perform virtual generation of IC textures using a CycleGAN.
This allows them to generate IC-images artificially from non-IC images, and they demonstrate how their GAN-based method outperforms the non-augmented images for reconstruction applications~\citep{widyaStomach3DReconstruction2020}.

\screenshot[widya2019]{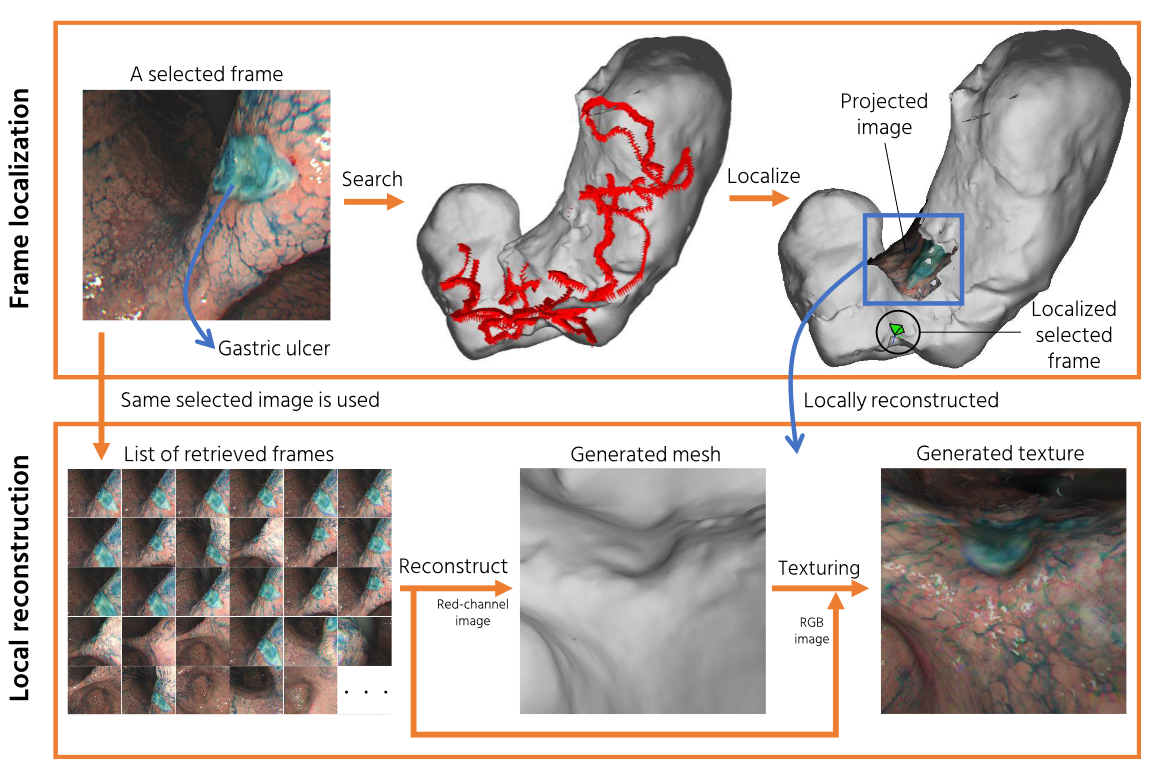}{Top: Camera localization using SfM as a map. Bottom: Local mesh reconstruction using the frames with the closest visual similarity for reconstruction. From~\citet{widyaWholeStomach3D2019} \CCBY{}}

\subsubsection{Nonrigid Structure from Motion}
\label{sec:nrsfm}

Nonrigid Structure from Motion (NRSfM) does not assume that the world has a rigid state.
This means there are many more parameters to be solved for in optimization, adding to both the computational expense and modelling difficulty.
To make it so not every point in the map is a degree of freedom, these methods need to make assumptions about tissue motion.
Since we do not have an underlying model to fit to, by assuming priors on the types of motion that can happen, we provide a way to regularize.
Two ways in which this is performed are via low rank shape models (LRSM)~\citep{torresaniNonrigidStructurefromMotionEstimating2008}, or isometric priors.
Isometric priors depend on assuming that locally connected (nearby) points are isometric (distance-preserving), and enforce this constraint between point neighbors during optimization.
As an example, a sheet of paper is isometric, while an exercise band is not.
Low rank shape models assume shape can be represented as a linear combination of multiple basis shapes.
A 3D shape at time \(t\), \(\mathbf{X}(t) \in \mathbb{R}^{3n}\) can be represented as a mean shape \(\mathbf{\Bar{X}}\) plus \(M\) basis shapes \(\mathbf{v}_m \in \mathbb{R}^{3n}\).
At each point in time, the shape is represented as a linear combination of these shapes with a set of weights, \(z(t) \in \mathbb{R}^M\):
\begin{equation}
	\mathbf{X}(t) = \mathbf{\Bar{X}} + \sum_m z_m(t) v_m
\end{equation}

Using LRSM,~\citet{huNonrigidReconstructionBeating2009a} reconstruct a beating heart model.
To improve their results, they take advantage of the periodic motion of the heart, and use the same times in the heart cycle as additional samples to reduce dimensionality (i.e. 10ms into a heartbeat should look the same every time). 

Although we do not have the same periodic motion in colonoscopy, we do have priors on the colon being a tubular structure.
To utilize this,~\citet{senguptaColonoscopic3DReconstruction2021} add an underlying model to NRSfM and demonstrate improved performance on simulated tubular structures.
They begin with calculating 3D point locations by performing NRSfM using an isometric prior. 
After calculating 3D point locations, they fit a tubular model to these locations.
They model the tubular structure with harmonic splines.
Optimization considers the tradeoff between being close to the 3D locations and smoothness regularization of the model. This is actually an example of a mixture of NRSfM with Shape-from-Template (SfT), which will be detailed in more detail in the following section.
In a similar vein (mixing NRSfM and SfT),~\citet{golyanikIntrinsicDynamicShape2020} learn a dynamic shape prior using NRSfM.
They collect this prior over a fixed representative set of frames, collecting a set of shape states.
That is, they have \(N\) different instances of what the shape can look like.
Then for performing tracking of their dynamic shape prior, they match images to the nearest pre-calculated state.

The choice of prior that NRSfM methods rely on is particularly important in MCV since the priors dictate the transformations that the map can undergo, and the motion that can be accurately represented.
The following section will detail further information on priors that come in the form of templates, rather than the regularization that is used in NRSfM via low rank or isometry constraints.

\subsubsection{Shape-from-Template (SfT)}
\label{sec:sft}

In Shape-from-Template (SfT), we first calculate a template of the scene or design a predetermined canonical one, e.g., a sheet or tube.
Then, in the following frames we align the template to match the current frame.
\citet{maltiTemplatebasedConformalShapefrommotionandshading2012} construct a template using rigid SfM and then combine Deformable Shape from Motion (DSfM) with SfS.
Their template is initialized using video from a rigid scene.
Then, they calculate the albedo of this template by using a Lambertian model as their bidirectional reflectance distribution function (BRDF).
A BRDF model captures how a surface emits incoming light at varying angles.
To initialize a coarse reconstruction of their shape at a certain time they match sparse points between the template and the current image.
They perform the coarse matching step using SIFT correspondences.
Then, with the calculated albedo, they can refine the coarsely aligned shape by matching lighting effects to their lighting model and using SfS.
\citet{maltiCombiningConformalDeformation2014} extend this work and use a more realistic lighting model than a Lambertian one.
They select a Cook-Torrance model with a Beckmann distribution rather than Lambertian shading for the SfS refinement step.
They show that this performs better than modelling using Lambertian or Oren-Nayar distributions.

\citet{cheemaImagealignedDynamicLiver2019} use SfS as well, but with an additional incorporation of a pre-operative CT model of the liver as a prior.
In the colon,~\citet{zhang3DReconstructionDeformable2021} also use a known 3D template generated using a CT scan.
They use SGM (semi-global matching) for disparity estimation from stereo camera data.
They generate a large video dataset from a colonoscopy simulator for evaluation.
They choose SIFT for feature description and represent deformation using an embedded deformation (ED) model~\citep{sumnerEmbeddedDeformationShape2007}.
In embedded deformation, the motion of any point, \(\mathbf{p}_i\), is a function of $m$ neighboring control-point nodes and their positions \(\mathbf{g}_j, j \in {1...m}\).
Each node essentially controls a rotation and translation \((R_j, \textbf{d}_j)\).
A weight function is used to increase influence of nodes that are closer, with weights, \(w_j\), that are normalized to sum to 1.
As can be seen from the equation, this model supports a smooth deformation:
\begin{equation}
\mathbf{p}_i = \sum_{j=1}^m w_j(\mathbf{p}_i) \left[R_j(\mathbf{p}_i-\mathbf{g}_j)+\mathbf{g}_j+\mathbf{d}_j\right]
\end{equation}

As neural networks have become popular, they have also taken a hold of SfT research.
The general idea of using neural networks in SfT is that a template can also be represented by a set of neural network weights or latent codes.
\citet{golyanikHDMNetMonocularNonRigid2018} train a CNN on images using varying known ground truth deformations.
Given an input image, their CNN estimates \((73 \times 73 \times 3)\)-sized sets of 3D points to represent a triangular mesh grid.
This can be seen as a form of template-based reconstruction, where the learning step learns the template, and an inference model estimates the current template position given an image.
For fitting novel views, they simply input a new image.
The difficulty is that this model requires ground truth training data along with full training on this dataset to solve for the parameters in the template-fitting network.
A step in this direction that no longer requires ground truth artificial data and uses optical flow as training signals is proposed by~\citet{sidhuNeuralDenseNonRigid2020a}.
The authors propose Neural NRSfM by learning a latent space function that adds CNN-estimated offsets to a mean shape using an autodecoder, similarly to an LRSM.
They additionally enforce a high-frequency regularization on the Fourier transform of the latent codes over time which -- in addition to regularizing -- allows for latent measurement of periodic signals.
This helps recognize motions such as a heartbeat.
The drawbacks for possible clinical application are the sensitivity to optical flow outliers, the need to initialize a rigid mean shape, and the model training taking multiple hours.

\subsection{Simultaneous Localization and Mapping (SLAM)}
\label{sec:slam}

In visual Simultaneous Localization and Mapping (SLAM, in this review, but denoted as VSLAM when differentiating between non-camera-based methods), the goal is to create a map of the environment, often a set of points, \(\{\mathbf{x}_i\}\), and at the same time localize the sensor position within said environment.
The sensor position is represented as poses over time, \(\{T(t)\}\).
In this section, we will review methods that perform SLAM given video data.
We note that for different environments, the means of mapping can vary.

In implementation, SLAM is often implemented with multi-threaded methods that both optimize a map over a large set of keyframes, along with a faster localization thread that estimates the position of the camera relative to the most current map state.
By having separate threads, this enables real-time operation since the slower (bundle adjustment and re-localization) optimizations will not affect near-term pose estimation. A map can be represented by anything from a point cloud with features to a triangular mesh.
The optimization is often performed using a nonlinear least squares (NLLS) method.
Levenberg-Marquardt is frequently used as the NLLS optimization method, whereby a set of error terms are minimized over what is called a pose graph.
The pose graph acts as overarching graph that connects nodes (poses) with data association terms (losses) (e.g., co-visible camera views are connected with feature matches).
The components that change between methods in MCV are primarily the underlying map representation, the error terms used, and the means for re-localization.
Re-localization is the process of finding out where in the environment the camera or the features are once they have been lost.
We will investigate rigid SLAM in Section~\ref{sec:rigid}, and nonrigid SLAM in Section~\ref{sec:nonrigid}.
In the rigid SLAM section we will additionally mention some work that addresses a subset of the SLAM problem, such as relative pose estimation.
In the nonrigid SLAM section, we also include the problem of nonrigid mapping (SLAM without the camera localization) since these works are closely related.
We will provide an overview, with a specific focus on the evolution of algorithms over time along with the particular reasons for different proposed solutions.

\metrics{}
Metrics used here include photometric reconstruction errors (PSNR, SSIM), particularly so in deformable environments.
Additionally, tracking errors such as epe. are used when labelled points are available.
For algorithms that are estimating pose, ATE and RPE are used.
Finally, when depth maps are available, RMSE is used between the ground truth and the reconstructed depths.

\subsubsection{Rigid SLAM}
\label{sec:rigid}

Rigid SLAM has been applied in endoscopy for decades, with initial applications in fields such as CT-guided sinus surgery~\citep{burschkaScaleinvariantRegistrationMonocular2004}.
In the rigid SLAM problem, we are looking to estimate a set of camera poses along with a map of the environment.
Thankfully, if we would like to track individual points in the environment, we can calculate their motions easily as the entire motion is explained by the rigid 6DoF transformation the camera undergoes.
A sparse map (set of points in space) can provide us with localization, rigid 6DoF motion estimation, and sparse measurements.
Sometimes we would also like a dense map rather than a sparse one; the primary reasons for this are to enable visualisation or dense surface reconstruction for applications such as tissue scanning.
Here we will summarize the methods in rigid SLAM for MCV, followed by some sections on map densification, localization, and dealing with texture.
Table~\ref{tab:rigidslam} provides an overarching summary of rigid SLAM methods.

\begin{table*}[tbp]
	\caption{Rigid SLAM Methods used in medical computer vision. Abbreviations: Provides Loop Closure (LC), Provides a dense reconstruction (Dens.), Uses NN (whether the model uses a neural network).}
\centering
\begin{tabular}{lllll}\toprule
	Authors & LC & Dens. & Base & Uses NN\\
	\cmidrule{1-5}
\cite{burschkaScaleinvariantRegistrationMonocular2004} & Y & N & N/A & N\\ \cite{mountneySimultaneousStereoscopeLocalization2006} & N & N & EKF-SLAM & N\\
\cite{grasaVisualSlamHandheld2014} & N & N & EKF-SLAM & N\\
\cite{mahmoudORBSLAMbasedEndoscopeTracking2017} & Y & Y & ORB-SLAM (dens. off keyframes) & N\\
\cite{chenSLAMbasedDenseSurface2018} & Y & Y & ORB-SLAM & N\\
\cite{mahmoudLiveTrackingDense2019} & Y & Y & ORB-SLAM & N\\ \cite{maRealTime3DReconstruction2019} & N & Y & DSO & For dens. \& pose-init~\citep{wangRecurrentNeuralNetwork2019}\\ \cite{vasconcelosRCMSLAMVisualLocalisation2019} & Y & N & N/A & N\\
\cite{zhouRealTimeDenseReconstruction2020} & N & Y & N/A & N\\
\cite{wangVisualSLAMbasedBronchoscope2020} & N & N & ORB-SLAM & N\\ \cite{jiaLongTermRobust2021} & Y & N & N/A & N\\
\cite{liuSAGESLAMAppearance2022} & Y & Y & N/A & For depth and im. features\\ \cite{huoRealTimeDenseReconstruction2023} & Y & Y & ORB-SLAM & \cite{khamisStereoNetGuidedHierarchical2018}\\
	\bottomrule
\end{tabular}
\label{tab:rigidslam}
\end{table*}

ORB-SLAM~\citep{mur-artalORBSLAMVersatileAccurate2015} is the most commonly used approach and will be seen throughout these MCV works, with some modifications for each environment.
ORB-SLAM is appealing as it uses sparse ORB features to maintain real-time performance.
It provides relocalization using the same features by using a bag-of-words method called DBoW2, and refines pose with bundle adjustment in both a local and global manner.
Since many video frames look similar, SLAM methods use keyframes (individual frames with separation in time or poses) so bundle adjustment and loop closure (recognizing when we have looped back to the same location, thus reducing drift) steps only have to look at a smaller set of frames.
For estimating camera pose in the real-time thread, the prior pose along with a constant velocity estimation can be used for efficiency.
Points are detected and then matched to the map, and the current pose is refined to best match them.

\citet{mountneySimultaneousStereoscopeLocalization2006} initially approach SLAM for estimating laparascope motion using patch-based Shi-Tomasi features~\citep{shiGoodFeaturesTrack1994} and a Kalman filter (called EKF-SLAM, which is another SLAM methodology that uses Kalman filtering).
\citet{grasaVisualSlamHandheld2014} extend EKF-SLAM in a monocular environment, using randomized list relocalization (RLR) to estimate pose after tracking loss.
For training the RLR, features of warped patches are used to train the classifier at initialization, and patches are also sampled online during operation.  The RLR patches are classified by performing binary comparison tests, and creating the class given the binary result, similarly to the process used for the ORB descriptor.
They use a feature matching method that searches for feature matches between images using image correlation which enables reasonable performance in low texture environments. The difficulty with using EKF-SLAM is that it can be limited in efficiency on cases with large numbers of landmarks due to requiring a full map update.
Thus, recent methods often use ORB-SLAM to track over longer periods.

\textit{Densifying a map:} 
The widely used ORB-SLAM uses a sparse point cloud for both tracking and mapping, which limits applications to surface reconstruction, since the map does not store measurements other than these points and their features.
To address this,~\citet{mahmoudORBSLAMbasedEndoscopeTracking2017} extend ORB-SLAM to create a denser map by using an epipolar NCC search along with Lucas-Kanade optical flow between paired keyframe images. This method allows them to add ORB points that were previously unmatched by using the depth calculated by their algorithm.
Extending this further, they~\citep{mahmoudLiveTrackingDense2019} add in a separate thread to densify over whole sets of keyframes.
They validate their method on \exvivo{} porcine liver samples using a surface extracted from CT scan ground truth.
The CT surface is aligned to their map using ICP.
Their densification method can provide a happy medium between sparse tracking and dense reconstruction.

In arthroscopic navigation, ArthroSLAM~\citep{marmolArthroSLAMMultisensorRobust2018} uses an external camera in addition to an arthroscopic camera and odometry to localize the arthroscope in space.
This method still results in a sparse reconstruction, so in DenseArthroSLAM~\citep{marmolDenseArthroSLAMDenseIntraArticular2019}, they extend ArthroSLAM to densely optimize a multi-view stereo method that estimates normals and points (\cite{schonbergerPixelwiseViewSelection2016}).
They can then create a mesh from this oriented point cloud using a screened Poisson reconstruction~\citep{kazhdanScreenedPoissonSurface2013}.
In endoscopy, another reconstruction method~\citep{chenSLAMbasedDenseSurface2018} uses ORB-SLAM with a Poisson surface reconstruction to create a dense surface.
The mesh surface can then be used for annotation and measurement (Fig.~\ref{fig:chen2018}).

\screenshot[chen2018]{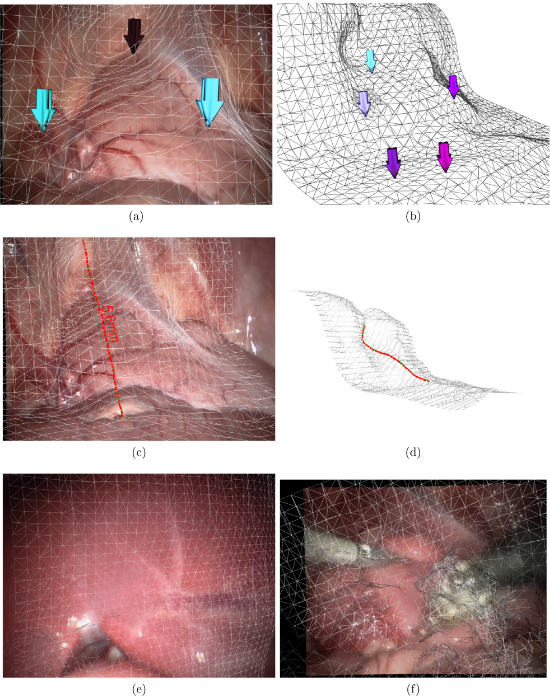}{ORB-SLAM with added mesh reconstruction for allowing annotations (arrows) intraoperatively (a, b from alternative view), or measurements (c, d from mesh view). (e). Mesh displayed on a liver, and (f) under failure in large deformations. Reprinted from~\citet{chenSLAMbasedDenseSurface2018} with permission from Elsevier} 

Using a different SLAM method called direct sparse odometry (DSO~\cite{engelDirectSparseOdometry2018}), which directly uses depth rather than image features,~\citet{maRealTime3DReconstruction2019} aim to densify maps in colonoscopy coverage estimation. They extend DSO by adding in a depth estimation RNN (recurrent neural network) which is trained on SfM reconstructions from the colon. They only can reconstruct partial regions of the colon, likely due to limitations in long term mapping using rigid models, but they do provide dense reconstructions for some regions.
In order to obtain a denser surface representation,~\cite{huoRealTimeDenseReconstruction2023} utilize ORB-SLAM in combination with a StereoNet~\citep{khamisStereoNetGuidedHierarchical2018} method to infill depth in a dense manner.

\textit{Localization:}
Here, we cover methods which look to improve localization through improved pose estimation, bundle adjustment, or outlier rejection.
Some are not necessarily SLAM on their own, since they do not construct a map and instead focus on subproblems like pose estimation.
These methods are designed to be usable as components in SLAM models.
In laparoscopy, we know the camera has specific motion constraints as it needs to pass through a fixed trocar hole.
\citet{vasconcelosRCMSLAMVisualLocalisation2019} integrate these known constraints to constrain possible poses, creating RCM-SLAM.

To estimate pose or depth between frames as a part of a network,~\citet{ozyorukEndoSLAMDatasetUnsupervised2021} introduce a large dataset along with a pose and depth learning neural network, Endo-SFMLearner, which learns relative pose and monocular depth using unsupervised losses. Specifically, they use an affine-adjusted photometric loss along with geometric losses for training.
Fig.~\ref{fig:ozyorukEndo2021b} shows an example from the dataset they use for evaluation.

\screenshot[ozyorukEndo2021b]{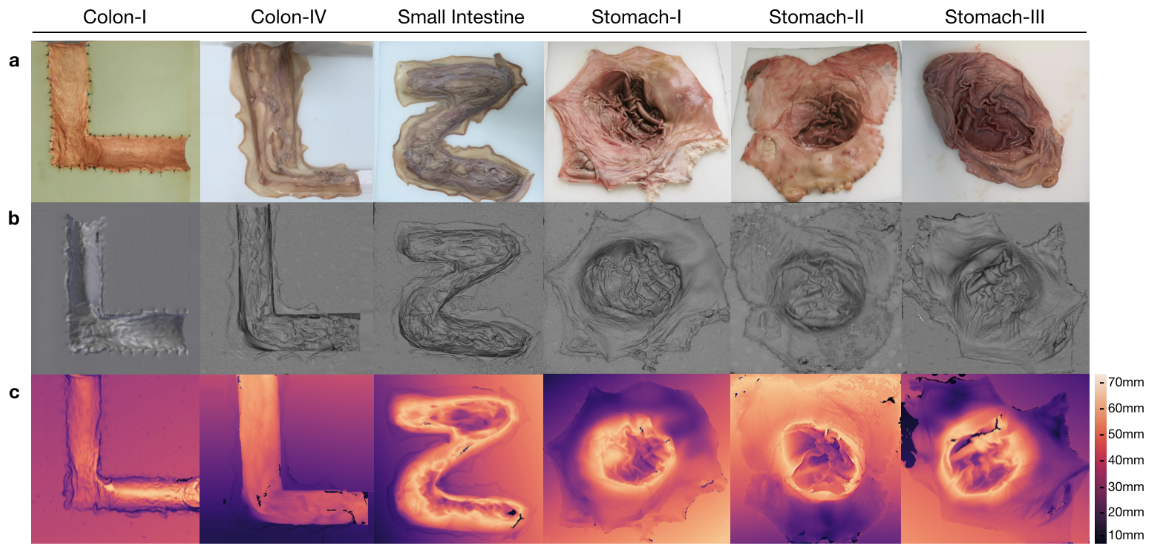}{Ground truth reconstructions from the EndoSLAM dataset, generated using a 3D scanner. From~\citet{ozyorukEndoSLAMDatasetUnsupervised2021} with permission from Elsevier.}

\textit{Poor texture:}
The following articles deal with using poorly textured regions for reconstruction and localization.
To address the low texture present in endoscopic environments,~\citet{zhouRealTimeDenseReconstruction2020} propose a rigid SLAM methodology with modifications for low texture.
They use ZNCC for stereo estimation, and ORB for feature matching.
They discard points using a RANSAC outlier rejection method and use an ICP loss for matching frames.
A Truncated Signed Distance Field (TSDF) is used for visualization of the point cloud as a watertight surface.
Again, addressing poor matches/texture,~\citet{wangVisualSLAMbasedBronchoscope2020} extend ORB-SLAM by proposing a specific feature matching criteria for new frames based on bronchoscopic priors (e.g., by limiting inliers to a smaller filtering window), and evaluate their method on \exvivo{} bronchoscopies.

\citet{jiaLongTermRobust2021}, address stereo endoscopy by matching stereo ORB features using an epipolar search in combination with GMSMatch~\citep{bianGMSGridBasedMotion2020}. They assume a rigid environment, and thus can use keyframes to re-localize, ORB features to match points, and ICP to find pose for new frames.
Their method performs bundle adjustment in a background thread to refine the poses and 3D map point locations.
Specifically, their method operates on a masked (separated from the background) kidney surface under the assumption it is rigid enough to track.

SAGESLAM~\citep{liuSAGESLAMAppearance2022} designs depth, feature, and descriptor networks to improve monocular SLAM methods in weakly textured regions.
They propose feature and depth estimation models that can then be integrated into SLAM.
They estimate a depth map as a combination of depth bases parameters similarly to how LRSMs reduce parameters.
These parameters can be optimized in the SLAM process.
They train their model using a differentiable Levenberg Marquardt relative-pose estimation method with ground truth generated by rigid SfM.
They use a bundle adjustment network (BA-Net,~\cite{tangBANetDenseBundle2019b}) and extremely dense point correspondences for a feature metric and sparse keypoint based loss.
For evaluation, they evaluate relative camera pose estimation performance.
\citet{hayozLearningHowRobustly2023} learn to estimate relative camera pose using back-projected stereo and optical flow using the recent Recurrent All-Pairs Field Transform (RAFT,~\cite{teedRaftRecurrentAllpairs2020}).
Their model addresses poor texture by learning a confidence map to discard pixels that are not useful for pose estimation.
Their optimization uses a Deep Declarative Network (DDN,~\citet{gouldDeepDeclarativeNetworks2021}).
The DDN enables easier embedding of optimization problems into neural networks, with the goal being pose optimization.
Their network architecture is shown in Fig.~\ref{fig:hayoz2023}.

\screenshot[hayoz2023]{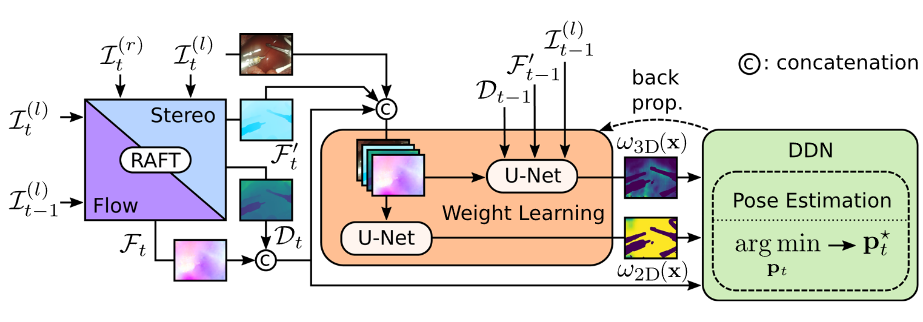}{Pose optimization using a Deep Declarative Network (DDN). The model takes in stereo disparity and optical flow estimations from RAFT. It learns to mask the image, so only informative regions (e.g. non-specular and rigid) can be used to estimate pose. From~\cite{hayozLearningHowRobustly2023} \CCBY}

\subsubsection{Non-Rigid SLAM}
\label{sec:nonrigid}

Non-rigid SLAM and NRSfM have the same difficulty, which is constraining their models to represent motion but not noise.
However, NRSLAM focuses on performing in real-time.
Differences between camera motion and object scaling cannot be resolved in a totally unconstrained environment.
Thus, these approaches must regularize.
As an example, camera pose and tissue movement cannot be decoupled without a fixed reference, so one way that methods can deal with this is to model the camera as accounting for the rigid movement in the scene.
Alternatively, measurements of camera pose can be made using other means such as robot kinematics.
Another difficulty in NRSLAM is that the environment can move in multiple different ways, so there needs to be a model of the underlying motion.
This can be done using a mesh, FEM, etc., or an implicit model that regularizes map points (e.g., point sets should be As-Rigid-As-Possible, similar to the regularization presented in Section~\ref{sec:nrsfm}).
This makes non-rigid SLAM an algorithm design problem that depends on the specific application needs.
Finally, the problem of localization and bundle adjustment are much more difficult in NRSLAM, as points in the images are not fixed and the environment can move even when it is out of view.
We will summarize models that make contributions to NRSLAM, grouping by methods and dependence on one another.
We begin with motion fields, then address mesh models, sparse methods, and finally tracking and mapping without localization.
Specific drawbacks of each method are noted in brief as well.
See Table~\ref{tab:nonrigidslam} for an overarching summary of all the methods introduced that work for deformable environments.

\begin{table*}[tbp]
	\caption{Methods for deformable tracking (SLAM, Surgical Perception, Tissue Tracking) in 3D.
	We omit CNN-based 2D optical flow methods (Section~\ref{sec:tissuetracking}).
	Top: Deformable SLAM, Middle: Tracking and mapping without pose, Bottom: Tissue tracking.
	Abbrevations used: Dis.: supports discontinuities, FEM: Finite Element method, LC: Loop Closure, Map: creates a map of the environment, TT: Tracks tissue, Tri.: Triangle, Per.: periodic environments only, kpts.: tracks only at sparse locations, ED: embedded deformation, R.: only rigidly, Pts.: Points, Uses NN: method uses a neural network. For tissue tracking we include 3D means that could be integrated with mapping frameworks.}
\centering
\begin{tabular}{llllllll}\toprule
	Authors & TT & Map & LC & Dis. & Uses NN & Base & Repr.\\
	\cmidrule{2-8}
	& \multicolumn{7}{c}{NRSLAM}\\
    \cmidrule{2-8}
	\cite{mountneyMotionCompensatedSLAM2010} & Per. & Y & N & N & N & EKFSlam & Pts.\\ \cite{songMISSLAMRealTimeLargeScale2018} & Y & Y (R.) & Y (R.) & N (ED)& N & ORB-SLAM & Pts.\\ \cite{schuleModelbasedSimultaneousLocalization2022} & Y & Y & Y & N & N & ORB-SLAM & Pts.+FEM\\ \cite{lamarcaDefSLAMTrackingMapping2021} & Y & Y & N & N & N & N/A & Pts.+Tri. Mesh\\ \cite{gomez-rodriguezSDDefSLAMSemiDirectMonocular2021} & Y & Y & Y & N & N & N/A & Pts.+Tri. Mesh\\ \cite{lamarcaDirectSparseDeformable2022} & kpts. & Y & N & Y & N & N/A & Pts.\\
	\cite{gomezrodriguezTrackingMonocularCamera2022} & kpts. & Y & N & Y & N & N/A & Pts.\\
	\cite{zhouEMDQSLAMRealTimeHighResolution2021} & Y & Y & N & N & N & N/A & Pts. + EMDQ\\ \cmidrule{2-8}
	& \multicolumn{7}{c}{Surgical Perception (No localization)}\\
    \cmidrule{2-8}
	\cite{schoobStereoVisionbasedTracking2017} & Y & Y & N & N & N & LK & Tri. Mesh\\
	\cite{liSuperSurgicalPerception2020} & Y & Y & N & N & N & \cite{gaoSurfelWarpEfficientNonVolumetric2019}  & Pts./Surfels\\
	\cite{luSuPerDeepSurgical2021} & Y & Y & N & N & Y & \cite{gaoSurfelWarpEfficientNonVolumetric2019} & Pts./Surfels\\
	\cite{longEDSSREfficientDynamic2021} & Y & Y & N & N & Y & \cite{gaoSurfelWarpEfficientNonVolumetric2019} & Pts./Surfels\\
	\cite{linSemanticSuPerSemanticawareSurgical2023} & Y & Y & N & N & Y & \cite{gaoSurfelWarpEfficientNonVolumetric2019} & Pts./Surfels\\
	\cmidrule{2-8}
	& \multicolumn{7}{c}{Tissue Tracking}\\
    \cmidrule{2-8}
	\cite{yipTissueTrackingRegistration2012} & Y & N & N & N & N & N/A & Pts.\\
	\cite{yeOnlineTrackingRetargeting2016} & Y & N & N & Y & N & N/A & Pts.\\
	\cite{liuSurfaceDeformationTracking2023} & Y & N & N & N & N & MRF & Mesh\\
	\cite{schmidtSENDDSparseEfficient2023} & Y & N & N & Y & Y & \cite{schmidtFastGraphRefinement2022a} & Pts.\\
	\bottomrule
\end{tabular}
\label{tab:nonrigidslam}
\end{table*}

A preliminary solution that frames NRSLAM in MCV (\discussed{},~\citet{mountneyMotionCompensatedSLAM2010}) uses a periodic motion model to account for deformation along with a learned tracker and EKF SLAM.
This prescribes specific motion to the environment, and requires exact periodicity, so later methods work to better accommodate changes and track all deformable tissue.

Far later on,~\citet{schuleModelbasedSimultaneousLocalization2022} (\discussed{}) introduce a model-based method integrated with ORB-SLAM.
They constrain map points by projecting them onto a FEM (finite element method) mesh using the Simulation Open Framework Architecture~\citep{faureSOFAMultiModelFramework2012}.
They create a map in 3D under the assumption of fully known forces and physical models.
This 3D map can then be passed into the SLAM algorithm.
The limitation of this approach is that the forces exerted on tissue and finite-element models are often unknown in surgery.
In order to escape these limitations, there needs to be a more flexible underlying representation.

\textbf{\textit{Motion Fields:}}

Instead of ascribing a physical model, we can instead represent the motion as a function of underlying control points with regularization.
ED~\citep{sumnerEmbeddedDeformationShape2007}, and Expectation Maximization Dual Quaternion (EMDQ,~\cite{zhouEMDQRemovalImage2022}) are two examples of this.

MIS-SLAM (\discussed{},~\citet{songMISSLAMRealTimeLargeScale2018}) performs rigid ORB-SLAM along with a separate deformable tracking thread.
The deformable tracking uses ED~\citep{sumnerEmbeddedDeformationShape2007} as the underlying motion model.
These methods can deform a separate model of back-projected stereo points but are unable to track or re-localize under large deformation since ORB-SLAM map points are not warped, only the live model is.
Thus, they are of limited use in applications that involve loop closure and longer term tracking since the underlying map does not deform and would not be able to be localized under visual changes.

EMDQ-SLAM (\discussed{},~\citet{zhouEMDQSLAMRealTimeHighResolution2021}) uses SURF features with an underlying motion field represented with dual quaternions.
A truncated signed distance field is used for surface visualization, where surface color is calculated by blending multiple images.
No re-localization or bundle adjustment is provided, and tracking is frame-to-frame.
They evaluate their method qualitatively.

\textbf{\textit{Mesh models:}}

We will now cover models that use a surface mesh as the underlying map representation.
Such models create a dense surface but can have difficulties in modelling discontinuous motion.

DefSLAM (\discussed{},~\citet{lamarcaDefSLAMTrackingMapping2021}) use a triangular mesh template for surface representation.
To fit this template, they minimize the 2D image re-projection error over detected keypoints.
Bending and stretching energy is used on said points as a regularization.
They perform data association using ORB matches within a local search region.
For near-term warping estimation, they use SfT.
In a slower mapping thread, they re-estimate the template as needed.
A NRSfM optimization is performed on co-visible frames to calculate template updates.
This method uses ORB features for data association along with a mesh representation that can limit the possible deformations and does not provide loop closure.
The authors use stereo methods to evaluate performance. 
These do not provide direct information on tracking accuracy in deformable environments.

SD-DefSLAM (\discussed{},~\citet{gomez-rodriguezSDDefSLAMSemiDirectMonocular2021}) adds LK optical flow to DefSLAM.
They add an ORB bag-of-words model for enabling relocalization under mild deformation conditions.
To estimate initial pose, they use the Perspective-n-Point algorithm.
Limitations again lie in the use of a mesh for the map, and quantification using RMSE on stereo maps.
Additionally, LK optical flow can fail in cases with large displacement.

\textbf{\textit{Sparse Models:}}

Recently, to escape the limitations of discontinuity, methods have been proposed that track sparse map points only.

Direct and sparse deformable tracking (\discussed{},~\citet{lamarcaDirectSparseDeformable2022}) proposes a mostly monocular (requires stereo for initialization) method to track surfel points independently in a deformable manner in space.
It detects surfels with Shi-Tomasi features~\citep{shiGoodFeaturesTrack1994} and tracks their re-projections in 3D space.
To deal with ambiguity, they regularize surfels to lie near an equilibrium position.
Covariance parameters can then be tuned to adjust regularization.
The benefit of sparse tracking is that it enables processing scenes with discontinuity.
Some limitations are that there are no constraints between surfels, the requirement of stereo for initialization, and the cost of tracking \(23\times 23\) sized patches.
The regularizers used can cause issues in representing motion according to~\citet{gomezrodriguezTrackingMonocularCamera2022}.

\cite{gomezrodriguezTrackingMonocularCamera2022} (\discussed{}) present a method to track a sparse set of detected points using a monocular camera.
For data association, they use LK optical flow and photometric error on patches. 
For regularization, they use a deformation graph connecting nearby points with a radial basis weight (see embedded deformation,~\citet{sumnerEmbeddedDeformationShape2007}).
Additional temporal regularization limits the size of motion over time.
This allows for discontinuities, which is particularly demonstrated in a video from the Hamlyn dataset where a liver lobe moves separately from the background.
The authors quantify performance using LibELAS~\citep{geigerEfficientLargeScaleStereo2011} as stereo ground truth. 
The method does not localize new map points, nor does it recover points, making it difficult to apply for long term tracking.

\textbf{\textit{Tracking and Mapping:}} 

Here we will summarize works that perform tracking and mapping, but either assume a static camera, or do not estimate camera position.

The work SuPer (\discussed{},~\citet{liSuperSurgicalPerception2020}) addresses surgical perception, which entails tracking instruments and tissue along with creating a map.
They track tissue with a surfel ED model~\citep{gaoSurfelWarpEfficientNonVolumetric2019}, using point-to-plane error and SURF features for data association. Painted markers are applied to the surgical instruments to aid in instrument detection and masking.

SuPerDeep (\discussed{},~\citet{luSuPerDeepSurgical2021}) extends the flexibility of SuPer by using CNNs to detect instrument keypoints and calculate disparity maps.
The instrument is then masked out by rendering a 3D CAD model given the instrument keypoints.
The primary limitation of this work is computational cost and efficiency from having to run two neural networks along with a deformation optimization.

Efficient Dynamic Surgical Scene Reconstruction (E-DSSR,\discussed{},~\citet{longEDSSREfficientDynamic2021}) uses a similar model, with a learned tool segmentation.
Instead of using an instrument rendering for masking, they mask using their learned tool masks.
They evaluate using photometric reconstruction errors of PSNR and SSIM.

Semantic SuPer (\discussed{},~\citet{linSemanticSuPerSemanticawareSurgical2023}) approaches the same problem, but they remove the keypoint matching loss.
They add in constraints for semantic segmentation regions to match, in addition to considering image features by using differentiable rendering in their optimization.
With a semantic separation of the embedded deformation warp field, this work could be extended to support discontinuity with some changes.

\begin{table*}[tb]
	\caption{\discussed{}, Selected Features of Nonrigid Methods with 3D tracking.}
\centering
	\begin{tabular}{lp{0.72\linewidth}}\toprule
		Authors & \multicolumn{1}{c}{Details}\\
    \cmidrule{1-2}
		& \multicolumn{1}{c}{NRSLAM}\\
	\cmidrule{2-2}
	\cite{mountneyMotionCompensatedSLAM2010} & 
\discuss{\pro Uses an online learned tracker. \con Only represents periodic motion.}\\
	\cite{songMISSLAMRealTimeLargeScale2018} &
		\discuss{\pro Tracks deformation efficiently using ED. \con Map thread does not deform map.}\\
	\cite{schuleModelbasedSimultaneousLocalization2022} & 
\discuss{\pro Integrates FEM with ORB-SLAM. \con Requires known forces and physical model.}\\
	\cite{lamarcaDefSLAMTrackingMapping2021} &
\discuss{ \pro Densely represents points. \con Limited to triangular mesh, and cannot represent discontinuity.}\\
	\cite{gomez-rodriguezSDDefSLAMSemiDirectMonocular2021} &
\discuss{\pro Extends \citet{lamarcaDefSLAMTrackingMapping2021} with LK and BoW relocalization. \con Constrained to triangular mesh. LK fails with large displacement}\\
	\cite{lamarcaDirectSparseDeformable2022} & 
\discuss{\pro Supports discontinuous motion. \con Tracks only sparse detected points. Motion is limited due to regularization.}\\
	\cite{gomezrodriguezTrackingMonocularCamera2022} &
\discuss{\pro Regularizes motion coherence with ED. \con Does not relocalize or recover map.}\\
	\cite{zhouEMDQSLAMRealTimeHighResolution2021} &
\discuss{ \pro Represents motion anywhere using EMDQ model. \con No relocalization or bundle adjustment. Frame-to-frame tracking.}\\
	\cmidrule{1-2}
		& \multicolumn{1}{c}{Surgical Perception (No localization of camera)}\\
	\cmidrule{2-2}
	\cite{schoobStereoVisionbasedTracking2017} & \discuss{\pro Efficient, refines match locations using texture information. \con Uses underlying triangular mesh representation.}\\
	\cite{liSuperSurgicalPerception2020} &
\discuss{ \pro Estimates motion anywhere in space. \con Limited to represent motion with ED.}\\
	\cite{luSuPerDeepSurgical2021} &
\discuss{ \pro Detects instrument keypoints and scene depth using CNNs. Estimates motion anywhere in space. \con Slow for evaluation.  Limited by ED.}\\
	\cite{longEDSSREfficientDynamic2021} &
\discuss{ \pro Efficiently calculates instrument mask directly with a CNN. \con Limited by ED.}\\
	\cite{linSemanticSuPerSemanticawareSurgical2023} &
\discuss{ \pro Uses segmentation to align semantic structures. \con Does not support discontinuity off-the-shelf. Requires image rendering loss for inference.}\\
	\cmidrule{1-2}
		& \multicolumn{1}{c}{Tissue Tracking (No map)}\\
	\cmidrule{2-2}
	\cite{yipTissueTrackingRegistration2012} & 
\discuss{ \pro Manages features over time. \con Limited to tracking affine regions and deformations. }\\
	\cite{yeOnlineTrackingRetargeting2016} & 
\discuss{\pro Efficient tracking-by-detection. Trained in an unsupervised manner. \con Frame-to-frame only. No feature management.}\\
	\cite{liuSurfaceDeformationTracking2023} & 
\discuss{ \pro Includes instrument masking. \con Limited to a tri. mesh representation.}\\
	\cite{schmidtSENDDSparseEfficient2023} & 
\discuss{ \pro Supports discontinuous motion in 3D with a implicit neural GNN. Real-time efficiency. \con No occlusion management. Frame-to-frame.}\\
	\bottomrule
\end{tabular}
\label{tab:proscons}
\end{table*}

\section{Discussion}
\label{sec:discussion}

We will begin with a summary discussion of datasets (Section~\ref{sec:discussdata}), followed by a discussion of methods (Section~\ref{sec:discussallmethods}).
These sections provide guidance for specific needs and limitations in datasets and methods.
We then discuss some additional challenges and limitations that are shared between methods in Section~\ref{sec:discusskeypoints}.
After which, we pose potential future research directions motivated by a snapshot of modern computer vision in Section~\ref{sec:discussuseful}.
We conclude with some questions for the field as a whole in Section~\ref{sec:conclusion}.

\subsection{Datasets}
\label{sec:discussdata}
In summary, datasets can appear in all manners, from entirely unlabelled, to meticulously hand-labeled or modelled.
All types are useful depending on the application and stage of algorithm development and evaluation.
Dataset generation in MCV is difficult since intra-operative environments are difficult to measure and simulate.
If the ground truth comprises tissue phantoms or algorithmically generated data, then it must also be validated on real tissue.
Furthermore, algorithms that learn on this data must be able to generalize to clinical images of real tissue.

Thus, even though difficult to achieve, having a dataset with real tissue is important.
While there are also differences between animal models and humans, using animal models for validation is a step in the right direction, because pre-clinical studies and training are performed using similar models.
A similar argument goes for removing visible markers.
In short, the question comes down to, how can we reduce both the bias and difficulty that occurs in generating ground truth while still enabling generalizability.
Some methods use non-medical datasets to evaluate their algorithms~\citep{duPatchbasedAdaptiveWeighting2019}, which can be useful for evaluation, but questionable for generalization. Concepts such as crowd sourcing, or means to ease the labelling process with markers or IR tattoos are relevant here.

Data generated using phantoms or algorithms are helpful, but this data does not remove the necessity of having labelled data for evaluating performance clinically.
Specifically, synthetic data can be helpful for training algorithms more efficiently by providing guiding loss terms to enable quicker convergence.
For evaluation and testing, synthetic data can provide both verification of algorithm performance along with enabling easier examination of edge cases for algorithm failure such as discontinuities or tracking losses.

The realism of the data used is important.
As an example, SfM can only reconstruct partial regions of the colon in methods proposed thus far, so the capability of ground truth is limited by the size of reconstruction and the quality of the classical-methods for generating ground truth.
Focusing on increasing the realism of these environments will improve generalizability.

Additionally, the data that is currently available for tracking and mapping is biased towards general surgery and colonoscopy, with fewer available datasets in mosaicking and fetoscopy.
No relevant datasets are found in neurosurgery, orthopedics, and plastic surgery.
Thus, in order to facilitate the application of new methods to MCV, we need to move to release and collect data for these other specialties.

As is often the case, we emphasize the importance of having both better data and more data in this field.
Methods often have to resort to quantifying on small datasets that do not directly address their problem (e.g., using photometric reconstruction or depth for quantifying deformable tracking), or using data that is outside their domain.
Of course, together with this is the problem of quantifying methods that can train either in an unsupervised manner or on synthetic data and then perform on in-domain data.
More work needs to be performed with relevant clinical experiments to know what data we need for training and evaluating algorithms in MCV.
In the future, sensors such as LiDAR could help to provide more ground truth, and have been demonstrated recently in medical environments~\citep{caccianigaDense3DReconstruction2024}.

\subsection{Discussions of Algorithms}

\label{sec:discussallmethods}

In this summary discussion, we discuss the algorithms covered in this review.
Alongside this discussion, we note some possible extensions for future work in each section.
We will discuss, in order, the following: sparse and dense features, and matching (Section~\ref{sec:discussfeatures});
mosaicking (Section~\ref{sec:discussmosaicking}); depth estimation (Section~\ref{sec:discussdepth}); tissue tracking (Section~\ref{sec:discusstracking}); offline reconstruction (SfM, Section~\ref{sec:discusssfm}); and SLAM (Section~\ref{sec:discussslam}).

\subsubsection{Feature Detection and Description}
\label{sec:discussfeatures}
Although many still use classical descriptors in the surgical environment, the newer methods which incorporate machine learning improve performance.
They do this by using medical data and train in an unsupervised manner, or on algorithmic reconstructions.
A question here is if there are better ways to reconstruct training data for this environment.
Since the algorithmic reconstructions depend on classical algorithms, we could be limited in the capability of the descriptors we are learning.
Training on deformable models or neural reconstructions generated offline could be a promising avenue here.

Even with good descriptors, matching can still be difficult, and better constraints could be generated instead of using classical methods such as GMS.
One way to improve this would be to design a means to train a GNN like SuperGlue~\citep{sarlinSuperGlueLearningFeature2020} on surgical scenes.
Alternatively we can avoid discarding matches and losing information, in the manner described in~\cite{schmidtFastGraphRefinement2022a}.
This still leaves us possibly limited by feature detection.
With good matching, there is no way to ensure that the matches are pixel accurate (e.g., the same detection may be a couple pixels off due to lighting conditions).
Thus a follow-up question to be answered is how to deal with slight inconsistencies.
At this point, these small errors are likely not realizable as there are more assumptions to be addressed in downstream processes (e.g., spline/mesh modelling) that can add to the error.
Once these improve, an investigation into detection quality is likely to further increase performance.
We note that in the case of rigid methods, this issue becomes less relevant under the assumption that detection noise is normal, as the least-squares fitting should average this error out.
Dense description~\citep{liuExtremelyDensePoint2020} provides a possible solution, as the matches can be searched over all possible pixel (or subpixel) locations.
How to fuse the benefits of sparse matching and dense descriptors in an efficient manner forms another research direction.

\subsubsection{Mosaicking}
\label{sec:discussmosaicking}
Methods from mosaicking share many difficulties also encountered in tracking and mapping.
These include means for matching and dealing with artifacts, blending images, and optimizing underlying maps (or pose graphs) for better global state estimation.
In mosaicking in the medical environment, many of the articles reviewed aim to deal with poor texture, artifacts, or specularity.
To address these, combinations of dense methods with sparse keypoint-based methods look promising, allowing for hybrid benefits from both.
Some questions that remain are how to better merge images that have different light distances or angles, along with how best to learn from machine learning methods for better matching of points (e.g., SuperGlue).
Additionally, mosaicking methods do not train custom descriptors, which could be due to the need for more data.
Note that as far as tracking or 3D mapping are concerned, mosaicking does not aim to estimate the underlying 3D state or motion as the primary goal.

\subsubsection{Depth estimation}
\label{sec:discussdepth}
Since depth estimation is very important for enabling accurate reconstruction, evaluating how these methods perform is critical.
Datasets such as SCARED~\citep{allanStereoCorrespondenceReconstruction2021} provide means to better quantify depth methods, although classical algorithms such as LibELAS~\citep{geigerEfficientLargeScaleStereo2011} are still frequently used as ground truth.
This is often due to other environments or domains not having the same ground truth data available, so evaluating on a different dataset is required, e.g., different cameras, or surgical field.
Thus, we believe it is important to generate more data in this space, or if using generated data, to provide a strong proof of artificial data generation being able to generalize, which would still require \textit{some} in-domain data.

Most of the methodological improvements to depth estimation in MCV are through in-painting, removing artifacts, or noting lighting priors.
Recent work in computer vision has shown high performance with data augmentation~\citep{yangDepthAnythingUnleashing2024}.
Physical priors such as smoothness have a similar impact.
Methods that use these specifics of the MCV environment to improve algorithm performance via masking, lighting adjustment, etc., are seen throughout this discussion (Section~\ref{sec:discussallmethods}).
Future work can take into account temporal consistency~\citep{liTemporallyConsistentOnline2023a}, or combine temporal consistency with a map-based consistency using a point cloud~\citep{khanTemporallyConsistentOnline2023}.

\subsubsection{Tissue Tracking}
\label{sec:discusstracking}
For tissue tracking, as in mosaicking, some methods choose to use hybrids of both dense optical flow and sparse feature matching.
For the same reason as provided in mosaicking, this enables matches in salient locations (sparse) along with performance in ill-textured regions (dense optical flow).
When using frame-to-frame motion estimates, some articles propose means for filtering and managing features using classical methods.
An avenue for improvement would be to bring this feature management into a neural paradigm, but feature management can be a difficult problem to formulate using a differentiable cost function.
Indeed, for discrete operations such as deciding whether or not to keep a keypoint, it can be difficult to generate good proxy gradients.
The straight-through estimator~\citep{bengioEstimatingPropagatingGradients2013} is one such heuristic that deals with the difficulty of enabling a gradient in discrete operations, such as binarization, by passing an identity gradient directly through.

In organized tissue tracking challenges, as we mentioned in Section~\ref{sec:tissuetracking}, methods are correlation-based, and the classical methods were close in performance to the winning deep learning based method.
In the future, they recommend that algorithms utilize the stereo data present for better performance.
Even though classical methods were accurate, the CSRT tracker~\citep{lukezicDiscriminativeCorrelationFilter2017} can be slow for tracking multiple points, so this is important to consider in challenges and applications as well.

Again in tissue tracking, we see that ways to deal with small amounts of data available for training is important.
For example,~\citet{ihlerSelfsupervisedDomainAdaptation2020} use a pre-trained non-MCV model followed by fine-tuning on medical images to improve their results.
In many classical methods of online training, using nearby patches as positive correspondences and far ones as negatives, provides another reasonable physical prior for designing loss functions.
Adaptation of such an approach to machine learning holds promise. 
More physical priors, such as enforcing stability or diffeomorphism in the loss function or model, could help improve performance.
This is especially important when we have small quantities of training data.

Performance is very important in tissue tracking, as downstream applications will have to use additional methods and computation, and large CNNs can be prohibitive in cost.
Methods that are informed by sparse feature-based matching show promise here.
Seeing machine learning models influence this field, and slowly accommodate classical ideas, is an exciting development.

Broader literature computer vision trackers that can manage occlusion~\citep{doerschTAPIRTrackingAny2023,harleyParticleVideoRevisited2022, neoralMFTLongTermTracking2024, rajicSegmentAnythingMeets2023, wangTrackingEverythingEverywhere2023}
provide additional avenues for MCV.
Both real-world hand-labeled~\citep{doerschTAPVidBenchmarkTracking2022} and synthetic~\citep{zhengPointOdysseyLargeScaleSynthetic2023,butlerNaturalisticOpenSource2012} datasets are of note for quantifying future tracking algorithm design in other domains.

\subsubsection{Structure from Motion, Nonrigid Structure from Motion, and Shape from Template}
\label{sec:discusssfm}
In addition to being used for reconstruction of anatomy and surgical planning, Structure from Motion (SfM) has also been used to create ground truth datasets~\citep{liuSelfsupervisedLearningDense2018}.
The primary contributions in MCV for SfM entail compensating for lighting or better integrating models, priors and applications~\citep{widyaWholeStomach3D2019}.
For rigid SfM, methods that perform regularization or outlier filtering are often used to manage artifacts.
Neural networks are used for estimating image similarity in re-localization, but do not yet extend into replacing ways to represent a map.

In Nonrigid SfM, simplifying assumptions are frequently used to ease the problem.
These include only modelling periodic motion or isometry, or using linear bases for deformation.
An avenue for research here is the development of methods that no longer require these assumptions, but would still benefit from parameter reduction without constraining the modelling capability.

These limiting assumptions also occur in SfT.
For example, to initialize a template model, many methods need a rigid sequence for initialization.
Antother possible limitation is that lighting models are simplified to approximate the template using few parameters (e.g., Lambertian or Cook-Torrance).
Neural networks can also be used for template reconstruction without the same hand-engineered limitations, but still have to manage other implicit regularization provided by loss, gradient descent, and weight decay.
These neural SfT models still depend on calculating an underlying representation and require this representation to be predefined or obtained as a mean shape by fitting.

As mentioned in Section~\ref{sec:discussfeatures}, the use of offline methods for the sake of training online methods is an interesting research avenue that could likely be extended to NRSfM methods (rather than just SfM) if reconstruction performance issues are solved.
In conclusion, NRSfM methods bring us closer to deformable reconstruction, but they can still be limited by the underlying representation.

\subsubsection{SLAM and Nonrigid SLAM}
\label{sec:discussslam}
For rigid SLAM, most works focus on means to address outliers and poor texture via removing the outliers or adjusting their models to use multiple correspondence methods (hybrids of both sparse and dense, etc.) that can measure motion in non-featured regions.
Some also add a goal of densifying sparse maps to better enable clinical application.
They do this via providing a parallel step that does not modify the sparse map or slow down the SLAM process.

Like seen in the other sections, the priors we have in this environment can have large effects, and integrating knowledge such as camera motion constraints can better improve performance of localization~\citep{vasconcelosRCMSLAMVisualLocalisation2019,batlleLightNeuSNeuralSurface2023b}.
Neural networks have begun to make an impact, with RNNs improving depth estimation by using the temporal state of the 2D disparity images.
Taking temporal visual state of the 3D map points into account using neural networks has yet to be approached.
Matching in SLAM has primarily been classical, but there are methods we saw in Section~\ref{sec:tissuetracking} that could better refine matches.

For Nonrigid SLAM, methods often quantify non-rigid methods using stereo data or photometric reconstruction, when the ideal goal is to quantify deformation and map accuracy.

For future work, being able to optimize non-physical parameters (e.g., a depth basis as in~\citet{liuSAGESLAMAppearance2022}) is of high interest.
For example, performing the same optimization for a map represented by a neural network can lead to great efficiency since it inherits the benefits of neural networks, with the efficiency of Levenberg-Marquardt optimization (or other NLLS methods).
Since we do not have accurate models for underlying deformation thus far, we seldom see re-localization/loop closure done in non-rigid SLAM, but an underlying neural representation with few control parameters is one possible way to address that problem.
The question comes down to how we define these priors while maintaining efficiency.

\subsection{Challenges and Limitations}
\label{sec:discusskeypoints}

Here we will discuss some of the important takeaways and challenges in tracking and mapping for MCV such as the importance of modelling lighting (Section~\ref{sec:discusslighting}), and having models that can represent the scene faithfully (Section~\ref{sec:discussmodels}).
We discuss the importance of efficiency in Section~\ref{sec:efficiency}, and the importance of understanding uncertainty in Section~\ref{sec:uncertainty}.

\subsubsection{Lighting}
\label{sec:discusslighting}
Here we briefly summarize some notes on the appearance of MCV environments.
Lighting models can be improved, and work has begun to do so in MCV: for example we can use GGX in BRDFs as a distribution instead of Beckmann~\citep{maltiCombiningConformalDeformation2014}, or even design other learned versions~\citep{batlleLightNeuSNeuralSurface2023b}.
There are many different options for choosing losses for unsupervised training~\citep{jonschkowskiWhatMattersUnsupervised2020}, but combining a loss with a learned lighting model is important for being able to better train models to recreate the environment correctly.
We often have to mask specularities or artifacts, but then after the fact we use SfS to continue to make use of these useful signals.
Detecting and exploiting specularities for estimating normals is another useful avenue~\citep{makkiEllipticalSpecularityDetection2023}.
Methods could benefit from integrating masked specularities and artifacts directly into the loss rather than having to combine different algorithms.
Most methods rely on hand-tuned mixtures of image-intensity based losses for training--combined L1 and SSIM for example (see~\cite{jonschkowskiWhatMattersUnsupervised2020} for recommendations), or classical algorithms that estimate using similar losses.
Discovering novel ways to take surface light transport into account could be useful not only in disparity estimation and deformable tracking, but also for mosaicking and learning blending methodologies.

\subsubsection{Underlying models}
\label{sec:discussmodels}
\begin{figure}[tb]\includegraphics[width=0.9\linewidth]{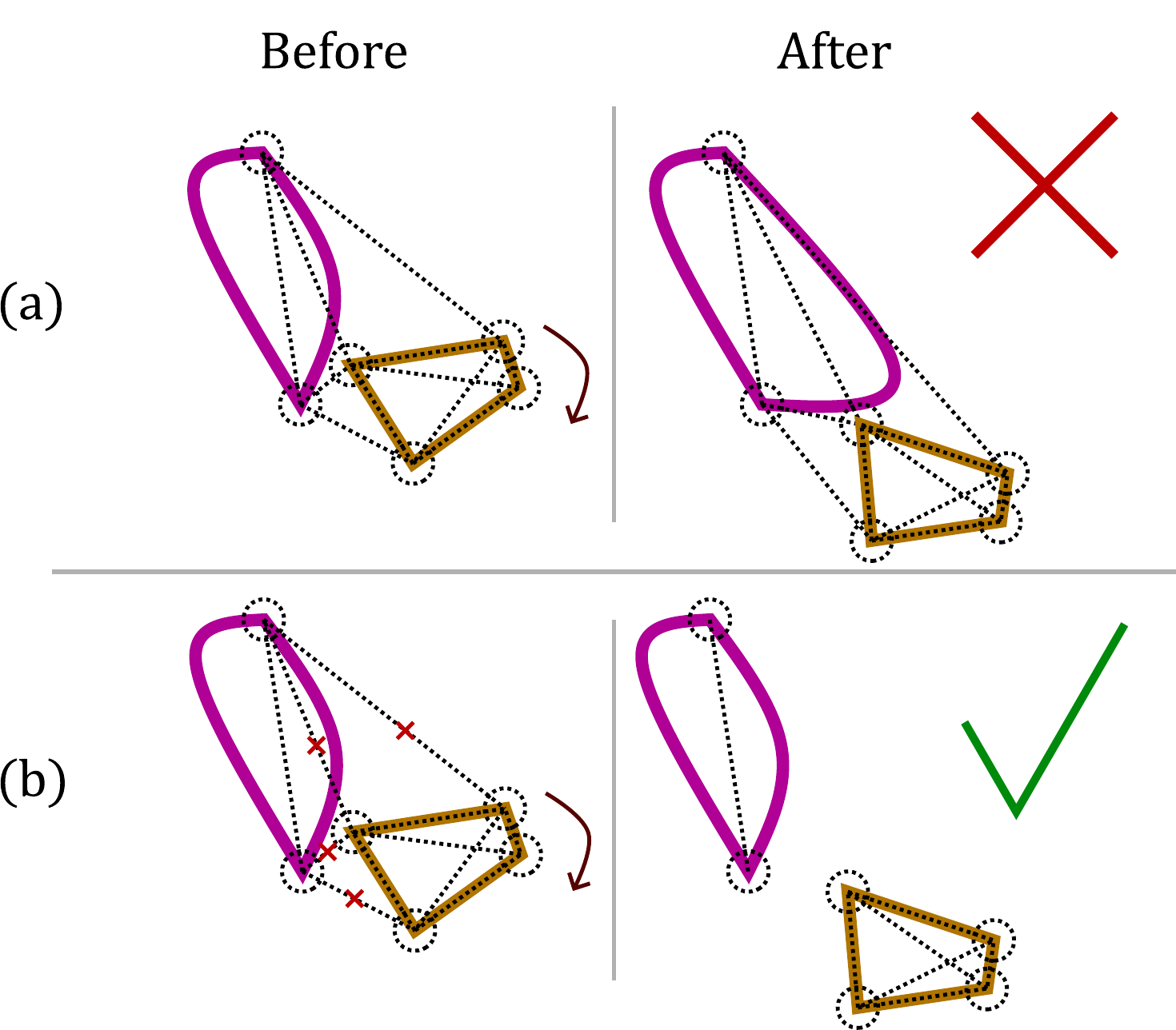}
	\caption{A toy example of representing deformation and allowing for discontinuity where  control points (dotted circles) determine motion of points in space. (a) shape transformation under a smooth model e.g., embedded deformation. (b) an ideal discontinuous model which does not connect disparate regions.}
    \label{fig:deformation}
\end{figure}

If we would like to map tissue, defining underlying models (in both NRSfM and NRSLAM) for how we map this tissue is extremely important.
There are gaps in how these models are represented, in that classical methods cannot faithfully represent tissue deformation (see Fig.~\ref{fig:deformation}), and modern machine learning is either not yet real-time capable, or does not enable tasks such as re-localization or bundle adjustment.
Designing ways to represent tissue or organs in an environment such as a persistent map that is also capable of adapting and changing is a very difficult problem.
In the near term, even using better data association terms (e.g., frame-to-frame tissue tracking) in lieu of classical descriptors such as SIFT or ORB is a closer, albeit still useful task.

In terms of how to do this with SLAM, there has been recent work using neural implicit functions~\citep{zhuNICESLAMNeuralImplicit2022} or Gaussian splatting for SLAM~\citep{yanGSSLAMDenseVisual2023,keethaSplaTAMSplatTrack2023}.
To bring this into a dynamic environment, movable map points with an implicit SDF representation are feasible~\citep{panPINSLAMLiDARSLAM2024}.
Ideally, the map points could represent deformation and texture information as well, as a neural sort of surfel.
Physically based real-time deformation modeling is another avenue with more efficient alternatives that take into account physical priors~\citep{linSuPerPMLargeDeformationRobust2023}.

\subsubsection{Efficiency}
\label{sec:efficiency}

Many downstream applications run alongside other applications on the system they are deployed on (e.g. a surgical robot, or a computer connected colonoscope).
Due to cost constraints, and needs for other possible applications on the system, algorithms will benefit from being efficient in terms of both inference time and memory usage.
Reporting FLOPs (floating point operations) and model size is a step in the right direction, but this does not take into account efficiencies of operations such as memory copies or random accesses.
Standardizing and benchmarking is difficult due to the varying systems, implementations, and batch sizes.
The best way to benchmark models continues to be an open problem.
That said, important metrics to report in publications focused on efficiency include: computational time for both training and inference, FLOPs, and memory usage.
In this review, we often found it difficult to determine a method's efficiency, and hope that in the future even offline methods will report this.

\subsubsection{Uncertainty}
\label{sec:uncertainty}

In order to be deployed clinically, a tracking algorithm should be able to determine failure.
This means points that drift should be discarded, and lost tracks should be detected.
This has been done before, but the question comes down to how do we enable guarantees of detecting failure cases based on heuristics such as feature quality.
The solution here could come down to creating an evaluation pipeline for evaluating tracking losses and uncertainty, or finding a means to do this in an unsupervised manner.

\subsection{Recent Developments in Computer Vision}
\label{sec:discussuseful}

In this section we will summarize some additional select concepts from computer vision and their relevance for Medical Computer Vision.
We note this is a snapshot of the field at this point in time, and should illustrate concepts that have not yet been fully translated into being used for tracking and mapping in MCV.
We begin with neural rendering in Section~\ref{sec:suggestedmodels}.
We detail its current use in medical computer vision, and then some future directions for it in the medical environment.
We then cover detection and point matching in Section~\ref{sec:suggestedmatching}.

\subsubsection{Neural Rendering}
\label{sec:suggestedmodels}

Neural networks for rendering such as neural radiance fields (NeRFs,~\citet{mildenhallNeRFRepresentingScenes2022}) parameterize 3D space volumetrically using an implicit neural representation.
Although they have been used for reconstruction, they have not yet been used in endoscopy for tracking and mapping applications in which point motion is used.
These can be trained in a per-scene manner using multiple views for reconstruction of deformable scenes, with recent applications in endoscopy~\citep{wangNeuralRenderingStereo2022}.
EndoNeRF~\citep{wangNeuralRenderingStereo2022} does this via training a neural deformation field along with a neural radiance field.
This framework is shown in Fig.~\ref{fig:wang2022}.
They demonstrate performance for scene reconstruction using image quality metrics of LPIPS, PSNR, and SSIM. Endoscopic neural radiance fields have also been approached with custom stereo neural networks~\citep{sunDynamicSurfaceReconstruction2023}.
More recently, EndoSurf~\citep{zhaEndoSurfNeuralSurface2023a} have approached the problem using signed distance fields which allows for better surface reconstruction.
This is likely because a SDF helps to enforce the fact that surfaces are often watertight and have edges rather than having possible density anywhere (like a cloud or fog).
They demonstrate performance with image quality metrics and depth reconstruction error as well.

\screenshot[wang2022]{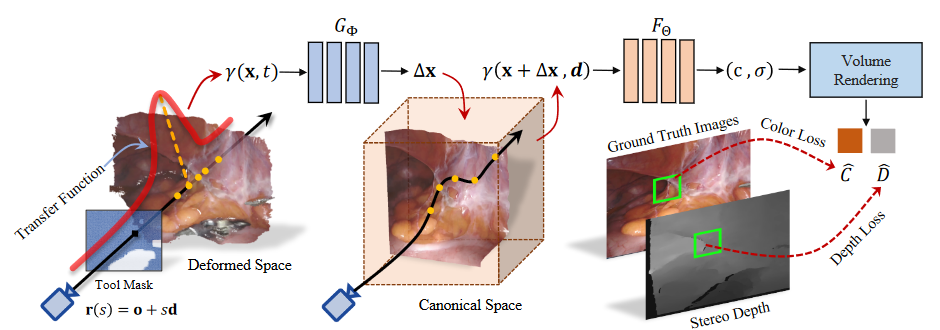}{Neural Radiance Fields (NERF) applied to deformable surgical scenes. A canonical volume is estimated along with a warp function to optimize a per-scene 3D reconstruction function. From~\cite{wangNeuralRenderingStereo2022} licensed under CC BY-NC-ND 4.0}

Also using signed distance fields (SDFs) for better surface representation,~\cite{batlleLightNeuSNeuralSurface2023b}, account for distance-based lighting decay models in rigid endoscopy.
Optimizing the training process (by conditioning on projected features from 2D planes),~\cite{yangNeuralLerPlaneRepresentations2023} (Neural Lerplane) propose a method that trains in minutes.
This efficient model is extended by~\cite{yangEfficientDeformableTissue2023} (Forplane), with optimized ray  marching, additional evaluation, and a monocular version using an off-the shelf depth predictor and scale based loss.  

That said, these presume a static camera.
BASED~\citep{sahaBASEDBundleAdjustingSurgical2023b} addresses this by optimizing camera pose in the initial NERF optimization before optimizing only deformation.
This is similar in principle to how ICP (iterative closest point) can be used to estimate a relative transformation before nonrigid fitting in classical point cloud models.

Methods such as Neural Graphics Primitives (NGP,~\citet{mullerInstantNeuralGraphics2022b}) or Gaussian splatting could be used to enable faster rendering and training~\citep{kerbl3DGaussianSplatting2023}.
Neural Radial Basis Functions (NeuRBF,~\citet{chenNeuRBFNeuralFields2023a}) provide promising directions for adaptive, non-voxelized representations.

In Gaussian splatting, multiple Gaussian density distributions are used along with spherical harmonics~\citep{kerbl3DGaussianSplatting2023} to represent a spatial volume.
This can be optimized with volume rendering in the same way as NeRFs.
Gaussian splats allow fast rendering using GPU graphics pipelines.
These have been extended to deformable environments by fitting positional parameters for the Gaussians over time~\citep{luitenDynamic3DGaussians2023a,wu4DGaussianSplatting2023}.

Bringing these works into endoscopy, \cite{zhuDeformableEndoscopicTissues2024} extend 4D Gaussian Splatting~\citep{wu4DGaussianSplatting2023}.
They adjust it by adding depth guidance for training and demonstrate high performance.
They mention that there is still possibility for artifacts and ambiguities in novel views, and recommend surface-alignment~\citep{guedonSuGaRSurfaceAlignedGaussian2023} for future work.
\cite{liuEndoGaussianGaussianSplatting2024} is another work to use 4D Gaussian splatting.
\cite{chenEndoGaussiansSingleView2024} do similarly with image inpainting and depth regularization.
\cite{huangEndo4DGSDistillingDepth2024} also use 4D Gaussian splatting, and train efficiently, using Depth-Anything~\citep{yangDepthAnythingUnleashing2024} for depth supervision via a ranked loss scheme.
That said, inference can still be slow, and these require manually masking instruments.

\metrics{}
In neural rendering for MCV, algorithms evaluate on depth reconstruction, or image reconstruction accuracy.
These do not necessarily measure deformation reconstruction accuracy.
For depth accuracy, they use Median Absolute Error (MedAE),  Mean Absolute Error (MAE), and Root Mean Squared Error (RMSE) between pixel estimates backprojected to 3D, and their true 3D positions.
For photometric accuracy, which measures how similar the reconstructed images look to the true images, they use LPIPS, PSNR, and SSIM.
See Section~\ref{sec:metrics} for more details on these.
These metrics are evaluated on scenes fully reconstructed with the neural rendering framework, against images from the same scene that are removed from the training set.

\textbf{Limitations:}
Although these are limited currently by having to train on the same scene, the next reasonable question is if we can generalize them to be conditional implicit functions that can be conditioned on the image for inference (such as is performed for flow in~\citet{schmidtFastGraphRefinement2022a}).
For NeRFs specifically, in addition to being limited by requiring training on the same dataset, they can also be slow for inference (each pixel requires multiple evaluations of a multi-layer perceptron (MLP) over samples along rays in 3D space).
For Gaussian splatting, inference is fast, but the question then comes down to how well spherical harmonics can represent lighting in the surgical environment.

If these are to be used in clinical applications, we need to begin quantifying them on real tissue deformation data.
Thus far, these methods use image reconstruction or stereo depth loss (on SCARED~\citep{allanStereoCorrespondenceReconstruction2021}) to qualify their loss, and require pre-defined instrument masks.
Due to this, there is no way to determine yet if they perform well on tracking problems, so we emphasize that for any clinical application, it is important to have these measurements.

\textbf{Directions:}
We will briefly summarize three possible ways in which neural rendering can be used for medical computer vision.

\textit{Representing Dynamic Scenes} is important for being able to faithfully represent the environment, and has been addressed in computer vision~\citep{liDynIBaRNeuralDynamic2023}.
More efficient methods are also being proposed~\citep{xu4K4DRealTime4D2023}.
Some of these include point-based and conditioning-based representations~\citep{zhouDynPointDynamicNeural2024,chenNeuRBFNeuralFields2023a}, and point-conditioned methods for monocular reconstruction~\citep{dasNeuralParametricGaussians2023}.
Separately, dynamic neural rendering has been used for creating offline methods for tracking points using full videos~\citep{wangTrackingEverythingEverywhere2023}.
Using NeRFs can even be used for segmenting regions based on unsupervised segmentation of motion~\citep{yangEmerNeRFEmergentSpatialTemporal2023}, and could be used for methods such as instrument segmentation.

\textit{Conditioning} on foundational features has shown to be useful for unsupervised semantic correspondence~\citep{zhangTaleTwoFeatures}.
NeRFs using conditional information from prior data could be useful to enable generalization, with diffusion NeRFs~\citep{chenSingleStageDiffusionNeRF2023, wynnDiffusioNeRFRegularizingNeural2023,guNerfDiffSingleimageView2023} providing directions for this.

\textit{Training Data:}
Reconstructing scenes using offline methods for neural rendering can provide those in MCV with a means to have high quality pseudo ground truth data that can be used for training algorithms.
This has been done using SfM to create data for training feature detection, description and reconstruction.
Dynamic neural rendering could ideally be used in the same manner.

\subsubsection{Detection and Matching:}
\label{sec:suggestedmatching}

Progress has been made recently in terms of feature detection and matching for computer vision as a whole.
For matching points between images, feature-metric refinement, which has been used for pose in SfM~\citep{sarlinPixelPerfectStructureFromMotionFeaturemetric2023} is an appetizing alternative to having detections be repeatable.
It is promising since it offers the best of both worlds - they detect in a sparse manner, but are not limited to matching sparse points.
Feature transformers can also be used for finding correspondences~\citep{sunLoFTRDetectorFreeLocal2021,jiangCOTRCorrespondenceTransformer2021a}, although at higher computational cost since they require a full-image search.
Alternatives can improve efficiency by conditioning motion on surrounding detected or tracked points in 2D~\citep{schmidtFastGraphRefinement2022a,moingDenseOpticalTracking2023} or 3D~\citep{schmidtSENDDSparseEfficient2023}.

With neural networks, we can learn point matching as a graph function of two point sets.
For finding point-to-point matching between point sets, there are works such as SuperGlue~\citep{sarlinSuperGlueLearningFeature2020} and LightGlue~\citep{lindenbergerLightGlueLocalFeature2023} that use a cross attention graph neural network to estimate correspondence between two point sets.
These are useful in the cases where the detections are accurate and have direct correspondences.

\begin{figure}[tb]\centering
    \includegraphics[width=0.9\linewidth]{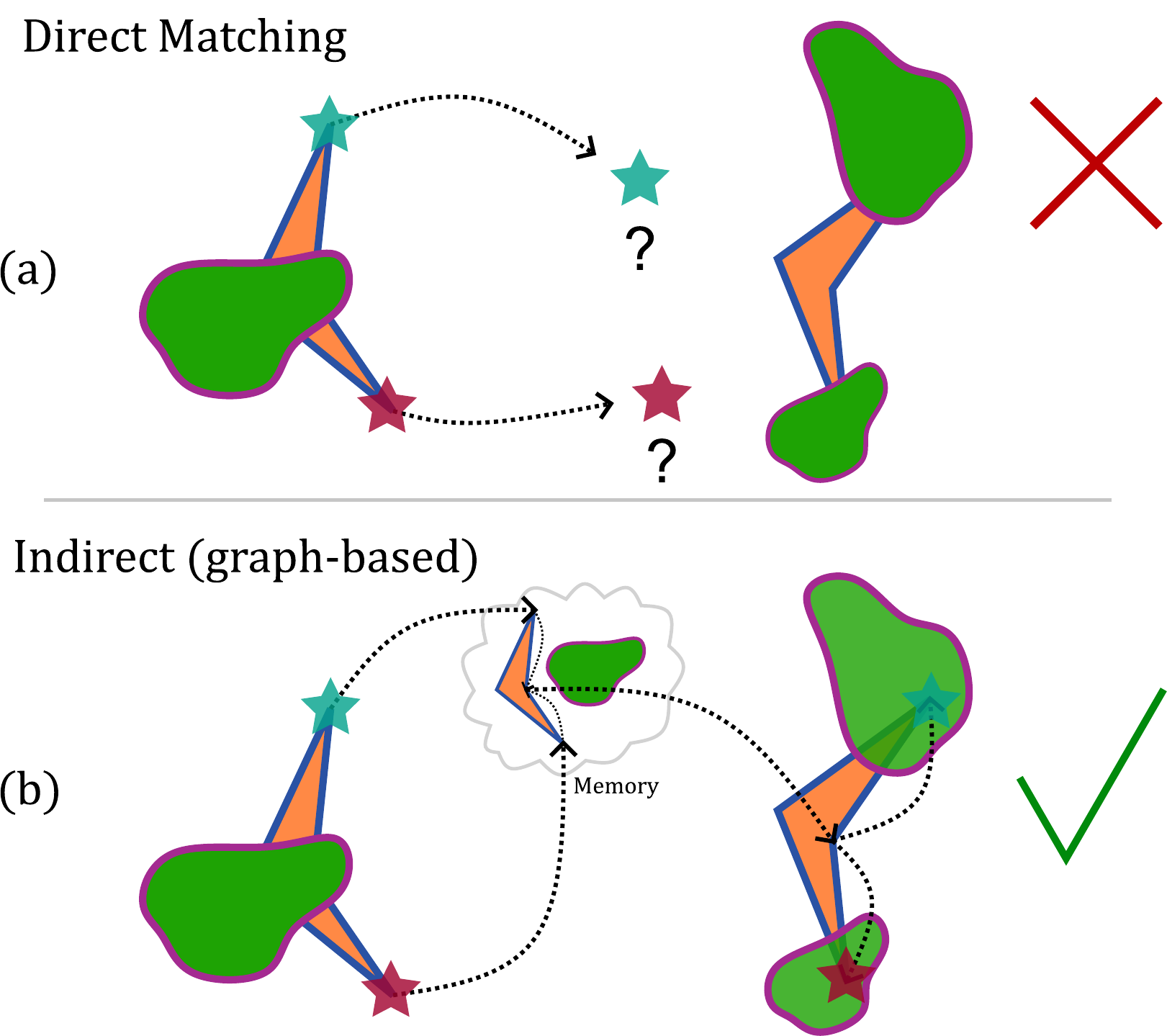}
	\caption{A toy example representing the matching process under occlusion. The orange chevron is the target object, occluded by green shapes (e.g., blur/smudge). (a) When trying to directly match points, if they are occluded or disappear, we are unable to find them. (b) If we have an idea of what the objects look like, a graph model should be able to take this into consideration, and use surroundings to locate the point.}
    \label{fig:matchinggraph}
\end{figure}

Alternative means to use detections is to not treat matching as such a one-to-one problem, in particular since points are not always detected, or visible.
If an object midpoint is only detected in one image, but the ends are detected in another, we should be able to softly match and use the end points to determine the midpoint in the other image (see Fig.~\ref{fig:matchinggraph}).
Transformer-based trackers that use cross-attention~\citep{karaevCoTrackerItBetter2023a} might be able to address this problem indirectly.

\section{Conclusion}
\label{sec:conclusion}
Computer vision and machine learning is taking a larger hold in  Tracking and Mapping in Medical Computer Vision, but there are still many difficulties we must account for.

We conclude with three main points.
Datasets for evaluation and training are only increasing in importance in this field.
Many of the challenges in each of the subtopics we cover have shared difficulties that should be used for crossover between these research topics (e.g., lighting, poor texture, and dynamic scenes).
In order to support deformable tracking, novel models still need to be designed.

We close out with some questions for the field as a whole.

\subsection{Questions for the Field}

\textbf{How do we localize in changing environments?}
If an environment is changing, is it still helpful to try to localize against a map that is out of date?
For example if tissue has been cut, we should ignore the map at all points near that region.
Thus, we need to design new ways to adaptively manipulate maps.

\textbf{How do we robustly deal with drift and feature change over time?}
For features that are initially defined to be a point, how do we track them if the feature changes -- e.g., blood splatter or translucent shifting of mesentery layers.
Some of these are ill-defined, but the question comes to when do we decide to update a feature's state and acknowledge a change in visible appearance.
Of course, this depends on the application.
For example on a mesentery layer, a feature could be anywhere in the translucent layer and is not easy to semantically define.

\textbf{How (or should) we quantify what is happening out of frame?}
For maintaining a map of the environment, assuming that motion happens out of frame can be useful.
That said, we cannot ever truly estimate the state of tissue.
The question comes down to how do we deform the map in order to best improve performance and correctly update the state when tissue comes back into view.
For maintaining performance in deformable bundle adjustment, loop closure, and  drift-correction, maintaining a map is important, so the question comes to how do we keep these without making overly coarse assumptions.

\textbf{SfM is to SLAM as NERF is to ...? or, How do we represent a map in a neural manner?}
The works of iMAP~\citep{sucarIMAPImplicitMapping2021} and NICE-SLAM~\citep{zhuNICESLAMNeuralImplicit2022} bring live NERF-optimization into a SLAM field.
The question is how we can do this with an underlying deformable motion model as a primary goal.
This again moves back to the question of how best to define models (Section~\ref{sec:discussmodels}).

\bibliographystyle{model2-names.bst}\biboptions{authoryear}
\bibliography{reviewpaper}

\begin{thebibliography}{265}
\expandafter\ifx\csname natexlab\endcsname\relax\def\natexlab#1{#1}\fi
\providecommand{\url}[1]{\texttt{#1}}
\providecommand{\href}[2]{#2}
\providecommand{\path}[1]{#1}
\providecommand{\DOIprefix}{doi:}
\providecommand{\ArXivprefix}{arXiv:}
\providecommand{\URLprefix}{URL: }
\providecommand{\Pubmedprefix}{pmid:}
\providecommand{\doi}[1]{\href{http://dx.doi.org/#1}{\path{#1}}}
\providecommand{\Pubmed}[1]{\href{pmid:#1}{\path{#1}}}
\providecommand{\bibinfo}[2]{#2}
\ifx\xfnm\relax \def\xfnm[#1]{\unskip,\space#1}\fi
\bibitem[{2AI(2024)}]{2AI2024}
, \bibinfo{year}{2024}.
\newblock \bibinfo{title}{{{2AI}}}.
\newblock \bibinfo{howpublished}{https://2ai.ipca.pt/about/}.
\bibitem[{Jme(2024)}]{Jmees2024}
, \bibinfo{year}{2024}.
\newblock \bibinfo{title}{Jmees}.
\newblock \bibinfo{howpublished}{https://www.jmees-inc.com/en}.
\bibitem[{Sco(2024)}]{Scopus2024}
, \bibinfo{year}{2024}.
\newblock \bibinfo{title}{Scopus}.
\newblock \bibinfo{howpublished}{https://www.scopus.com/home.uri}.
\bibitem[{Acidi et~al.(2023)Acidi, Ghallab, Cotin, Vibert and
  Golse}]{acidiAugmentedRealityLiver2023}
\bibinfo{author}{Acidi, B.}, \bibinfo{author}{Ghallab, M.},
  \bibinfo{author}{Cotin, S.}, \bibinfo{author}{Vibert, E.},
  \bibinfo{author}{Golse, N.}, \bibinfo{year}{2023}.
\newblock \bibinfo{title}{Augmented reality in liver surgery}.
\newblock \bibinfo{journal}{Journal of Visceral Surgery} \bibinfo{volume}{160},
  \bibinfo{pages}{118--126}.
\newblock \DOIprefix\doi{10.1016/j.jviscsurg.2023.01.008}.
\bibitem[{Agrawal et~al.(2008)Agrawal, Konolige and
  Blas}]{agrawalCenSurECenterSurround2008}
\bibinfo{author}{Agrawal, M.}, \bibinfo{author}{Konolige, K.},
  \bibinfo{author}{Blas, M.R.}, \bibinfo{year}{2008}.
\newblock \bibinfo{title}{{{CenSurE}}: {{Center Surround Extremas}} for
  {{Realtime Feature Detection}} and {{Matching}}}, in:
  \bibinfo{editor}{Forsyth, D.}, \bibinfo{editor}{Torr, P.},
  \bibinfo{editor}{Zisserman, A.} (Eds.), \bibinfo{booktitle}{Computer
  {{Vision}} {\textendash} {{ECCV}} 2008}, \bibinfo{publisher}{{Springer}},
  \bibinfo{address}{{Berlin, Heidelberg}}. pp. \bibinfo{pages}{102--115}.
\newblock \DOIprefix\doi{10.1007/978-3-540-88693-8_8}.
\bibitem[{Allan et~al.(2021)Allan, Mcleod, Wang, Rosenthal, Hu, Gard, Eisert,
  Fu, Zeffiro, Xia, Zhu, Luo, Jia, Zhang, Li, Sharan, Kurmann, Schmid,
  Sznitman, Psychogyios, Azizian, Stoyanov, {Maier-Hein} and
  Speidel}]{allanStereoCorrespondenceReconstruction2021}
\bibinfo{author}{Allan, M.}, \bibinfo{author}{Mcleod, J.},
  \bibinfo{author}{Wang, C.}, \bibinfo{author}{Rosenthal, J.C.},
  \bibinfo{author}{Hu, Z.}, \bibinfo{author}{Gard, N.},
  \bibinfo{author}{Eisert, P.}, \bibinfo{author}{Fu, K.X.},
  \bibinfo{author}{Zeffiro, T.}, \bibinfo{author}{Xia, W.},
  \bibinfo{author}{Zhu, Z.}, \bibinfo{author}{Luo, H.}, \bibinfo{author}{Jia,
  F.}, \bibinfo{author}{Zhang, X.}, \bibinfo{author}{Li, X.},
  \bibinfo{author}{Sharan, L.}, \bibinfo{author}{Kurmann, T.},
  \bibinfo{author}{Schmid, S.}, \bibinfo{author}{Sznitman, R.},
  \bibinfo{author}{Psychogyios, D.}, \bibinfo{author}{Azizian, M.},
  \bibinfo{author}{Stoyanov, D.}, \bibinfo{author}{{Maier-Hein}, L.},
  \bibinfo{author}{Speidel, S.}, \bibinfo{year}{2021}.
\newblock \bibinfo{title}{Stereo {{Correspondence}} and {{Reconstruction}} of
  {{Endoscopic Data Challenge}}}.
\newblock \bibinfo{journal}{arXiv:2101.01133 [cs]}
  \href{http://arxiv.org/abs/2101.01133}{\tt arXiv:2101.01133}.
\bibitem[{Arandjelovi{\'c} et~al.(2016)Arandjelovi{\'c}, Gronat, Torii, Pajdla
  and Sivic}]{arandjelovicNetVLADCNNArchitecture2016}
\bibinfo{author}{Arandjelovi{\'c}, R.}, \bibinfo{author}{Gronat, P.},
  \bibinfo{author}{Torii, A.}, \bibinfo{author}{Pajdla, T.},
  \bibinfo{author}{Sivic, J.}, \bibinfo{year}{2016}.
\newblock \bibinfo{title}{{{NetVLAD}}: {{CNN}} architecture for weakly
  supervised place recognition}, in: \bibinfo{booktitle}{{{arXiv}}:1511.07247
  [Cs]}.
\newblock \href{http://arxiv.org/abs/1511.07247}{\tt arXiv:1511.07247}.
\bibitem[{Atasoy et~al.(2008)Atasoy, Noonan, Benhimane, Navab and
  Yang}]{atasoyGlobalApproachAutomatic2008a}
\bibinfo{author}{Atasoy, S.}, \bibinfo{author}{Noonan, D.},
  \bibinfo{author}{Benhimane, S.}, \bibinfo{author}{Navab, N.},
  \bibinfo{author}{Yang, G.Z.}, \bibinfo{year}{2008}.
\newblock \bibinfo{title}{A Global Approach for Automatic Fibroscopic Video
  Mosaicing in Minimally Invasive Diagnosis}. volume \bibinfo{volume}{5241
  LNCS}.
\newblock \DOIprefix\doi{10.1007/978-3-540-85988-8_101}.
\bibitem[{Azagra et~al.(2022)Azagra, Sostres, Ferrandez, Riazuelo, Tomasini,
  Barbed, Morlana, Recasens, Batlle, {G{\'o}mez-Rodr{\'i}guez}, Elvira,
  L{\'o}pez, Oriol, Civera, Tard{\'o}s, Murillo, Lanas and
  Montiel}]{azagraEndoMapperDatasetComplete2022}
\bibinfo{author}{Azagra, P.}, \bibinfo{author}{Sostres, C.},
  \bibinfo{author}{Ferrandez, {\'A}.}, \bibinfo{author}{Riazuelo, L.},
  \bibinfo{author}{Tomasini, C.}, \bibinfo{author}{Barbed, O.L.},
  \bibinfo{author}{Morlana, J.}, \bibinfo{author}{Recasens, D.},
  \bibinfo{author}{Batlle, V.M.}, \bibinfo{author}{{G{\'o}mez-Rodr{\'i}guez},
  J.J.}, \bibinfo{author}{Elvira, R.}, \bibinfo{author}{L{\'o}pez, J.},
  \bibinfo{author}{Oriol, C.}, \bibinfo{author}{Civera, J.},
  \bibinfo{author}{Tard{\'o}s, J.D.}, \bibinfo{author}{Murillo, A.C.},
  \bibinfo{author}{Lanas, A.}, \bibinfo{author}{Montiel, J.M.M.},
  \bibinfo{year}{2022}.
\newblock \bibinfo{title}{{{EndoMapper}} dataset of complete calibrated
  endoscopy procedures}.
\newblock \href{http://arxiv.org/abs/2204.14240}{\tt arXiv:2204.14240}.
\bibitem[{Bano et~al.(2021)Bano, Casella, Vasconcelos, Moccia, Attilakos,
  Wimalasundera, David, Paladini, Deprest, De~Momi, Mattos and
  Stoyanov}]{banoFetRegPlacentalVessel2021}
\bibinfo{author}{Bano, S.}, \bibinfo{author}{Casella, A.},
  \bibinfo{author}{Vasconcelos, F.}, \bibinfo{author}{Moccia, S.},
  \bibinfo{author}{Attilakos, G.}, \bibinfo{author}{Wimalasundera, R.},
  \bibinfo{author}{David, A.L.}, \bibinfo{author}{Paladini, D.},
  \bibinfo{author}{Deprest, J.}, \bibinfo{author}{De~Momi, E.},
  \bibinfo{author}{Mattos, L.S.}, \bibinfo{author}{Stoyanov, D.},
  \bibinfo{year}{2021}.
\newblock \bibinfo{title}{{{FetReg}}: {{Placental Vessel Segmentation}} and
  {{Registration}} in {{Fetoscopy Challenge Dataset}}}.
\newblock \href{http://arxiv.org/abs/2106.05923}{\tt arXiv:2106.05923}.
\bibitem[{Bano and Stoyanov(2024)}]{banoChapter15Image2024}
\bibinfo{author}{Bano, S.}, \bibinfo{author}{Stoyanov, D.},
  \bibinfo{year}{2024}.
\newblock \bibinfo{title}{Chapter 15 - {{Image}} mosaicking}, in:
  \bibinfo{editor}{Frangi, A.F.}, \bibinfo{editor}{Prince, J.L.},
  \bibinfo{editor}{Sonka, M.} (Eds.), \bibinfo{booktitle}{Medical {{Image
  Analysis}}}. \bibinfo{publisher}{{Academic Press}}. The {{MICCAI Society}}
  Book {{Series}}, pp. \bibinfo{pages}{387--411}.
\newblock \DOIprefix\doi{10.1016/B978-0-12-813657-7.00030-3}.
\bibitem[{Bano et~al.(2023)Bano, Vasconcelos, David, Deprest and
  Stoyanov}]{banoPlacentalVesselguidedHybrid2023}
\bibinfo{author}{Bano, S.}, \bibinfo{author}{Vasconcelos, F.},
  \bibinfo{author}{David, A.}, \bibinfo{author}{Deprest, J.},
  \bibinfo{author}{Stoyanov, D.}, \bibinfo{year}{2023}.
\newblock \bibinfo{title}{Placental vessel-guided hybrid framework for
  fetoscopic mosaicking}.
\newblock \bibinfo{journal}{Computer Methods in Biomechanics and Biomedical
  Engineering: Imaging and Visualization} \bibinfo{volume}{11},
  \bibinfo{pages}{1166--1171}.
\newblock \DOIprefix\doi{10.1080/21681163.2022.2154278}.
\bibitem[{Bano et~al.(2020a)Bano, Vasconcelos, Shepherd, Poorten, Vercauteren,
  Ourselin, David, Deprest and Stoyanov}]{banoDeepPlacentalVessel2020}
\bibinfo{author}{Bano, S.}, \bibinfo{author}{Vasconcelos, F.},
  \bibinfo{author}{Shepherd, L.M.}, \bibinfo{author}{Poorten, E.V.},
  \bibinfo{author}{Vercauteren, T.}, \bibinfo{author}{Ourselin, S.},
  \bibinfo{author}{David, A.L.}, \bibinfo{author}{Deprest, J.},
  \bibinfo{author}{Stoyanov, D.}, \bibinfo{year}{2020}a.
\newblock \bibinfo{title}{Deep {{Placental Vessel Segmentation}} for
  {{Fetoscopic Mosaicking}}}, volume \bibinfo{volume}{12263}, pp.
  \bibinfo{pages}{763--773}.
\newblock \DOIprefix\doi{10.1007/978-3-030-59716-0_73},
  \href{http://arxiv.org/abs/2007.04349}{\tt arXiv:2007.04349}.
\bibitem[{Bano et~al.(2019)Bano, Vasconcelos, Tella~Amo, Dwyer, Gruijthuijsen,
  Deprest, Ourselin, Vander~Poorten, Vercauteren and
  Stoyanov}]{banoDeepSequentialMosaicking2019a}
\bibinfo{author}{Bano, S.}, \bibinfo{author}{Vasconcelos, F.},
  \bibinfo{author}{Tella~Amo, M.}, \bibinfo{author}{Dwyer, G.},
  \bibinfo{author}{Gruijthuijsen, C.}, \bibinfo{author}{Deprest, J.},
  \bibinfo{author}{Ourselin, S.}, \bibinfo{author}{Vander~Poorten, E.},
  \bibinfo{author}{Vercauteren, T.}, \bibinfo{author}{Stoyanov, D.},
  \bibinfo{year}{2019}.
\newblock \bibinfo{title}{Deep Sequential Mosaicking of Fetoscopic Videos}.
  volume \bibinfo{volume}{11764 LNCS}.
\newblock \DOIprefix\doi{10.1007/978-3-030-32239-7_35}.
\bibitem[{Bano et~al.(2020b)Bano, Vasconcelos, {Tella-Amo}, Dwyer,
  Gruijthuijsen, Vander~Poorten, Vercauteren, Ourselin, Deprest and
  Stoyanov}]{banoDeepLearningbasedFetoscopic2020}
\bibinfo{author}{Bano, S.}, \bibinfo{author}{Vasconcelos, F.},
  \bibinfo{author}{{Tella-Amo}, M.}, \bibinfo{author}{Dwyer, G.},
  \bibinfo{author}{Gruijthuijsen, C.}, \bibinfo{author}{Vander~Poorten, E.},
  \bibinfo{author}{Vercauteren, T.}, \bibinfo{author}{Ourselin, S.},
  \bibinfo{author}{Deprest, J.}, \bibinfo{author}{Stoyanov, D.},
  \bibinfo{year}{2020}b.
\newblock \bibinfo{title}{Deep learning-based fetoscopic mosaicking for
  field-of-view expansion}.
\newblock \bibinfo{journal}{International Journal of Computer Assisted
  Radiology and Surgery} \bibinfo{volume}{15}, \bibinfo{pages}{1807--1816}.
\newblock \DOIprefix\doi{10.1007/s11548-020-02242-8}.
\bibitem[{Barbed et~al.(2023)Barbed, Montiel, Fua and
  Murillo}]{barbedTrackingAdaptationImprove2023}
\bibinfo{author}{Barbed, O.L.}, \bibinfo{author}{Montiel, J.M.M.},
  \bibinfo{author}{Fua, P.}, \bibinfo{author}{Murillo, A.C.},
  \bibinfo{year}{2023}.
\newblock \bibinfo{title}{Tracking {{Adaptation}} to~{{Improve SuperPoint}}
  for~{{3D Reconstruction}} in~{{Endoscopy}}}, in: \bibinfo{editor}{Greenspan,
  H.}, \bibinfo{editor}{Madabhushi, A.}, \bibinfo{editor}{Mousavi, P.},
  \bibinfo{editor}{Salcudean, S.}, \bibinfo{editor}{Duncan, J.},
  \bibinfo{editor}{{Syeda-Mahmood}, T.}, \bibinfo{editor}{Taylor, R.} (Eds.),
  \bibinfo{booktitle}{Medical {{Image Computing}} and {{Computer Assisted
  Intervention}} {\textendash} {{MICCAI}} 2023}, \bibinfo{publisher}{{Springer
  Nature Switzerland}}, \bibinfo{address}{{Cham}}. pp.
  \bibinfo{pages}{583--593}.
\newblock \DOIprefix\doi{10.1007/978-3-031-43907-0_56}.
\bibitem[{Bardozzo et~al.(2022)Bardozzo, Collins, Forgione, Hostettler and
  Tagliaferri}]{bardozzoStaSiSNetStackedSiamese2022}
\bibinfo{author}{Bardozzo, F.}, \bibinfo{author}{Collins, T.},
  \bibinfo{author}{Forgione, A.}, \bibinfo{author}{Hostettler, A.},
  \bibinfo{author}{Tagliaferri, R.}, \bibinfo{year}{2022}.
\newblock \bibinfo{title}{{{StaSiS-Net}}: {{A}} stacked and siamese disparity
  estimation network for depth reconstruction in modern {{3D}} laparoscopy}.
\newblock \bibinfo{journal}{Medical Image Analysis} \bibinfo{volume}{77}.
\newblock \DOIprefix\doi{10.1016/j.media.2022.102380}.
\bibitem[{Baserga et~al.(2020)Baserga, Cappella, Gibelli, Sacco, Dolci,
  Cullati, Giann{\`i} and Sforza}]{basergaEfficacyAutologousFat2020}
\bibinfo{author}{Baserga, C.}, \bibinfo{author}{Cappella, A.},
  \bibinfo{author}{Gibelli, D.}, \bibinfo{author}{Sacco, R.},
  \bibinfo{author}{Dolci, C.}, \bibinfo{author}{Cullati, F.},
  \bibinfo{author}{Giann{\`i}, A.}, \bibinfo{author}{Sforza, C.},
  \bibinfo{year}{2020}.
\newblock \bibinfo{title}{Efficacy of autologous fat grafting in restoring
  facial symmetry in linear morphea-associated lesions}.
\newblock \bibinfo{journal}{Symmetry} \bibinfo{volume}{12},
  \bibinfo{pages}{1--13}.
\newblock \DOIprefix\doi{10.3390/sym12122098}.
\bibitem[{Batlle et~al.(2022)Batlle, Montiel and
  Tardos}]{batllePhotometricSingleviewDense2022}
\bibinfo{author}{Batlle, V.}, \bibinfo{author}{Montiel, J.},
  \bibinfo{author}{Tardos, J.}, \bibinfo{year}{2022}.
\newblock \bibinfo{title}{Photometric single-view dense {{3D}} reconstruction
  in endoscopy}, in: \bibinfo{booktitle}{{{IEEE International Conference}} on
  {{Intelligent Robots}} and {{Systems}}}, pp. \bibinfo{pages}{4904--4910}.
\newblock \DOIprefix\doi{10.1109/IROS47612.2022.9981742}.
\bibitem[{Batlle et~al.(2023)Batlle, Montiel, Fua and
  Tard{\'o}s}]{batlleLightNeuSNeuralSurface2023b}
\bibinfo{author}{Batlle, V.M.}, \bibinfo{author}{Montiel, J.M.M.},
  \bibinfo{author}{Fua, P.}, \bibinfo{author}{Tard{\'o}s, J.D.},
  \bibinfo{year}{2023}.
\newblock \bibinfo{title}{{{LightNeuS}}: {{Neural Surface Reconstruction}}
  in~{{Endoscopy Using Illumination Decline}}}, in: \bibinfo{editor}{Greenspan,
  H.}, \bibinfo{editor}{Madabhushi, A.}, \bibinfo{editor}{Mousavi, P.},
  \bibinfo{editor}{Salcudean, S.}, \bibinfo{editor}{Duncan, J.},
  \bibinfo{editor}{{Syeda-Mahmood}, T.}, \bibinfo{editor}{Taylor, R.} (Eds.),
  \bibinfo{booktitle}{Medical {{Image Computing}} and {{Computer Assisted
  Intervention}} {\textendash} {{MICCAI}} 2023}, \bibinfo{publisher}{{Springer
  Nature Switzerland}}, \bibinfo{address}{{Cham}}. pp.
  \bibinfo{pages}{502--512}.
\newblock \DOIprefix\doi{10.1007/978-3-031-43999-5_48}.
\bibitem[{Bay et~al.(2008)Bay, Ess, Tuytelaars and
  Van~Gool}]{baySpeededUpRobustFeatures2008}
\bibinfo{author}{Bay, H.}, \bibinfo{author}{Ess, A.},
  \bibinfo{author}{Tuytelaars, T.}, \bibinfo{author}{Van~Gool, L.},
  \bibinfo{year}{2008}.
\newblock \bibinfo{title}{Speeded-{{Up Robust Features}} ({{SURF}})}.
\newblock \bibinfo{journal}{Computer Vision and Image Understanding}
  \bibinfo{volume}{110}, \bibinfo{pages}{346--359}.
\newblock \DOIprefix\doi{10.1016/j.cviu.2007.09.014}.
\bibitem[{Bay et~al.(2006)Bay, Tuytelaars and
  Van~Gool}]{baySURFSpeededRobust2006}
\bibinfo{author}{Bay, H.}, \bibinfo{author}{Tuytelaars, T.},
  \bibinfo{author}{Van~Gool, L.}, \bibinfo{year}{2006}.
\newblock \bibinfo{title}{{{SURF}}: {{Speeded Up Robust Features}}}, in:
  \bibinfo{editor}{Leonardis, A.}, \bibinfo{editor}{Bischof, H.},
  \bibinfo{editor}{Pinz, A.} (Eds.), \bibinfo{booktitle}{Computer {{Vision}}
  {\textendash} {{ECCV}} 2006}. \bibinfo{publisher}{{Springer Berlin
  Heidelberg}}, \bibinfo{address}{{Berlin, Heidelberg}}. volume
  \bibinfo{volume}{3951}, pp. \bibinfo{pages}{404--417}.
\newblock \DOIprefix\doi{10.1007/11744023_32}.
\bibitem[{Behrens et~al.(2011)Behrens, Bommes, Stehle, Gross, Leonhardt and
  Aach}]{behrensRealtimeImageComposition2011}
\bibinfo{author}{Behrens, A.}, \bibinfo{author}{Bommes, M.},
  \bibinfo{author}{Stehle, T.}, \bibinfo{author}{Gross, S.},
  \bibinfo{author}{Leonhardt, S.}, \bibinfo{author}{Aach, T.},
  \bibinfo{year}{2011}.
\newblock \bibinfo{title}{Real-time image composition of bladder mosaics in
  fluorescence endoscopy}.
\newblock \bibinfo{journal}{Computer Science - Research and Development}
  \bibinfo{volume}{26}, \bibinfo{pages}{51--64}.
\newblock \DOIprefix\doi{10.1007/s00450-010-0135-z}.
\bibitem[{Bengio et~al.(2013)Bengio, L{\'e}onard and
  Courville}]{bengioEstimatingPropagatingGradients2013}
\bibinfo{author}{Bengio, Y.}, \bibinfo{author}{L{\'e}onard, N.},
  \bibinfo{author}{Courville, A.}, \bibinfo{year}{2013}.
\newblock \bibinfo{title}{Estimating or {{Propagating Gradients Through
  Stochastic Neurons}} for {{Conditional Computation}}}.
\newblock \href{http://arxiv.org/abs/1308.3432}{\tt arXiv:1308.3432}.
\bibitem[{Bergen et~al.(2009)Bergen, Ruthotto, M{\"u}nzenmayer, Rupp, Paulus
  and Winter}]{bergenFeaturebasedRealtimeEndoscopic2009}
\bibinfo{author}{Bergen, T.}, \bibinfo{author}{Ruthotto, S.},
  \bibinfo{author}{M{\"u}nzenmayer, C.}, \bibinfo{author}{Rupp, S.},
  \bibinfo{author}{Paulus, D.}, \bibinfo{author}{Winter, C.},
  \bibinfo{year}{2009}.
\newblock \bibinfo{title}{Feature-based real-time endoscopic mosaicking}, in:
  \bibinfo{booktitle}{{{ISPA}} 2009 - {{Proceedings}} of the 6th
  {{International Symposium}} on {{Image}} and {{Signal Processing}} and
  {{Analysis}}}, pp. \bibinfo{pages}{695--700}.
\newblock \DOIprefix\doi{10.1109/ispa.2009.5297633}.
\bibitem[{Bergen and
  Wittenberg(2016)}]{bergenStitchingSurfaceReconstruction2016}
\bibinfo{author}{Bergen, T.}, \bibinfo{author}{Wittenberg, T.},
  \bibinfo{year}{2016}.
\newblock \bibinfo{title}{Stitching and surface reconstruction from endoscopic
  image sequences: {{A}} review of applications and methods}.
\newblock \bibinfo{journal}{IEEE Journal of Biomedical and Health Informatics}
  \bibinfo{volume}{20}, \bibinfo{pages}{304--321}.
\newblock \DOIprefix\doi{10.1109/JBHI.2014.2384134}.
\bibitem[{Bernhardt et~al.(2017)Bernhardt, Nicolau, Soler and
  Doignon}]{bernhardtStatusAugmentedReality2017}
\bibinfo{author}{Bernhardt, S.}, \bibinfo{author}{Nicolau, S.A.},
  \bibinfo{author}{Soler, L.}, \bibinfo{author}{Doignon, C.},
  \bibinfo{year}{2017}.
\newblock \bibinfo{title}{The status of augmented reality in laparoscopic
  surgery as of 2016}.
\newblock \bibinfo{journal}{Medical Image Analysis} \bibinfo{volume}{37},
  \bibinfo{pages}{66--90}.
\newblock \DOIprefix\doi{10.1016/j.media.2017.01.007}.
\bibitem[{Bian et~al.(2020)Bian, Lin, Liu, Zhang, Yeung, Cheng and
  Reid}]{bianGMSGridBasedMotion2020}
\bibinfo{author}{Bian, J.W.}, \bibinfo{author}{Lin, W.Y.},
  \bibinfo{author}{Liu, Y.}, \bibinfo{author}{Zhang, L.},
  \bibinfo{author}{Yeung, S.K.}, \bibinfo{author}{Cheng, M.M.},
  \bibinfo{author}{Reid, I.}, \bibinfo{year}{2020}.
\newblock \bibinfo{title}{{{GMS}}: {{Grid-Based Motion Statistics}} for
  {{Fast}}, {{Ultra-robust Feature Correspondence}}}.
\newblock \bibinfo{journal}{International Journal of Computer Vision}
  \bibinfo{volume}{128}, \bibinfo{pages}{1580--1593}.
\bibitem[{Bobrow et~al.(2022)Bobrow, Golhar, Vijayan, Akshintala, Garcia and
  Durr}]{bobrowColonoscopy3DVideo2022}
\bibinfo{author}{Bobrow, T.L.}, \bibinfo{author}{Golhar, M.},
  \bibinfo{author}{Vijayan, R.}, \bibinfo{author}{Akshintala, V.S.},
  \bibinfo{author}{Garcia, J.R.}, \bibinfo{author}{Durr, N.J.},
  \bibinfo{year}{2022}.
\newblock \bibinfo{title}{Colonoscopy {{3D Video Dataset}} with {{Paired
  Depth}} from {{2D-3D Registration}}}.
\newblock \DOIprefix\doi{10.48550/arXiv.2206.08903},
  \href{http://arxiv.org/abs/2206.08903}{\tt arXiv:2206.08903}.
\bibitem[{{Borrego-Carazo} et~al.(2023){Borrego-Carazo}, Sanchez,
  {Castells-Rufas}, Carrabina and
  Gil}]{borrego-carazoBronchoPoseAnalysisData2023}
\bibinfo{author}{{Borrego-Carazo}, J.}, \bibinfo{author}{Sanchez, C.},
  \bibinfo{author}{{Castells-Rufas}, D.}, \bibinfo{author}{Carrabina, J.},
  \bibinfo{author}{Gil, D.}, \bibinfo{year}{2023}.
\newblock \bibinfo{title}{{{BronchoPose}}: An analysis of data and model
  configuration for vision-based bronchoscopy pose estimation}.
\newblock \bibinfo{journal}{Computer Methods and Programs in Biomedicine}
  \bibinfo{volume}{228}, \bibinfo{pages}{107241}.
\newblock \DOIprefix\doi{10.1016/j.cmpb.2022.107241}.
\bibitem[{Buchart et~al.(2009)Buchart, Vicente, Amundarain and
  Borro}]{buchartHybridVisualizationMaxillofacial2009}
\bibinfo{author}{Buchart, C.}, \bibinfo{author}{Vicente, G.},
  \bibinfo{author}{Amundarain, A.}, \bibinfo{author}{Borro, D.},
  \bibinfo{year}{2009}.
\newblock \bibinfo{title}{Hybrid visualization for maxillofacial surgery
  planning and simulation}, in: \bibinfo{booktitle}{Proceedings of the
  {{International Conference}} on {{Information Visualisation}}}, pp.
  \bibinfo{pages}{266--273}.
\newblock \DOIprefix\doi{10.1109/IV.2009.98}.
\bibitem[{Burschka et~al.(2005)Burschka, Li, Ishii, Taylor and
  Hager}]{burschkaScaleinvariantRegistrationMonocular2005}
\bibinfo{author}{Burschka, D.}, \bibinfo{author}{Li, M.},
  \bibinfo{author}{Ishii, M.}, \bibinfo{author}{Taylor, R.H.},
  \bibinfo{author}{Hager, G.D.}, \bibinfo{year}{2005}.
\newblock \bibinfo{title}{Scale-invariant registration of monocular endoscopic
  images to {{CT-scans}} for sinus surgery}.
\newblock \bibinfo{journal}{Medical Image Analysis} \bibinfo{volume}{9},
  \bibinfo{pages}{413--426}.
\newblock \DOIprefix\doi{10.1016/j.media.2005.05.005}.
\bibitem[{Burschka et~al.(2004)Burschka, Li, Taylor and
  Hager}]{burschkaScaleinvariantRegistrationMonocular2004}
\bibinfo{author}{Burschka, D.}, \bibinfo{author}{Li, M.},
  \bibinfo{author}{Taylor, R.}, \bibinfo{author}{Hager, G.},
  \bibinfo{year}{2004}.
\newblock \bibinfo{title}{Scale-invariant registration of monocular stereo
  images to {{3D}} surface models}.
\newblock \bibinfo{journal}{2004 IEEE/RSJ International Conference on
  Intelligent Robots and Systems (IROS)} \bibinfo{volume}{3},
  \bibinfo{pages}{2581--2586}.
\bibitem[{Burt and Adelson(1983)}]{burtMultiresolutionSplineApplication1983}
\bibinfo{author}{Burt, P.J.}, \bibinfo{author}{Adelson, E.H.},
  \bibinfo{year}{1983}.
\newblock \bibinfo{title}{A multiresolution spline with application to image
  mosaics}.
\newblock \bibinfo{journal}{ACM Transactions on Graphics (TOG)}
  \bibinfo{volume}{2}, \bibinfo{pages}{217--236}.
\bibitem[{Butler et~al.(2012)Butler, Wulff, Stanley and
  Black}]{butlerNaturalisticOpenSource2012}
\bibinfo{author}{Butler, D.J.}, \bibinfo{author}{Wulff, J.},
  \bibinfo{author}{Stanley, G.B.}, \bibinfo{author}{Black, M.J.},
  \bibinfo{year}{2012}.
\newblock \bibinfo{title}{A {{Naturalistic Open Source Movie}} for {{Optical
  Flow Evaluation}}}, in: \bibinfo{editor}{Hutchison, D.},
  \bibinfo{editor}{Kanade, T.}, \bibinfo{editor}{Kittler, J.},
  \bibinfo{editor}{Kleinberg, J.M.}, \bibinfo{editor}{Mattern, F.},
  \bibinfo{editor}{Mitchell, J.C.}, \bibinfo{editor}{Naor, M.},
  \bibinfo{editor}{Nierstrasz, O.}, \bibinfo{editor}{Pandu~Rangan, C.},
  \bibinfo{editor}{Steffen, B.}, \bibinfo{editor}{Sudan, M.},
  \bibinfo{editor}{Terzopoulos, D.}, \bibinfo{editor}{Tygar, D.},
  \bibinfo{editor}{Vardi, M.Y.}, \bibinfo{editor}{Weikum, G.},
  \bibinfo{editor}{Fitzgibbon, A.}, \bibinfo{editor}{Lazebnik, S.},
  \bibinfo{editor}{Perona, P.}, \bibinfo{editor}{Sato, Y.},
  \bibinfo{editor}{Schmid, C.} (Eds.), \bibinfo{booktitle}{Computer {{Vision}}
  {\textendash} {{ECCV}} 2012}. \bibinfo{publisher}{{Springer Berlin
  Heidelberg}}, \bibinfo{address}{{Berlin, Heidelberg}}. volume
  \bibinfo{volume}{7577}, pp. \bibinfo{pages}{611--625}.
\newblock \DOIprefix\doi{10.1007/978-3-642-33783-3_44}.
\bibitem[{Caccianiga et~al.(2024)Caccianiga, Nubert, Hutter and
  Kuchenbecker}]{caccianigaDense3DReconstruction2024}
\bibinfo{author}{Caccianiga, G.}, \bibinfo{author}{Nubert, J.},
  \bibinfo{author}{Hutter, M.}, \bibinfo{author}{Kuchenbecker, K.J.},
  \bibinfo{year}{2024}.
\newblock \bibinfo{title}{Dense {{3D Reconstruction Through Lidar}}: {{A
  Comparative Study}} on {{Ex-vivo Porcine Tissue}}}.
\bibitem[{Cao et~al.(2022)Cao, Wang, Zheng, Yin, Tang, Miao, Liu and
  Yang}]{caoAlgorithmStereoVision2022}
\bibinfo{author}{Cao, Z.}, \bibinfo{author}{Wang, Y.}, \bibinfo{author}{Zheng,
  W.}, \bibinfo{author}{Yin, L.}, \bibinfo{author}{Tang, Y.},
  \bibinfo{author}{Miao, W.}, \bibinfo{author}{Liu, S.}, \bibinfo{author}{Yang,
  B.}, \bibinfo{year}{2022}.
\newblock \bibinfo{title}{The algorithm of stereo vision and shape from shading
  based on endoscope imaging}.
\newblock \bibinfo{journal}{Biomedical Signal Processing and Control}
  \bibinfo{volume}{76}.
\newblock \DOIprefix\doi{10.1016/j.bspc.2022.103658}.
\bibitem[{Cartucho et~al.(2021)Cartucho, Tukra, Li, S.~Elson and
  Giannarou}]{cartuchoVisionBlenderToolEfficiently2021}
\bibinfo{author}{Cartucho, J.}, \bibinfo{author}{Tukra, S.},
  \bibinfo{author}{Li, Y.}, \bibinfo{author}{S.~Elson, D.},
  \bibinfo{author}{Giannarou, S.}, \bibinfo{year}{2021}.
\newblock \bibinfo{title}{{{VisionBlender}}: A tool to efficiently generate
  computer vision datasets for robotic surgery}.
\newblock \bibinfo{journal}{Computer Methods in Biomechanics and Biomedical
  Engineering: Imaging \& Visualization} \bibinfo{volume}{9},
  \bibinfo{pages}{331--338}.
\newblock \DOIprefix\doi{10.1080/21681163.2020.1835546}.
\bibitem[{Cartucho et~al.(2024)Cartucho, Weld, Tukra, Xu, Matsuzaki, Ishikawa,
  Kwon, Jang, Kim, Lee, Bai, Kahrs, Boecking, Allmendinger, M{\"u}ller, Zhang,
  Jin, Bano, Vasconcelos, Reiter, Hajek, Silva, Lima, Vila{\c c}a, Queir{\'o}s
  and Giannarou}]{cartuchoSurgTChallengeBenchmark2024}
\bibinfo{author}{Cartucho, J.}, \bibinfo{author}{Weld, A.},
  \bibinfo{author}{Tukra, S.}, \bibinfo{author}{Xu, H.},
  \bibinfo{author}{Matsuzaki, H.}, \bibinfo{author}{Ishikawa, T.},
  \bibinfo{author}{Kwon, M.}, \bibinfo{author}{Jang, Y.E.},
  \bibinfo{author}{Kim, K.J.}, \bibinfo{author}{Lee, G.}, \bibinfo{author}{Bai,
  B.}, \bibinfo{author}{Kahrs, L.A.}, \bibinfo{author}{Boecking, L.},
  \bibinfo{author}{Allmendinger, S.}, \bibinfo{author}{M{\"u}ller, L.},
  \bibinfo{author}{Zhang, Y.}, \bibinfo{author}{Jin, Y.},
  \bibinfo{author}{Bano, S.}, \bibinfo{author}{Vasconcelos, F.},
  \bibinfo{author}{Reiter, W.}, \bibinfo{author}{Hajek, J.},
  \bibinfo{author}{Silva, B.}, \bibinfo{author}{Lima, E.},
  \bibinfo{author}{Vila{\c c}a, J.L.}, \bibinfo{author}{Queir{\'o}s, S.},
  \bibinfo{author}{Giannarou, S.}, \bibinfo{year}{2024}.
\newblock \bibinfo{title}{{{SurgT}} challenge: {{Benchmark}} of soft-tissue
  trackers for robotic surgery}.
\newblock \bibinfo{journal}{Medical Image Analysis} \bibinfo{volume}{91},
  \bibinfo{pages}{102985}.
\newblock \DOIprefix\doi{10.1016/j.media.2023.102985}.
\bibitem[{Chadebecq et~al.(2023)Chadebecq, Lovat and
  Stoyanov}]{chadebecqArtificialIntelligenceAutomation2023}
\bibinfo{author}{Chadebecq, F.}, \bibinfo{author}{Lovat, L.B.},
  \bibinfo{author}{Stoyanov, D.}, \bibinfo{year}{2023}.
\newblock \bibinfo{title}{Artificial intelligence and automation in endoscopy
  and surgery}.
\newblock \bibinfo{journal}{Nat Rev Gastroenterol Hepatol}
  \bibinfo{volume}{20}, \bibinfo{pages}{171--182}.
\newblock \DOIprefix\doi{10.1038/s41575-022-00701-y}.
\bibitem[{Chang and Chen(2018)}]{changPyramidStereoMatching2018}
\bibinfo{author}{Chang, J.R.}, \bibinfo{author}{Chen, Y.S.},
  \bibinfo{year}{2018}.
\newblock \bibinfo{title}{Pyramid {{Stereo Matching Network}}}, in:
  \bibinfo{booktitle}{2018 {{IEEE}}/{{CVF Conference}} on {{Computer Vision}}
  and {{Pattern Recognition}}}, \bibinfo{publisher}{{IEEE}},
  \bibinfo{address}{{Salt Lake City, UT}}. pp. \bibinfo{pages}{5410--5418}.
\newblock \DOIprefix\doi{10.1109/CVPR.2018.00567}.
\bibitem[{Chang et~al.(2013)Chang, Stoyanov, Davison and
  Edwards}]{changRealtimeDenseStereo2013}
\bibinfo{author}{Chang, P.L.}, \bibinfo{author}{Stoyanov, D.},
  \bibinfo{author}{Davison, A.}, \bibinfo{author}{Edwards, P.},
  \bibinfo{year}{2013}.
\newblock \bibinfo{title}{Real-time dense stereo reconstruction using convex
  optimisation with a cost-volume for image-guided robotic surgery}.
\newblock \bibinfo{journal}{Lecture Notes in Computer Science (including
  subseries Lecture Notes in Artificial Intelligence and Lecture Notes in
  Bioinformatics)} \bibinfo{volume}{8149 LNCS}, \bibinfo{pages}{42--49}.
\newblock \DOIprefix\doi{10.1007/978-3-642-40811-3_6}.
\bibitem[{Cheema et~al.(2019)Cheema, Nazir, Sheng, Li, Qin, Kim and
  Feng}]{cheemaImagealignedDynamicLiver2019}
\bibinfo{author}{Cheema, M.}, \bibinfo{author}{Nazir, A.},
  \bibinfo{author}{Sheng, B.}, \bibinfo{author}{Li, P.}, \bibinfo{author}{Qin,
  J.}, \bibinfo{author}{Kim, J.}, \bibinfo{author}{Feng, D.},
  \bibinfo{year}{2019}.
\newblock \bibinfo{title}{Image-aligned dynamic liver reconstruction using
  intra-operative field of views for minimal invasive surgery}.
\newblock \bibinfo{journal}{IEEE Transactions on Biomedical Engineering}
  \bibinfo{volume}{66}, \bibinfo{pages}{2163--2173}.
\newblock \DOIprefix\doi{10.1109/TBME.2018.2884319}.
\bibitem[{Chen et~al.(2023a)Chen, Gu, Chen, Tian, Tu, Liu and
  Su}]{chenSingleStageDiffusionNeRF2023}
\bibinfo{author}{Chen, H.}, \bibinfo{author}{Gu, J.}, \bibinfo{author}{Chen,
  A.}, \bibinfo{author}{Tian, W.}, \bibinfo{author}{Tu, Z.},
  \bibinfo{author}{Liu, L.}, \bibinfo{author}{Su, H.}, \bibinfo{year}{2023}a.
\newblock \bibinfo{title}{Single-{{Stage Diffusion NeRF}}: {{A Unified
  Approach}} to {{3D Generation}} and {{Reconstruction}}}.
\newblock \href{http://arxiv.org/abs/2304.06714}{\tt arXiv:2304.06714}.
\bibitem[{Chen et~al.(2018)Chen, Tang, John, Wan and
  Zhang}]{chenSLAMbasedDenseSurface2018}
\bibinfo{author}{Chen, L.}, \bibinfo{author}{Tang, W.}, \bibinfo{author}{John,
  N.}, \bibinfo{author}{Wan, T.}, \bibinfo{author}{Zhang, J.},
  \bibinfo{year}{2018}.
\newblock \bibinfo{title}{{{SLAM-based}} dense surface reconstruction in
  monocular {{Minimally Invasive Surgery}} and its application to {{Augmented
  Reality}}}.
\newblock \bibinfo{journal}{Computer Methods and Programs in Biomedicine}
  \bibinfo{volume}{158}, \bibinfo{pages}{135--146}.
\newblock \DOIprefix\doi{10.1016/j.cmpb.2018.02.006}.
\bibitem[{Chen and Wang(2024)}]{chenEndoGaussiansSingleView2024}
\bibinfo{author}{Chen, Y.}, \bibinfo{author}{Wang, H.}, \bibinfo{year}{2024}.
\newblock \bibinfo{title}{{{EndoGaussians}}: {{Single View Dynamic Gaussian
  Splatting}} for {{Deformable Endoscopic Tissues Reconstruction}}}.
\newblock \href{http://arxiv.org/abs/2401.13352}{\tt arXiv:2401.13352}.
\bibitem[{Chen et~al.(2023b)Chen, Li, Song, Chen, Yu, Yuan and
  Xu}]{chenNeuRBFNeuralFields2023a}
\bibinfo{author}{Chen, Z.}, \bibinfo{author}{Li, Z.}, \bibinfo{author}{Song,
  L.}, \bibinfo{author}{Chen, L.}, \bibinfo{author}{Yu, J.},
  \bibinfo{author}{Yuan, J.}, \bibinfo{author}{Xu, Y.}, \bibinfo{year}{2023}b.
\newblock \bibinfo{title}{{{NeuRBF}}: {{A Neural Fields Representation}} with
  {{Adaptive Radial Basis Functions}}}, in: \bibinfo{booktitle}{Proceedings of
  the {{IEEE}}/{{CVF International Conference}} on {{Computer Vision}}}, pp.
  \bibinfo{pages}{4182--4194}.
\bibitem[{Chu et~al.(2020)Chu, Li, Li, Ding, Yang, Ai, Chen, Wang and
  Yang}]{chuEndoscopicImageFeature2020}
\bibinfo{author}{Chu, Y.}, \bibinfo{author}{Li, H.}, \bibinfo{author}{Li, X.},
  \bibinfo{author}{Ding, Y.}, \bibinfo{author}{Yang, X.}, \bibinfo{author}{Ai,
  D.}, \bibinfo{author}{Chen, X.}, \bibinfo{author}{Wang, Y.},
  \bibinfo{author}{Yang, J.}, \bibinfo{year}{2020}.
\newblock \bibinfo{title}{Endoscopic image feature matching via motion
  consensus and global bilateral regression}.
\newblock \bibinfo{journal}{Computer Methods and Programs in Biomedicine}
  \bibinfo{volume}{190}.
\newblock \DOIprefix\doi{10.1016/j.cmpb.2020.105370}.
\bibitem[{Collins et~al.(2016)Collins, Bartoli, Bourdel and
  Canis}]{collinsRobustRealtimeDense2016}
\bibinfo{author}{Collins, T.}, \bibinfo{author}{Bartoli, A.},
  \bibinfo{author}{Bourdel, N.}, \bibinfo{author}{Canis, M.},
  \bibinfo{year}{2016}.
\newblock \bibinfo{title}{Robust,Real-Time,Dense and Deformable {{3D}} Organ
  Tracking in Laparoscopic Videos}. volume \bibinfo{volume}{9900 LNCS}.
\newblock \DOIprefix\doi{10.1007/978-3-319-46720-7_47}.
\bibitem[{Cui et~al.(2024)Cui, Islam, Bai and
  Ren}]{cuiSurgicalDINOAdapterLearning2024}
\bibinfo{author}{Cui, B.}, \bibinfo{author}{Islam, M.}, \bibinfo{author}{Bai,
  L.}, \bibinfo{author}{Ren, H.}, \bibinfo{year}{2024}.
\newblock \bibinfo{title}{Surgical-{{DINO}}: {{Adapter Learning}} of
  {{Foundation Models}} for {{Depth Estimation}} in {{Endoscopic Surgery}}}.
\bibitem[{Das et~al.(2023)Das, Wewer, Yunus, Ilg and
  Lenssen}]{dasNeuralParametricGaussians2023}
\bibinfo{author}{Das, D.}, \bibinfo{author}{Wewer, C.}, \bibinfo{author}{Yunus,
  R.}, \bibinfo{author}{Ilg, E.}, \bibinfo{author}{Lenssen, J.E.},
  \bibinfo{year}{2023}.
\newblock \bibinfo{title}{Neural {{Parametric Gaussians}} for {{Monocular
  Non-Rigid Object Reconstruction}}}.
\newblock \href{http://arxiv.org/abs/2312.01196}{\tt arXiv:2312.01196}.
\bibitem[{De~Momi et~al.(2016)De~Momi, Ferrigno, Bosoni, Bassanini, Blasi,
  Casaceli, Fuschillo, Castana, Cossu, Lo~Russo and
  Cardinale}]{demomiMethodAssessmentTimevarying2016}
\bibinfo{author}{De~Momi, E.}, \bibinfo{author}{Ferrigno, G.},
  \bibinfo{author}{Bosoni, G.}, \bibinfo{author}{Bassanini, P.},
  \bibinfo{author}{Blasi, P.}, \bibinfo{author}{Casaceli, G.},
  \bibinfo{author}{Fuschillo, D.}, \bibinfo{author}{Castana, L.},
  \bibinfo{author}{Cossu, M.}, \bibinfo{author}{Lo~Russo, G.},
  \bibinfo{author}{Cardinale, F.}, \bibinfo{year}{2016}.
\newblock \bibinfo{title}{A method for the assessment of time-varying brain
  shift during navigated epilepsy surgery}.
\newblock \bibinfo{journal}{International Journal of Computer Assisted
  Radiology and Surgery} \bibinfo{volume}{11}, \bibinfo{pages}{473--481}.
\newblock \DOIprefix\doi{10.1007/s11548-015-1259-1}.
\bibitem[{De~Smet et~al.(2019)De~Smet, Poorten, Poliakov, Niu, Chesterman,
  Fornier, Ahmad, Ourak, Voros and
  Deprest}]{desmetEvaluatingPotentialBenefit2019}
\bibinfo{author}{De~Smet, J.}, \bibinfo{author}{Poorten, E.},
  \bibinfo{author}{Poliakov, V.}, \bibinfo{author}{Niu, K.},
  \bibinfo{author}{Chesterman, F.}, \bibinfo{author}{Fornier, J.},
  \bibinfo{author}{Ahmad, M.}, \bibinfo{author}{Ourak, M.},
  \bibinfo{author}{Voros, V.}, \bibinfo{author}{Deprest, J.},
  \bibinfo{year}{2019}.
\newblock \bibinfo{title}{Evaluating the potential benefit of autostereoscopy
  in laparoscopic sacrocolpopexy through {{VR}} simulation}, in:
  \bibinfo{booktitle}{2019 19th {{International Conference}} on {{Advanced
  Robotics}}, {{ICAR}} 2019}, pp. \bibinfo{pages}{566--571}.
\newblock \DOIprefix\doi{10.1109/ICAR46387.2019.8981553}.
\bibitem[{DeTone et~al.(2018)DeTone, Malisiewicz and
  Rabinovich}]{detoneSuperPointSelfSupervisedInterest2018}
\bibinfo{author}{DeTone, D.}, \bibinfo{author}{Malisiewicz, T.},
  \bibinfo{author}{Rabinovich, A.}, \bibinfo{year}{2018}.
\newblock \bibinfo{title}{{{SuperPoint}}: {{Self-Supervised Interest Point
  Detection}} and {{Description}}}, in: \bibinfo{booktitle}{Proceedings of the
  {{IEEE Conference}} on {{Computer Vision}} and {{Pattern Recognition
  Workshops}}}.
\bibitem[{Doersch et~al.(2022)Doersch, Gupta, Markeeva, Recasens, Smaira,
  Aytar, Carreira, Zisserman and Yang}]{doerschTAPVidBenchmarkTracking2022}
\bibinfo{author}{Doersch, C.}, \bibinfo{author}{Gupta, A.},
  \bibinfo{author}{Markeeva, L.}, \bibinfo{author}{Recasens, A.},
  \bibinfo{author}{Smaira, L.}, \bibinfo{author}{Aytar, Y.},
  \bibinfo{author}{Carreira, J.}, \bibinfo{author}{Zisserman, A.},
  \bibinfo{author}{Yang, Y.}, \bibinfo{year}{2022}.
\newblock \bibinfo{title}{{{TAP-Vid}}: {{A Benchmark}} for {{Tracking Any
  Point}} in a {{Video}}}, in: \bibinfo{booktitle}{Advances in {{Neural
  Information Processing Systems}}}, pp. \bibinfo{pages}{13610--13626}.
\bibitem[{Doersch et~al.(2023)Doersch, Yang, Vecerik, Gokay, Gupta, Aytar,
  Carreira and Zisserman}]{doerschTAPIRTrackingAny2023}
\bibinfo{author}{Doersch, C.}, \bibinfo{author}{Yang, Y.},
  \bibinfo{author}{Vecerik, M.}, \bibinfo{author}{Gokay, D.},
  \bibinfo{author}{Gupta, A.}, \bibinfo{author}{Aytar, Y.},
  \bibinfo{author}{Carreira, J.}, \bibinfo{author}{Zisserman, A.},
  \bibinfo{year}{2023}.
\newblock \bibinfo{title}{{{TAPIR}}: {{Tracking Any Point}} with per-frame
  {{Initialization}} and temporal {{Refinement}}}.
\newblock \href{http://arxiv.org/abs/2306.08637}{\tt arXiv:2306.08637}.
\bibitem[{Du et~al.(2019)Du, Allan, Bodenstedt, {Maier-Hein}, Speidel, Dore and
  Stoyanov}]{duPatchbasedAdaptiveWeighting2019}
\bibinfo{author}{Du, X.}, \bibinfo{author}{Allan, M.},
  \bibinfo{author}{Bodenstedt, S.}, \bibinfo{author}{{Maier-Hein}, L.},
  \bibinfo{author}{Speidel, S.}, \bibinfo{author}{Dore, A.},
  \bibinfo{author}{Stoyanov, D.}, \bibinfo{year}{2019}.
\newblock \bibinfo{title}{Patch-based adaptive weighting with segmentation and
  scale ({{PAWSS}}) for visual tracking in surgical video}.
\newblock \bibinfo{journal}{Medical Image Analysis} \bibinfo{volume}{57},
  \bibinfo{pages}{120--135}.
\newblock \DOIprefix\doi{10.1016/j.media.2019.07.002}.
\bibitem[{Du et~al.(2015)Du, Clancy, Arya, Hanna, Kelly, Elson and
  Stoyanov}]{duRobustSurfaceTracking2015}
\bibinfo{author}{Du, X.}, \bibinfo{author}{Clancy, N.}, \bibinfo{author}{Arya,
  S.}, \bibinfo{author}{Hanna, G.}, \bibinfo{author}{Kelly, J.},
  \bibinfo{author}{Elson, D.}, \bibinfo{author}{Stoyanov, D.},
  \bibinfo{year}{2015}.
\newblock \bibinfo{title}{Robust surface tracking combining features, intensity
  and illumination compensation}.
\newblock \bibinfo{journal}{International Journal of Computer Assisted
  Radiology and Surgery} \bibinfo{volume}{10}, \bibinfo{pages}{1915--1926}.
\newblock \DOIprefix\doi{10.1007/s11548-015-1243-9}.
\bibitem[{Dusmanu et~al.(2019)Dusmanu, Rocco, Pajdla, Pollefeys, Sivic, Torii
  and Sattler}]{dusmanuD2NetTrainableCNN2019}
\bibinfo{author}{Dusmanu, M.}, \bibinfo{author}{Rocco, I.},
  \bibinfo{author}{Pajdla, T.}, \bibinfo{author}{Pollefeys, M.},
  \bibinfo{author}{Sivic, J.}, \bibinfo{author}{Torii, A.},
  \bibinfo{author}{Sattler, T.}, \bibinfo{year}{2019}.
\newblock \bibinfo{title}{D2-{{Net}}: {{A Trainable CNN}} for {{Joint
  Description}} and {{Detection}} of {{Local Features}}}, in:
  \bibinfo{booktitle}{2019 {{IEEE}}/{{CVF Conference}} on {{Computer Vision}}
  and {{Pattern Recognition}} ({{CVPR}})}, \bibinfo{publisher}{{IEEE}},
  \bibinfo{address}{{Long Beach, CA, USA}}. pp. \bibinfo{pages}{8084--8093}.
\newblock \DOIprefix\doi{10.1109/CVPR.2019.00828}.
\bibitem[{Edwards et~al.(2022)Edwards, Psychogyios, Speidel, {Maier-Hein} and
  Stoyanov}]{edwardsSERVCTDisparityDataset2022}
\bibinfo{author}{Edwards, P.}, \bibinfo{author}{Psychogyios, D.},
  \bibinfo{author}{Speidel, S.}, \bibinfo{author}{{Maier-Hein}, L.},
  \bibinfo{author}{Stoyanov, D.}, \bibinfo{year}{2022}.
\newblock \bibinfo{title}{{{SERV-CT}}: {{A}} disparity dataset from cone-beam
  {{CT}} for validation of endoscopic {{3D}} reconstruction}.
\newblock \bibinfo{journal}{Medical Image Analysis} \bibinfo{volume}{76}.
\newblock \DOIprefix\doi{10.1016/j.media.2021.102302}.
\bibitem[{Engel et~al.(2018)Engel, Koltun and
  Cremers}]{engelDirectSparseOdometry2018}
\bibinfo{author}{Engel, J.}, \bibinfo{author}{Koltun, V.},
  \bibinfo{author}{Cremers, D.}, \bibinfo{year}{2018}.
\newblock \bibinfo{title}{Direct {{Sparse Odometry}}}.
\newblock \bibinfo{journal}{IEEE Transactions on Pattern Analysis and Machine
  Intelligence} \bibinfo{volume}{40}, \bibinfo{pages}{611--625}.
\newblock \DOIprefix\doi{10.1109/TPAMI.2017.2658577}.
\bibitem[{Faure et~al.(2012)Faure, Duriez, Delingette, Allard, Gilles,
  Marchesseau, Talbot, Courtecuisse, Bousquet, Peterlik and
  Cotin}]{faureSOFAMultiModelFramework2012}
\bibinfo{author}{Faure, F.}, \bibinfo{author}{Duriez, C.},
  \bibinfo{author}{Delingette, H.}, \bibinfo{author}{Allard, J.},
  \bibinfo{author}{Gilles, B.}, \bibinfo{author}{Marchesseau, S.},
  \bibinfo{author}{Talbot, H.}, \bibinfo{author}{Courtecuisse, H.},
  \bibinfo{author}{Bousquet, G.}, \bibinfo{author}{Peterlik, I.},
  \bibinfo{author}{Cotin, S.}, \bibinfo{year}{2012}.
\newblock \bibinfo{title}{{{SOFA}}: {{A Multi-Model Framework}} for
  {{Interactive Physical Simulation}}}, in: \bibinfo{editor}{Payan, Y.} (Ed.),
  \bibinfo{booktitle}{Soft {{Tissue Biomechanical Modeling}} for {{Computer
  Assisted Surgery}}}. \bibinfo{publisher}{{Springer}},
  \bibinfo{address}{{Berlin, Heidelberg}}. Studies in {{Mechanobiology}},
  {{Tissue Engineering}} and {{Biomaterials}}, pp. \bibinfo{pages}{283--321}.
\newblock \DOIprefix\doi{10.1007/8415_2012_125}.
\bibitem[{Frangi et~al.(1998)Frangi, Niessen, Vincken and
  Viergever}]{frangiMultiscaleVesselEnhancement1998}
\bibinfo{author}{Frangi, A.F.}, \bibinfo{author}{Niessen, W.J.},
  \bibinfo{author}{Vincken, K.L.}, \bibinfo{author}{Viergever, M.A.},
  \bibinfo{year}{1998}.
\newblock \bibinfo{title}{Multiscale vessel enhancement filtering}, in:
  \bibinfo{editor}{Wells, W.M.}, \bibinfo{editor}{Colchester, A.},
  \bibinfo{editor}{Delp, S.} (Eds.), \bibinfo{booktitle}{Medical {{Image
  Computing}} and {{Computer-Assisted Intervention}} {\textemdash}
  {{MICCAI}}'98}, \bibinfo{publisher}{{Springer}}, \bibinfo{address}{{Berlin,
  Heidelberg}}. pp. \bibinfo{pages}{130--137}.
\newblock \DOIprefix\doi{10.1007/BFb0056195}.
\bibitem[{Fu et~al.(2021a)Fu, Jin, Zhang, Dai, Gao, Wang, Li, Ding, Hu, Wang
  and Ye}]{fuVisualelectromagneticSystemNovel2021}
\bibinfo{author}{Fu, Z.}, \bibinfo{author}{Jin, Z.}, \bibinfo{author}{Zhang,
  C.}, \bibinfo{author}{Dai, Y.}, \bibinfo{author}{Gao, X.},
  \bibinfo{author}{Wang, Z.}, \bibinfo{author}{Li, L.}, \bibinfo{author}{Ding,
  G.}, \bibinfo{author}{Hu, H.}, \bibinfo{author}{Wang, P.},
  \bibinfo{author}{Ye, X.}, \bibinfo{year}{2021}a.
\newblock \bibinfo{title}{Visual-electromagnetic system: {{A}} novel
  fusion-based monocular localization, reconstruction, and measurement for
  flexible ureteroscopy}.
\newblock \bibinfo{journal}{International Journal of Medical Robotics and
  Computer Assisted Surgery} \bibinfo{volume}{17}.
\newblock \DOIprefix\doi{10.1002/rcs.2274}.
\bibitem[{Fu et~al.(2021b)Fu, Jin, Zhang, He, Zha, Hu, Gan, Yan, Wang and
  Ye}]{fuFutureEndoscopicNavigation2021}
\bibinfo{author}{Fu, Z.}, \bibinfo{author}{Jin, Z.}, \bibinfo{author}{Zhang,
  C.}, \bibinfo{author}{He, Z.}, \bibinfo{author}{Zha, Z.},
  \bibinfo{author}{Hu, C.}, \bibinfo{author}{Gan, T.}, \bibinfo{author}{Yan,
  Q.}, \bibinfo{author}{Wang, P.}, \bibinfo{author}{Ye, X.},
  \bibinfo{year}{2021}b.
\newblock \bibinfo{title}{The {{Future}} of {{Endoscopic Navigation}}: {{A
  Review}} of {{Advanced Endoscopic Vision Technology}}}.
\newblock \bibinfo{journal}{IEEE Access} .
\bibitem[{Fulton et~al.(2020)Fulton, Micah~Prendergast, Ditommaso and
  Rentschler}]{fultonComparingVisualOdometry2020}
\bibinfo{author}{Fulton, M.}, \bibinfo{author}{Micah~Prendergast, J.},
  \bibinfo{author}{Ditommaso, E.}, \bibinfo{author}{Rentschler, M.},
  \bibinfo{year}{2020}.
\newblock \bibinfo{title}{Comparing visual odometry systems in actively
  deforming simulated colon environments}, in: \bibinfo{booktitle}{{{IEEE
  International Conference}} on {{Intelligent Robots}} and {{Systems}}}, pp.
  \bibinfo{pages}{4988--4995}.
\newblock \DOIprefix\doi{10.1109/IROS45743.2020.9341159}.
\bibitem[{Gao and Tedrake(2019)}]{gaoSurfelWarpEfficientNonVolumetric2019}
\bibinfo{author}{Gao, W.}, \bibinfo{author}{Tedrake, R.}, \bibinfo{year}{2019}.
\newblock \bibinfo{title}{{{SurfelWarp}}: {{Efficient Non-Volumetric Single
  View Dynamic Reconstruction}}}.
\newblock \DOIprefix\doi{10.48550/arXiv.1904.13073},
  \href{http://arxiv.org/abs/1904.13073}{\tt arXiv:1904.13073}.
\bibitem[{Geiger et~al.(2011)Geiger, Roser and
  Urtasun}]{geigerEfficientLargeScaleStereo2011}
\bibinfo{author}{Geiger, A.}, \bibinfo{author}{Roser, M.},
  \bibinfo{author}{Urtasun, R.}, \bibinfo{year}{2011}.
\newblock \bibinfo{title}{Efficient {{Large-Scale Stereo Matching}}}, in:
  \bibinfo{editor}{Kimmel, R.}, \bibinfo{editor}{Klette, R.},
  \bibinfo{editor}{Sugimoto, A.} (Eds.), \bibinfo{booktitle}{Computer
  {{Vision}} {\textendash} {{ACCV}} 2010}. \bibinfo{publisher}{{Springer Berlin
  Heidelberg}}, \bibinfo{address}{{Berlin, Heidelberg}}. volume
  \bibinfo{volume}{6492}, pp. \bibinfo{pages}{25--38}.
\bibitem[{Giannarou et~al.(2009)Giannarou, {Visentini-Scarzanella} and
  Yang}]{giannarouAffineinvariantAnisotropicDetector2009}
\bibinfo{author}{Giannarou, S.}, \bibinfo{author}{{Visentini-Scarzanella}, M.},
  \bibinfo{author}{Yang, G.Z.}, \bibinfo{year}{2009}.
\newblock \bibinfo{title}{Affine-invariant anisotropic detector for soft tissue
  tracking in minimally invasive surgery}, in: \bibinfo{booktitle}{Proceedings
  - 2009 {{IEEE International Symposium}} on {{Biomedical Imaging}}: {{From
  Nano}} to {{Macro}}, {{ISBI}} 2009}, pp. \bibinfo{pages}{1059--1062}.
\newblock \DOIprefix\doi{10.1109/ISBI.2009.5193238}.
\bibitem[{Giannarou et~al.(2013)Giannarou, {Visentini-Scarzanella} and
  Yang}]{giannarouProbabilisticTrackingAffineinvariant2013}
\bibinfo{author}{Giannarou, S.}, \bibinfo{author}{{Visentini-Scarzanella}, M.},
  \bibinfo{author}{Yang, G.Z.}, \bibinfo{year}{2013}.
\newblock \bibinfo{title}{Probabilistic tracking of affine-invariant
  anisotropic regions}.
\newblock \bibinfo{journal}{IEEE Transactions on Pattern Analysis and Machine
  Intelligence} \bibinfo{volume}{35}, \bibinfo{pages}{130--143}.
\newblock \DOIprefix\doi{10.1109/TPAMI.2012.81}.
\bibitem[{Girerd et~al.(2020)Girerd, Kudryavtsev, Rougeot, Renaud, Rabenorosoa
  and Tamadazte}]{girerdAutomaticTipSteeringConcentric2020}
\bibinfo{author}{Girerd, C.}, \bibinfo{author}{Kudryavtsev, A.},
  \bibinfo{author}{Rougeot, P.}, \bibinfo{author}{Renaud, P.},
  \bibinfo{author}{Rabenorosoa, K.}, \bibinfo{author}{Tamadazte, B.},
  \bibinfo{year}{2020}.
\newblock \bibinfo{title}{Automatic {{Tip-Steering}} of {{Concentric Tube
  Robots}} in the {{Trachea Based}} on {{Visual SLAM}}}.
\newblock \bibinfo{journal}{IEEE Transactions on Medical Robotics and Bionics}
  \bibinfo{volume}{2}, \bibinfo{pages}{582--585}.
\newblock \DOIprefix\doi{10.1109/TMRB.2020.3034720}.
\bibitem[{Golyanik et~al.(2020)Golyanik, Jonas, Stricker and
  Theobalt}]{golyanikIntrinsicDynamicShape2020}
\bibinfo{author}{Golyanik, V.}, \bibinfo{author}{Jonas, A.},
  \bibinfo{author}{Stricker, D.}, \bibinfo{author}{Theobalt, C.},
  \bibinfo{year}{2020}.
\newblock \bibinfo{title}{Intrinsic {{Dynamic Shape Prior}} for {{Fast}},
  {{Sequential}} and {{Dense Non-Rigid Structure}} from {{Motion}} with
  {{Detection}} of {{Temporally-Disjoint Rigidity}}}.
\newblock \bibinfo{journal}{arXiv:1909.02468 [cs]}
  \href{http://arxiv.org/abs/1909.02468}{\tt arXiv:1909.02468}.
\bibitem[{Golyanik et~al.(2018)Golyanik, Shimada, Varanasi and
  Stricker}]{golyanikHDMNetMonocularNonRigid2018}
\bibinfo{author}{Golyanik, V.}, \bibinfo{author}{Shimada, S.},
  \bibinfo{author}{Varanasi, K.}, \bibinfo{author}{Stricker, D.},
  \bibinfo{year}{2018}.
\newblock \bibinfo{title}{{{HDM-Net}}: {{Monocular Non-Rigid 3D
  Reconstruction}} with {{Learned Deformation Model}}}.
\newblock \bibinfo{journal}{EuroVR}
  \DOIprefix\doi{10.1007/978-3-030-01790-3_4},
  \href{http://arxiv.org/abs/1803.10193}{\tt arXiv:1803.10193}.
\bibitem[{Gomez~Rodriguez et~al.(2022)Gomez~Rodriguez, Montiel and
  Tardos}]{gomezrodriguezTrackingMonocularCamera2022}
\bibinfo{author}{Gomez~Rodriguez, J.}, \bibinfo{author}{Montiel, J.},
  \bibinfo{author}{Tardos, J.}, \bibinfo{year}{2022}.
\newblock \bibinfo{title}{Tracking monocular camera pose and deformation for
  {{SLAM}} inside the human body}, in: \bibinfo{booktitle}{{{IEEE International
  Conference}} on {{Intelligent Robots}} and {{Systems}}}, pp.
  \bibinfo{pages}{5278--5285}.
\newblock \DOIprefix\doi{10.1109/IROS47612.2022.9981203}.
\bibitem[{{G{\'o}mez-Rodr{\'i}guez} et~al.(2021){G{\'o}mez-Rodr{\'i}guez},
  Lamarca, Morlana, Tard{\'o}s and
  Montiel}]{gomez-rodriguezSDDefSLAMSemiDirectMonocular2021}
\bibinfo{author}{{G{\'o}mez-Rodr{\'i}guez}, J.J.}, \bibinfo{author}{Lamarca,
  J.}, \bibinfo{author}{Morlana, J.}, \bibinfo{author}{Tard{\'o}s, J.D.},
  \bibinfo{author}{Montiel, J.M.M.}, \bibinfo{year}{2021}.
\newblock \bibinfo{title}{{{SD-DefSLAM}}: {{Semi-Direct Monocular SLAM}} for
  {{Deformable}} and {{Intracorporeal Scenes}}}, in: \bibinfo{booktitle}{2021
  {{IEEE International Conference}} on {{Robotics}} and {{Automation}}
  ({{ICRA}})}, pp. \bibinfo{pages}{5170--5177}.
\bibitem[{Gould et~al.(2021)Gould, Hartley and
  Campbell}]{gouldDeepDeclarativeNetworks2021}
\bibinfo{author}{Gould, S.}, \bibinfo{author}{Hartley, R.},
  \bibinfo{author}{Campbell, D.}, \bibinfo{year}{2021}.
\newblock \bibinfo{title}{Deep {{Declarative Networks}}: {{A New Hope}}}.
\newblock \bibinfo{journal}{IEEE Trans. Pattern Anal. Mach. Intell.} ,
  \bibinfo{pages}{1--1}\href{http://arxiv.org/abs/1909.04866}{\tt
  arXiv:1909.04866}.
\bibitem[{Grasa et~al.(2014)Grasa, Bernal, Casado, Gil and
  Montiel}]{grasaVisualSlamHandheld2014}
\bibinfo{author}{Grasa, O.}, \bibinfo{author}{Bernal, E.},
  \bibinfo{author}{Casado, S.}, \bibinfo{author}{Gil, I.},
  \bibinfo{author}{Montiel, J.}, \bibinfo{year}{2014}.
\newblock \bibinfo{title}{Visual slam for handheld monocular endoscope}.
\newblock \bibinfo{journal}{IEEE Transactions on Medical Imaging}
  \bibinfo{volume}{33}, \bibinfo{pages}{135--146}.
\newblock \DOIprefix\doi{10.1109/TMI.2013.2282997}.
\bibitem[{Gu et~al.(2023)Gu, Trevithick, Lin, Susskind, Theobalt, Liu and
  Ramamoorthi}]{guNerfDiffSingleimageView2023}
\bibinfo{author}{Gu, J.}, \bibinfo{author}{Trevithick, A.},
  \bibinfo{author}{Lin, K.E.}, \bibinfo{author}{Susskind, J.M.},
  \bibinfo{author}{Theobalt, C.}, \bibinfo{author}{Liu, L.},
  \bibinfo{author}{Ramamoorthi, R.}, \bibinfo{year}{2023}.
\newblock \bibinfo{title}{{{NerfDiff}}: {{Single-image View Synthesis}} with
  {{NeRF-guided Distillation}} from {{3D-aware Diffusion}}}, in:
  \bibinfo{booktitle}{Proceedings of the 40th {{International Conference}} on
  {{Machine Learning}}}, \bibinfo{publisher}{{PMLR}}. pp.
  \bibinfo{pages}{11808--11826}.
\bibitem[{Gu{\'e}don and Lepetit(2023)}]{guedonSuGaRSurfaceAlignedGaussian2023}
\bibinfo{author}{Gu{\'e}don, A.}, \bibinfo{author}{Lepetit, V.},
  \bibinfo{year}{2023}.
\newblock \bibinfo{title}{{{SuGaR}}: {{Surface-Aligned Gaussian Splatting}} for
  {{Efficient 3D Mesh Reconstruction}} and {{High-Quality Mesh Rendering}}}.
\newblock \DOIprefix\doi{10.48550/arXiv.2311.12775},
  \href{http://arxiv.org/abs/2311.12775}{\tt arXiv:2311.12775}.
\bibitem[{Guy et~al.(2022)Guy, Haberbusch, Promayon, Mancini and
  Voros}]{guyQualitativeComparisonImage2022}
\bibinfo{author}{Guy, S.}, \bibinfo{author}{Haberbusch, J.L.},
  \bibinfo{author}{Promayon, E.}, \bibinfo{author}{Mancini, S.},
  \bibinfo{author}{Voros, S.}, \bibinfo{year}{2022}.
\newblock \bibinfo{title}{Qualitative {{Comparison}} of {{Image Stitching
  Algorithms}} for {{Multi-Camera Systems}} in {{Laparoscopy}}}.
\newblock \bibinfo{journal}{Journal of Imaging} \bibinfo{volume}{8},
  \bibinfo{pages}{52}.
\newblock \DOIprefix\doi{10.3390/jimaging8030052}.
\bibitem[{Han et~al.(2024)Han, Acar, Henry and
  Wu}]{hanDepthAnythingMedical2024}
\bibinfo{author}{Han, J.J.}, \bibinfo{author}{Acar, A.},
  \bibinfo{author}{Henry, C.}, \bibinfo{author}{Wu, J.Y.},
  \bibinfo{year}{2024}.
\newblock \bibinfo{title}{Depth {{Anything}} in {{Medical Images}}: {{A
  Comparative Study}}}.
\newblock \href{http://arxiv.org/abs/2401.16600}{\tt arXiv:2401.16600}.
\bibitem[{Harley et~al.(2022)Harley, Fang and
  Fragkiadaki}]{harleyParticleVideoRevisited2022}
\bibinfo{author}{Harley, A.W.}, \bibinfo{author}{Fang, Z.},
  \bibinfo{author}{Fragkiadaki, K.}, \bibinfo{year}{2022}.
\newblock \bibinfo{title}{Particle {{Video Revisited}}: {{Tracking Through
  Occlusions Using Point Trajectories}}}.
\newblock \href{http://arxiv.org/abs/2204.04153}{\tt arXiv:2204.04153}.
\bibitem[{Hartkens et~al.(2003)Hartkens, Hill, {Castellano-Smith}, Hawkes,
  Maurer, Martin, Hall, Liu and Truwit}]{hartkensMeasurementAnalysisBrain2003a}
\bibinfo{author}{Hartkens, T.}, \bibinfo{author}{Hill, D.L.G.},
  \bibinfo{author}{{Castellano-Smith}, A.D.}, \bibinfo{author}{Hawkes, D.J.},
  \bibinfo{author}{Maurer, C.R.}, \bibinfo{author}{Martin, A.J.},
  \bibinfo{author}{Hall, W.A.}, \bibinfo{author}{Liu, H.},
  \bibinfo{author}{Truwit, C.L.}, \bibinfo{year}{2003}.
\newblock \bibinfo{title}{Measurement and analysis of brain deformation during
  neurosurgery}.
\newblock \bibinfo{journal}{IEEE Trans Med Imaging} \bibinfo{volume}{22},
  \bibinfo{pages}{82--92}.
\newblock \DOIprefix\doi{10.1109/TMI.2002.806596}.
\bibitem[{Hartwig et~al.(2022)Hartwig, Ostler, Rosenthal, Feu{\ss}ner, Wilhelm
  and Wollherr}]{hartwigMITISLAMBenchmark2022}
\bibinfo{author}{Hartwig, R.}, \bibinfo{author}{Ostler, D.},
  \bibinfo{author}{Rosenthal, J.C.}, \bibinfo{author}{Feu{\ss}ner, H.},
  \bibinfo{author}{Wilhelm, D.}, \bibinfo{author}{Wollherr, D.},
  \bibinfo{year}{2022}.
\newblock \bibinfo{title}{{{MITI}}: {{SLAM Benchmark}} for {{Laparoscopic
  Surgery}}}.
\newblock \href{http://arxiv.org/abs/2202.11496}{\tt arXiv:2202.11496}.
\bibitem[{Hayoz et~al.(2023)Hayoz, Hahne, Gallardo, Candinas, Kurmann, Allan
  and Sznitman}]{hayozLearningHowRobustly2023}
\bibinfo{author}{Hayoz, M.}, \bibinfo{author}{Hahne, C.},
  \bibinfo{author}{Gallardo, M.}, \bibinfo{author}{Candinas, D.},
  \bibinfo{author}{Kurmann, T.}, \bibinfo{author}{Allan, M.},
  \bibinfo{author}{Sznitman, R.}, \bibinfo{year}{2023}.
\newblock \bibinfo{title}{Learning how to robustly estimate camera pose in
  endoscopic videos}.
\newblock \bibinfo{journal}{International Journal of Computer Assisted
  Radiology and Surgery} \DOIprefix\doi{10.1007/s11548-023-02919-w}.
\bibitem[{{Hern{\'a}ndez-Mier} et~al.(2010){Hern{\'a}ndez-Mier}, Blondel, Daul,
  Wolf and Guillemin}]{hernandez-mierFastConstructionPanoramic2010}
\bibinfo{author}{{Hern{\'a}ndez-Mier}, Y.}, \bibinfo{author}{Blondel, W.},
  \bibinfo{author}{Daul, C.}, \bibinfo{author}{Wolf, D.},
  \bibinfo{author}{Guillemin, F.}, \bibinfo{year}{2010}.
\newblock \bibinfo{title}{Fast construction of panoramic images for cystoscopic
  exploration}.
\newblock \bibinfo{journal}{Computerized Medical Imaging and Graphics}
  \bibinfo{volume}{34}, \bibinfo{pages}{579--592}.
\newblock \DOIprefix\doi{10.1016/j.compmedimag.2010.02.002}.
\bibitem[{Hu et~al.(2021)Hu, Shen, Wallis, {Allen-Zhu}, Li, Wang, Wang and
  Chen}]{huLoRALowRankAdaptation2021}
\bibinfo{author}{Hu, E.J.}, \bibinfo{author}{Shen, Y.},
  \bibinfo{author}{Wallis, P.}, \bibinfo{author}{{Allen-Zhu}, Z.},
  \bibinfo{author}{Li, Y.}, \bibinfo{author}{Wang, S.}, \bibinfo{author}{Wang,
  L.}, \bibinfo{author}{Chen, W.}, \bibinfo{year}{2021}.
\newblock \bibinfo{title}{{{LoRA}}: {{Low-Rank Adaptation}} of {{Large Language
  Models}}}.
\newblock \href{http://arxiv.org/abs/2106.09685}{\tt arXiv:2106.09685}.
\bibitem[{Hu et~al.(2007)Hu, Penney, Edwards, Figl and
  Hawkes}]{hu3DReconstructionInternal2007}
\bibinfo{author}{Hu, M.}, \bibinfo{author}{Penney, G.},
  \bibinfo{author}{Edwards, P.}, \bibinfo{author}{Figl, M.},
  \bibinfo{author}{Hawkes, D.}, \bibinfo{year}{2007}.
\newblock \bibinfo{title}{{{3D}} reconstruction of internal organ surfaces for
  minimal invasive surgery} \bibinfo{volume}{4791 LNCS}, \bibinfo{pages}{77}.
\newblock \DOIprefix\doi{10.1007/978-3-540-75757-3_9}.
\bibitem[{Hu et~al.(2012)Hu, Penney, Figl, Edwards, Bello, Casula, Rueckert and
  Hawkes}]{huReconstruction3DSurface2012a}
\bibinfo{author}{Hu, M.}, \bibinfo{author}{Penney, G.}, \bibinfo{author}{Figl,
  M.}, \bibinfo{author}{Edwards, P.}, \bibinfo{author}{Bello, F.},
  \bibinfo{author}{Casula, R.}, \bibinfo{author}{Rueckert, D.},
  \bibinfo{author}{Hawkes, D.}, \bibinfo{year}{2012}.
\newblock \bibinfo{title}{Reconstruction of a {{3D}} surface from video that is
  robust to missing data and outliers: {{Application}} to minimally invasive
  surgery using stereo and mono endoscopes}.
\newblock \bibinfo{journal}{Medical Image Analysis} \bibinfo{volume}{16},
  \bibinfo{pages}{597--611}.
\newblock \DOIprefix\doi{10.1016/j.media.2010.11.002}.
\bibitem[{Hu et~al.(2009)Hu, Penney, Rueckert, Edwards, Bello, Casula, Figl and
  Hawkes}]{huNonrigidReconstructionBeating2009a}
\bibinfo{author}{Hu, M.}, \bibinfo{author}{Penney, G.},
  \bibinfo{author}{Rueckert, D.}, \bibinfo{author}{Edwards, P.},
  \bibinfo{author}{Bello, F.}, \bibinfo{author}{Casula, R.},
  \bibinfo{author}{Figl, M.}, \bibinfo{author}{Hawkes, D.},
  \bibinfo{year}{2009}.
\newblock \bibinfo{title}{Non-Rigid Reconstruction of the Beating Heart Surface
  for Minimally Invasive Cardiac Surgery}. volume \bibinfo{volume}{5761 LNCS}.
\newblock \DOIprefix\doi{10.1007/978-3-642-04268-3_5}.
\bibitem[{Huang et~al.(2024)Huang, Cui, Bai, Guo, Xu and
  Ren}]{huangEndo4DGSDistillingDepth2024}
\bibinfo{author}{Huang, Y.}, \bibinfo{author}{Cui, B.}, \bibinfo{author}{Bai,
  L.}, \bibinfo{author}{Guo, Z.}, \bibinfo{author}{Xu, M.},
  \bibinfo{author}{Ren, H.}, \bibinfo{year}{2024}.
\newblock \bibinfo{title}{Endo-{{4DGS}}: {{Distilling Depth Ranking}} for
  {{Endoscopic Monocular Scene Reconstruction}} with {{4D Gaussian
  Splatting}}}.
\newblock \href{http://arxiv.org/abs/2401.16416}{\tt arXiv:2401.16416}.
\bibitem[{Huo et~al.(2023)Huo, Zhou, Yuan, Yang and
  Wang}]{huoRealTimeDenseReconstruction2023}
\bibinfo{author}{Huo, J.}, \bibinfo{author}{Zhou, C.}, \bibinfo{author}{Yuan,
  B.}, \bibinfo{author}{Yang, Q.}, \bibinfo{author}{Wang, L.},
  \bibinfo{year}{2023}.
\newblock \bibinfo{title}{Real-{{Time Dense Reconstruction}} with {{Binocular
  Endoscopy Based}} on {{StereoNet}} and {{ORB-SLAM}}}.
\newblock \bibinfo{journal}{Sensors} \bibinfo{volume}{23}.
\newblock \DOIprefix\doi{10.3390/s23042074}.
\bibitem[{Hutchison et~al.(2010)Hutchison, Kanade, Kittler, Kleinberg, Mattern,
  Mitchell, Naor, Nierstrasz, Pandu~Rangan, Steffen, Sudan, Terzopoulos, Tygar,
  Vardi, Weikum, Calonder, Lepetit, Strecha and
  Fua}]{hutchisonBRIEFBinaryRobust2010}
\bibinfo{author}{Hutchison, D.}, \bibinfo{author}{Kanade, T.},
  \bibinfo{author}{Kittler, J.}, \bibinfo{author}{Kleinberg, J.M.},
  \bibinfo{author}{Mattern, F.}, \bibinfo{author}{Mitchell, J.C.},
  \bibinfo{author}{Naor, M.}, \bibinfo{author}{Nierstrasz, O.},
  \bibinfo{author}{Pandu~Rangan, C.}, \bibinfo{author}{Steffen, B.},
  \bibinfo{author}{Sudan, M.}, \bibinfo{author}{Terzopoulos, D.},
  \bibinfo{author}{Tygar, D.}, \bibinfo{author}{Vardi, M.Y.},
  \bibinfo{author}{Weikum, G.}, \bibinfo{author}{Calonder, M.},
  \bibinfo{author}{Lepetit, V.}, \bibinfo{author}{Strecha, C.},
  \bibinfo{author}{Fua, P.}, \bibinfo{year}{2010}.
\newblock \bibinfo{title}{{{BRIEF}}: {{Binary Robust Independent Elementary
  Features}}}, in: \bibinfo{editor}{Daniilidis, K.}, \bibinfo{editor}{Maragos,
  P.}, \bibinfo{editor}{Paragios, N.} (Eds.), \bibinfo{booktitle}{Computer
  {{Vision}} {\textendash} {{ECCV}} 2010}. \bibinfo{publisher}{{Springer Berlin
  Heidelberg}}, \bibinfo{address}{{Berlin, Heidelberg}}. volume
  \bibinfo{volume}{6314}, pp. \bibinfo{pages}{778--792}.
\newblock \DOIprefix\doi{10.1007/978-3-642-15561-1_56}.
\bibitem[{Ihler et~al.(2020)Ihler, Kuhnke, Laves and
  Ortmaier}]{ihlerSelfsupervisedDomainAdaptation2020}
\bibinfo{author}{Ihler, S.}, \bibinfo{author}{Kuhnke, F.},
  \bibinfo{author}{Laves, M.H.}, \bibinfo{author}{Ortmaier, T.},
  \bibinfo{year}{2020}.
\newblock \bibinfo{title}{Self-supervised domain adaptation for
  patient-specific, real-time tissue tracking}.
\newblock \bibinfo{journal}{Lecture Notes in Computer Science (including
  subseries Lecture Notes in Artificial Intelligence and Lecture Notes in
  Bioinformatics)} \bibinfo{volume}{12263 LNCS}, \bibinfo{pages}{54--64}.
\bibitem[{Ji et~al.(2014)Ji, Fan, Roberts, Hartov and
  Paulsen}]{jiCorticalSurfaceShift2014}
\bibinfo{author}{Ji, S.}, \bibinfo{author}{Fan, X.}, \bibinfo{author}{Roberts,
  D.}, \bibinfo{author}{Hartov, A.}, \bibinfo{author}{Paulsen, K.},
  \bibinfo{year}{2014}.
\newblock \bibinfo{title}{Cortical surface shift estimation using stereovision
  and optical flow motion tracking via projection image registration}.
\newblock \bibinfo{journal}{Medical Image Analysis} \bibinfo{volume}{18},
  \bibinfo{pages}{1169--1183}.
\newblock \DOIprefix\doi{10.1016/j.media.2014.07.001}.
\bibitem[{Jia et~al.(2021)Jia, Taylor and Chen}]{jiaLongTermRobust2021}
\bibinfo{author}{Jia, T.}, \bibinfo{author}{Taylor, Z.}, \bibinfo{author}{Chen,
  X.}, \bibinfo{year}{2021}.
\newblock \bibinfo{title}{Long term and robust {{6DoF}} motion tracking for
  highly dynamic stereo endoscopy videos}.
\newblock \bibinfo{journal}{Computerized Medical Imaging and Graphics}
  \bibinfo{volume}{94}.
\newblock \DOIprefix\doi{10.1016/j.compmedimag.2021.101995}.
\bibitem[{Jiang et~al.(2016)Jiang, Nakajima, Sohma, Saito, Kin, Oyama and
  Saito}]{jiangMarkerlessTrackingBrain2016}
\bibinfo{author}{Jiang, J.}, \bibinfo{author}{Nakajima, Y.},
  \bibinfo{author}{Sohma, Y.}, \bibinfo{author}{Saito, T.},
  \bibinfo{author}{Kin, T.}, \bibinfo{author}{Oyama, H.},
  \bibinfo{author}{Saito, N.}, \bibinfo{year}{2016}.
\newblock \bibinfo{title}{Marker-less tracking of brain surface deformations by
  non-rigid registration integrating surface and vessel/sulci features}.
\newblock \bibinfo{journal}{International Journal of Computer Assisted
  Radiology and Surgery} \bibinfo{volume}{11}, \bibinfo{pages}{1687--1701}.
\newblock \DOIprefix\doi{10.1007/s11548-016-1358-7}.
\bibitem[{Jiang et~al.(2015)Jiang, Zhang, Yang, Zhuang, Zhang and
  Gu}]{jiangRobustAutomatedMarkerless2015}
\bibinfo{author}{Jiang, L.}, \bibinfo{author}{Zhang, S.},
  \bibinfo{author}{Yang, J.}, \bibinfo{author}{Zhuang, X.},
  \bibinfo{author}{Zhang, L.}, \bibinfo{author}{Gu, L.}, \bibinfo{year}{2015}.
\newblock \bibinfo{title}{A robust automated markerless registration framework
  for neurosurgery navigation}.
\newblock \bibinfo{journal}{International Journal of Medical Robotics and
  Computer Assisted Surgery} \bibinfo{volume}{11}, \bibinfo{pages}{436--447}.
\newblock \DOIprefix\doi{10.1002/rcs.1626}.
\bibitem[{Jiang et~al.(2021)Jiang, Trulls, Hosang, Tagliasacchi and
  Yi}]{jiangCOTRCorrespondenceTransformer2021a}
\bibinfo{author}{Jiang, W.}, \bibinfo{author}{Trulls, E.},
  \bibinfo{author}{Hosang, J.}, \bibinfo{author}{Tagliasacchi, A.},
  \bibinfo{author}{Yi, K.M.}, \bibinfo{year}{2021}.
\newblock \bibinfo{title}{{{COTR}}: {{Correspondence Transformer}} for
  {{Matching Across Images}}}, in: \bibinfo{booktitle}{Proceedings of the
  {{IEEE}}/{{CVF International Conference}} on {{Computer Vision}}}, pp.
  \bibinfo{pages}{6207--6217}.
\bibitem[{Jin et~al.(2021)Jin, Mishkin, Mishchuk, Matas, Fua, Yi and
  Trulls}]{jinImageMatchingWide2021}
\bibinfo{author}{Jin, Y.}, \bibinfo{author}{Mishkin, D.},
  \bibinfo{author}{Mishchuk, A.}, \bibinfo{author}{Matas, J.},
  \bibinfo{author}{Fua, P.}, \bibinfo{author}{Yi, K.M.},
  \bibinfo{author}{Trulls, E.}, \bibinfo{year}{2021}.
\newblock \bibinfo{title}{Image {{Matching Across Wide Baselines}}: {{From
  Paper}} to {{Practice}}}.
\newblock \bibinfo{journal}{Int J Comput Vis} \bibinfo{volume}{129},
  \bibinfo{pages}{517--547}.
\newblock \DOIprefix\doi{10.1007/s11263-020-01385-0},
  \href{http://arxiv.org/abs/2003.01587}{\tt arXiv:2003.01587}.
\bibitem[{Jonschkowski et~al.(2020)Jonschkowski, Stone, Barron, Gordon,
  Konolige and Angelova}]{jonschkowskiWhatMattersUnsupervised2020}
\bibinfo{author}{Jonschkowski, R.}, \bibinfo{author}{Stone, A.},
  \bibinfo{author}{Barron, J.T.}, \bibinfo{author}{Gordon, A.},
  \bibinfo{author}{Konolige, K.}, \bibinfo{author}{Angelova, A.},
  \bibinfo{year}{2020}.
\newblock \bibinfo{title}{What {{Matters}} in {{Unsupervised Optical Flow}}},
  in: \bibinfo{editor}{Vedaldi, A.}, \bibinfo{editor}{Bischof, H.},
  \bibinfo{editor}{Brox, T.}, \bibinfo{editor}{Frahm, J.M.} (Eds.),
  \bibinfo{booktitle}{Computer {{Vision}} {\textendash} {{ECCV}} 2020},
  \bibinfo{publisher}{{Springer International Publishing}},
  \bibinfo{address}{{Cham}}. pp. \bibinfo{pages}{557--572}.
\bibitem[{Kalal et~al.(2010)Kalal, Mikolajczyk and
  Matas}]{kalalForwardBackwardErrorAutomatic2010}
\bibinfo{author}{Kalal, Z.}, \bibinfo{author}{Mikolajczyk, K.},
  \bibinfo{author}{Matas, J.}, \bibinfo{year}{2010}.
\newblock \bibinfo{title}{Forward-{{Backward Error}}: {{Automatic Detection}}
  of {{Tracking Failures}}}, in: \bibinfo{booktitle}{2010 20th {{International
  Conference}} on {{Pattern Recognition}} ({{ICPR}})}.
\bibitem[{Kam et~al.(2023)Kam, Wei, Opfermann, Saeidi, Hsieh, Wang, Kang and
  Krieger}]{kamAutonomousSystemVaginal2023}
\bibinfo{author}{Kam, M.}, \bibinfo{author}{Wei, S.},
  \bibinfo{author}{Opfermann, J.}, \bibinfo{author}{Saeidi, H.},
  \bibinfo{author}{Hsieh, M.}, \bibinfo{author}{Wang, K.},
  \bibinfo{author}{Kang, J.}, \bibinfo{author}{Krieger, A.},
  \bibinfo{year}{2023}.
\newblock \bibinfo{title}{Autonomous {{System}} for {{Vaginal Cuff Closure}}
  via {{Model-Based Planning}} and {{Markerless Tracking Techniques}}}.
\newblock \bibinfo{journal}{IEEE Robotics and Automation Letters}
  \bibinfo{volume}{8}, \bibinfo{pages}{3915--3922}.
\newblock \DOIprefix\doi{10.1109/LRA.2023.3273416}.
\bibitem[{Karaev et~al.(2023)Karaev, Rocco, Graham, Neverova, Vedaldi and
  Rupprecht}]{karaevCoTrackerItBetter2023a}
\bibinfo{author}{Karaev, N.}, \bibinfo{author}{Rocco, I.},
  \bibinfo{author}{Graham, B.}, \bibinfo{author}{Neverova, N.},
  \bibinfo{author}{Vedaldi, A.}, \bibinfo{author}{Rupprecht, C.},
  \bibinfo{year}{2023}.
\newblock \bibinfo{title}{{{CoTracker}}: {{It}} is {{Better}} to {{Track
  Together}}} \DOIprefix\doi{10.48550/ARXIV.2307.07635}.
\bibitem[{Karaoglu et~al.(2023)Karaoglu, Markova, Navab, Busam and
  Ladikos}]{karaogluRIDESelfSupervisedLearning2023}
\bibinfo{author}{Karaoglu, M.A.}, \bibinfo{author}{Markova, V.},
  \bibinfo{author}{Navab, N.}, \bibinfo{author}{Busam, B.},
  \bibinfo{author}{Ladikos, A.}, \bibinfo{year}{2023}.
\newblock \bibinfo{title}{{{RIDE}}: {{Self-Supervised Learning}} of
  {{Rotation-Equivariant Keypoint Detection}} and {{Invariant Description}} for
  {{Endoscopy}}}.
\newblock \href{http://arxiv.org/abs/2309.09563}{\tt arXiv:2309.09563}.
\bibitem[{Kazhdan and Hoppe(2013)}]{kazhdanScreenedPoissonSurface2013}
\bibinfo{author}{Kazhdan, M.}, \bibinfo{author}{Hoppe, H.},
  \bibinfo{year}{2013}.
\newblock \bibinfo{title}{Screened poisson surface reconstruction}.
\newblock \bibinfo{journal}{ACM Trans. Graph.} \bibinfo{volume}{32},
  \bibinfo{pages}{29:1--29:13}.
\newblock \DOIprefix\doi{10.1145/2487228.2487237}.
\bibitem[{Keetha et~al.(2023)Keetha, Karhade, Jatavallabhula, Yang, Scherer,
  Ramanan and Luiten}]{keethaSplaTAMSplatTrack2023}
\bibinfo{author}{Keetha, N.}, \bibinfo{author}{Karhade, J.},
  \bibinfo{author}{Jatavallabhula, K.M.}, \bibinfo{author}{Yang, G.},
  \bibinfo{author}{Scherer, S.}, \bibinfo{author}{Ramanan, D.},
  \bibinfo{author}{Luiten, J.}, \bibinfo{year}{2023}.
\newblock \bibinfo{title}{{{SplaTAM}}: {{Splat}}, {{Track}} \& {{Map 3D
  Gaussians}} for {{Dense RGB-D SLAM}}}.
\newblock \DOIprefix\doi{10.48550/arXiv.2312.02126},
  \href{http://arxiv.org/abs/2312.02126}{\tt arXiv:2312.02126}.
\bibitem[{Kerbl et~al.(2023)Kerbl, Kopanas, Leimkuehler and
  Drettakis}]{kerbl3DGaussianSplatting2023}
\bibinfo{author}{Kerbl, B.}, \bibinfo{author}{Kopanas, G.},
  \bibinfo{author}{Leimkuehler, T.}, \bibinfo{author}{Drettakis, G.},
  \bibinfo{year}{2023}.
\newblock \bibinfo{title}{{{3D Gaussian Splatting}} for {{Real-Time Radiance
  Field Rendering}}}.
\newblock \bibinfo{journal}{ACM Trans. Graph.} \bibinfo{volume}{42},
  \bibinfo{pages}{1--14}.
\newblock \DOIprefix\doi{10.1145/3592433}.
\bibitem[{Khamis et~al.(2018)Khamis, Fanello, Rhemann, Kowdle, Valentin and
  Izadi}]{khamisStereoNetGuidedHierarchical2018}
\bibinfo{author}{Khamis, S.}, \bibinfo{author}{Fanello, S.},
  \bibinfo{author}{Rhemann, C.}, \bibinfo{author}{Kowdle, A.},
  \bibinfo{author}{Valentin, J.}, \bibinfo{author}{Izadi, S.},
  \bibinfo{year}{2018}.
\newblock \bibinfo{title}{{{StereoNet}}: {{Guided Hierarchical Refinement}} for
  {{Real-Time Edge-Aware Depth Prediction}}}, in:
  \bibinfo{booktitle}{Proceedings of the {{European Conference}} on {{Computer
  Vision}} ({{ECCV}})}, pp. \bibinfo{pages}{573--590}.
\bibitem[{Khan et~al.(2023)Khan, Penner, Lanman and
  Xiao}]{khanTemporallyConsistentOnline2023}
\bibinfo{author}{Khan, N.}, \bibinfo{author}{Penner, E.},
  \bibinfo{author}{Lanman, D.}, \bibinfo{author}{Xiao, L.},
  \bibinfo{year}{2023}.
\newblock \bibinfo{title}{Temporally {{Consistent Online Depth Estimation Using
  Point-Based Fusion}}}, in: \bibinfo{booktitle}{2023 {{IEEE}}/{{CVF
  Conference}} on {{Computer Vision}} and {{Pattern Recognition}} ({{CVPR}})},
  \bibinfo{publisher}{{IEEE}}, \bibinfo{address}{{Vancouver, BC, Canada}}. pp.
  \bibinfo{pages}{9119--9129}.
\newblock \DOIprefix\doi{10.1109/CVPR52729.2023.00880}.
\bibitem[{Lamarca et~al.(2022)Lamarca, G{\'o}mez~Rodr{\'i}guez, Tard{\'o}s and
  Montiel}]{lamarcaDirectSparseDeformable2022}
\bibinfo{author}{Lamarca, J.}, \bibinfo{author}{G{\'o}mez~Rodr{\'i}guez, J.J.},
  \bibinfo{author}{Tard{\'o}s, J.D.}, \bibinfo{author}{Montiel, J.},
  \bibinfo{year}{2022}.
\newblock \bibinfo{title}{Direct and {{Sparse Deformable Tracking}}}.
\newblock \bibinfo{journal}{IEEE Robotics and Automation Letters}
  \bibinfo{volume}{7}, \bibinfo{pages}{11450--11457}.
\newblock \DOIprefix\doi{10.1109/LRA.2022.3201253}.
\bibitem[{Lamarca et~al.(2021)Lamarca, Parashar, Bartoli and
  Montiel}]{lamarcaDefSLAMTrackingMapping2021}
\bibinfo{author}{Lamarca, J.}, \bibinfo{author}{Parashar, S.},
  \bibinfo{author}{Bartoli, A.}, \bibinfo{author}{Montiel, J.},
  \bibinfo{year}{2021}.
\newblock \bibinfo{title}{{{DefSLAM}}: {{Tracking}} and mapping of deforming
  scenes from monocular sequences}.
\newblock \bibinfo{journal}{IEEE Transactions on Robotics}
  \bibinfo{volume}{37}, \bibinfo{pages}{291--303}.
\newblock \DOIprefix\doi{10.1109/TRO.2020.3020739}.
\bibitem[{Li et~al.(2021)Li, Bano, Deprest, David, Stoyanov and
  Vasconcelos}]{liGloballyOptimalFetoscopic2021}
\bibinfo{author}{Li, L.}, \bibinfo{author}{Bano, S.}, \bibinfo{author}{Deprest,
  J.}, \bibinfo{author}{David, A.}, \bibinfo{author}{Stoyanov, D.},
  \bibinfo{author}{Vasconcelos, F.}, \bibinfo{year}{2021}.
\newblock \bibinfo{title}{Globally optimal fetoscopic mosaicking based on pose
  graph optimisation with affine constraints}.
\newblock \bibinfo{journal}{IEEE Robotics and Automation Letters}
  \bibinfo{volume}{6}, \bibinfo{pages}{7831--7838}.
\newblock \DOIprefix\doi{10.1109/LRA.2021.3100938}.
\bibitem[{Li et~al.(2023a)Li, Mazomenos, Chandler, Obstein, Valdastri, Stoyanov
  and Vasconcelos}]{liRobustEndoscopicImage2023}
\bibinfo{author}{Li, L.}, \bibinfo{author}{Mazomenos, E.},
  \bibinfo{author}{Chandler, J.}, \bibinfo{author}{Obstein, K.},
  \bibinfo{author}{Valdastri, P.}, \bibinfo{author}{Stoyanov, D.},
  \bibinfo{author}{Vasconcelos, F.}, \bibinfo{year}{2023}a.
\newblock \bibinfo{title}{Robust endoscopic image mosaicking via fusion of
  multimodal estimation}.
\newblock \bibinfo{journal}{Medical Image Analysis} \bibinfo{volume}{84}.
\newblock \DOIprefix\doi{10.1016/j.media.2022.102709}.
\bibitem[{Li et~al.(2020)Li, Richter, Lu, Funk, Orosco, Zhu and
  Yip}]{liSuperSurgicalPerception2020}
\bibinfo{author}{Li, Y.}, \bibinfo{author}{Richter, F.}, \bibinfo{author}{Lu,
  J.}, \bibinfo{author}{Funk, E.}, \bibinfo{author}{Orosco, R.},
  \bibinfo{author}{Zhu, J.}, \bibinfo{author}{Yip, M.}, \bibinfo{year}{2020}.
\newblock \bibinfo{title}{Super: {{A}} surgical perception framework for
  endoscopic tissue manipulation with surgical robotics}.
\newblock \bibinfo{journal}{IEEE Robotics and Automation Letters}
  \bibinfo{volume}{5}, \bibinfo{pages}{2294--2301}.
\newblock \DOIprefix\doi{10.1109/LRA.2020.2970659}.
\bibitem[{Li et~al.(2023b)Li, Wang, Cole, Tucker and
  Snavely}]{liDynIBaRNeuralDynamic2023}
\bibinfo{author}{Li, Z.}, \bibinfo{author}{Wang, Q.}, \bibinfo{author}{Cole,
  F.}, \bibinfo{author}{Tucker, R.}, \bibinfo{author}{Snavely, N.},
  \bibinfo{year}{2023}b.
\newblock \bibinfo{title}{{{DynIBaR}}: {{Neural Dynamic Image-Based
  Rendering}}}, in: \bibinfo{booktitle}{2023 {{IEEE}}/{{CVF Conference}} on
  {{Computer Vision}} and {{Pattern Recognition}} ({{CVPR}})},
  \bibinfo{publisher}{{IEEE}}, \bibinfo{address}{{Vancouver, BC, Canada}}. pp.
  \bibinfo{pages}{4273--4284}.
\newblock \DOIprefix\doi{10.1109/CVPR52729.2023.00416}.
\bibitem[{Li et~al.(2023c)Li, Ye, Wang, Creighton, Taylor, Venkatesh and
  Unberath}]{liTemporallyConsistentOnline2023a}
\bibinfo{author}{Li, Z.}, \bibinfo{author}{Ye, W.}, \bibinfo{author}{Wang, D.},
  \bibinfo{author}{Creighton, F.X.}, \bibinfo{author}{Taylor, R.H.},
  \bibinfo{author}{Venkatesh, G.}, \bibinfo{author}{Unberath, M.},
  \bibinfo{year}{2023}c.
\newblock \bibinfo{title}{Temporally {{Consistent Online Depth Estimation}} in
  {{Dynamic Scenes}}}, in: \bibinfo{booktitle}{2023 {{IEEE}}/{{CVF Winter
  Conference}} on {{Applications}} of {{Computer Vision}} ({{WACV}})},
  \bibinfo{publisher}{{IEEE}}, \bibinfo{address}{{Waikoloa, HI, USA}}. pp.
  \bibinfo{pages}{3017--3026}.
\newblock \DOIprefix\doi{10.1109/WACV56688.2023.00303}.
\bibitem[{Lin et~al.(2016)Lin, Sun, Qian, Goldgof, Gitlin and
  You}]{linVideobased3DReconstruction2016}
\bibinfo{author}{Lin, B.}, \bibinfo{author}{Sun, Y.}, \bibinfo{author}{Qian,
  X.}, \bibinfo{author}{Goldgof, D.}, \bibinfo{author}{Gitlin, R.},
  \bibinfo{author}{You, Y.}, \bibinfo{year}{2016}.
\newblock \bibinfo{title}{Video-based {{3D}} reconstruction, laparoscope
  localization and deformation recovery for abdominal minimally invasive
  surgery: A survey}.
\newblock \bibinfo{journal}{The International Journal of Medical Robotics and
  Computer Assisted Surgery} \bibinfo{volume}{12}, \bibinfo{pages}{158--178}.
\newblock \DOIprefix\doi{10.1002/rcs.1661}.
\bibitem[{Lin et~al.(2023a)Lin, Miao, Alabiad, Liu, Wang, Lu, Richter and
  Yip}]{linSuPerPMLargeDeformationRobust2023}
\bibinfo{author}{Lin, S.}, \bibinfo{author}{Miao, A.J.},
  \bibinfo{author}{Alabiad, A.}, \bibinfo{author}{Liu, F.},
  \bibinfo{author}{Wang, K.}, \bibinfo{author}{Lu, J.},
  \bibinfo{author}{Richter, F.}, \bibinfo{author}{Yip, M.C.},
  \bibinfo{year}{2023}a.
\newblock \bibinfo{title}{{{SuPerPM}}: {{A Large Deformation-Robust Surgical
  Perception Framework Based}} on {{Deep Point Matching Learned}} from
  {{Physical Constrained Simulation Data}}}.
\newblock \href{http://arxiv.org/abs/2309.13863}{\tt arXiv:2309.13863}.
\bibitem[{Lin et~al.(2023b)Lin, Miao, Lu, Yu, Chiu, Richter and
  Yip}]{linSemanticSuPerSemanticawareSurgical2023}
\bibinfo{author}{Lin, S.}, \bibinfo{author}{Miao, A.J.}, \bibinfo{author}{Lu,
  J.}, \bibinfo{author}{Yu, S.}, \bibinfo{author}{Chiu, Z.Y.},
  \bibinfo{author}{Richter, F.}, \bibinfo{author}{Yip, M.C.},
  \bibinfo{year}{2023}b.
\newblock \bibinfo{title}{Semantic-{{SuPer}}: {{A Semantic-aware Surgical
  Perception Framework}} for {{Endoscopic Tissue Identification}},
  {{Reconstruction}}, and {{Tracking}}}, in: \bibinfo{booktitle}{2023 {{IEEE
  International Conference}} on {{Robotics}} and {{Automation}} ({{ICRA}})},
  pp. \bibinfo{pages}{4739--4746}.
\newblock \DOIprefix\doi{10.1109/ICRA48891.2023.10160746}.
\bibitem[{Lindenberger et~al.(2023)Lindenberger, Sarlin and
  Pollefeys}]{lindenbergerLightGlueLocalFeature2023}
\bibinfo{author}{Lindenberger, P.}, \bibinfo{author}{Sarlin, P.E.},
  \bibinfo{author}{Pollefeys, M.}, \bibinfo{year}{2023}.
\newblock \bibinfo{title}{{{LightGlue}}: {{Local Feature Matching}} at {{Light
  Speed}}}, in: \bibinfo{booktitle}{2023 {{IEEE}}/{{CVF International
  Conference}} on {{Computer Vision}} ({{ICCV}})}, \bibinfo{publisher}{{IEEE}},
  \bibinfo{address}{{Paris, France}}. pp. \bibinfo{pages}{17581--17592}.
\newblock \DOIprefix\doi{10.1109/ICCV51070.2023.01616}.
\bibitem[{Lipson et~al.(2021)Lipson, Teed and
  Deng}]{lipsonRAFTStereoMultilevelRecurrent2021}
\bibinfo{author}{Lipson, L.}, \bibinfo{author}{Teed, Z.},
  \bibinfo{author}{Deng, J.}, \bibinfo{year}{2021}.
\newblock \bibinfo{title}{{{RAFT-Stereo}}: {{Multilevel Recurrent Field
  Transforms}} for {{Stereo Matching}}}, in: \bibinfo{booktitle}{2021
  {{International Conference}} on {{3D Vision}} ({{3DV}})}, pp.
  \bibinfo{pages}{218--227}.
\newblock \DOIprefix\doi{10.1109/3DV53792.2021.00032}.
\bibitem[{Liu et~al.(2020a)Liu, Zhang, He, Liu, Wang, Tai, Luo, Wang, Li and
  Huang}]{liuLearningAnalogyReliable2020}
\bibinfo{author}{Liu, L.}, \bibinfo{author}{Zhang, J.}, \bibinfo{author}{He,
  R.}, \bibinfo{author}{Liu, Y.}, \bibinfo{author}{Wang, Y.},
  \bibinfo{author}{Tai, Y.}, \bibinfo{author}{Luo, D.}, \bibinfo{author}{Wang,
  C.}, \bibinfo{author}{Li, J.}, \bibinfo{author}{Huang, F.},
  \bibinfo{year}{2020}a.
\newblock \bibinfo{title}{Learning by {{Analogy}}: {{Reliable Supervision From
  Transformations}} for {{Unsupervised Optical Flow Estimation}}}, in:
  \bibinfo{booktitle}{2020 {{IEEE}}/{{CVF Conference}} on {{Computer Vision}}
  and {{Pattern Recognition}} ({{CVPR}})}, \bibinfo{publisher}{{IEEE}},
  \bibinfo{address}{{Seattle, WA, USA}}. pp. \bibinfo{pages}{6488--6497}.
\newblock \DOIprefix\doi{10.1109/CVPR42600.2020.00652}.
\bibitem[{Liu et~al.(2022)Liu, Li, Ishii, Hager, Taylor and
  Unberath}]{liuSAGESLAMAppearance2022}
\bibinfo{author}{Liu, X.}, \bibinfo{author}{Li, Z.}, \bibinfo{author}{Ishii,
  M.}, \bibinfo{author}{Hager, G.}, \bibinfo{author}{Taylor, R.},
  \bibinfo{author}{Unberath, M.}, \bibinfo{year}{2022}.
\newblock \bibinfo{title}{{{SAGE}}: {{SLAM}} with {{Appearance}} and {{Geometry
  Prior}} for {{Endoscopy}}}, in: \bibinfo{booktitle}{Proceedings - {{IEEE
  International Conference}} on {{Robotics}} and {{Automation}}}, pp.
  \bibinfo{pages}{5587--5593}.
\newblock \DOIprefix\doi{10.1109/ICRA46639.2022.9812257}.
\bibitem[{Liu et~al.(2020b)Liu, Sinha, Ishii, Hager, Reiter, Taylor and
  Unberath}]{liuDenseDepthEstimation2020a}
\bibinfo{author}{Liu, X.}, \bibinfo{author}{Sinha, A.}, \bibinfo{author}{Ishii,
  M.}, \bibinfo{author}{Hager, G.}, \bibinfo{author}{Reiter, A.},
  \bibinfo{author}{Taylor, R.}, \bibinfo{author}{Unberath, M.},
  \bibinfo{year}{2020}b.
\newblock \bibinfo{title}{Dense {{Depth Estimation}} in {{Monocular Endoscopy}}
  with {{Self-Supervised Learning Methods}}}.
\newblock \bibinfo{journal}{IEEE Transactions on Medical Imaging}
  \bibinfo{volume}{39}, \bibinfo{pages}{1438--1447}.
\newblock \DOIprefix\doi{10.1109/TMI.2019.2950936}.
\bibitem[{Liu et~al.(2018)Liu, Sinha, Unberath, Ishii, Hager, Taylor and
  Reiter}]{liuSelfsupervisedLearningDense2018}
\bibinfo{author}{Liu, X.}, \bibinfo{author}{Sinha, A.},
  \bibinfo{author}{Unberath, M.}, \bibinfo{author}{Ishii, M.},
  \bibinfo{author}{Hager, G.}, \bibinfo{author}{Taylor, R.},
  \bibinfo{author}{Reiter, A.}, \bibinfo{year}{2018}.
\newblock \bibinfo{title}{Self-supervised learning for dense depth estimation
  in monocular endoscopy}.
\newblock \bibinfo{journal}{Lecture Notes in Computer Science (including
  subseries Lecture Notes in Artificial Intelligence and Lecture Notes in
  Bioinformatics)} \bibinfo{volume}{11041 LNCS}, \bibinfo{pages}{128--138}.
\newblock \DOIprefix\doi{10.1007/978-3-030-01201-4_15}.
\bibitem[{Liu et~al.(2020c)Liu, Zheng, Killeen, Ishii, Hager, Taylor and
  Unberath}]{liuExtremelyDensePoint2020}
\bibinfo{author}{Liu, X.}, \bibinfo{author}{Zheng, Y.},
  \bibinfo{author}{Killeen, B.}, \bibinfo{author}{Ishii, M.},
  \bibinfo{author}{Hager, G.}, \bibinfo{author}{Taylor, R.},
  \bibinfo{author}{Unberath, M.}, \bibinfo{year}{2020}c.
\newblock \bibinfo{title}{Extremely dense point correspondences using a learned
  feature descriptor}, in: \bibinfo{booktitle}{Proceedings of the {{IEEE
  Computer Society Conference}} on {{Computer Vision}} and {{Pattern
  Recognition}}}, pp. \bibinfo{pages}{4846--4855}.
\newblock \DOIprefix\doi{10.1109/CVPR42600.2020.00490}.
\bibitem[{Liu et~al.(2024)Liu, Li, Yang and
  Yuan}]{liuEndoGaussianGaussianSplatting2024}
\bibinfo{author}{Liu, Y.}, \bibinfo{author}{Li, C.}, \bibinfo{author}{Yang,
  C.}, \bibinfo{author}{Yuan, Y.}, \bibinfo{year}{2024}.
\newblock \bibinfo{title}{{{EndoGaussian}}: {{Gaussian Splatting}} for
  {{Deformable Surgical Scene Reconstruction}}}.
\newblock \href{http://arxiv.org/abs/2401.12561}{\tt arXiv:2401.12561}.
\bibitem[{Liu et~al.(2023)Liu, Gao, Zhu, Yu and
  Fu}]{liuSurfaceDeformationTracking2023}
\bibinfo{author}{Liu, Z.}, \bibinfo{author}{Gao, W.}, \bibinfo{author}{Zhu,
  J.}, \bibinfo{author}{Yu, Z.}, \bibinfo{author}{Fu, Y.},
  \bibinfo{year}{2023}.
\newblock \bibinfo{title}{Surface deformation tracking in monocular
  laparoscopic video}.
\newblock \bibinfo{journal}{Medical Image Analysis} \bibinfo{volume}{86}.
\newblock \DOIprefix\doi{10.1016/j.media.2023.102775}.
\bibitem[{Lo et~al.(2008)Lo, Scarzanella, Stoyanov and
  Yang}]{loBeliefPropagationDepth2008}
\bibinfo{author}{Lo, B.}, \bibinfo{author}{Scarzanella, M.},
  \bibinfo{author}{Stoyanov, D.}, \bibinfo{author}{Yang, G.Z.},
  \bibinfo{year}{2008}.
\newblock \bibinfo{title}{Belief propagation for depth cue fusion in minimally
  invasive surgery}, p. \bibinfo{pages}{112}.
\newblock \DOIprefix\doi{10.1007/978-3-540-85990-1_13}.
\bibitem[{Long et~al.(2021)Long, Li, Yee, Ng, Taylor, Unberath and
  Dou}]{longEDSSREfficientDynamic2021}
\bibinfo{author}{Long, Y.}, \bibinfo{author}{Li, Z.}, \bibinfo{author}{Yee,
  C.}, \bibinfo{author}{Ng, C.}, \bibinfo{author}{Taylor, R.H.},
  \bibinfo{author}{Unberath, M.}, \bibinfo{author}{Dou, Q.},
  \bibinfo{year}{2021}.
\newblock \bibinfo{title}{E-{{DSSR}}: {{Efficient Dynamic Surgical Scene
  Reconstruction}} with {{Transformer-based Stereoscopic Depth Perception}}},
  in: \bibinfo{booktitle}{{{MICCAI}}}.
\newblock \DOIprefix\doi{10.1007/978-3-030-87202-1_40}.
\bibitem[{Lowe(1999)}]{loweObjectRecognitionLocal1999a}
\bibinfo{author}{Lowe, D.}, \bibinfo{year}{1999}.
\newblock \bibinfo{title}{Object recognition from local scale-invariant
  features}, in: \bibinfo{booktitle}{Proceedings of the {{Seventh IEEE
  International Conference}} on {{Computer Vision}}}, pp.
  \bibinfo{pages}{1150--1157 vol.2}.
\newblock \DOIprefix\doi{10.1109/ICCV.1999.790410}.
\bibitem[{Lu et~al.(2021)Lu, Jayakumari, Richter, Li and
  Yip}]{luSuPerDeepSurgical2021}
\bibinfo{author}{Lu, J.}, \bibinfo{author}{Jayakumari, A.},
  \bibinfo{author}{Richter, F.}, \bibinfo{author}{Li, Y.},
  \bibinfo{author}{Yip, M.}, \bibinfo{year}{2021}.
\newblock \bibinfo{title}{{{SuPer Deep}}: {{A Surgical Perception Framework}}
  for {{Robotic Tissue Manipulation}} using {{Deep Learning}} for {{Feature
  Extraction}}}, in: \bibinfo{booktitle}{Proceedings - {{IEEE International
  Conference}} on {{Robotics}} and {{Automation}}}, pp.
  \bibinfo{pages}{4783--4789}.
\newblock \DOIprefix\doi{10.1109/ICRA48506.2021.9561249}.
\bibitem[{Lucas and Kanade(1981)}]{lucasIterativeImageRegistration1981}
\bibinfo{author}{Lucas, B.D.}, \bibinfo{author}{Kanade, T.},
  \bibinfo{year}{1981}.
\newblock \bibinfo{title}{An iterative image registration technique with an
  application to stereo vision}, in: \bibinfo{booktitle}{Proceedings of the 7th
  International Joint Conference on {{Artificial}} Intelligence - {{Volume}}
  2}, \bibinfo{publisher}{{Morgan Kaufmann Publishers Inc.}},
  \bibinfo{address}{{San Francisco, CA, USA}}. pp. \bibinfo{pages}{674--679}.
\bibitem[{Luiten et~al.(2023)Luiten, Kopanas, Leibe and
  Ramanan}]{luitenDynamic3DGaussians2023a}
\bibinfo{author}{Luiten, J.}, \bibinfo{author}{Kopanas, G.},
  \bibinfo{author}{Leibe, B.}, \bibinfo{author}{Ramanan, D.},
  \bibinfo{year}{2023}.
\newblock \bibinfo{title}{Dynamic {{3D Gaussians}}: {{Tracking}} by
  {{Persistent Dynamic View Synthesis}}}.
\newblock \href{http://arxiv.org/abs/2308.09713}{\tt arXiv:2308.09713}.
\bibitem[{Lukezic et~al.(2017)Lukezic, Vojir, Zajc, Matas and
  Kristan}]{lukezicDiscriminativeCorrelationFilter2017}
\bibinfo{author}{Lukezic, A.}, \bibinfo{author}{Vojir, T.},
  \bibinfo{author}{Zajc, L.C.}, \bibinfo{author}{Matas, J.},
  \bibinfo{author}{Kristan, M.}, \bibinfo{year}{2017}.
\newblock \bibinfo{title}{Discriminative {{Correlation Filter}} with
  {{Channel}} and {{Spatial Reliability}}}, in: \bibinfo{booktitle}{2017 {{IEEE
  Conference}} on {{Computer Vision}} and {{Pattern Recognition}} ({{CVPR}})},
  \bibinfo{publisher}{{IEEE}}, \bibinfo{address}{{Honolulu, HI}}. pp.
  \bibinfo{pages}{4847--4856}.
\newblock \DOIprefix\doi{10.1109/CVPR.2017.515}.
\bibitem[{Luo et~al.(2019)Luo, Hu and
  Jia}]{luoDetailsPreservedUnsupervised2019}
\bibinfo{author}{Luo, H.}, \bibinfo{author}{Hu, Q.}, \bibinfo{author}{Jia, F.},
  \bibinfo{year}{2019}.
\newblock \bibinfo{title}{Details preserved unsupervised depth estimation by
  fusing traditional stereo knowledge from laparoscopic images}.
\newblock \bibinfo{journal}{Healthc Technol Lett} \bibinfo{volume}{6},
  \bibinfo{pages}{154--158}.
\newblock \DOIprefix\doi{10.1049/htl.2019.0063}.
\bibitem[{Luo et~al.(2022)Luo, Wang, Duan, Liu, Wang, Hu and
  Jia}]{luoUnsupervisedLearningDepth2022}
\bibinfo{author}{Luo, H.}, \bibinfo{author}{Wang, C.}, \bibinfo{author}{Duan,
  X.}, \bibinfo{author}{Liu, H.}, \bibinfo{author}{Wang, P.},
  \bibinfo{author}{Hu, Q.}, \bibinfo{author}{Jia, F.}, \bibinfo{year}{2022}.
\newblock \bibinfo{title}{Unsupervised learning of depth estimation from
  imperfect rectified stereo laparoscopic images}.
\newblock \bibinfo{journal}{Computers in Biology and Medicine}
  \bibinfo{volume}{140}.
\newblock \DOIprefix\doi{10.1016/j.compbiomed.2021.105109}.
\bibitem[{Ma et~al.(2020)Ma, Cui, Chen, Ma, Xin and
  Liao}]{maKneeArthroscopicNavigation2020}
\bibinfo{author}{Ma, C.}, \bibinfo{author}{Cui, X.}, \bibinfo{author}{Chen,
  F.}, \bibinfo{author}{Ma, L.}, \bibinfo{author}{Xin, S.},
  \bibinfo{author}{Liao, H.}, \bibinfo{year}{2020}.
\newblock \bibinfo{title}{Knee arthroscopic navigation using virtual-vision
  rendering and self-positioning technology}.
\newblock \bibinfo{journal}{International Journal of Computer Assisted
  Radiology and Surgery} \bibinfo{volume}{15}, \bibinfo{pages}{467--477}.
\newblock \DOIprefix\doi{10.1007/s11548-019-02099-6}.
\bibitem[{Ma et~al.(2019)Ma, Wang, Pizer, Rosenman, McGill and
  Frahm}]{maRealTime3DReconstruction2019}
\bibinfo{author}{Ma, R.}, \bibinfo{author}{Wang, R.}, \bibinfo{author}{Pizer,
  S.}, \bibinfo{author}{Rosenman, J.}, \bibinfo{author}{McGill, S.},
  \bibinfo{author}{Frahm, J.M.}, \bibinfo{year}{2019}.
\newblock \bibinfo{title}{Real-{{Time 3D Reconstruction}} of {{Colonoscopic
  Surfaces}} for {{Determining Missing Regions}}} \bibinfo{volume}{11768 LNCS},
  \bibinfo{pages}{582}.
\newblock \DOIprefix\doi{10.1007/978-3-030-32254-0_64}.
\bibitem[{Ma et~al.(2021)Ma, Wang, Zhang, Pizer, McGill, Rosenman and
  Frahm}]{maRNNSLAMReconstructing3D2021}
\bibinfo{author}{Ma, R.}, \bibinfo{author}{Wang, R.}, \bibinfo{author}{Zhang,
  Y.}, \bibinfo{author}{Pizer, S.}, \bibinfo{author}{McGill, S.},
  \bibinfo{author}{Rosenman, J.}, \bibinfo{author}{Frahm, J.M.},
  \bibinfo{year}{2021}.
\newblock \bibinfo{title}{{{RNNSLAM}}: {{Reconstructing}} the {{3D}} colon to
  visualize missing regions during a colonoscopy}.
\newblock \bibinfo{journal}{Medical Image Analysis} \bibinfo{volume}{72}.
\newblock \DOIprefix\doi{10.1016/j.media.2021.102100}.
\bibitem[{Mahmoud et~al.(2017)Mahmoud, Cirauqui, Hostettler, Doignon, Soler,
  Marescaux and Montiel}]{mahmoudORBSLAMbasedEndoscopeTracking2017}
\bibinfo{author}{Mahmoud, N.}, \bibinfo{author}{Cirauqui, I.},
  \bibinfo{author}{Hostettler, A.}, \bibinfo{author}{Doignon, C.},
  \bibinfo{author}{Soler, L.}, \bibinfo{author}{Marescaux, J.},
  \bibinfo{author}{Montiel, J.}, \bibinfo{year}{2017}.
\newblock \bibinfo{title}{{{ORBSLAM-based}} endoscope tracking and 3d
  reconstruction} \bibinfo{volume}{10170 LNCS}, \bibinfo{pages}{83}.
\newblock \DOIprefix\doi{10.1007/978-3-319-54057-3_7}.
\bibitem[{Mahmoud et~al.(2019)Mahmoud, Collins, Hostettler, Soler, Doignon and
  Montiel}]{mahmoudLiveTrackingDense2019}
\bibinfo{author}{Mahmoud, N.}, \bibinfo{author}{Collins, T.},
  \bibinfo{author}{Hostettler, A.}, \bibinfo{author}{Soler, L.},
  \bibinfo{author}{Doignon, C.}, \bibinfo{author}{Montiel, J.},
  \bibinfo{year}{2019}.
\newblock \bibinfo{title}{Live tracking and dense reconstruction for handheld
  monocular endoscopy}.
\newblock \bibinfo{journal}{IEEE Transactions on Medical Imaging}
  \bibinfo{volume}{38}, \bibinfo{pages}{79--89}.
\newblock \DOIprefix\doi{10.1109/TMI.2018.2856109}.
\bibitem[{{Maier-Hein} et~al.(2022){Maier-Hein}, Eisenmann, Sarikaya, M{\"a}rz,
  Collins, Malpani, Fallert, Feussner, Giannarou, Mascagni, Nakawala, Park,
  Pugh, Stoyanov, Vedula, Cleary, Fichtinger, Forestier, Gibaud, Grantcharov,
  Hashizume, {Heckmann-N{\"o}tzel}, Kenngott, Kikinis, M{\"u}ndermann, Navab,
  Onogur, Ro{\ss}, Sznitman, Taylor, Tizabi, Wagner, Hager, Neumuth, Padoy,
  Collins, Gockel, Goedeke, Hashimoto, Joyeux, Lam, Leff, Madani, Marcus,
  Meireles, Seitel, Teber, {\"U}ckert, {M{\"u}ller-Stich}, Jannin and
  Speidel}]{maier-heinSurgicalDataScience2022}
\bibinfo{author}{{Maier-Hein}, L.}, \bibinfo{author}{Eisenmann, M.},
  \bibinfo{author}{Sarikaya, D.}, \bibinfo{author}{M{\"a}rz, K.},
  \bibinfo{author}{Collins, T.}, \bibinfo{author}{Malpani, A.},
  \bibinfo{author}{Fallert, J.}, \bibinfo{author}{Feussner, H.},
  \bibinfo{author}{Giannarou, S.}, \bibinfo{author}{Mascagni, P.},
  \bibinfo{author}{Nakawala, H.}, \bibinfo{author}{Park, A.},
  \bibinfo{author}{Pugh, C.}, \bibinfo{author}{Stoyanov, D.},
  \bibinfo{author}{Vedula, S.S.}, \bibinfo{author}{Cleary, K.},
  \bibinfo{author}{Fichtinger, G.}, \bibinfo{author}{Forestier, G.},
  \bibinfo{author}{Gibaud, B.}, \bibinfo{author}{Grantcharov, T.},
  \bibinfo{author}{Hashizume, M.}, \bibinfo{author}{{Heckmann-N{\"o}tzel}, D.},
  \bibinfo{author}{Kenngott, H.G.}, \bibinfo{author}{Kikinis, R.},
  \bibinfo{author}{M{\"u}ndermann, L.}, \bibinfo{author}{Navab, N.},
  \bibinfo{author}{Onogur, S.}, \bibinfo{author}{Ro{\ss}, T.},
  \bibinfo{author}{Sznitman, R.}, \bibinfo{author}{Taylor, R.H.},
  \bibinfo{author}{Tizabi, M.D.}, \bibinfo{author}{Wagner, M.},
  \bibinfo{author}{Hager, G.D.}, \bibinfo{author}{Neumuth, T.},
  \bibinfo{author}{Padoy, N.}, \bibinfo{author}{Collins, J.},
  \bibinfo{author}{Gockel, I.}, \bibinfo{author}{Goedeke, J.},
  \bibinfo{author}{Hashimoto, D.A.}, \bibinfo{author}{Joyeux, L.},
  \bibinfo{author}{Lam, K.}, \bibinfo{author}{Leff, D.R.},
  \bibinfo{author}{Madani, A.}, \bibinfo{author}{Marcus, H.J.},
  \bibinfo{author}{Meireles, O.}, \bibinfo{author}{Seitel, A.},
  \bibinfo{author}{Teber, D.}, \bibinfo{author}{{\"U}ckert, F.},
  \bibinfo{author}{{M{\"u}ller-Stich}, B.P.}, \bibinfo{author}{Jannin, P.},
  \bibinfo{author}{Speidel, S.}, \bibinfo{year}{2022}.
\newblock \bibinfo{title}{Surgical data science {\textendash} from concepts
  toward clinical translation}.
\newblock \bibinfo{journal}{Medical Image Analysis} \bibinfo{volume}{76},
  \bibinfo{pages}{102306}.
\newblock \DOIprefix\doi{10.1016/j.media.2021.102306}.
\bibitem[{{Maier-Hein} et~al.(2014){Maier-Hein}, Groch, Bartoli, Bodenstedt,
  Boissonnat, Chang, Clancy, Elson, Haase, Heim, Hornegger, Jannin, Kenngott,
  Kilgus, {Muller-Stich}, Oladokun, Rohl, Dos~Santos, Schlemmer, Seitel,
  Speidel, Wagner and Stoyanov}]{maier-heinComparativeValidationSingleshot2014}
\bibinfo{author}{{Maier-Hein}, L.}, \bibinfo{author}{Groch, A.},
  \bibinfo{author}{Bartoli, A.}, \bibinfo{author}{Bodenstedt, S.},
  \bibinfo{author}{Boissonnat, G.}, \bibinfo{author}{Chang, P.L.},
  \bibinfo{author}{Clancy, N.}, \bibinfo{author}{Elson, D.},
  \bibinfo{author}{Haase, S.}, \bibinfo{author}{Heim, E.},
  \bibinfo{author}{Hornegger, J.}, \bibinfo{author}{Jannin, P.},
  \bibinfo{author}{Kenngott, H.}, \bibinfo{author}{Kilgus, T.},
  \bibinfo{author}{{Muller-Stich}, B.}, \bibinfo{author}{Oladokun, D.},
  \bibinfo{author}{Rohl, S.}, \bibinfo{author}{Dos~Santos, T.},
  \bibinfo{author}{Schlemmer, H.P.}, \bibinfo{author}{Seitel, A.},
  \bibinfo{author}{Speidel, S.}, \bibinfo{author}{Wagner, M.},
  \bibinfo{author}{Stoyanov, D.}, \bibinfo{year}{2014}.
\newblock \bibinfo{title}{Comparative validation of single-shot optical
  techniques for laparoscopic 3-d surface reconstruction}.
\newblock \bibinfo{journal}{IEEE Transactions on Medical Imaging}
  \bibinfo{volume}{33}, \bibinfo{pages}{1913--1930}.
\newblock \DOIprefix\doi{10.1109/TMI.2014.2325607}.
\bibitem[{{Maier-Hein} et~al.(2015){Maier-Hein}, Kondermann, Ro{\ss}, Mersmann,
  Heim, Bodenstedt, Kenngott, Sanchez, Wagner, Preukschas, Wekerle, Helfert,
  M{\"a}rz, Mehrabi, Speidel and Stock}]{maier-heinCrowdtruthValidationNew2015}
\bibinfo{author}{{Maier-Hein}, L.}, \bibinfo{author}{Kondermann, D.},
  \bibinfo{author}{Ro{\ss}, T.}, \bibinfo{author}{Mersmann, S.},
  \bibinfo{author}{Heim, E.}, \bibinfo{author}{Bodenstedt, S.},
  \bibinfo{author}{Kenngott, H.}, \bibinfo{author}{Sanchez, A.},
  \bibinfo{author}{Wagner, M.}, \bibinfo{author}{Preukschas, A.},
  \bibinfo{author}{Wekerle, A.L.}, \bibinfo{author}{Helfert, S.},
  \bibinfo{author}{M{\"a}rz, K.}, \bibinfo{author}{Mehrabi, A.},
  \bibinfo{author}{Speidel, S.}, \bibinfo{author}{Stock, C.},
  \bibinfo{year}{2015}.
\newblock \bibinfo{title}{Crowdtruth validation: A new paradigm for validating
  algorithms that rely on image correspondences}.
\newblock \bibinfo{journal}{International Journal of Computer Assisted
  Radiology and Surgery} \bibinfo{volume}{10}, \bibinfo{pages}{1201--1212}.
\newblock \DOIprefix\doi{10.1007/s11548-015-1168-3}.
\bibitem[{{Maier-Hein} et~al.(2013){Maier-Hein}, Mountney, Bartoli, Elhawary,
  Elson, Groch, Kolb, Rodrigues, Sorger, Speidel and
  Stoyanov}]{maier-heinOpticalTechniques3D2013}
\bibinfo{author}{{Maier-Hein}, L.}, \bibinfo{author}{Mountney, P.},
  \bibinfo{author}{Bartoli, A.}, \bibinfo{author}{Elhawary, H.},
  \bibinfo{author}{Elson, D.}, \bibinfo{author}{Groch, A.},
  \bibinfo{author}{Kolb, A.}, \bibinfo{author}{Rodrigues, M.},
  \bibinfo{author}{Sorger, J.}, \bibinfo{author}{Speidel, S.},
  \bibinfo{author}{Stoyanov, D.}, \bibinfo{year}{2013}.
\newblock \bibinfo{title}{Optical techniques for {{3D}} surface reconstruction
  in computer-assisted laparoscopic surgery}.
\newblock \bibinfo{journal}{Medical Image Analysis} \bibinfo{volume}{17},
  \bibinfo{pages}{974--996}.
\newblock \DOIprefix\doi{10.1016/j.media.2013.04.003}.
\bibitem[{{Maier-Hein} et~al.(2023){Maier-Hein}, Reinke, Godau, Tizabi,
  B{\"u}ttner, Christodoulou, Glocker, Isensee, Kleesiek, Kozubek, Reyes,
  Riegler, Wiesenfarth, Kavur, Sudre, Baumgartner, Eisenmann,
  {Heckmann-N{\"o}tzel}, R{\"a}dsch, Acion, Antonelli, Arbel, Bakas, Benis,
  Blaschko, Cardoso, Cheplygina, Cimini, Collins, Farahani, Ferrer, Galdran,
  {van Ginneken}, Haase, Hashimoto, Hoffman, Huisman, Jannin, Kahn,
  Kainmueller, Kainz, Karargyris, Karthikesalingam, Kenngott, Kofler,
  {Kopp-Schneider}, Kreshuk, Kurc, Landman, Litjens, Madani, {Maier-Hein},
  Martel, Mattson, Meijering, Menze, Moons, M{\"u}ller, Nichyporuk, Nickel,
  Petersen, Rajpoot, Rieke, {Saez-Rodriguez}, S{\'a}nchez, Shetty, {van
  Smeden}, Summers, Taha, Tiulpin, Tsaftaris, Van~Calster, Varoquaux and
  J{\"a}ger}]{maier-heinMetricsReloadedRecommendations2023}
\bibinfo{author}{{Maier-Hein}, L.}, \bibinfo{author}{Reinke, A.},
  \bibinfo{author}{Godau, P.}, \bibinfo{author}{Tizabi, M.D.},
  \bibinfo{author}{B{\"u}ttner, F.}, \bibinfo{author}{Christodoulou, E.},
  \bibinfo{author}{Glocker, B.}, \bibinfo{author}{Isensee, F.},
  \bibinfo{author}{Kleesiek, J.}, \bibinfo{author}{Kozubek, M.},
  \bibinfo{author}{Reyes, M.}, \bibinfo{author}{Riegler, M.A.},
  \bibinfo{author}{Wiesenfarth, M.}, \bibinfo{author}{Kavur, A.E.},
  \bibinfo{author}{Sudre, C.H.}, \bibinfo{author}{Baumgartner, M.},
  \bibinfo{author}{Eisenmann, M.}, \bibinfo{author}{{Heckmann-N{\"o}tzel}, D.},
  \bibinfo{author}{R{\"a}dsch, A.T.}, \bibinfo{author}{Acion, L.},
  \bibinfo{author}{Antonelli, M.}, \bibinfo{author}{Arbel, T.},
  \bibinfo{author}{Bakas, S.}, \bibinfo{author}{Benis, A.},
  \bibinfo{author}{Blaschko, M.}, \bibinfo{author}{Cardoso, M.J.},
  \bibinfo{author}{Cheplygina, V.}, \bibinfo{author}{Cimini, B.A.},
  \bibinfo{author}{Collins, G.S.}, \bibinfo{author}{Farahani, K.},
  \bibinfo{author}{Ferrer, L.}, \bibinfo{author}{Galdran, A.},
  \bibinfo{author}{{van Ginneken}, B.}, \bibinfo{author}{Haase, R.},
  \bibinfo{author}{Hashimoto, D.A.}, \bibinfo{author}{Hoffman, M.M.},
  \bibinfo{author}{Huisman, M.}, \bibinfo{author}{Jannin, P.},
  \bibinfo{author}{Kahn, C.E.}, \bibinfo{author}{Kainmueller, D.},
  \bibinfo{author}{Kainz, B.}, \bibinfo{author}{Karargyris, A.},
  \bibinfo{author}{Karthikesalingam, A.}, \bibinfo{author}{Kenngott, H.},
  \bibinfo{author}{Kofler, F.}, \bibinfo{author}{{Kopp-Schneider}, A.},
  \bibinfo{author}{Kreshuk, A.}, \bibinfo{author}{Kurc, T.},
  \bibinfo{author}{Landman, B.A.}, \bibinfo{author}{Litjens, G.},
  \bibinfo{author}{Madani, A.}, \bibinfo{author}{{Maier-Hein}, K.},
  \bibinfo{author}{Martel, A.L.}, \bibinfo{author}{Mattson, P.},
  \bibinfo{author}{Meijering, E.}, \bibinfo{author}{Menze, B.},
  \bibinfo{author}{Moons, K.G.M.}, \bibinfo{author}{M{\"u}ller, H.},
  \bibinfo{author}{Nichyporuk, B.}, \bibinfo{author}{Nickel, F.},
  \bibinfo{author}{Petersen, J.}, \bibinfo{author}{Rajpoot, N.},
  \bibinfo{author}{Rieke, N.}, \bibinfo{author}{{Saez-Rodriguez}, J.},
  \bibinfo{author}{S{\'a}nchez, C.I.}, \bibinfo{author}{Shetty, S.},
  \bibinfo{author}{{van Smeden}, M.}, \bibinfo{author}{Summers, R.M.},
  \bibinfo{author}{Taha, A.A.}, \bibinfo{author}{Tiulpin, A.},
  \bibinfo{author}{Tsaftaris, S.A.}, \bibinfo{author}{Van~Calster, B.},
  \bibinfo{author}{Varoquaux, G.}, \bibinfo{author}{J{\"a}ger, P.F.},
  \bibinfo{year}{2023}.
\newblock \bibinfo{title}{Metrics reloaded: {{Recommendations}} for image
  analysis validation}.
\newblock \href{http://arxiv.org/abs/2206.01653}{\tt arXiv:2206.01653}.
\bibitem[{Makki et~al.(2023)Makki, Chandelon and
  Bartoli}]{makkiEllipticalSpecularityDetection2023}
\bibinfo{author}{Makki, K.}, \bibinfo{author}{Chandelon, K.},
  \bibinfo{author}{Bartoli, A.}, \bibinfo{year}{2023}.
\newblock \bibinfo{title}{Elliptical specularity detection in endoscopy with
  application to normal reconstruction}.
\newblock \bibinfo{journal}{International Journal of Computer Assisted
  Radiology and Surgery} \DOIprefix\doi{10.1007/s11548-023-02904-3}.
\bibitem[{Malhotra et~al.(2023)Malhotra, Halabi, Dakua, Padhan, Paul and
  Palliyali}]{malhotraAugmentedRealitySurgical2023}
\bibinfo{author}{Malhotra, S.}, \bibinfo{author}{Halabi, O.},
  \bibinfo{author}{Dakua, S.P.}, \bibinfo{author}{Padhan, J.},
  \bibinfo{author}{Paul, S.}, \bibinfo{author}{Palliyali, W.},
  \bibinfo{year}{2023}.
\newblock \bibinfo{title}{Augmented {{Reality}} in {{Surgical Navigation}}: {{A
  Review}} of {{Evaluation}} and {{Validation Metrics}}}.
\newblock \bibinfo{journal}{Applied Sciences} \bibinfo{volume}{13},
  \bibinfo{pages}{1629}.
\newblock \DOIprefix\doi{10.3390/app13031629}.
\bibitem[{Malti and Bartoli(2014)}]{maltiCombiningConformalDeformation2014}
\bibinfo{author}{Malti, A.}, \bibinfo{author}{Bartoli, A.},
  \bibinfo{year}{2014}.
\newblock \bibinfo{title}{Combining conformal deformation and cook-torrance
  shading for 3-{{D}} reconstruction in laparoscopy}.
\newblock \bibinfo{journal}{IEEE Transactions on Biomedical Engineering}
  \bibinfo{volume}{61}, \bibinfo{pages}{1684--1692}.
\newblock \DOIprefix\doi{10.1109/TBME.2014.2300237}.
\bibitem[{Malti et~al.(2012)Malti, Bartoli and
  Collins}]{maltiTemplatebasedConformalShapefrommotionandshading2012}
\bibinfo{author}{Malti, A.}, \bibinfo{author}{Bartoli, A.},
  \bibinfo{author}{Collins, T.}, \bibinfo{year}{2012}.
\newblock \bibinfo{title}{Template-Based Conformal
  Shape-from-Motion-and-Shading for Laparoscopy}. volume \bibinfo{volume}{7330
  LNCS}.
\newblock \DOIprefix\doi{10.1007/978-3-642-30618-1_1}.
\bibitem[{Marmol et~al.(2019)Marmol, Banach and
  Peynot}]{marmolDenseArthroSLAMDenseIntraArticular2019}
\bibinfo{author}{Marmol, A.}, \bibinfo{author}{Banach, A.},
  \bibinfo{author}{Peynot, T.}, \bibinfo{year}{2019}.
\newblock \bibinfo{title}{Dense-{{ArthroSLAM}}: {{Dense Intra-Articular}} 3-{{D
  Reconstruction With Robust Localization Prior}} for {{Arthroscopy}}}.
\newblock \bibinfo{journal}{IEEE Robotics and Automation Letters}
  \bibinfo{volume}{4}, \bibinfo{pages}{918--925}.
\newblock \DOIprefix\doi{10.1109/LRA.2019.2892199}.
\bibitem[{Marmol et~al.(2018)Marmol, Corke and
  Peynot}]{marmolArthroSLAMMultisensorRobust2018}
\bibinfo{author}{Marmol, A.}, \bibinfo{author}{Corke, P.},
  \bibinfo{author}{Peynot, T.}, \bibinfo{year}{2018}.
\newblock \bibinfo{title}{{{ArthroSLAM}}: {{Multi-sensor}} robust visual
  localization for minimally invasive orthopedic surgery}, pp.
  \bibinfo{pages}{3882--3889}.
\newblock \DOIprefix\doi{10.1109/IROS.2018.8593501}.
\bibitem[{Marmol et~al.(2017)Marmol, Peynot, Eriksson, Jaiprakash, Roberts and
  Crawford}]{marmolEvaluationKeypointDetectors2017}
\bibinfo{author}{Marmol, A.}, \bibinfo{author}{Peynot, T.},
  \bibinfo{author}{Eriksson, A.}, \bibinfo{author}{Jaiprakash, A.},
  \bibinfo{author}{Roberts, J.}, \bibinfo{author}{Crawford, R.},
  \bibinfo{year}{2017}.
\newblock \bibinfo{title}{Evaluation of {{Keypoint Detectors}} and
  {{Descriptors}} in {{Arthroscopic Images}} for {{Feature-Based Matching
  Applications}}}.
\newblock \bibinfo{journal}{IEEE Robotics and Automation Letters}
  \bibinfo{volume}{2}, \bibinfo{pages}{2135--2142}.
\newblock \DOIprefix\doi{10.1109/LRA.2017.2714150}.
\bibitem[{Martin et~al.(2023)Martin, El~Hage, Shedid and
  Bojanowski}]{martinUsingArtificialIntelligence2023}
\bibinfo{author}{Martin, T.}, \bibinfo{author}{El~Hage, G.},
  \bibinfo{author}{Shedid, D.}, \bibinfo{author}{Bojanowski, M.},
  \bibinfo{year}{2023}.
\newblock \bibinfo{title}{Using artificial intelligence to quantify dynamic
  retraction of brain tissue and the manipulation of instruments in
  neurosurgery}.
\newblock \bibinfo{journal}{International Journal of Computer Assisted
  Radiology and Surgery} \DOIprefix\doi{10.1007/s11548-022-02824-8}.
\bibitem[{Mildenhall et~al.(2022)Mildenhall, Srinivasan, Tancik, Barron,
  Ramamoorthi and Ng}]{mildenhallNeRFRepresentingScenes2022}
\bibinfo{author}{Mildenhall, B.}, \bibinfo{author}{Srinivasan, P.P.},
  \bibinfo{author}{Tancik, M.}, \bibinfo{author}{Barron, J.T.},
  \bibinfo{author}{Ramamoorthi, R.}, \bibinfo{author}{Ng, R.},
  \bibinfo{year}{2022}.
\newblock \bibinfo{title}{{{NeRF}}: Representing scenes as neural radiance
  fields for view synthesis}.
\newblock \bibinfo{journal}{Commun. ACM} \bibinfo{volume}{65},
  \bibinfo{pages}{99--106}.
\newblock \DOIprefix\doi{10.1145/3503250}.
\bibitem[{{Miranda-Luna} et~al.(2008){Miranda-Luna}, Daul, Blondel,
  {Hernandez-Mier}, Wolf and
  Guillemin}]{miranda-lunaMosaicingBladderEndoscopic2008a}
\bibinfo{author}{{Miranda-Luna}, R.}, \bibinfo{author}{Daul, C.},
  \bibinfo{author}{Blondel, W.}, \bibinfo{author}{{Hernandez-Mier}, Y.},
  \bibinfo{author}{Wolf, D.}, \bibinfo{author}{Guillemin, F.},
  \bibinfo{year}{2008}.
\newblock \bibinfo{title}{Mosaicing of bladder endoscopic image sequences:
  {{Distortion}} calibration and registration algorithm}.
\newblock \bibinfo{journal}{IEEE Transactions on Biomedical Engineering}
  \bibinfo{volume}{55}, \bibinfo{pages}{541--553}.
\newblock \DOIprefix\doi{10.1109/TBME.2007.903520}.
\bibitem[{Moing et~al.(2023)Moing, Ponce and
  Schmid}]{moingDenseOpticalTracking2023}
\bibinfo{author}{Moing, G.L.}, \bibinfo{author}{Ponce, J.},
  \bibinfo{author}{Schmid, C.}, \bibinfo{year}{2023}.
\newblock \bibinfo{title}{Dense {{Optical Tracking}}: {{Connecting}} the
  {{Dots}}}.
\bibitem[{Mountney et~al.(2006)Mountney, Stoyanov, Davison and
  Yang}]{mountneySimultaneousStereoscopeLocalization2006}
\bibinfo{author}{Mountney, P.}, \bibinfo{author}{Stoyanov, D.},
  \bibinfo{author}{Davison, A.}, \bibinfo{author}{Yang, G.Z.},
  \bibinfo{year}{2006}.
\newblock \bibinfo{title}{Simultaneous Stereoscope Localization and Soft-Tissue
  Mapping for Minimal Invasive Surgery}. volume \bibinfo{volume}{4190 LNCS -
  I}.
\newblock \DOIprefix\doi{10.1007/11866565_43}.
\bibitem[{Mountney et~al.(2010)Mountney, Stoyanov and
  Yang}]{mountneyThreedimensionalTissueDeformation2010}
\bibinfo{author}{Mountney, P.}, \bibinfo{author}{Stoyanov, D.},
  \bibinfo{author}{Yang, G.Z.}, \bibinfo{year}{2010}.
\newblock \bibinfo{title}{Three-dimensional tissue deformation recovery and
  tracking}.
\newblock \bibinfo{journal}{IEEE Signal Processing Magazine}
  \bibinfo{volume}{27}, \bibinfo{pages}{14--24}.
\newblock \DOIprefix\doi{10.1109/MSP.2010.936728}.
\bibitem[{Mountney and Yang(2008)}]{mountneySoftTissueTracking2008}
\bibinfo{author}{Mountney, P.}, \bibinfo{author}{Yang, G.Z.},
  \bibinfo{year}{2008}.
\newblock \bibinfo{title}{Soft Tissue Tracking for Minimally Invasive Surgery:
  {{Learning}} Local Deformation Online}. volume \bibinfo{volume}{5242 LNCS}.
\newblock \DOIprefix\doi{10.1007/978-3-540-85990-1_44}.
\bibitem[{Mountney and Yang(2010)}]{mountneyMotionCompensatedSLAM2010}
\bibinfo{author}{Mountney, P.}, \bibinfo{author}{Yang, G.Z.},
  \bibinfo{year}{2010}.
\newblock \bibinfo{title}{Motion compensated {{SLAM}} for image guided
  surgery}, p. \bibinfo{pages}{504}.
\newblock \DOIprefix\doi{10.1007/978-3-642-15745-5_61}.
\bibitem[{M{\"u}ller et~al.(2022)M{\"u}ller, Evans, Schied and
  Keller}]{mullerInstantNeuralGraphics2022b}
\bibinfo{author}{M{\"u}ller, T.}, \bibinfo{author}{Evans, A.},
  \bibinfo{author}{Schied, C.}, \bibinfo{author}{Keller, A.},
  \bibinfo{year}{2022}.
\newblock \bibinfo{title}{Instant neural graphics primitives with a
  multiresolution hash encoding}.
\newblock \bibinfo{journal}{ACM Trans. Graph.} \bibinfo{volume}{41},
  \bibinfo{pages}{1--15}.
\newblock \DOIprefix\doi{10.1145/3528223.3530127}.
\bibitem[{M{\"u}nzer et~al.(2018)M{\"u}nzer, Schoeffmann and
  B{\"o}sz{\"o}rmenyi}]{munzerContentbasedProcessingAnalysis2018}
\bibinfo{author}{M{\"u}nzer, B.}, \bibinfo{author}{Schoeffmann, K.},
  \bibinfo{author}{B{\"o}sz{\"o}rmenyi, L.}, \bibinfo{year}{2018}.
\newblock \bibinfo{title}{Content-based processing and analysis of endoscopic
  images and videos: {{A}} survey}.
\newblock \bibinfo{journal}{Multimed Tools Appl} \bibinfo{volume}{77},
  \bibinfo{pages}{1323--1362}.
\newblock \DOIprefix\doi{10.1007/s11042-016-4219-z}.
\bibitem[{{Mur-Artal} et~al.(2015){Mur-Artal}, Montiel and
  Tardos}]{mur-artalORBSLAMVersatileAccurate2015}
\bibinfo{author}{{Mur-Artal}, R.}, \bibinfo{author}{Montiel, J.M.M.},
  \bibinfo{author}{Tardos, J.D.}, \bibinfo{year}{2015}.
\newblock \bibinfo{title}{{{ORB-SLAM}}: {{A Versatile}} and {{Accurate
  Monocular SLAM System}}}.
\newblock \bibinfo{journal}{IEEE Transactions on Robotics}
  \bibinfo{volume}{31}, \bibinfo{pages}{1147--1163}.
\newblock \DOIprefix\doi{10.1109/TRO.2015.2463671}.
\bibitem[{Neoral et~al.(2024)Neoral, {\v S}er{\'y}ch and
  Matas}]{neoralMFTLongTermTracking2024}
\bibinfo{author}{Neoral, M.}, \bibinfo{author}{{\v S}er{\'y}ch, J.},
  \bibinfo{author}{Matas, J.}, \bibinfo{year}{2024}.
\newblock \bibinfo{title}{{{MFT}}: {{Long-Term Tracking}} of {{Every Pixel}}},
  in: \bibinfo{booktitle}{Proceedings of the {{IEEE}}/{{CVF Winter Conference}}
  on {{Applications}} of {{Computer Vision}}}, pp. \bibinfo{pages}{6837--6847}.
\bibitem[{Oliva~Maza et~al.(2023)Oliva~Maza, Steidle, Klodmann, Strobl and
  Triebel}]{olivamazaORBSLAM3basedApproachSurgical2023}
\bibinfo{author}{Oliva~Maza, L.}, \bibinfo{author}{Steidle, F.},
  \bibinfo{author}{Klodmann, J.}, \bibinfo{author}{Strobl, K.},
  \bibinfo{author}{Triebel, R.}, \bibinfo{year}{2023}.
\newblock \bibinfo{title}{An {{ORB-SLAM3-based Approach}} for {{Surgical
  Navigation}} in {{Ureteroscopy}}}.
\newblock \bibinfo{journal}{Computer Methods in Biomechanics and Biomedical
  Engineering: Imaging and Visualization} \bibinfo{volume}{11},
  \bibinfo{pages}{1005--1011}.
\newblock \DOIprefix\doi{10.1080/21681163.2022.2156392}.
\bibitem[{Oquab et~al.(2023)Oquab, Darcet, Moutakanni, Vo, Szafraniec,
  Khalidov, Fernandez, Haziza, Massa, {El-Nouby}, Assran, Ballas, Galuba,
  Howes, Huang, Li, Misra, Rabbat, Sharma, Synnaeve, Xu, Jegou, Mairal,
  Labatut, Joulin and Bojanowski}]{oquabDINOv2LearningRobust2023}
\bibinfo{author}{Oquab, M.}, \bibinfo{author}{Darcet, T.},
  \bibinfo{author}{Moutakanni, T.}, \bibinfo{author}{Vo, H.},
  \bibinfo{author}{Szafraniec, M.}, \bibinfo{author}{Khalidov, V.},
  \bibinfo{author}{Fernandez, P.}, \bibinfo{author}{Haziza, D.},
  \bibinfo{author}{Massa, F.}, \bibinfo{author}{{El-Nouby}, A.},
  \bibinfo{author}{Assran, M.}, \bibinfo{author}{Ballas, N.},
  \bibinfo{author}{Galuba, W.}, \bibinfo{author}{Howes, R.},
  \bibinfo{author}{Huang, P.Y.}, \bibinfo{author}{Li, S.W.},
  \bibinfo{author}{Misra, I.}, \bibinfo{author}{Rabbat, M.},
  \bibinfo{author}{Sharma, V.}, \bibinfo{author}{Synnaeve, G.},
  \bibinfo{author}{Xu, H.}, \bibinfo{author}{Jegou, H.},
  \bibinfo{author}{Mairal, J.}, \bibinfo{author}{Labatut, P.},
  \bibinfo{author}{Joulin, A.}, \bibinfo{author}{Bojanowski, P.},
  \bibinfo{year}{2023}.
\newblock \bibinfo{title}{{{DINOv2}}: {{Learning Robust Visual Features}}
  without {{Supervision}}}.
\newblock \href{http://arxiv.org/abs/2304.07193}{\tt arXiv:2304.07193}.
\bibitem[{Ozyoruk et~al.(2021)Ozyoruk, Gokceler, Bobrow, Coskun, Incetan,
  Almalioglu, Mahmood, Curto, Perdigoto, Oliveira, Sahin, Araujo, Alexandrino,
  Durr, Gilbert and Turan}]{ozyorukEndoSLAMDatasetUnsupervised2021}
\bibinfo{author}{Ozyoruk, K.}, \bibinfo{author}{Gokceler, G.},
  \bibinfo{author}{Bobrow, T.}, \bibinfo{author}{Coskun, G.},
  \bibinfo{author}{Incetan, K.}, \bibinfo{author}{Almalioglu, Y.},
  \bibinfo{author}{Mahmood, F.}, \bibinfo{author}{Curto, E.},
  \bibinfo{author}{Perdigoto, L.}, \bibinfo{author}{Oliveira, M.},
  \bibinfo{author}{Sahin, H.}, \bibinfo{author}{Araujo, H.},
  \bibinfo{author}{Alexandrino, H.}, \bibinfo{author}{Durr, N.},
  \bibinfo{author}{Gilbert, H.}, \bibinfo{author}{Turan, M.},
  \bibinfo{year}{2021}.
\newblock \bibinfo{title}{{{EndoSLAM}} dataset and an unsupervised monocular
  visual odometry and depth estimation approach for endoscopic videos}.
\newblock \bibinfo{journal}{Medical Image Analysis} \bibinfo{volume}{71}.
\newblock \DOIprefix\doi{10.1016/j.media.2021.102058}.
\bibitem[{Pan et~al.(2024)Pan, Zhong, Wiesmann, Posewsky, Behley and
  Stachniss}]{panPINSLAMLiDARSLAM2024}
\bibinfo{author}{Pan, Y.}, \bibinfo{author}{Zhong, X.},
  \bibinfo{author}{Wiesmann, L.}, \bibinfo{author}{Posewsky, T.},
  \bibinfo{author}{Behley, J.}, \bibinfo{author}{Stachniss, C.},
  \bibinfo{year}{2024}.
\newblock \bibinfo{title}{{{PIN-SLAM}}: {{LiDAR SLAM Using}} a {{Point-Based
  Implicit Neural Representation}} for {{Achieving Global Map Consistency}}}.
\newblock \href{http://arxiv.org/abs/2401.09101}{\tt arXiv:2401.09101}.
\bibitem[{Penza et~al.(2018a)Penza, Ciullo, Moccia, Mattos and
  De~Momi}]{penzaEndoAbSDatasetEndoscopic2018}
\bibinfo{author}{Penza, V.}, \bibinfo{author}{Ciullo, A.},
  \bibinfo{author}{Moccia, S.}, \bibinfo{author}{Mattos, L.},
  \bibinfo{author}{De~Momi, E.}, \bibinfo{year}{2018}a.
\newblock \bibinfo{title}{{{EndoAbS}} dataset: {{Endoscopic}} abdominal stereo
  image dataset for benchmarking {{3D}} stereo reconstruction algorithms}.
\newblock \bibinfo{journal}{International Journal of Medical Robotics and
  Computer Assisted Surgery} \bibinfo{volume}{14}.
\newblock \DOIprefix\doi{10.1002/rcs.1926}.
\bibitem[{Penza et~al.(2018b)Penza, Du, Stoyanov, Forgione, Mattos and
  De~Momi}]{penzaLongTermSafety2018}
\bibinfo{author}{Penza, V.}, \bibinfo{author}{Du, X.},
  \bibinfo{author}{Stoyanov, D.}, \bibinfo{author}{Forgione, A.},
  \bibinfo{author}{Mattos, L.}, \bibinfo{author}{De~Momi, E.},
  \bibinfo{year}{2018}b.
\newblock \bibinfo{title}{Long {{Term Safety Area Tracking}} ({{LT-SAT}}) with
  online failure detection and recovery for robotic minimally invasive
  surgery}.
\newblock \bibinfo{journal}{Medical Image Analysis} \bibinfo{volume}{45},
  \bibinfo{pages}{13--23}.
\newblock \DOIprefix\doi{10.1016/j.media.2017.12.010}.
\bibitem[{Potje et~al.(2023)Potje, Cadar, Araujo, Martins and
  Nascimento}]{potjeEnhancingDeformableLocal2023}
\bibinfo{author}{Potje, G.}, \bibinfo{author}{Cadar, F.},
  \bibinfo{author}{Araujo, A.}, \bibinfo{author}{Martins, R.},
  \bibinfo{author}{Nascimento, E.R.}, \bibinfo{year}{2023}.
\newblock \bibinfo{title}{Enhancing {{Deformable Local Features}} by {{Jointly
  Learning}} to {{Detect}} and {{Describe Keypoints}}}, in:
  \bibinfo{booktitle}{2023 {{IEEE}}/{{CVF Conference}} on {{Computer Vision}}
  and {{Pattern Recognition}} ({{CVPR}})}, \bibinfo{publisher}{{IEEE}},
  \bibinfo{address}{{Vancouver, BC, Canada}}. pp. \bibinfo{pages}{1306--1315}.
\newblock \DOIprefix\doi{10.1109/CVPR52729.2023.00132}.
\bibitem[{Pratt et~al.(2010)Pratt, Stoyanov, {Visentini-Scarzanella} and
  Yang}]{prattDynamicGuidanceRobotic2010}
\bibinfo{author}{Pratt, P.}, \bibinfo{author}{Stoyanov, D.},
  \bibinfo{author}{{Visentini-Scarzanella}, M.}, \bibinfo{author}{Yang, G.Z.},
  \bibinfo{year}{2010}.
\newblock \bibinfo{title}{Dynamic {{Guidance}} for {{Robotic Surgery Using
  Image-Constrained Biomechanical Models}}}, in: \bibinfo{editor}{Jiang, T.},
  \bibinfo{editor}{Navab, N.}, \bibinfo{editor}{Pluim, J.P.W.},
  \bibinfo{editor}{Viergever, M.A.} (Eds.), \bibinfo{booktitle}{Medical {{Image
  Computing}} and {{Computer-Assisted Intervention}} {\textendash} {{MICCAI}}
  2010}, \bibinfo{publisher}{{Springer}}, \bibinfo{address}{{Berlin,
  Heidelberg}}. pp. \bibinfo{pages}{77--85}.
\newblock \DOIprefix\doi{10.1007/978-3-642-15705-9_10}.
\bibitem[{Psychogyios et~al.(2022)Psychogyios, Mazomenos, Vasconcelos and
  Stoyanov}]{psychogyiosMSDESISMultitaskStereo2022}
\bibinfo{author}{Psychogyios, D.}, \bibinfo{author}{Mazomenos, E.},
  \bibinfo{author}{Vasconcelos, F.}, \bibinfo{author}{Stoyanov, D.},
  \bibinfo{year}{2022}.
\newblock \bibinfo{title}{{{MSDESIS}}: {{Multitask Stereo Disparity
  Estimation}} and {{Surgical Instrument Segmentation}}}.
\newblock \bibinfo{journal}{IEEE Trans. Med. Imaging} \bibinfo{volume}{41},
  \bibinfo{pages}{3218--3230}.
\newblock \DOIprefix\doi{10.1109/TMI.2022.3181229}.
\bibitem[{Qian et~al.(2020)Qian, Wu, DiMaio, Navab and
  Kazanzides}]{qianReviewAugmentedReality2020}
\bibinfo{author}{Qian, L.}, \bibinfo{author}{Wu, J.Y.},
  \bibinfo{author}{DiMaio, S.P.}, \bibinfo{author}{Navab, N.},
  \bibinfo{author}{Kazanzides, P.}, \bibinfo{year}{2020}.
\newblock \bibinfo{title}{A {{Review}} of {{Augmented Reality}} in
  {{Robotic-Assisted Surgery}}}.
\newblock \bibinfo{journal}{IEEE Transactions on Medical Robotics and Bionics}
  \bibinfo{volume}{2}, \bibinfo{pages}{1--16}.
\newblock \DOIprefix\doi{10.1109/TMRB.2019.2957061}.
\bibitem[{Raji{\v c} et~al.(2023)Raji{\v c}, Ke, Tai, Tang, Danelljan and
  Yu}]{rajicSegmentAnythingMeets2023}
\bibinfo{author}{Raji{\v c}, F.}, \bibinfo{author}{Ke, L.},
  \bibinfo{author}{Tai, Y.W.}, \bibinfo{author}{Tang, C.K.},
  \bibinfo{author}{Danelljan, M.}, \bibinfo{author}{Yu, F.},
  \bibinfo{year}{2023}.
\newblock \bibinfo{title}{Segment {{Anything Meets Point Tracking}}}.
\newblock \href{http://arxiv.org/abs/2307.01197}{\tt arXiv:2307.01197}.
\bibitem[{Rau et~al.(2022)Rau, Bhattarai, Agapito and
  Stoyanov}]{rauBimodalCameraPose2022}
\bibinfo{author}{Rau, A.}, \bibinfo{author}{Bhattarai, B.},
  \bibinfo{author}{Agapito, L.}, \bibinfo{author}{Stoyanov, D.},
  \bibinfo{year}{2022}.
\newblock \bibinfo{title}{Bimodal {{Camera Pose Prediction}} for
  {{Endoscopy}}}.
\newblock \href{http://arxiv.org/abs/2204.04968}{\tt arXiv:2204.04968}.
\bibitem[{Rau et~al.(2019)Rau, Edwards, Ahmad, Riordan, Janatka, Lovat and
  Stoyanov}]{rauImplicitDomainAdaptation2019}
\bibinfo{author}{Rau, A.}, \bibinfo{author}{Edwards, P.J.E.},
  \bibinfo{author}{Ahmad, O.F.}, \bibinfo{author}{Riordan, P.},
  \bibinfo{author}{Janatka, M.}, \bibinfo{author}{Lovat, L.B.},
  \bibinfo{author}{Stoyanov, D.}, \bibinfo{year}{2019}.
\newblock \bibinfo{title}{Implicit domain adaptation with conditional
  generative adversarial networks for depth prediction in endoscopy}.
\newblock \bibinfo{journal}{Int J CARS} \bibinfo{volume}{14},
  \bibinfo{pages}{1167--1176}.
\newblock \DOIprefix\doi{10.1007/s11548-019-01962-w}.
\bibitem[{Recasens et~al.(2021)Recasens, Lamarca, F{\'a}cil, Montiel and
  Civera}]{recasensEndoDepthandMotionReconstructionTracking2021}
\bibinfo{author}{Recasens, D.}, \bibinfo{author}{Lamarca, J.},
  \bibinfo{author}{F{\'a}cil, J.M.}, \bibinfo{author}{Montiel, J.M.M.},
  \bibinfo{author}{Civera, J.}, \bibinfo{year}{2021}.
\newblock \bibinfo{title}{Endo-{{Depth-and-Motion}}: {{Reconstruction}} and
  {{Tracking}} in {{Endoscopic Videos Using Depth Networks}} and {{Photometric
  Constraints}}}.
\newblock \bibinfo{journal}{IEEE Robotics and Automation Letters}
  \bibinfo{volume}{6}, \bibinfo{pages}{7225--7232}.
\newblock \DOIprefix\doi{10.1109/LRA.2021.3095528}.
\bibitem[{Richa et~al.(2011)Richa, B{\'o} and
  Poignet}]{richaRobust3DVisual2011}
\bibinfo{author}{Richa, R.}, \bibinfo{author}{B{\'o}, A.},
  \bibinfo{author}{Poignet, P.}, \bibinfo{year}{2011}.
\newblock \bibinfo{title}{Towards robust {{3D}} visual tracking for motion
  compensation in beating heart surgery}.
\newblock \bibinfo{journal}{Medical Image Analysis} \bibinfo{volume}{15},
  \bibinfo{pages}{302--315}.
\newblock \DOIprefix\doi{10.1016/j.media.2010.12.002}.
\bibitem[{Richa et~al.(2008)Richa, Poignet and
  Liu}]{richaEfficient3DTracking2008}
\bibinfo{author}{Richa, R.}, \bibinfo{author}{Poignet, P.},
  \bibinfo{author}{Liu, C.}, \bibinfo{year}{2008}.
\newblock \bibinfo{title}{Efficient {{3D}} Tracking for Motion Compensation in
  Beating Heart Surgery}. volume \bibinfo{volume}{5242 LNCS}.
\newblock \DOIprefix\doi{10.1007/978-3-540-85990-1_82}.
\bibitem[{Richter et~al.(2021)Richter, Shen, Liu, Huang, Funk, Orosco and
  Yip}]{richterAutonomousRoboticSuction2021}
\bibinfo{author}{Richter, F.}, \bibinfo{author}{Shen, S.},
  \bibinfo{author}{Liu, F.}, \bibinfo{author}{Huang, J.},
  \bibinfo{author}{Funk, E.K.}, \bibinfo{author}{Orosco, R.K.},
  \bibinfo{author}{Yip, M.C.}, \bibinfo{year}{2021}.
\newblock \bibinfo{title}{Autonomous {{Robotic Suction}} to {{Clear}} the
  {{Surgical Field}} for {{Hemostasis Using Image-Based Blood Flow
  Detection}}}.
\newblock \bibinfo{journal}{IEEE Robotics and Automation Letters}
  \bibinfo{volume}{6}, \bibinfo{pages}{1383--1390}.
\newblock \DOIprefix\doi{10.1109/LRA.2021.3056057}.
\bibitem[{Rublee et~al.(2011)Rublee, Rabaud, Konolige and
  Bradski}]{rubleeORBEfficientAlternative2011a}
\bibinfo{author}{Rublee, E.}, \bibinfo{author}{Rabaud, V.},
  \bibinfo{author}{Konolige, K.}, \bibinfo{author}{Bradski, G.},
  \bibinfo{year}{2011}.
\newblock \bibinfo{title}{{{ORB}}: {{An}} efficient alternative to {{SIFT}} or
  {{SURF}}}, in: \bibinfo{booktitle}{2011 {{International Conference}} on
  {{Computer Vision}}}, pp. \bibinfo{pages}{2564--2571}.
\newblock \DOIprefix\doi{10.1109/ICCV.2011.6126544}.
\bibitem[{Saha et~al.(2023)Saha, Liu, Lin, Lu and
  Yip}]{sahaBASEDBundleAdjustingSurgical2023b}
\bibinfo{author}{Saha, S.}, \bibinfo{author}{Liu, S.}, \bibinfo{author}{Lin,
  S.}, \bibinfo{author}{Lu, J.}, \bibinfo{author}{Yip, M.},
  \bibinfo{year}{2023}.
\newblock \bibinfo{title}{{{BASED}}: {{Bundle-Adjusting Surgical Endoscopic
  Dynamic Video Reconstruction}} using {{Neural Radiance Fields}}}
  \DOIprefix\doi{10.48550/ARXIV.2309.15329}.
\bibitem[{Sarlin et~al.(2020)Sarlin, DeTone, Malisiewicz and
  Rabinovich}]{sarlinSuperGlueLearningFeature2020}
\bibinfo{author}{Sarlin, P.E.}, \bibinfo{author}{DeTone, D.},
  \bibinfo{author}{Malisiewicz, T.}, \bibinfo{author}{Rabinovich, A.},
  \bibinfo{year}{2020}.
\newblock \bibinfo{title}{{{SuperGlue}}: {{Learning Feature Matching With Graph
  Neural Networks}}}, in: \bibinfo{booktitle}{{{CVPR}}}.
\newblock \href{http://arxiv.org/abs/1911.11763}{\tt arXiv:1911.11763}.
\bibitem[{Sarlin et~al.(2023)Sarlin, Lindenberger, Larsson and
  Pollefeys}]{sarlinPixelPerfectStructureFromMotionFeaturemetric2023}
\bibinfo{author}{Sarlin, P.E.}, \bibinfo{author}{Lindenberger, P.},
  \bibinfo{author}{Larsson, V.}, \bibinfo{author}{Pollefeys, M.},
  \bibinfo{year}{2023}.
\newblock \bibinfo{title}{Pixel-{{Perfect Structure-From-Motion With
  Featuremetric Refinement}}}.
\newblock \bibinfo{journal}{IEEE Transactions on Pattern Analysis and Machine
  Intelligence} ,
  \bibinfo{pages}{1--12}\DOIprefix\doi{10.1109/TPAMI.2023.3237269}.
\bibitem[{Schmidt et~al.(2022a)Schmidt, Mohareri, Dimaio and
  Salcudean}]{schmidtFastGraphRefinement2022a}
\bibinfo{author}{Schmidt, A.}, \bibinfo{author}{Mohareri, O.},
  \bibinfo{author}{Dimaio, S.}, \bibinfo{author}{Salcudean, S.},
  \bibinfo{year}{2022}a.
\newblock \bibinfo{title}{Fast {{Graph Refinement}} and {{Implicit Neural
  Representation}} for {{Tissue Tracking}}}, in:
  \bibinfo{booktitle}{Proceedings - {{IEEE International Conference}} on
  {{Robotics}} and {{Automation}}}, pp. \bibinfo{pages}{1281--1288}.
\newblock \DOIprefix\doi{10.1109/ICRA46639.2022.9811742}.
\bibitem[{Schmidt et~al.(2022b)Schmidt, Mohareri, DiMaio and
  Salcudean}]{schmidtRecurrentImplicitNeural2022}
\bibinfo{author}{Schmidt, A.}, \bibinfo{author}{Mohareri, O.},
  \bibinfo{author}{DiMaio, S.}, \bibinfo{author}{Salcudean, S.E.},
  \bibinfo{year}{2022}b.
\newblock \bibinfo{title}{Recurrent {{Implicit Neural Graph}} for~{{Deformable
  Tracking}} in~{{Endoscopic Videos}}}, in: \bibinfo{editor}{Wang, L.},
  \bibinfo{editor}{Dou, Q.}, \bibinfo{editor}{Fletcher, P.T.},
  \bibinfo{editor}{Speidel, S.}, \bibinfo{editor}{Li, S.} (Eds.),
  \bibinfo{booktitle}{Medical {{Image Computing}} and {{Computer Assisted
  Intervention}} {\textendash} {{MICCAI}} 2022}, \bibinfo{publisher}{{Springer
  Nature Switzerland}}, \bibinfo{address}{{Cham}}. pp.
  \bibinfo{pages}{478--488}.
\newblock \DOIprefix\doi{10.1007/978-3-031-16440-8_46}.
\bibitem[{Schmidt et~al.(2023a)Schmidt, Mohareri, DiMaio and
  Salcudean}]{schmidtSENDDSparseEfficient2023}
\bibinfo{author}{Schmidt, A.}, \bibinfo{author}{Mohareri, O.},
  \bibinfo{author}{DiMaio, S.}, \bibinfo{author}{Salcudean, S.E.},
  \bibinfo{year}{2023}a.
\newblock \bibinfo{title}{{{SENDD}}: {{Sparse Efficient Neural Depth}}
  and~{{Deformation}} for~{{Tissue Tracking}}}, in: \bibinfo{booktitle}{Medical
  {{Image Computing}} and {{Computer Assisted Intervention}} {\textendash}
  {{MICCAI}} 2023: 26th {{International Conference}}, {{Vancouver}}, {{BC}},
  {{Canada}}, {{October}} 8{\textendash}12, 2023, {{Proceedings}}, {{Part
  IX}}}, \bibinfo{publisher}{{Springer-Verlag}}, \bibinfo{address}{{Berlin,
  Heidelberg}}. pp. \bibinfo{pages}{238--248}.
\newblock \DOIprefix\doi{10.1007/978-3-031-43996-4_23}.
\bibitem[{Schmidt et~al.(2023b)Schmidt, Mohareri, DiMaio and
  Salcudean}]{schmidtSTIRSurgicalTattoos2023}
\bibinfo{author}{Schmidt, A.}, \bibinfo{author}{Mohareri, O.},
  \bibinfo{author}{DiMaio, S.}, \bibinfo{author}{Salcudean, S.E.},
  \bibinfo{year}{2023}b.
\newblock \bibinfo{title}{{{STIR}}: {{Surgical Tattoos}} in {{Infrared}}}.
\newblock \DOIprefix\doi{10.48550/arXiv.2309.16782},
  \href{http://arxiv.org/abs/2309.16782}{\tt arXiv:2309.16782}.
\bibitem[{Schmidt and
  Salcudean(2021)}]{schmidtRealTimeRotatedConvolutional2021a}
\bibinfo{author}{Schmidt, A.}, \bibinfo{author}{Salcudean, S.},
  \bibinfo{year}{2021}.
\newblock \bibinfo{title}{Real-{{Time Rotated Convolutional Descriptor}} for
  {{Surgical Environments}}}. volume \bibinfo{volume}{12904 LNCS}.
\newblock \DOIprefix\doi{10.1007/978-3-030-87202-1_27}.
\bibitem[{Schneider et~al.(2021)Schneider, Allam, Stoyanov, Hawkes, Gurusamy
  and Davidson}]{schneiderPerformanceImageGuided2021}
\bibinfo{author}{Schneider, C.}, \bibinfo{author}{Allam, M.},
  \bibinfo{author}{Stoyanov, D.}, \bibinfo{author}{Hawkes, D.},
  \bibinfo{author}{Gurusamy, K.}, \bibinfo{author}{Davidson, B.},
  \bibinfo{year}{2021}.
\newblock \bibinfo{title}{Performance of image guided navigation in
  laparoscopic liver surgery {\textendash} {{A}} systematic review}.
\newblock \bibinfo{journal}{Surgical Oncology} \bibinfo{volume}{38},
  \bibinfo{pages}{101637}.
\newblock \DOIprefix\doi{10.1016/j.suronc.2021.101637}.
\bibitem[{Schonberger and
  Frahm(2016)}]{schonbergerStructureFromMotionRevisited2016}
\bibinfo{author}{Schonberger, J.L.}, \bibinfo{author}{Frahm, J.M.},
  \bibinfo{year}{2016}.
\newblock \bibinfo{title}{Structure-{{From-Motion Revisited}}}, in:
  \bibinfo{booktitle}{Proceedings of the {{IEEE Conference}} on {{Computer
  Vision}} and {{Pattern Recognition}}}, pp. \bibinfo{pages}{4104--4113}.
\bibitem[{Sch{\"o}nberger et~al.(2016a)Sch{\"o}nberger, Zheng, Frahm and
  Pollefeys}]{schonbergerPixelwiseViewSelection2016a}
\bibinfo{author}{Sch{\"o}nberger, J.L.}, \bibinfo{author}{Zheng, E.},
  \bibinfo{author}{Frahm, J.M.}, \bibinfo{author}{Pollefeys, M.},
  \bibinfo{year}{2016}a.
\newblock \bibinfo{title}{Pixelwise {{View Selection}} for {{Unstructured
  Multi-View Stereo}}}, in: \bibinfo{editor}{Leibe, B.},
  \bibinfo{editor}{Matas, J.}, \bibinfo{editor}{Sebe, N.},
  \bibinfo{editor}{Welling, M.} (Eds.), \bibinfo{booktitle}{Computer {{Vision}}
  {\textendash} {{ECCV}} 2016}, \bibinfo{publisher}{{Springer International
  Publishing}}, \bibinfo{address}{{Cham}}. pp. \bibinfo{pages}{501--518}.
\newblock \DOIprefix\doi{10.1007/978-3-319-46487-9_31}.
\bibitem[{Sch{\"o}nberger et~al.(2016b)Sch{\"o}nberger, Zheng, Frahm and
  Pollefeys}]{schonbergerPixelwiseViewSelection2016}
\bibinfo{author}{Sch{\"o}nberger, J.L.}, \bibinfo{author}{Zheng, E.},
  \bibinfo{author}{Frahm, J.M.}, \bibinfo{author}{Pollefeys, M.},
  \bibinfo{year}{2016}b.
\newblock \bibinfo{title}{Pixelwise {{View Selection}} for {{Unstructured
  Multi-View Stereo}}}, in: \bibinfo{editor}{Leibe, B.},
  \bibinfo{editor}{Matas, J.}, \bibinfo{editor}{Sebe, N.},
  \bibinfo{editor}{Welling, M.} (Eds.), \bibinfo{booktitle}{Computer {{Vision}}
  {\textendash} {{ECCV}} 2016}. \bibinfo{publisher}{{Springer International
  Publishing}}, \bibinfo{address}{{Cham}}. volume \bibinfo{volume}{9907}, pp.
  \bibinfo{pages}{501--518}.
\newblock \DOIprefix\doi{10.1007/978-3-319-46487-9_31}.
\bibitem[{Schoob et~al.(2017)Schoob, Kundrat, Kahrs and
  Ortmaier}]{schoobStereoVisionbasedTracking2017}
\bibinfo{author}{Schoob, A.}, \bibinfo{author}{Kundrat, D.},
  \bibinfo{author}{Kahrs, L.}, \bibinfo{author}{Ortmaier, T.},
  \bibinfo{year}{2017}.
\newblock \bibinfo{title}{Stereo vision-based tracking of soft tissue motion
  with application to online ablation control in laser microsurgery}.
\newblock \bibinfo{journal}{Medical Image Analysis} \bibinfo{volume}{40},
  \bibinfo{pages}{80--95}.
\newblock \DOIprefix\doi{10.1016/j.media.2017.06.004}.
\bibitem[{Schule et~al.(2022)Schule, Haag, Somers, Veil, Tarin and
  Sawodny}]{schuleModelbasedSimultaneousLocalization2022}
\bibinfo{author}{Schule, J.}, \bibinfo{author}{Haag, J.},
  \bibinfo{author}{Somers, P.}, \bibinfo{author}{Veil, C.},
  \bibinfo{author}{Tarin, C.}, \bibinfo{author}{Sawodny, O.},
  \bibinfo{year}{2022}.
\newblock \bibinfo{title}{A {{Model-based Simultaneous Localization}} and
  {{Mapping Approach}} for {{Deformable Bodies}}}, in:
  \bibinfo{booktitle}{{{IEEE}}/{{ASME International Conference}} on {{Advanced
  Intelligent Mechatronics}}, {{AIM}}}, pp. \bibinfo{pages}{607--612}.
\bibitem[{Sengupta and
  Bartoli(2021)}]{senguptaColonoscopic3DReconstruction2021}
\bibinfo{author}{Sengupta, A.}, \bibinfo{author}{Bartoli, A.},
  \bibinfo{year}{2021}.
\newblock \bibinfo{title}{Colonoscopic {{3D}} reconstruction by tubular
  non-rigid structure-from-motion}.
\newblock \bibinfo{journal}{International Journal of Computer Assisted
  Radiology and Surgery} \bibinfo{volume}{16}, \bibinfo{pages}{1237--1241}.
\newblock \DOIprefix\doi{10.1007/s11548-021-02409-x}.
\bibitem[{Seshamani et~al.(2006)Seshamani, Lau and
  Hager}]{seshamaniRealtimeEndoscopicMosaicking2006a}
\bibinfo{author}{Seshamani, S.}, \bibinfo{author}{Lau, W.},
  \bibinfo{author}{Hager, G.}, \bibinfo{year}{2006}.
\newblock \bibinfo{title}{Real-Time Endoscopic Mosaicking}. volume
  \bibinfo{volume}{4190 LNCS - I}.
\newblock \DOIprefix\doi{10.1007/11866565_44}.
\bibitem[{Shi and {Tomasi}(1994)}]{shiGoodFeaturesTrack1994}
\bibinfo{author}{Shi, J.}, \bibinfo{author}{{Tomasi}}, \bibinfo{year}{1994}.
\newblock \bibinfo{title}{Good features to track}, in: \bibinfo{booktitle}{1994
  {{Proceedings}} of {{IEEE Conference}} on {{Computer Vision}} and {{Pattern
  Recognition}}}, pp. \bibinfo{pages}{593--600}.
\newblock \DOIprefix\doi{10.1109/CVPR.1994.323794}.
\bibitem[{Sidhu et~al.(2020)Sidhu, Tretschk, Golyanik, Agudo and
  Theobalt}]{sidhuNeuralDenseNonRigid2020a}
\bibinfo{author}{Sidhu, V.}, \bibinfo{author}{Tretschk, E.},
  \bibinfo{author}{Golyanik, V.}, \bibinfo{author}{Agudo, A.},
  \bibinfo{author}{Theobalt, C.}, \bibinfo{year}{2020}.
\newblock \bibinfo{title}{Neural {{Dense Non-Rigid Structure}} from {{Motion}}
  with {{Latent Space Constraints}}}, in: \bibinfo{editor}{Vedaldi, A.},
  \bibinfo{editor}{Bischof, H.}, \bibinfo{editor}{Brox, T.},
  \bibinfo{editor}{Frahm, J.M.} (Eds.), \bibinfo{booktitle}{Computer {{Vision}}
  {\textendash} {{ECCV}} 2020}. \bibinfo{publisher}{{Springer International
  Publishing}}, \bibinfo{address}{{Cham}}. volume \bibinfo{volume}{12361}, pp.
  \bibinfo{pages}{204--222}.
\newblock \DOIprefix\doi{10.1007/978-3-030-58517-4_13}.
\bibitem[{Song et~al.(2018)Song, Wang, Zhao, Huang and
  Dissanayake}]{songMISSLAMRealTimeLargeScale2018}
\bibinfo{author}{Song, J.}, \bibinfo{author}{Wang, J.}, \bibinfo{author}{Zhao,
  L.}, \bibinfo{author}{Huang, S.}, \bibinfo{author}{Dissanayake, G.},
  \bibinfo{year}{2018}.
\newblock \bibinfo{title}{{{MIS-SLAM}}: {{Real-Time Large-Scale Dense
  Deformable SLAM System}} in {{Minimal Invasive Surgery Based}} on
  {{Heterogeneous Computing}}}.
\newblock \bibinfo{journal}{IEEE Robotics and Automation Letters}
  \bibinfo{volume}{3}, \bibinfo{pages}{4068--4075}.
\newblock \DOIprefix\doi{10.1109/LRA.2018.2856519}.
\bibitem[{Song et~al.(2023)Song, Zhu, Lin and
  Ghaffari}]{songBDISBayesianDense2023}
\bibinfo{author}{Song, J.}, \bibinfo{author}{Zhu, Q.}, \bibinfo{author}{Lin,
  J.}, \bibinfo{author}{Ghaffari, M.}, \bibinfo{year}{2023}.
\newblock \bibinfo{title}{{{BDIS}}: {{Bayesian Dense Inverse Searching Method}}
  for {{Real-Time Stereo Surgical Image Matching}}}.
\newblock \bibinfo{journal}{IEEE Transactions on Robotics}
  \bibinfo{volume}{39}, \bibinfo{pages}{1388--1406}.
\bibitem[{Soper et~al.(2012)Soper, Porter and
  Seibel}]{soperSurfaceMosaicsBladder2012}
\bibinfo{author}{Soper, T.}, \bibinfo{author}{Porter, M.},
  \bibinfo{author}{Seibel, E.}, \bibinfo{year}{2012}.
\newblock \bibinfo{title}{Surface mosaics of the bladder reconstructed from
  endoscopic video for automated surveillance}.
\newblock \bibinfo{journal}{IEEE Transactions on Biomedical Engineering}
  \bibinfo{volume}{59}, \bibinfo{pages}{1670--1680}.
\newblock \DOIprefix\doi{10.1109/TBME.2012.2191783}.
\bibitem[{Stoyanov et~al.(2004)Stoyanov, Darzi and
  Yang}]{stoyanovDense3DDepth2004}
\bibinfo{author}{Stoyanov, D.}, \bibinfo{author}{Darzi, A.},
  \bibinfo{author}{Yang, G.}, \bibinfo{year}{2004}.
\newblock \bibinfo{title}{Dense {{3D}} depth recovery for soft tissue
  deformation during robotically assisted laparoscopic surgery}, in:
  \bibinfo{booktitle}{Lecture {{Notes}} in {{Computer Science}}}, pp.
  \bibinfo{pages}{41--48}.
\newblock \DOIprefix\doi{10.1007/978-3-540-30136-3_6}.
\bibitem[{Stoyanov et~al.(2005)Stoyanov, Mylonas, Deligianni, Darzi and
  Yang}]{stoyanovSoftTissueMotionTracking2005}
\bibinfo{author}{Stoyanov, D.}, \bibinfo{author}{Mylonas, G.P.},
  \bibinfo{author}{Deligianni, F.}, \bibinfo{author}{Darzi, A.},
  \bibinfo{author}{Yang, G.Z.}, \bibinfo{year}{2005}.
\newblock \bibinfo{title}{Soft-{{Tissue Motion Tracking}} and {{Structure
  Estimation}} for {{Robotic Assisted MIS Procedures}}},
  \bibinfo{publisher}{{Springer Berlin Heidelberg}}, \bibinfo{address}{{Berlin,
  Heidelberg}}, pp. \bibinfo{pages}{139--146}.
\newblock \DOIprefix\doi{10.1007/11566489_18}.
\bibitem[{Stoyanov et~al.(2010)Stoyanov, Scarzanella, Pratt and
  Yang}]{stoyanovRealtimeStereoReconstruction2010}
\bibinfo{author}{Stoyanov, D.}, \bibinfo{author}{Scarzanella, M.},
  \bibinfo{author}{Pratt, P.}, \bibinfo{author}{Yang, G.Z.},
  \bibinfo{year}{2010}.
\newblock \bibinfo{title}{Real-Time Stereo Reconstruction in Robotically
  Assisted Minimally Invasive Surgery}. volume \bibinfo{volume}{6361 LNCS}.
\newblock \DOIprefix\doi{10.1007/978-3-642-15705-9_34}.
\bibitem[{Sturm et~al.(2012)Sturm, Engelhard, Endres, Burgard and
  Cremers}]{sturmBenchmarkEvaluationRGBD2012}
\bibinfo{author}{Sturm, J.}, \bibinfo{author}{Engelhard, N.},
  \bibinfo{author}{Endres, F.}, \bibinfo{author}{Burgard, W.},
  \bibinfo{author}{Cremers, D.}, \bibinfo{year}{2012}.
\newblock \bibinfo{title}{A benchmark for the evaluation of {{RGB-D SLAM}}
  systems}, in: \bibinfo{booktitle}{2012 {{IEEE}}/{{RSJ International
  Conference}} on {{Intelligent Robots}} and {{Systems}}},
  \bibinfo{publisher}{{IEEE}}, \bibinfo{address}{{Vilamoura-Algarve,
  Portugal}}. pp. \bibinfo{pages}{573--580}.
\newblock \DOIprefix\doi{10.1109/IROS.2012.6385773}.
\bibitem[{Sucar et~al.(2021)Sucar, Liu, Ortiz and
  Davison}]{sucarIMAPImplicitMapping2021}
\bibinfo{author}{Sucar, E.}, \bibinfo{author}{Liu, S.}, \bibinfo{author}{Ortiz,
  J.}, \bibinfo{author}{Davison, A.J.}, \bibinfo{year}{2021}.
\newblock \bibinfo{title}{{{iMAP}}: {{Implicit Mapping}} and {{Positioning}} in
  {{Real-Time}}}.
\newblock \bibinfo{journal}{arXiv:2103.12352 [cs]}
  \href{http://arxiv.org/abs/2103.12352}{\tt arXiv:2103.12352}.
\bibitem[{Sumner et~al.(2007)Sumner, Schmid and
  Pauly}]{sumnerEmbeddedDeformationShape2007}
\bibinfo{author}{Sumner, R.W.}, \bibinfo{author}{Schmid, J.},
  \bibinfo{author}{Pauly, M.}, \bibinfo{year}{2007}.
\newblock \bibinfo{title}{Embedded deformation for shape manipulation}, in:
  \bibinfo{booktitle}{{{ACM SIGGRAPH}} 2007 Papers},
  \bibinfo{publisher}{{Association for Computing Machinery}},
  \bibinfo{address}{{New York, NY, USA}}. pp. \bibinfo{pages}{80--es}.
\newblock \DOIprefix\doi{10.1145/1275808.1276478}.
\bibitem[{Sun et~al.(2018)Sun, Yang, Liu and Kautz}]{sunPWCNetCNNsOptical2018}
\bibinfo{author}{Sun, D.}, \bibinfo{author}{Yang, X.}, \bibinfo{author}{Liu,
  M.Y.}, \bibinfo{author}{Kautz, J.}, \bibinfo{year}{2018}.
\newblock \bibinfo{title}{{{PWC-Net}}: {{CNNs}} for {{Optical Flow Using
  Pyramid}}, {{Warping}}, and {{Cost Volume}}}, in: \bibinfo{booktitle}{2018
  {{IEEE}}/{{CVF Conference}} on {{Computer Vision}} and {{Pattern
  Recognition}}}, \bibinfo{publisher}{{IEEE}}, \bibinfo{address}{{Salt Lake
  City, UT, USA}}. pp. \bibinfo{pages}{8934--8943}.
\newblock \DOIprefix\doi{10.1109/CVPR.2018.00931}.
\bibitem[{Sun et~al.(2021)Sun, Shen, Wang, Bao and
  Zhou}]{sunLoFTRDetectorFreeLocal2021}
\bibinfo{author}{Sun, J.}, \bibinfo{author}{Shen, Z.}, \bibinfo{author}{Wang,
  Y.}, \bibinfo{author}{Bao, H.}, \bibinfo{author}{Zhou, X.},
  \bibinfo{year}{2021}.
\newblock \bibinfo{title}{{{LoFTR}}: {{Detector-Free Local Feature Matching
  With Transformers}}}, in: \bibinfo{booktitle}{Proceedings of the
  {{IEEE}}/{{CVF Conference}} on {{Computer Vision}} and {{Pattern
  Recognition}}}, pp. \bibinfo{pages}{8922--8931}.
\bibitem[{Sun et~al.(2023)Sun, Wang, Ma and
  Su}]{sunDynamicSurfaceReconstruction2023}
\bibinfo{author}{Sun, X.}, \bibinfo{author}{Wang, F.}, \bibinfo{author}{Ma,
  Z.}, \bibinfo{author}{Su, H.}, \bibinfo{year}{2023}.
\newblock \bibinfo{title}{Dynamic surface reconstruction in robot-assisted
  minimally invasive surgery based on neural radiance fields}.
\newblock \bibinfo{journal}{Int J CARS}
  \DOIprefix\doi{10.1007/s11548-023-03016-8}.
\bibitem[{Suputra et~al.(2020)Suputra, Sensusiati, Yuniarno, Purnomo and
  Purnama}]{suputra3DLaplacianSurface2020}
\bibinfo{author}{Suputra, P.}, \bibinfo{author}{Sensusiati, A.},
  \bibinfo{author}{Yuniarno, E.}, \bibinfo{author}{Purnomo, M.},
  \bibinfo{author}{Purnama, I.}, \bibinfo{year}{2020}.
\newblock \bibinfo{title}{{{3D}} laplacian surface deformation for template
  fitting on craniofacial reconstruction}, in: \bibinfo{booktitle}{{{ACM
  International Conference Proceeding Series}}}, pp. \bibinfo{pages}{27--32}.
\newblock \DOIprefix\doi{10.1145/3411174.3411175}.
\bibitem[{Tang and Tan(2019)}]{tangBANetDenseBundle2019b}
\bibinfo{author}{Tang, C.}, \bibinfo{author}{Tan, P.}, \bibinfo{year}{2019}.
\newblock \bibinfo{title}{{{BA-Net}}: {{Dense Bundle Adjustment Network}}}.
\newblock \DOIprefix\doi{10.48550/arXiv.1806.04807},
  \href{http://arxiv.org/abs/1806.04807}{\tt arXiv:1806.04807}.
\bibitem[{Teed and Deng(2020)}]{teedRaftRecurrentAllpairs2020}
\bibinfo{author}{Teed, Z.}, \bibinfo{author}{Deng, J.}, \bibinfo{year}{2020}.
\newblock \bibinfo{title}{Raft: {{Recurrent}} all-pairs field transforms for
  optical flow}, in: \bibinfo{booktitle}{European Conference on Computer
  Vision}. \bibinfo{publisher}{{Springer}}, pp. \bibinfo{pages}{402--419}.
\bibitem[{Tomasi and Kanade(1991)}]{tomasi1991detection}
\bibinfo{author}{Tomasi, C.}, \bibinfo{author}{Kanade, T.},
  \bibinfo{year}{1991}.
\newblock \bibinfo{title}{Detection and tracking of point}.
\newblock \bibinfo{journal}{Int J Comput Vis} \bibinfo{volume}{9},
  \bibinfo{pages}{3}.
\bibitem[{Torresani et~al.(2008)Torresani, Hertzmann and
  Bregler}]{torresaniNonrigidStructurefromMotionEstimating2008}
\bibinfo{author}{Torresani, L.}, \bibinfo{author}{Hertzmann, A.},
  \bibinfo{author}{Bregler, C.}, \bibinfo{year}{2008}.
\newblock \bibinfo{title}{Nonrigid {{Structure-from-Motion}}: {{Estimating
  Shape}} and {{Motion}} with {{Hierarchical Priors}}}.
\newblock \bibinfo{journal}{IEEE Transactions on Pattern Analysis and Machine
  Intelligence} \bibinfo{volume}{30}, \bibinfo{pages}{878--892}.
\newblock \DOIprefix\doi{10.1109/TPAMI.2007.70752}.
\bibitem[{Tretschk et~al.(2022)Tretschk, Kairanda, R, Dabral, Kortylewski,
  Egger, Habermann, Fua, Theobalt and Golyanik}]{tretschkStateArtDense2022}
\bibinfo{author}{Tretschk, E.}, \bibinfo{author}{Kairanda, N.},
  \bibinfo{author}{R, M.B.}, \bibinfo{author}{Dabral, R.},
  \bibinfo{author}{Kortylewski, A.}, \bibinfo{author}{Egger, B.},
  \bibinfo{author}{Habermann, M.}, \bibinfo{author}{Fua, P.},
  \bibinfo{author}{Theobalt, C.}, \bibinfo{author}{Golyanik, V.},
  \bibinfo{year}{2022}.
\newblock \bibinfo{title}{State of the {{Art}} in {{Dense Monocular Non-Rigid
  3D Reconstruction}}}.
\newblock \href{http://arxiv.org/abs/2210.15664}{\tt arXiv:2210.15664}.
\bibitem[{Tukra and Giannarou(2022)}]{tukraStereoDepthEstimation2022}
\bibinfo{author}{Tukra, S.}, \bibinfo{author}{Giannarou, S.},
  \bibinfo{year}{2022}.
\newblock \bibinfo{title}{Stereo {{Depth Estimation}} via~{{Self-supervised
  Contrastive Representation Learning}}}. volume \bibinfo{volume}{13437 LNCS}.
\newblock \DOIprefix\doi{10.1007/978-3-031-16449-1_58}.
\bibitem[{Turan et~al.(2017)Turan, Almalioglu, Araujo, Konukoglu and
  Sitti}]{turanNonrigidMapFusionbased2017}
\bibinfo{author}{Turan, M.}, \bibinfo{author}{Almalioglu, Y.},
  \bibinfo{author}{Araujo, H.}, \bibinfo{author}{Konukoglu, E.},
  \bibinfo{author}{Sitti, M.}, \bibinfo{year}{2017}.
\newblock \bibinfo{title}{A non-rigid map fusion-based direct {{SLAM}} method
  for endoscopic capsule robots}.
\newblock \bibinfo{journal}{International Journal of Intelligent Robotics and
  Applications} \bibinfo{volume}{1}, \bibinfo{pages}{399--409}.
\newblock \DOIprefix\doi{10.1007/s41315-017-0036-4}.
\bibitem[{Vasconcelos et~al.(2019)Vasconcelos, Mazomenos, Kelly and
  Stoyanov}]{vasconcelosRCMSLAMVisualLocalisation2019}
\bibinfo{author}{Vasconcelos, F.}, \bibinfo{author}{Mazomenos, E.},
  \bibinfo{author}{Kelly, J.}, \bibinfo{author}{Stoyanov, D.},
  \bibinfo{year}{2019}.
\newblock \bibinfo{title}{{{RCM-SLAM}}: {{Visual}} localisation and mapping
  under remote centre of motion constraints}, in:
  \bibinfo{booktitle}{Proceedings - {{IEEE International Conference}} on
  {{Robotics}} and {{Automation}}}, pp. \bibinfo{pages}{9278--9284}.
\newblock \DOIprefix\doi{10.1109/ICRA.2019.8793931}.
\bibitem[{Viola and Jones(2001)}]{violaRapidObjectDetection2001}
\bibinfo{author}{Viola, P.}, \bibinfo{author}{Jones, M.}, \bibinfo{year}{2001}.
\newblock \bibinfo{title}{Rapid object detection using a boosted cascade of
  simple features}, in: \bibinfo{booktitle}{Proceedings of the 2001 {{IEEE
  Computer Society Conference}} on {{Computer Vision}} and {{Pattern
  Recognition}}. {{CVPR}} 2001}, pp. \bibinfo{pages}{I--I}.
\newblock \DOIprefix\doi{10.1109/CVPR.2001.990517}.
\bibitem[{{Visentini-Scarzanella} et~al.(2017){Visentini-Scarzanella}, Sugiura,
  Kaneko and Koto}]{visentini-scarzanellaDeepMonocular3D2017}
\bibinfo{author}{{Visentini-Scarzanella}, M.}, \bibinfo{author}{Sugiura, T.},
  \bibinfo{author}{Kaneko, T.}, \bibinfo{author}{Koto, S.},
  \bibinfo{year}{2017}.
\newblock \bibinfo{title}{Deep monocular {{3D}} reconstruction for assisted
  navigation in bronchoscopy}.
\newblock \bibinfo{journal}{International Journal of Computer Assisted
  Radiology and Surgery} \bibinfo{volume}{12}, \bibinfo{pages}{1089--1099}.
\newblock \DOIprefix\doi{10.1007/s11548-017-1609-2}.
\bibitem[{Wang et~al.(2020a)Wang, Oda, Hayashi, Villard, Kitasaka, Takabatake,
  Mori, Honma, Natori and Mori}]{wangVisualSLAMbasedBronchoscope2020}
\bibinfo{author}{Wang, C.}, \bibinfo{author}{Oda, M.},
  \bibinfo{author}{Hayashi, Y.}, \bibinfo{author}{Villard, B.},
  \bibinfo{author}{Kitasaka, T.}, \bibinfo{author}{Takabatake, H.},
  \bibinfo{author}{Mori, M.}, \bibinfo{author}{Honma, H.},
  \bibinfo{author}{Natori, H.}, \bibinfo{author}{Mori, K.},
  \bibinfo{year}{2020}a.
\newblock \bibinfo{title}{A visual {{SLAM-based}} bronchoscope tracking scheme
  for bronchoscopic navigation}.
\newblock \bibinfo{journal}{International Journal of Computer Assisted
  Radiology and Surgery} \bibinfo{volume}{15}, \bibinfo{pages}{1619--1630}.
\newblock \DOIprefix\doi{10.1007/s11548-020-02241-9}.
\bibitem[{Wang et~al.(2023)Wang, Chang, Cai, Li, Hariharan, Holynski and
  Snavely}]{wangTrackingEverythingEverywhere2023}
\bibinfo{author}{Wang, Q.}, \bibinfo{author}{Chang, Y.Y.},
  \bibinfo{author}{Cai, R.}, \bibinfo{author}{Li, Z.},
  \bibinfo{author}{Hariharan, B.}, \bibinfo{author}{Holynski, A.},
  \bibinfo{author}{Snavely, N.}, \bibinfo{year}{2023}.
\newblock \bibinfo{title}{Tracking {{Everything Everywhere All}} at {{Once}}},
  in: \bibinfo{booktitle}{Proceedings of the {{IEEE International Conference}}
  on {{Computer Vision}}}.
\bibitem[{Wang et~al.(2020b)Wang, Zhou, Hariharan and
  Snavely}]{wangLearningFeatureDescriptors2020}
\bibinfo{author}{Wang, Q.}, \bibinfo{author}{Zhou, X.},
  \bibinfo{author}{Hariharan, B.}, \bibinfo{author}{Snavely, N.},
  \bibinfo{year}{2020}b.
\newblock \bibinfo{title}{Learning {{Feature Descriptors Using Camera Pose
  Supervision}}}, in: \bibinfo{booktitle}{Computer {{Vision}} {\textendash}
  {{ECCV}} 2020}.
\newblock \href{http://arxiv.org/abs/2004.13324}{\tt arXiv:2004.13324}.
\bibitem[{Wang et~al.(2019)Wang, Pizer and
  Frahm}]{wangRecurrentNeuralNetwork2019}
\bibinfo{author}{Wang, R.}, \bibinfo{author}{Pizer, S.M.},
  \bibinfo{author}{Frahm, J.M.}, \bibinfo{year}{2019}.
\newblock \bibinfo{title}{Recurrent {{Neural Network}} for
  ({{Un-}}){{Supervised Learning}} of {{Monocular Video Visual Odometry}} and
  {{Depth}}}, in: \bibinfo{booktitle}{Proceedings of the {{IEEE}}/{{CVF
  Conference}} on {{Computer Vision}} and {{Pattern Recognition}}}.
\bibitem[{Wang et~al.(2022)Wang, Long, Fan and
  Dou}]{wangNeuralRenderingStereo2022}
\bibinfo{author}{Wang, Y.}, \bibinfo{author}{Long, Y.}, \bibinfo{author}{Fan,
  S.H.}, \bibinfo{author}{Dou, Q.}, \bibinfo{year}{2022}.
\newblock \bibinfo{title}{Neural {{Rendering}} for~{{Stereo 3D Reconstruction}}
  of~{{Deformable Tissues}} in~{{Robotic Surgery}}}, in: \bibinfo{editor}{Wang,
  L.}, \bibinfo{editor}{Dou, Q.}, \bibinfo{editor}{Fletcher, P.T.},
  \bibinfo{editor}{Speidel, S.}, \bibinfo{editor}{Li, S.} (Eds.),
  \bibinfo{booktitle}{Medical {{Image Computing}} and {{Computer Assisted
  Intervention}} {\textendash} {{MICCAI}} 2022}, \bibinfo{publisher}{{Springer
  Nature Switzerland}}, \bibinfo{address}{{Cham}}. pp.
  \bibinfo{pages}{431--441}.
\bibitem[{Wei et~al.(2023)Wei, Li, Mo, Lu, Long, Yang, Dou, Liu and
  Sun}]{weiStereoDenseScene2023}
\bibinfo{author}{Wei, R.}, \bibinfo{author}{Li, B.}, \bibinfo{author}{Mo, H.},
  \bibinfo{author}{Lu, B.}, \bibinfo{author}{Long, Y.}, \bibinfo{author}{Yang,
  B.}, \bibinfo{author}{Dou, Q.}, \bibinfo{author}{Liu, Y.},
  \bibinfo{author}{Sun, D.}, \bibinfo{year}{2023}.
\newblock \bibinfo{title}{Stereo {{Dense Scene Reconstruction}} and {{Accurate
  Localization}} for {{Learning-Based Navigation}} of {{Laparoscope}} in
  {{Minimally Invasive Surgery}}}.
\newblock \bibinfo{journal}{IEEE Transactions on Biomedical Engineering}
  \bibinfo{volume}{70}, \bibinfo{pages}{488--500}.
\newblock \DOIprefix\doi{10.1109/TBME.2022.3195027}.
\bibitem[{Weibel et~al.(2012)Weibel, Daul, Wolf, R{\"o}sch and
  Guillemin}]{weibelGraphBasedConstruction2012a}
\bibinfo{author}{Weibel, T.}, \bibinfo{author}{Daul, C.},
  \bibinfo{author}{Wolf, D.}, \bibinfo{author}{R{\"o}sch, R.},
  \bibinfo{author}{Guillemin, F.}, \bibinfo{year}{2012}.
\newblock \bibinfo{title}{Graph based construction of textured large field of
  view mosaics for bladder cancer diagnosis}.
\newblock \bibinfo{journal}{Pattern Recognition} \bibinfo{volume}{45},
  \bibinfo{pages}{4138--4150}.
\newblock \DOIprefix\doi{10.1016/j.patcog.2012.05.023}.
\bibitem[{Widya et~al.(2019)Widya, Monno, Okutomi, Suzuki, Gotoda and
  Miki}]{widyaWholeStomach3D2019}
\bibinfo{author}{Widya, A.}, \bibinfo{author}{Monno, Y.},
  \bibinfo{author}{Okutomi, M.}, \bibinfo{author}{Suzuki, S.},
  \bibinfo{author}{Gotoda, T.}, \bibinfo{author}{Miki, K.},
  \bibinfo{year}{2019}.
\newblock \bibinfo{title}{Whole {{Stomach 3D}} reconstruction and frame
  localization from monocular endoscope video}.
\newblock \bibinfo{journal}{IEEE Journal of Translational Engineering in Health
  and Medicine} \bibinfo{volume}{7}.
\newblock \DOIprefix\doi{10.1109/JTEHM.2019.2946802}.
\bibitem[{Widya et~al.(2020)Widya, Monno, Okutomi, Suzuki, Gotoda and
  Miki}]{widyaStomach3DReconstruction2020}
\bibinfo{author}{Widya, A.}, \bibinfo{author}{Monno, Y.},
  \bibinfo{author}{Okutomi, M.}, \bibinfo{author}{Suzuki, S.},
  \bibinfo{author}{Gotoda, T.}, \bibinfo{author}{Miki, K.},
  \bibinfo{year}{2020}.
\newblock \bibinfo{title}{Stomach {{3D Reconstruction Based}} on {{Virtual
  Chromoendoscopic Image Generation}}}, in: \bibinfo{booktitle}{Proceedings of
  the {{Annual International Conference}} of the {{IEEE Engineering}} in
  {{Medicine}} and {{Biology Society}}, {{EMBS}}}, pp.
  \bibinfo{pages}{1848--1852}.
\newblock \DOIprefix\doi{10.1109/EMBC44109.2020.9176016}.
\bibitem[{Widya et~al.(2021)Widya, Monno, Okutomi, Suzuki, Gotoda and
  Miki}]{widyaStomach3DReconstruction2021}
\bibinfo{author}{Widya, A.}, \bibinfo{author}{Monno, Y.},
  \bibinfo{author}{Okutomi, M.}, \bibinfo{author}{Suzuki, S.},
  \bibinfo{author}{Gotoda, T.}, \bibinfo{author}{Miki, K.},
  \bibinfo{year}{2021}.
\newblock \bibinfo{title}{Stomach {{3D Reconstruction Using Virtual
  Chromoendoscopic Images}}}.
\newblock \bibinfo{journal}{IEEE Journal of Translational Engineering in Health
  and Medicine} \bibinfo{volume}{9}.
\newblock \DOIprefix\doi{10.1109/JTEHM.2021.3062226}.
\bibitem[{Wu et~al.(2023)Wu, Yi, Fang, Xie, Zhang, Wei, Liu, Tian and
  Wang}]{wu4DGaussianSplatting2023}
\bibinfo{author}{Wu, G.}, \bibinfo{author}{Yi, T.}, \bibinfo{author}{Fang, J.},
  \bibinfo{author}{Xie, L.}, \bibinfo{author}{Zhang, X.}, \bibinfo{author}{Wei,
  W.}, \bibinfo{author}{Liu, W.}, \bibinfo{author}{Tian, Q.},
  \bibinfo{author}{Wang, X.}, \bibinfo{year}{2023}.
\newblock \bibinfo{title}{{{4D Gaussian Splatting}} for {{Real-Time Dynamic
  Scene Rendering}}}.
\newblock \href{http://arxiv.org/abs/2310.08528}{\tt arXiv:2310.08528}.
\bibitem[{Wynn and
  Turmukhambetov(2023)}]{wynnDiffusioNeRFRegularizingNeural2023}
\bibinfo{author}{Wynn, J.}, \bibinfo{author}{Turmukhambetov, D.},
  \bibinfo{year}{2023}.
\newblock \bibinfo{title}{{{DiffusioNeRF}}: {{Regularizing Neural Radiance
  Fields}} with {{Denoising Diffusion Models}}}, in: \bibinfo{booktitle}{2023
  {{IEEE}}/{{CVF Conference}} on {{Computer Vision}} and {{Pattern
  Recognition}} ({{CVPR}})}, \bibinfo{publisher}{{IEEE}},
  \bibinfo{address}{{Vancouver, BC, Canada}}. pp. \bibinfo{pages}{4180--4189}.
\newblock \DOIprefix\doi{10.1109/CVPR52729.2023.00407}.
\bibitem[{Xi et~al.(2021)Xi, Zhao, Chen, Gao, Tang, Wan and
  Xue}]{xiRecoveringDense3D2021}
\bibinfo{author}{Xi, L.}, \bibinfo{author}{Zhao, Y.}, \bibinfo{author}{Chen,
  L.}, \bibinfo{author}{Gao, Q.}, \bibinfo{author}{Tang, W.},
  \bibinfo{author}{Wan, T.}, \bibinfo{author}{Xue, T.}, \bibinfo{year}{2021}.
\newblock \bibinfo{title}{Recovering dense {{3D}} point clouds from single
  endoscopic image}.
\newblock \bibinfo{journal}{Computer Methods and Programs in Biomedicine}
  \bibinfo{volume}{205}.
\newblock \DOIprefix\doi{10.1016/j.cmpb.2021.106077}.
\bibitem[{Xu et~al.(2023)Xu, Peng, Lin, He, Sun, Shen, Bao and
  Zhou}]{xu4K4DRealTime4D2023}
\bibinfo{author}{Xu, Z.}, \bibinfo{author}{Peng, S.}, \bibinfo{author}{Lin,
  H.}, \bibinfo{author}{He, G.}, \bibinfo{author}{Sun, J.},
  \bibinfo{author}{Shen, Y.}, \bibinfo{author}{Bao, H.}, \bibinfo{author}{Zhou,
  X.}, \bibinfo{year}{2023}.
\newblock \bibinfo{title}{{{4K4D}}: {{Real-Time 4D View Synthesis}} at {{4K
  Resolution}}}.
\newblock \href{http://arxiv.org/abs/2310.11448}{\tt arXiv:2310.11448}.
\bibitem[{Yan et~al.(2023)Yan, Qu, Wang, Xu, Wang, Zhao and
  Li}]{yanGSSLAMDenseVisual2023}
\bibinfo{author}{Yan, C.}, \bibinfo{author}{Qu, D.}, \bibinfo{author}{Wang,
  D.}, \bibinfo{author}{Xu, D.}, \bibinfo{author}{Wang, Z.},
  \bibinfo{author}{Zhao, B.}, \bibinfo{author}{Li, X.}, \bibinfo{year}{2023}.
\newblock \bibinfo{title}{{{GS-SLAM}}: {{Dense Visual SLAM}} with {{3D Gaussian
  Splatting}}}.
\newblock \DOIprefix\doi{10.48550/arXiv.2311.11700},
  \href{http://arxiv.org/abs/2311.11700}{\tt arXiv:2311.11700}.
\bibitem[{Yang et~al.(2023a)Yang, Wang, Wang, Dou, Yang and
  Shen}]{yangEfficientDeformableTissue2023}
\bibinfo{author}{Yang, C.}, \bibinfo{author}{Wang, K.}, \bibinfo{author}{Wang,
  Y.}, \bibinfo{author}{Dou, Q.}, \bibinfo{author}{Yang, X.},
  \bibinfo{author}{Shen, W.}, \bibinfo{year}{2023}a.
\newblock \bibinfo{title}{Efficient {{Deformable Tissue Reconstruction}} via
  {{Orthogonal Neural Plane}}}.
\newblock \DOIprefix\doi{10.48550/arXiv.2312.15253},
  \href{http://arxiv.org/abs/2312.15253}{\tt arXiv:2312.15253}.
\bibitem[{Yang et~al.(2023b)Yang, Wang, Wang, Yang and
  Shen}]{yangNeuralLerPlaneRepresentations2023}
\bibinfo{author}{Yang, C.}, \bibinfo{author}{Wang, K.}, \bibinfo{author}{Wang,
  Y.}, \bibinfo{author}{Yang, X.}, \bibinfo{author}{Shen, W.},
  \bibinfo{year}{2023}b.
\newblock \bibinfo{title}{Neural {{LerPlane Representations}} for~{{Fast 4D
  Reconstruction}} of~{{Deformable Tissues}}}, in: \bibinfo{editor}{Greenspan,
  H.}, \bibinfo{editor}{Madabhushi, A.}, \bibinfo{editor}{Mousavi, P.},
  \bibinfo{editor}{Salcudean, S.}, \bibinfo{editor}{Duncan, J.},
  \bibinfo{editor}{{Syeda-Mahmood}, T.}, \bibinfo{editor}{Taylor, R.} (Eds.),
  \bibinfo{booktitle}{Medical {{Image Computing}} and {{Computer Assisted
  Intervention}} {\textendash} {{MICCAI}} 2023}, \bibinfo{publisher}{{Springer
  Nature Switzerland}}, \bibinfo{address}{{Cham}}. pp. \bibinfo{pages}{46--56}.
\newblock \DOIprefix\doi{10.1007/978-3-031-43996-4_5}.
\bibitem[{Yang et~al.(2023c)Yang, Ivanovic, Litany, Weng, Kim, Li, Che, Xu,
  Fidler, Pavone and Wang}]{yangEmerNeRFEmergentSpatialTemporal2023}
\bibinfo{author}{Yang, J.}, \bibinfo{author}{Ivanovic, B.},
  \bibinfo{author}{Litany, O.}, \bibinfo{author}{Weng, X.},
  \bibinfo{author}{Kim, S.W.}, \bibinfo{author}{Li, B.}, \bibinfo{author}{Che,
  T.}, \bibinfo{author}{Xu, D.}, \bibinfo{author}{Fidler, S.},
  \bibinfo{author}{Pavone, M.}, \bibinfo{author}{Wang, Y.},
  \bibinfo{year}{2023}c.
\newblock \bibinfo{title}{{{EmerNeRF}}: {{Emergent Spatial-Temporal Scene
  Decomposition}} via {{Self-Supervision}}}.
\newblock \DOIprefix\doi{10.48550/arXiv.2311.02077},
  \href{http://arxiv.org/abs/2311.02077}{\tt arXiv:2311.02077}.
\bibitem[{Yang et~al.(2024)Yang, Kang, Huang, Xu, Feng and
  Zhao}]{yangDepthAnythingUnleashing2024}
\bibinfo{author}{Yang, L.}, \bibinfo{author}{Kang, B.}, \bibinfo{author}{Huang,
  Z.}, \bibinfo{author}{Xu, X.}, \bibinfo{author}{Feng, J.},
  \bibinfo{author}{Zhao, H.}, \bibinfo{year}{2024}.
\newblock \bibinfo{title}{Depth {{Anything}}: {{Unleashing}} the {{Power}} of
  {{Large-Scale Unlabeled Data}}}.
\newblock \href{http://arxiv.org/abs/2401.10891}{\tt arXiv:2401.10891}.
\bibitem[{Yang et~al.(2021)Yang, Simon, Li and
  Linte}]{yangDenseDepthEstimation2021}
\bibinfo{author}{Yang, Z.}, \bibinfo{author}{Simon, R.}, \bibinfo{author}{Li,
  Y.}, \bibinfo{author}{Linte, C.A.}, \bibinfo{year}{2021}.
\newblock \bibinfo{title}{Dense {{Depth Estimation}} from {{Stereo Endoscopy
  Videos Using Unsupervised Optical Flow Methods}}}, in:
  \bibinfo{booktitle}{Medical {{Image Understanding}} and {{Analysis}}}.
\bibitem[{Ye et~al.(2016)Ye, Giannarou, Meining and
  Yang}]{yeOnlineTrackingRetargeting2016}
\bibinfo{author}{Ye, M.}, \bibinfo{author}{Giannarou, S.},
  \bibinfo{author}{Meining, A.}, \bibinfo{author}{Yang, G.Z.},
  \bibinfo{year}{2016}.
\newblock \bibinfo{title}{Online tracking and retargeting with applications to
  optical biopsy in gastrointestinal endoscopic examinations}.
\newblock \bibinfo{journal}{Medical Image Analysis} \bibinfo{volume}{30},
  \bibinfo{pages}{144--157}.
\newblock \DOIprefix\doi{10.1016/j.media.2015.10.003}.
\bibitem[{Ye et~al.(2017)Ye, Johns, Handa, Zhang, Pratt and
  Yang}]{yeSelfSupervisedSiameseLearning2017}
\bibinfo{author}{Ye, M.}, \bibinfo{author}{Johns, E.}, \bibinfo{author}{Handa,
  A.}, \bibinfo{author}{Zhang, L.}, \bibinfo{author}{Pratt, P.},
  \bibinfo{author}{Yang, G.Z.}, \bibinfo{year}{2017}.
\newblock \bibinfo{title}{Self-{{Supervised Siamese Learning}} on {{Stereo
  Image Pairs}} for {{Depth Estimation}} in {{Robotic Surgery}}}.
\newblock \bibinfo{journal}{arXiv:1705.08260 [cs]}
  \href{http://arxiv.org/abs/1705.08260}{\tt arXiv:1705.08260}.
\bibitem[{Yip et~al.(2012)Yip, Lowe, Salcudean, Rohling and
  Nguan}]{yipTissueTrackingRegistration2012}
\bibinfo{author}{Yip, M.}, \bibinfo{author}{Lowe, D.},
  \bibinfo{author}{Salcudean, S.}, \bibinfo{author}{Rohling, R.},
  \bibinfo{author}{Nguan, C.}, \bibinfo{year}{2012}.
\newblock \bibinfo{title}{Tissue tracking and registration for image-guided
  surgery}.
\newblock \bibinfo{journal}{IEEE Transactions on Medical Imaging}
  \bibinfo{volume}{31}, \bibinfo{pages}{2169--2182}.
\newblock \DOIprefix\doi{10.1109/TMI.2012.2212718}.
\bibitem[{Zha et~al.(2023)Zha, Cheng, Li, Harandi and
  Ge}]{zhaEndoSurfNeuralSurface2023a}
\bibinfo{author}{Zha, R.}, \bibinfo{author}{Cheng, X.}, \bibinfo{author}{Li,
  H.}, \bibinfo{author}{Harandi, M.}, \bibinfo{author}{Ge, Z.},
  \bibinfo{year}{2023}.
\newblock \bibinfo{title}{{{EndoSurf}}: {{Neural Surface Reconstruction}}
  of~{{Deformable Tissues}} with~{{Stereo Endoscope Videos}}}, in:
  \bibinfo{editor}{Greenspan, H.}, \bibinfo{editor}{Madabhushi, A.},
  \bibinfo{editor}{Mousavi, P.}, \bibinfo{editor}{Salcudean, S.},
  \bibinfo{editor}{Duncan, J.}, \bibinfo{editor}{{Syeda-Mahmood}, T.},
  \bibinfo{editor}{Taylor, R.} (Eds.), \bibinfo{booktitle}{Medical {{Image
  Computing}} and {{Computer Assisted Intervention}} {\textendash} {{MICCAI}}
  2023}, \bibinfo{publisher}{{Springer Nature Switzerland}},
  \bibinfo{address}{{Cham}}. pp. \bibinfo{pages}{13--23}.
\newblock \DOIprefix\doi{10.1007/978-3-031-43996-4_2}.
\bibitem[{Zhang et~al.(2019)Zhang, Prisacariu, Yang and
  Torr}]{zhangGANetGuidedAggregation2019}
\bibinfo{author}{Zhang, F.}, \bibinfo{author}{Prisacariu, V.},
  \bibinfo{author}{Yang, R.}, \bibinfo{author}{Torr, P.H.},
  \bibinfo{year}{2019}.
\newblock \bibinfo{title}{{{GA-Net}}: {{Guided Aggregation Net}} for
  {{End-To-End Stereo Matching}}}, in: \bibinfo{booktitle}{2019 {{IEEE}}/{{CVF
  Conference}} on {{Computer Vision}} and {{Pattern Recognition}} ({{CVPR}})},
  \bibinfo{publisher}{{IEEE}}, \bibinfo{address}{{Long Beach, CA, USA}}. pp.
  \bibinfo{pages}{185--194}.
\newblock \DOIprefix\doi{10.1109/CVPR.2019.00027}.
\bibitem[{Zhang et~al.(2023)Zhang, Feng, He and
  Jiang}]{zhangRobustFeatureMatching2023}
\bibinfo{author}{Zhang, G.}, \bibinfo{author}{Feng, G.}, \bibinfo{author}{He,
  F.}, \bibinfo{author}{Jiang, Z.}, \bibinfo{year}{2023}.
\newblock \bibinfo{title}{Robust {{Feature Matching}} for {{VSLAM}} in
  {{Non-Rigid Scenes}}}, in: \bibinfo{booktitle}{Proceedings of {{SPIE}} -
  {{The International Society}} for {{Optical Engineering}}}.
\newblock \DOIprefix\doi{10.1117/12.2668343}.
\bibitem[{Zhang et~al.()Zhang, Herrmann, Hur, Polan{\'i}a, Jampani, Sun and
  Yang}]{zhangTaleTwoFeatures}
\bibinfo{author}{Zhang, J.}, \bibinfo{author}{Herrmann, C.},
  \bibinfo{author}{Hur, J.}, \bibinfo{author}{Polan{\'i}a, L.F.},
  \bibinfo{author}{Jampani, V.}, \bibinfo{author}{Sun, D.},
  \bibinfo{author}{Yang, M.H.}, .
\newblock \bibinfo{title}{A {{Tale}} of {{Two Features}}: {{Stable Diffusion
  Complements DINO}} for {{Zero-Shot Semantic Correspondence}}} .
\bibitem[{Zhang et~al.(2018)Zhang, Isola, Efros, Shechtman and
  Wang}]{zhangUnreasonableEffectivenessDeep2018}
\bibinfo{author}{Zhang, R.}, \bibinfo{author}{Isola, P.},
  \bibinfo{author}{Efros, A.A.}, \bibinfo{author}{Shechtman, E.},
  \bibinfo{author}{Wang, O.}, \bibinfo{year}{2018}.
\newblock \bibinfo{title}{The {{Unreasonable Effectiveness}} of {{Deep
  Features}} as a {{Perceptual Metric}}}, in: \bibinfo{booktitle}{2018
  {{IEEE}}/{{CVF Conference}} on {{Computer Vision}} and {{Pattern
  Recognition}}}, \bibinfo{publisher}{{IEEE}}, \bibinfo{address}{{Salt Lake
  City, UT}}. pp. \bibinfo{pages}{586--595}.
\newblock \DOIprefix\doi{10.1109/CVPR.2018.00068}.
\bibitem[{Zhang et~al.(2021a)Zhang, Zhao, Huang, Ma, Hu and
  Hao}]{zhang3DReconstructionDeformable2021}
\bibinfo{author}{Zhang, S.}, \bibinfo{author}{Zhao, L.},
  \bibinfo{author}{Huang, S.}, \bibinfo{author}{Ma, R.}, \bibinfo{author}{Hu,
  B.}, \bibinfo{author}{Hao, Q.}, \bibinfo{year}{2021}a.
\newblock \bibinfo{title}{{{3D Reconstruction}} of {{Deformable Colon
  Structures}} based on {{Preoperative Model}} and {{Deep Neural Network}}},
  in: \bibinfo{booktitle}{Proceedings - {{IEEE International Conference}} on
  {{Robotics}} and {{Automation}}}, pp. \bibinfo{pages}{11457--11462}.
\newblock \DOIprefix\doi{10.1109/ICRA48506.2021.9561772}.
\bibitem[{Zhang et~al.(2022)Zhang, Zhao, Huang, Wang, Luo and
  Hao}]{zhangSLAMTKARealtimeIntraoperative2022}
\bibinfo{author}{Zhang, S.}, \bibinfo{author}{Zhao, L.},
  \bibinfo{author}{Huang, S.}, \bibinfo{author}{Wang, H.},
  \bibinfo{author}{Luo, Q.}, \bibinfo{author}{Hao, Q.}, \bibinfo{year}{2022}.
\newblock \bibinfo{title}{{{SLAM-TKA}}: {{Real-time Intra-operative
  Measurement}} of~{{Tibial Resection Plane}} in~{{Conventional Total Knee
  Arthroplasty}}}. volume \bibinfo{volume}{13437 LNCS}.
\newblock \DOIprefix\doi{10.1007/978-3-031-16449-1_13}.
\bibitem[{Zhang et~al.(2021b)Zhang, Zhao, Huang, Ye and
  Hao}]{zhangTemplateBased3DReconstruction2021}
\bibinfo{author}{Zhang, S.}, \bibinfo{author}{Zhao, L.},
  \bibinfo{author}{Huang, S.}, \bibinfo{author}{Ye, M.}, \bibinfo{author}{Hao,
  Q.}, \bibinfo{year}{2021}b.
\newblock \bibinfo{title}{A {{Template-Based 3D Reconstruction}} of {{Colon
  Structures}} and {{Textures}} from {{Stereo Colonoscopic Images}}}.
\newblock \bibinfo{journal}{IEEE Transactions on Medical Robotics and Bionics}
  \bibinfo{volume}{3}, \bibinfo{pages}{85--95}.
\newblock \DOIprefix\doi{10.1109/TMRB.2020.3044108}.
\bibitem[{Zhao et~al.(2022)Zhao, Wang, Wang, Liu and
  Zhou}]{zhao3DEndoscopicDepth2022}
\bibinfo{author}{Zhao, S.}, \bibinfo{author}{Wang, C.}, \bibinfo{author}{Wang,
  Q.}, \bibinfo{author}{Liu, Y.}, \bibinfo{author}{Zhou, S.K.},
  \bibinfo{year}{2022}.
\newblock \bibinfo{title}{{{3D}} endoscopic depth estimation using {{3D}}
  surface-aware constraints}.
\newblock \bibinfo{journal}{arXiv:2203.02131 [cs, eess]}
  \href{http://arxiv.org/abs/2203.02131}{\tt arXiv:2203.02131}.
\bibitem[{Zheng et~al.(2023)Zheng, Harley, Shen, Wetzstein and
  Guibas}]{zhengPointOdysseyLargeScaleSynthetic2023}
\bibinfo{author}{Zheng, Y.}, \bibinfo{author}{Harley, A.W.},
  \bibinfo{author}{Shen, B.}, \bibinfo{author}{Wetzstein, G.},
  \bibinfo{author}{Guibas, L.J.}, \bibinfo{year}{2023}.
\newblock \bibinfo{title}{{{PointOdyssey}}: {{A Large-Scale Synthetic Dataset}}
  for {{Long-Term Point Tracking}}}, in: \bibinfo{booktitle}{Proceedings of the
  {{IEEE}}/{{CVF International Conference}} on {{Computer Vision}}}, pp.
  \bibinfo{pages}{19855--19865}.
\bibitem[{Zhou and Jagadeesan(2020)}]{zhouRealTimeDenseReconstruction2020}
\bibinfo{author}{Zhou, H.}, \bibinfo{author}{Jagadeesan, J.},
  \bibinfo{year}{2020}.
\newblock \bibinfo{title}{Real-{{Time Dense Reconstruction}} of {{Tissue
  Surface}} from {{Stereo Optical Video}}}.
\newblock \bibinfo{journal}{IEEE Transactions on Medical Imaging}
  \bibinfo{volume}{39}, \bibinfo{pages}{400--412}.
\newblock \DOIprefix\doi{10.1109/TMI.2019.2927436}.
\bibitem[{Zhou and Jayender(2021)}]{zhouEMDQSLAMRealTimeHighResolution2021}
\bibinfo{author}{Zhou, H.}, \bibinfo{author}{Jayender, J.},
  \bibinfo{year}{2021}.
\newblock \bibinfo{title}{{{EMDQ-SLAM}}: {{Real-Time High-Resolution
  Reconstruction}} of {{Soft Tissue Surface}} from {{Stereo Laparoscopy
  Videos}}}, in: \bibinfo{booktitle}{{{MICCAI}} 2021}.
\bibitem[{Zhou and Jayender(2022)}]{zhouEMDQRemovalImage2022}
\bibinfo{author}{Zhou, H.}, \bibinfo{author}{Jayender, J.},
  \bibinfo{year}{2022}.
\newblock \bibinfo{title}{{{EMDQ}}: {{Removal}} of {{Image Feature Mismatches}}
  in {{Real-Time}}}.
\newblock \bibinfo{journal}{IEEE Transactions on Image Processing}
  \bibinfo{volume}{31}, \bibinfo{pages}{706--720}.
\newblock \DOIprefix\doi{10.1109/TIP.2021.3134456}.
\bibitem[{Zhou et~al.(2024)Zhou, Zhong, Shin, Lu, Yang, Markham and
  Trigoni}]{zhouDynPointDynamicNeural2024}
\bibinfo{author}{Zhou, K.}, \bibinfo{author}{Zhong, J.X.},
  \bibinfo{author}{Shin, S.}, \bibinfo{author}{Lu, K.}, \bibinfo{author}{Yang,
  Y.}, \bibinfo{author}{Markham, A.}, \bibinfo{author}{Trigoni, N.},
  \bibinfo{year}{2024}.
\newblock \bibinfo{title}{{{DynPoint}}: {{Dynamic Neural Point For View
  Synthesis}}}.
\newblock \href{http://arxiv.org/abs/2310.18999}{\tt arXiv:2310.18999}.
\bibitem[{Zhu et~al.(2024)Zhu, Wang, Jin, Lin and
  Yu}]{zhuDeformableEndoscopicTissues2024}
\bibinfo{author}{Zhu, L.}, \bibinfo{author}{Wang, Z.}, \bibinfo{author}{Jin,
  Z.}, \bibinfo{author}{Lin, G.}, \bibinfo{author}{Yu, L.},
  \bibinfo{year}{2024}.
\newblock \bibinfo{title}{Deformable {{Endoscopic Tissues Reconstruction}} with
  {{Gaussian Splatting}}}.
\newblock \href{http://arxiv.org/abs/2401.11535}{\tt arXiv:2401.11535}.
\bibitem[{Zhu et~al.(2022)Zhu, Peng, Larsson, Xu, Bao, Cui, Oswald and
  Pollefeys}]{zhuNICESLAMNeuralImplicit2022}
\bibinfo{author}{Zhu, Z.}, \bibinfo{author}{Peng, S.},
  \bibinfo{author}{Larsson, V.}, \bibinfo{author}{Xu, W.},
  \bibinfo{author}{Bao, H.}, \bibinfo{author}{Cui, Z.},
  \bibinfo{author}{Oswald, M.R.}, \bibinfo{author}{Pollefeys, M.},
  \bibinfo{year}{2022}.
\newblock \bibinfo{title}{{{NICE-SLAM}}: {{Neural Implicit Scalable Encoding}}
  for {{SLAM}}}, in: \bibinfo{booktitle}{Proceedings of the {{IEEE}}/{{CVF
  Conference}} on {{Computer Vision}} and {{Pattern Recognition}}}, pp.
  \bibinfo{pages}{12786--12796}.

\end{thebibliography}

\end{document}